\documentclass[
a4paper,
11pt,
oneside,
singlespacing,	
onecolumn,
headsepline,	
]{Latex/Classes/PhDthesisPSnPDF}
\usepackage[latin1]{inputenc}
\usepackage[T1]{fontenc}
\usepackage[english]{babel}
\usepackage{alltt}
\usepackage[plainpages=false]{hyperref}	
\usepackage[all]{hypcap}
\usepackage{amsmath}
\usepackage{amsfonts}
\usepackage{amssymb}
\usepackage{amsmath}
\usepackage{verbatim}
\usepackage{array}
\usepackage{graphicx,wrapfig}
\usepackage{caption}
\usepackage{subcaption}
\usepackage{algorithm}
\usepackage{algpseudocode}
\usepackage{times}
\usepackage{helvet}
\usepackage{courier}
\usepackage{comments}
\usepackage{todo}
\usepackage{filecontents}
\usepackage{lettrine}
\usepackage{epigraph}
\usepackage{tikz}
\usepackage[framemethod=tikz]{mdframed}
\usepackage{listings}
\usepackage{tabularx}
\usepackage{breakcites}
\usepackage{setspace}
\usepackage{fancyhdr}
\newcommand{\st}{\operatorname{start\_time}}
\newcommand{\et}{\operatorname{end\_time}}

\newcommand{\duration}{\operatorname{duration}}

\newcommand{\Obs}{\mbox{$\cal O$}}
\newcommand{\calD}{\mbox{$\cal D$}}
\newcommand{\calE}{\mbox{$\cal E$}}
\newcommand{\sbs}{\,\operatorname{start\_before\_start}}
\newcommand{\ebe}{\,\operatorname{end\_before\_end}}
\newcommand{\sbe}{\,\operatorname{start\_before\_end}}
\newcommand{\ebs}{\,\operatorname{end\_before\_start}}
\newcommand{\values}{\operatorname{values}}
\newcommand{\val}{\operatorname{val}}
\newcommand{\before}{\operatorname{before}}
\newcommand{\equals}{\operatorname{equals}}
\newcommand{\meets}{\operatorname{meets}}
\newcommand{\overlaps}{\operatorname{overlaps}}
\newcommand{\finishes}{\operatorname{finishes}}
\newcommand{\contains}{\operatorname{contains}}
\newcommand{\during}{\operatorname{during}}
\newcommand{\starts}{\operatorname{starts}}

\newcommand{\startat}{\operatorname{starts\_at}}
\newcommand{\xendat}{\operatorname{ends\_at}}
\newcommand{\xendbefore}{\operatorname{ends\_before}}
\newcommand{\startafter}{\operatorname{starts\_after}}
\newcommand{\xendafter}{\operatorname{ends\_after}}
\newcommand{\startbefore}{\operatorname{starts\_before}}

\newcommand{\defi}[1]{\mbox{\em #1}}
\newcommand{\comma}{,\allowbreak}
\newcommand{\tl}{\mbox{\bf STL}}
\newcommand{\ftl}{\mbox{\bf FTL}}
\newcommand{\C}{\mbox{$\cal C$}}
\renewcommand{\S}{\mbox{$\cal S$}}
\newcommand{\G}{\mbox{$\cal G$}}
\newcommand{\D}{\mbox{$\cal D$}}

\newcommand{\R}{\mbox{$\cal R$}}
\renewcommand{\P}{\mbox{$\cal P$}}
\newcommand{\Rel}{\mbox{\bf R}}
\newcommand{\tpRel}{\mbox{\bf R}'}
\newcommand{\rel}{\rho}

\newcommand{\esa}{\textsc{Esa}}

\newcommand{\apsi}{\textsc{APSI-TRF}}
\newcommand{\hsts}{\textsc{HSTS}}
\newcommand{\omps}{\textsc{OMPS}}
\newcommand{\ps}{\textsc{P\&S}}

\newcommand{\nop}[1]{}
\newcommand{\trex}{\textsc{T-Rex}}
\newcommand{\ixtet}{\textsc{IxTeT}}
\newcommand{\ixtetexec}{\textsc{IxTeT-eXeC}}
\newcommand{\idea}{\textsc{IDEA}}
\newcommand{\europa}{\textsc{EUROPA}}

\newcommand{\strips}{\textsc{STRIPS}}
\newcommand{\pddl}{\textsc{PDDL}}
\newcommand{\nddl}{\textsc{NDDL}}
\newcommand{\ddl}{\textsc{DDL}}
\newcommand{\pdl}{\textsc{PDL}}
\newcommand{\cbtp}{\textsc{CBTP}}

\newcommand{\epsl}{\textsc{EPSL}}
\newcommand{\kbcl}{\textsc{KBCL}}
\newcommand{\rover}{\textsc{Rover}}

\newcommand{\hatp}{\textsc{HATP}}
\newcommand{\ai}{\textsc{AI}}
\newcommand{\hrc}{\textsc{HRC}}

\newcommand{\nasa}{\textsc{NASA}}

\newcommand{\fape}{\textsc{FAPE}}
\newcommand{\chimp}{\textsc{CHIMP}}
\newcommand{\htn}{\textsc{HTN}}
\newcommand{\ppddl}{\textsc{PDDL2.1}}
\newcommand{\anml}{\textsc{ANML}}
\newcommand{\gecko}{\textsc{GECKO}}
\newcommand{\fbt}{\textsc{FourByThree}}
\newcommand{\shop}{\textsc{SHOP2}}
\newcommand{\oplan}{\textsc{O-Plan}}
\newcommand{\lpg}{\textsc{LPG}}
\newcommand{\lama}{\textsc{LAMA}}
\newcommand{\colin}{\textsc{COLIN}}
\newcommand{\optic}{\textsc{OPTIC}}
\newcommand{\popf}{\textsc{POPF}}
\newcommand{\ase}{\textsc{ASE}}
\newcommand{\ipem}{\textsc{IPEM}}
\newcommand{\cpef}{\textsc{CPEF}}
\newcommand{\ff}{\textsc{FF}}
\newcommand{\satplan}{\textsc{SATplan}}
\newcommand{\dolce}{\textsc{DOLCE}}
\newcommand{\edg}{\textsc{EDG}}
\newcommand{\protege}{\textsc{Protégé}}
\newcommand{\tga}{\textsc{TGA}}
\newcommand{\tiga}{\textsc{UPPAAL-TIGA}}
\newcommand{\laas}{\textsc{LAAS-CNRS}}
\newcommand{\ros}{\textsc{ROS}}
\newcommand{\ie}{i.e.}
\newcommand{\eg}{e.g.}
\newcommand{\binf}{s}
\newcommand{\bsup}{s'}
\newcommand{\einf}{e}
\newcommand{\esup}{e'}
\newcommand{\dinf}{d}
\newcommand{\dsup}{d'}
\newcommand{\chapquote}[3]{\begin{quotation} \textit{\small #1} \end{quotation} \begin{flushright} - {\small #2}, \textit{\small #3}\end{flushright} }
\graphicspath{{frontmatter/figures/}}
\newmdenv[tikzsetting={draw=lightgray,fill=lightgray,fill opacity=0.2}]{highlights}
\newtheorem{definition}{Definition}
\newtheorem{theorem}{Theorem}
\newtheorem{examplecontent}{Example}
\newenvironment{example}
{\begin{highlights}\small\begin{examplecontent}}
{\end{examplecontent}\end{highlights}}
\newenvironment{code}
{\begin{highlights}\footnotesize\begin{alltt}}
{\end{alltt}\end{highlights}}
\def\Nat{\mathbb{N}}
\def\tp{\mathbb{T}}
\def\reals{\mathbb{R}_{\geq 0}}
\begin{document}
%
%
%
\begin{titlepage}
\begin{center}
\includegraphics[width=.48\textwidth]{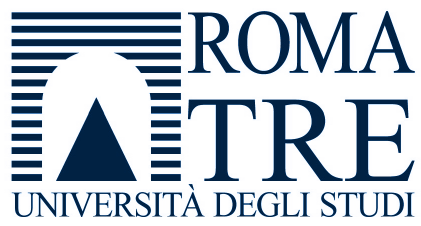}
\end{center}
\vspace{.1cm}
\begin{center}
\Large
{\sc Corso di Dottorato di Ricerca in Informatica e Automazione}\\[.5cm]
{\sc xxix ciclo}
\end{center}
%
%
\vfill
\begin{center}
\LARGE
{\sc Timeline-based Planning and Execution with Uncertainty:\\[.1cm]
Theory, Modeling Methodologies and Practice}
\end{center}
%
%
%
%
\vfill
\begin{center}
\begin{tabularx}{\textwidth}{lXXr}
{\em Dottorando:} & & & \\[.1cm]
~~Alessandro Umbrico & & & \_\_\_\_\_\_\_\_\_\_\_\_\_\_\_\_\_\_\\[.5cm]
{\em Docenti guida:} & & &\\[.1cm]
~~Prof.ssa Marta Cialdea Mayer & & & \_\_\_\_\_\_\_\_\_\_\_\_\_\_\_\_\_\_\\[.4cm]
~~Dr. Andrea Orlandini & & & \_\_\_\_\_\_\_\_\_\_\_\_\_\_\_\_\_\_\\[.5cm]
{\em Coordinatore:} & & & \\[.1cm]
~~Prof. Stefano Panzieri & & & \_\_\_\_\_\_\_\_\_\_\_\_\_\_\_\_\_\_
\end{tabularx}
\end{center}
\end{titlepage}
\setstretch{1.2}
%
%
\pagenumbering{Roman}
\mbox{}
\thispagestyle{empty}
\newpage


\begin{abstracts}       	 
Automated Planning is one of the main research field of Artificial Intelligence since 
its beginnings. Research in Automated Planning aims at developing general 
reasoners (\ie\ planners) capable of automatically solve complex problems.
Broadly speaking, planners rely on a general model characterizing the 
possible states of the {\em world} and the actions that can be 
performed in order to  change the status of the world. 
Given a model and an initial known state, the objective of a planner is to 
synthesize a set of actions needed to achieve a particular goal state.
The {\em classical approach} to planning roughly corresponds to the description given 
above. However, many planning techniques have been introduced in the literature relying 
on different formalisms and making different assumptions on the features of the model of the 
world.
%
The {\em timeline-based approach} is a particular planning paradigm capable 
of integrating causal and temporal reasoning within a unified solving process. 
This approach has been successfully applied in many real-world scenarios although 
a common {\em interpretation} of the related planning concepts is missing. 
Indeed, there are significant differences among the existing frameworks that 
apply this technique. Each framework relies on its own interpretation of 
timeline-based planning and therefore it is not easy to compare these systems.
Thus, the objective of this work is to investigate the timeline-based approach 
to planning by addressing several aspects ranging from the semantics of the 
related planning concepts to the modeling and solving techniques. 
Specifically, the main contributions of this PhD work consist of: 
(i) the proposal of a formal characterization of the timeline-based approach capable 
of dealing with {\em temporal uncertainty}; 
(ii) the proposal of a hierarchical modeling and solving approach; 
(iii) the development of a general purpose framework for planning and execution with timelines; 
(iv) the validation of this approach in real-world manufacturing scenarios.
\end{abstracts}


%

\begin{acknowledgements}      
{\em
I would like to express my sincere gratitude to my advisors Prof. Marta Cialdea Mayer and 
Dr. Andrea Orlandini for their continuous support of my Ph.D study and related research, 
for their patience, motivation and knowledge. Their guidance helped me in all the time of
research and writing of this thesis. I could not have imagined having better advisors and 
mentors for my Ph.D study.

My sincere thanks also goes to Prof. Joachim Hertzberg from Osnabruck University who 
provided me the opportunity to join his team as an intern. It has been a great professional 
and personal experience for me.

I thank my fellow labmates and all my friends in the Institute of Cognitive Science and 
Technology of National Research Council of Italy. 
Without their precious support, insightful comments and encouragement 
it would not be possible to conduct this research. In particular, I am grateful to Dr. Amedeo 
Cesta, for enlightening me the first glance of research, and to Dr. Stefano Borgo, for 
introducing me into the world of ontologies.
 
Last but not least, I would like to thank my parents and my wife for giving me all the 
love that I need. 
}
\end{acknowledgements}


%
%
\setcounter{tocdepth}{3}
\setcounter{secnumdepth}{3}
\newpage
\pagenumbering{roman}
\setcounter{page}{1}
%
\tableofcontents
%
%
\listoffigures
\listofalgorithms
%
%
%
%
%
\newpage
\pagenumbering{arabic}
\setcounter{page}{1}

\ifpdf
    \graphicspath{{chapters/1_introduction/figures/PNG/}{chapters/1_introduction/figures/PDF/}{chapters/1_introduction/figures/}}
\else
    \graphicspath{{chapters/1_introduction/figures/EPS/}{chapters/1_introduction/figures/}}
\fi


%
%
%
%
\chapter{Introduction}
\label{chap:intro}
\chapquote{"Maybe the only significant difference between a really smart simulation and a human being
was the noise they made when you punched them."}{Terry Pratchett}{The Long Earth}
\lettrine[lines=2]{A}{rtificial Intelligence} (\ai) is the field of Computer Science that deals with the development of 
techniques that aim at endowing machines with some sort of {\em intelligence}. There are different research fields 
in \ai\ that characterize {\em intelligence} in different ways and realize different types of intelligent machines accordingly. 
Broadly speaking, {\em "the term {\em artificial intelligence} is applied when a machine mimics {\em cognitive} functions 
that humans associate with other human minds, such as learning and problem solving"} as stated in \cite{russell-norvig}.
{\em Automated Planning} is one of the core fields of \ai\ since its beginnings. Its research objective is to endow a 
machine (an {\em artificial agent}) with the capability of autonomously carry out complex tasks. From a practical point 
of view, this is a key enabling feature in application scenarios where direct human involvement is neither possible nor 
safe, \eg\ space mission or deep sea exploration. Moreover, the recent and continuous improvements of robotic platforms 
with respect to reliability and efficiency represent a great opportunity for deploying \ai-based techniques in even more 
common application scenarios (\eg\ domestic care, manufacturing, rescue missions).

A {\em planner} is a general {\em problem solver} able to automatically {\em synthesize} a set of {\em actions} that allow 
an agent to achieve some objectives (\eg\ explore and gather scientific data about an unknown environment or accomplish
some complex task within the production process of a factory). A planning system usually relies on a {\em model} which 
represents a general description of the {\em world}. The model characterizes the {\em environment} the agent is supposed 
to operate in, and the {\em agent capabilities} in terms of the {\em actions} the agent can perform to interact with the 
environment. 
The {\em classical approach} to planning relies on a {\em logical} characterization of the model that focuses on the causal 
aspects  of the problem to solve. {\em States} consist of sets of {\em atoms} asserting known {\em facts} and {\em properties} 
about the world, \eg\ the position of a robot or an object in the environment. {\em Actions} encode transitions between states by 
specifying {\em preconditions} and {\em effects}. Preconditions specify a set of conditions that must be true (\ie\ atoms) in order 
to apply the action in a particular state. Effects specify conditions that become true (\ie\ positive effects) or false (\ie\ negative
effects) after the application of the action. For example, the action of {\em moving} an object from an {\em initial location} to a
{\em destination location} can be applied to all states in which the object to move is located at {\em initial location}. The 
states resulting from the application of the action are all those states in which the object is located at the {\em destination 
location}. A {\em planning goal} usually consists of a logical formula representing the {\em goal state} to achieve, \eg\ the 
state in which all objects of the domain are in a desired location. Thus, given an {\em initial state} containing a set 
of known facts about the world (\eg\ the initial locations of the objects), the planning system must synthesize a set of 
actions needed to reach the {\em goal state}.

\strips\ \cite{strips} is one of the first planning system introduced into the literature whose language inspired the \pddl\ 
\cite{mcdermott98pddl} which is the {\em standard modeling language} for planning. Many {\em planning systems} 
have been developed, \eg\ \satplan\ \cite{satplan}, \ff\ \cite{ff} or \lpg\ \cite{lpg}, that rely on the classical 
planning formalism and \pddl\ language. 
These planners have also shown relevant solving capabilities on {\em toy problems} during the 
International Planning Competitions. However, from a practical point of view, the classical planning formalism makes 
{\em strong assumptions} on the features of the problems to model. These assumptions {\em limit} the capabilities of classical 
planners to address real-world problems. For example, classical planning paradigms use an {\em implicit representation} of time 
which does not allow planners to deal with concurrency, temporal constraints or durative actions that are crucial in real-world 
scenarios. Consequently, several extensions to classical planning have been introduced into the literature in order to overcome
these limitations and address more realistic problems. These extensions lead to the definition of several planning paradigms 
that {\em relax} different assumptions of the classical approach. 
{\em Temporal planning} represents the class of planning paradigms that introduce an {\em explicit representation of time} into 
the modeling language.

The timeline-based approach is a particular {\em temporal planning} paradigm introduced in early 90's with \hsts\ 
\cite{hsts-94} which has been successfully applied to solve many real-world problems (in space-like contexts mainly). 
This approach takes inspiration from classical {\em control theory} and 
is characterized by a more practical than logical view of planning.
The timeline-based approach focuses on the {\em temporal behavior} of a system and the related features that must be 
controlled. Specifically, a complex system (\eg\ an exploration rover) is modeled by identifying a set of relevant features that 
must be controlled over time within a known {\em temporal horizon} (\eg\ the wheeled base of the robot or the communication 
facility). The control process consists in the synthesis of a set of temporal behaviors (\ie\ the {\em timelines}) that describe 
how the modeled features evolve over time. 
The main advantage of planning with timelines consists in the representation approach, which allows the planner to 
deal with time and temporal constraints while building a plan. Namely, the timeline-based representation fosters 
a hybrid solving procedure by means of which it is possible to integrate planning and scheduling in 
a unified reasoning mechanism. 
In general, {\em hybrid reasoning} is essential to effectively address real world problems. Indeed, the key factor 
influencing the successful application of planning technologies to real-world problems is the capability of simultaneously 
dealing with different aspects of the problem like causality, time, resources, concurrency or uncertainty at solving time.

Despite the practical success of timeline-based approach, formal frameworks characterizing this formalism have been 
proposed only recently. There are several planning systems that have been 
introduced into the literature, \eg\ \europa\ \cite{europa2012}, \ixtet\ \cite{ghallab1994representation}, \apsi\ \cite{apsigoac11}, 
each of which applies its own {\em interpretation} of timeline-based planning. Moreover, developed Planning and Scheduling (\ps) 
applications are strictly connected to the specific context they have been designed for. Thus, existing timeline-based 
applications are hard to adapt to different problems. In general, there is a lack of methodology in modeling and solving 
timeline-based problems. Given the elements that compose a particular domain, it is not easy to design a suited model 
in order to ralize effettive \ps\ solutions. In addition, different systems apply different solving approaches and generate
plans with different features. Thus, it is not simple to compare different timeline-based systems and it is even more 
difficult to compare such systems with other existing approaches.

\subsection*{Contribution}
The objective of this work is to investigate timeline-based planning by taking into account several aspects 
ranging from the semantics of the main planning concepts to the modeling and solving approach. Thus the contribution
of the work involves (i) the proposal of a formal characterization of the timeline-based approach which takes into account 
also {\em temporal uncertainty}, (ii) the proposal of a hierarchical modeling and solving approach, (iii) the 
development of a general-purpose framework for planning and execution with timelines 
(\epsl\ - {\em Extensible Planning and Scheduling Library}), which complies with the proposed formalization and 
implements the proposed hierarchical solving procedure and lastly (iv) the {\em validation} of \epsl\ and 
the envisaged approach to timeline-based planning in real-world manufacturing scenarios.

The proposed formalization defines a clear semantics of concepts like {\em timelines}, timeline-based {\em plans} and 
{\em state variables}, representing the basic building blocks of a planning domain. In particular, the formalization takes 
into account the {\em controllability properties} in order to model the {\em temporal uncertainty} concerning 
the {\em uncontrollable} features of a domain. This is particularly relevant for real-world problems, where not all the features 
of the domain are controllable with respect to the point of view of the artificial agent. Namely, the environment has 
{\em uncontrollable dynamics} that may affect the behavior of the system to control and the outcome of its operations 
(\eg\ the visibility of the ground station for the communication operations of a satellite). 
The timeline-based plans, generated according to the proposed formalization, contain information about the 
{\em uncertainty} of the domain that can be analyzed to characterize the {\em robustness} of the plan with respect to 
its execution.
There are several works in this field \cite{VidalF99,morris01,ker09} aiming at analyzing the plan in order to understand if, 
given the possible evolutions of the uncontrollable features of the domain, it is possible to complete the execution of the 
plan. With respect to planning, it is important to leverage the controllability information about 
the domain during the solving process in order to generate plans with some desired controllability properties (if possible) 
and therefore, have some information regarding their {\em executability}.

Given an agent to control, the proposed modeling approach follows a hierarchical specification of the domain which is 
similar to \htn\ planing \cite{htn-overview}.
%
Specifically the approach proposes a functional characterization of the agent at different levels of abstraction. 
Broadly speaking, a {\em primitive level} characterizes the functional behavior of the physical/logical elements composing 
the agent in terms of commands they can directly manage over time. 
{\em Functional levels} model complex functions/operations the agent could perform over time by leveraging its 
components.
Namely, functional levels model complex activities (\ie\ complex tasks) the agent can perform 
by combining the available commands (\ie\ primitive tasks). Domain rules, like {\em methods} in \htn\ planning, describe the 
operational constraints that allow the agent to implement tasks. They specify hierarchical decomposition of complex tasks in 
sets of constraints between primitive tasks. The resulting hierarchical structure encodes specific knowledge about the 
domain that the planning system can leverage during solving. Specifically, this work introduces {\em search heuristics} that 
leverages the hierarchical structure of the planning domain to support the plan generation process.

The \epsl\ framework complies with both the formalization and hierarchical modeling/solving approaches presented. \epsl\ 
is the major result of this study. It realizes a uniform framework for planning and execution with timelines under 
uncertainty. From the planning point of view, \epsl\ implements a hierarchical solving approach which is capable of dealing
with temporal uncertainty during plan generation. Specifically, the solving procedure leverages information about the 
{\em temporal uncertainty} of the planning domain in order to generate plans with some properties characterizing their 
{\em robustness} with respect to the execution in the real-world.
From the execution point of view, \epsl\ executes the timeline of a plan by taking into account the controllability properties 
of the related values and adapting the plans to the {\em unexpected} behaviors of the environment.
\epsl\ planning and execution capabilities have been successfully applied to real-world manufacturing scenarios showing 
the effectiveness of the proposed approach. 

%
%
%
%
\subsection*{Outline}
Chapter \ref{chap:planning} provides a brief description of the background of Automated Planning in \ai\ by describing the 
classical approaches to planning, the limit of these approaches in solving real-world problems and how they have been 
improved in order to address more realistic problems.
Chapter \ref{chap:timelines} provides a more detailed overview of the timeline-based planning approach and the related 
state of the art prior to this study. In particular, this chapter describes some of the most relevant timeline-based systems 
introduced into the literature (\europa, \ixtet\ and \apsi) together with a brief description of the temporal formalisms this kind 
of systems usually relies on.
Chapter \ref{chap:formalization} enters into the details of the contribution of the study by describing the proposed 
formalization of the timeline-based approach and the related {\em controllability problem}. 
Chapter \ref{chap:epsl} describes \epsl\ its structure and the implemented hierarchical modeling and solving approach.
Chapter \ref{chap:hrc} presents a relevant extension of \epsl\ that allows the framework to execute plans 
while managing {\em temporal uncertainty}. This chapter also describe the deployment of \epsl\ to an interesting real-world
manufacturing scenario of Human-Robot Collaboration (\hrc). In particular, \hrc\ applications represent 
well-suited contexts to leverage the \epsl\ capabilities of dealing with {\em temporal uncertainty} at planning and execution time.
Chapter \ref{chap:kbcl} presents another interesting application of the \epsl\ framework and its integration with {\em semantic
technologies} for realizing an extended plan-based control architecture.
Specifically, the chapter presents a flexible control architecture, called \kbcl\ ({\em Knowledge-based Control Loop}), which has 
been applied to a real-world scenario for controlling reconfigurable manufacturing systems. \kbcl\ aims at realizing a flexible 
control process able to dynamically adapt the control model to the different situations that may affect the capabilities of the system. \kbcl\ 
investigates the integration and the correlations of ontological analysis and knowledge processing with the 
timeline-based planning approach.
Finally chapter \ref{chap:conclusions} draws some conclusions by providing an assessment of the achieved results and 
illustrates some of the most relevant open points that must be addressed in the near future.

\ifpdf
    \graphicspath{{chapters/2_planning/figures/PNG/}{chapters/2_planning/figures/PDF/}{chapters/2_planning/figures/}}
\else
    \graphicspath{{chapters/2_planning/figures/EPS/}{chapters/2_planning/figures/}}
\fi


%
%
%
\chapter{Planning in Artificial Intelligence}
\label{chap:planning}
\lettrine[lines=2]{P}{lanning} is one of the most relevant research field of \ai\ since its beginnings. The objective
of a planning system is to automatically solve a problem by synthesizing a set of {\em operations} (\ie\ a plan) 
needed to reach a desired {\em goal} (\ie\ a desired state or configuration). There are many practical field 
like robotics or manufacturing where planning technologies have provided a significant contribution. Let us 
consider, for example, planetary exploration rovers that must operate in a context where direct human 
control is not possible. In such a context, planning technologies provide the rover with the {\em autonomy} 
needed for navigating an unknown environment and gathering scientific data to communicate.

There are different ways to describe the fundamental elements of a planning system. 
Such differences have lead to different {\em planning paradigms} ranging from those addressing fully observable, 
deterministic, static and discrete environments, to those that deal with partially observable stochastic environments. 
This chapter does not aim at presenting a complete background on all the planning technologies and systems that 
have been introduced into the literature. Thus, after a brief overview of some {\em classical} approaches to planning 
in section \ref{sec:classical-planning}, section \ref{sec:planning-real} explains the limits of these planning paradigms 
and the improvements needed to address real-world problems. 
Finally, section \ref{sec:temporal-planning} focuses on a particular class of planning paradigms (\ie\ {\em Temporal 
Planning}) which extends the classical approach by introducing an explicit representation of time. 

\section{Classical Planning}
\label{sec:classical-planning}
Broadly speaking, a planning system is a general problem solver whose aim is to synthesize a set of operations
that, given an initial state, allow the system to reach a desired goal state. The reasoning process relies on a 
{\em model} which represents a general description of the problem to solve. The model provides a representation of 
the {\em environment} in terms of the possible {\em states of the world} and the {\em actions} the system can perform 
to {\em interact} with the environment. Thus, a planning process starts from an {\em initial state} and iteratively moves 
to other states by {\em applying} the available actions until a desired {\em goal state} is reached.

An example of a simple planning problem is represented by the {\em Vacuum World} problem described in 
\cite{russell-norvig}. The problem consists of a set of rooms that can be either clean or dirty, and a vacuum cleaner which 
can {\em move} between (adjacent) rooms and {\em clean} the room the vacuum is located in. 
In this regard, a {\em state of the world} describes the set of rooms that compose the environment, their connections 
(\ie\ whether two rooms are adjacent or not), their states (\ie\ whether the rooms are clean or not), and the current room of 
the vacuum cleaner. The {\em goal state} is the state of the world where all rooms are clean. 
The {\em initial state} describes the status of all the rooms and the particular room the vacuum cleaner is initially located in.

\begin{figure}[ht]
\centering
\includegraphics[width=.98\textwidth]{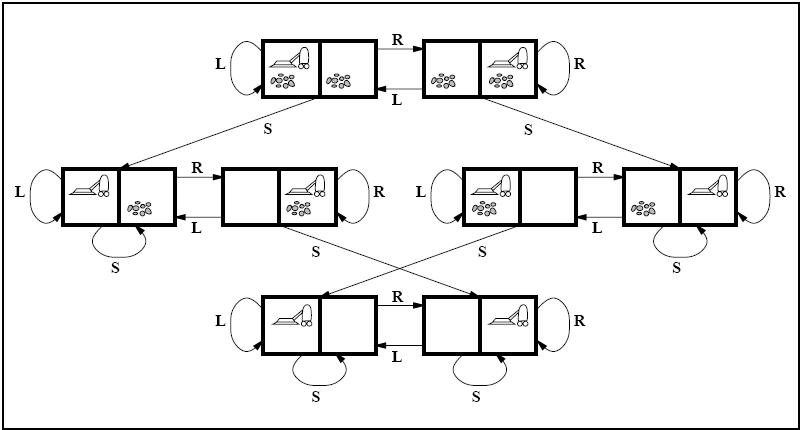}
\caption{\small{The state space of the Vacuum World domain}}
\label{fig:vacuum}
\end{figure}

Figure \ref{fig:vacuum} shows the {\em state space} for the {\em Vacuum World} problem with two adjacent rooms. The 
state space can be seen as a directed graph where the possible states of the world are the nodes and actions are the 
(directed) edges connecting two states of the world. Let us consider, for example, the state in Figure \ref{fig:vacuum} where both 
the rooms are dirty and the cleaner is located in the room on the left. The {\em execution} of action R (\ie\ move right) leads to the 
state where both the rooms are dirty and the cleaner is located in the room on the right. Similarly, the {\em execution} of action 
S (\ie\ clean/suck) leads to the state where the cleaner has not changed its position, the room on the left is clean and the room 
on the right is dirty.

Given a state space like the one depicted in Figure \ref{fig:vacuum} and a known initial state, the planning process 
must find a sequence of actions needed to reach the state where both rooms are clean. 
The {\em Vacuum World} problem described above is very simple because states are fully observable (\eg\ it is always possible to 
know whether a room is clean or dirty), actions are deterministic (\ie\ there is not {\em uncertainty} about the effects of actions) 
and the search space is small. Thus, the planning process must simply find a {\em path} on the graph (\ie\ the search space) 
connecting the initial state with the goal state. 
However, planning problems are not always fully observable or deterministic and typically entail huge search spaces that cannot 
be explicitly represented. A more compact and {\em expressive} representation/description of planning problems is needed 
and therefore several modeling languages and planning paradigms have been introduced into the literature.

\subsection{\strips}
\strips\ ({\em STanford Research Institute Problem Solver}) \cite{strips} is one of the first automated planner and 
language used in \ai. The \strips\ modeling language has represented the basic formalism for many planning paradigms 
that have been introduced successively. The formalism relies on the {\em first-order predicate calculus} to represent the 
space of {\em world models} the planning system must search in order to find a particular world model (\ie\ a state), where 
a desired goal formula is achieved. 
A world model consists of a set of {\em clauses}, \ie\ formulas of first-order predicate logic that describe a particular 
{\em situation} concerning the environment and the agent. For example, considering a robotic planning problem the related
 world models will contain a set of formulas concerning the position of the robot and all objects of the environment.
 {\em Operators} are particular transition functions that allow the planning system to move from one world model to 
others. It is supposed that for each world model there exists at least one operator which could be {\em applied} to 
"transform" the related world model into another. Thus the resulting problem solver must find the appropriate 
{\em composition} of operators that transform an initial world model to a "final" world model which satisfies a {\em goal 
condition} (\ie\ a particular logical formula).

The problem space of a \strips\ problem is defined by the initial world model, the set of available operators and the goal
states. Operators are grouped by {\em schema} which models a set of {\em instances} of applicable operators. Let us 
consider for example the operator {\em goto}, used for moving a robot between two points on a floor. In such a case, 
there is a distinct operator for each pair of points of the floor. Therefore it is more convenient to group all these possible instances into an 
operator schema {\em goto(m, n)} parametrized by the initial and final positions ({\em m} and {\em n} respectively).
Specifically, an operator schema describes the {\em effects} and the {\em conditions} under which the operator is 
applicable. Effects specify the list of formulas that must be added to the model (the {\em add list}) and a list of formulas that 
must be removed (the {\em delete list}). Let us consider the example described in \cite{strips} concerning a operator 
{\em push(k, m, n)} which models the action of pushing an object {\em k} from {\em m} to {\em n}.
Such an operation can be modeled by the code below where ATR(m) is a predicate stating that the robot is at 
location {\em m}, and AT(k, m) is a predicate stating that the object {\em k} is at location {\em m}.

\begin{code}
push(k, m,n)
   precondition:   ATR(m) \(\land\) AT(k, m)
   delete list:   ATR(m), AT(k, m)
   add list:   ATR(n), AT(k, n)
\end{code}

\subsection{\pddl}
The {\em Problem Domain Description Language} (\pddl) is an action-based language introduced in \cite{mcdermott98pddl} 
for the AIPS-98 planning competition. \pddl\ relies on the \strips\ formalism and aims at defining a standard syntax for 
expressing planning domains.
An early design decision was to separate the description of parametrized actions of the domain from the description of 
the objects, initial conditions and goals that characterize problem instances. Thus the domain description defines  the 
general rules and behaviors that characterize as specific application scenario/context. Given a domain description, a 
problem description instantiates a planning problem in terms of specific type and number of objects, initial conditions 
and goals. 
In this way, a particular domain description can be used to define many different problem descriptions.
\pddl\ defines parametrized actions by using variables denoting elements of a particular problem instance. Indeed, variables 
are instantiated to objects of the specific problem description when actions are grounded for applications. Preconditions and 
effects of actions are logical propositions constructed from predicates, arguments (\ie\ objects from a problem instance) and 
logical connectivities. 
Moreover, \pddl\ extends the expressive power of \strips\ formalism by including the ability to descirbe structured object types, 
specify types for action parameters, specify actions with negative preconditions and conditional effects, as well as  
introduce the use of quantification in expressing both pre- and post- conditions. The code below shows an example of a 
simple \pddl\ action which allows a rover to move between two locations.

\begin{code}
(: move
   :parameters (?r - rover ?from ?to - location)
   :precondition (and (at ?r ?from) 
         (path ?from ?to))
   :effect (and (not (at ?r ?from))
         (at ?r ?to))
)
\end{code} 

The action $move$ has one parameter denoting the particular rover which is moving, and two other parameters 
denoting the specific locations the rover moves from and to. Action preconditions specify the conditions that must hold 
to apply actions. An instance of the action $move$ can be applied if the rover, the action refers to, is at the starting 
location (\ie\ the location denoted by variable $?from$) and there exists a path connecting the starting location with the 
destination (\ie\ the location denoted by variable $?to$). 
Action effects specify the state resulting from the application of the action. Thus, once the action has been applied, the rover 
denoted by variable $?r$, is no longer at location $?from$ (negative effect) but is at location denoted by the variable $?to$.
Note that no temporal information is associated with action descriptions. Therefore, effects of actions become 
valid (\ie\ true) as soon as actions are applied. Namely, actions in \pddl\ are instantaneous and there is not an explicit 
representation of time.

\subsection{\htn}
{\em Hierarchical Task Network} (\htn) planning \cite{htn-overview} is a particular paradigm which relies on the 
\pddl-based formalism. Like \pddl, atoms represent states of the world and actions represent deterministic state transitions. 
However the objective of HTN planners like \shop\ \cite{shop2,shop} or \oplan\ \cite{oplan} is to generate 
a sequence of actions that perform some {\em tasks}. A task represents an activity to perform which can be either
{\em primitive} or {\em compound}. Primitive tasks are accomplished by {\em planning operators} that, like \pddl\ operators/actions, 
describe transitions between states of the world. Compound tasks represent complex activity that cannot be directly "executed" 
and need to be further decomposed into a set of "smaller" tasks.
In \htn\ planning, the objective is to synthesize a set of actions (\ie\ primitive tasks) realizing a complex activity 
(\ie\ a compound task) rather than reaching a desired goal state like classical planners. 
Thus, \htn\ domain description consists of a set of {\em operators} that describe the primitive tasks and a set of 
{\em methods} that specify how to decompose {\em complex tasks} into subtasks. Methods specify the 
hierarchical task decomposition \htn\ planners uses to recursively decompose tasks into a set of subtasks. 
Methods decompose tasks until {\em primitive tasks} are found and no further decomposition is needed. The resulting 
decomposition tree encode domain specific knowledge describing the {\em standard operating procedures} 
to use in order to perform tasks. Such a knowledge supports and guides the solving process of \htn\ planners. 
Although \htn\ solving procedure is general and domain independent, method specification is domain-dependent and 
characterizes the specific procedure to follow in order to realize complex tasks in the considered domain. 

\begin{code}
(:method
   ; head
      (transport-person ?p ?c2)
   ; precondition
      (and (at ?p ?c1) 
         (aircraft ?a) 
         (at ?a ?c3) 
         (different ?c1 ?c3))
   ; subtasks
      (:ordered
         (move-aircraft ?a ?c1)
         (board ?p ?a ?c1)
         (move-aircraft ?a ?c2)
         (debark ?p ?a ?c2))
)
\end{code}

The block of code above shows an simple example of a \shop\ method defined in \cite{shop2}, for a simplified versione 
of the \textsc{ZenoTravel} domain of the AIPS-2002 Planning Competition. The method describes how to transport a 
person $?p$ by aircraft from a location $?c1$ to another location $?c2$ in the case that the aircraft is not located at $?c1$. 
The $ordered$ keyword concerns task decomposition and specifies the order the planner must follow to expand subtasks.
Thus, first the aircraft moves to location $?c1$, then the aircraft boards the person $?p$, then the aircraft moves to location 
$?c2$ and finally the aircraft debarks the person $?p$.

\section{Planning in the Real-World}
\label{sec:planning-real}
The modeling features of classical planning approaches rely on a set of assumptions that make {\em strong simplifications} of 
the problems to address with respect to real-world scenarios. Indeed, classical planning mainly deals with {\em static}, {\em fully
observable} and {\em deterministic} domains. It means that given any state of the environment and a particular 
action, it is possible to know exactly which is the next state of the system. Such an assumption does not hold in real-world
contexts where the environment may be {\em partially observable} and something may be either unknown or unpredictable. 
In such a case the planning system should be able to handle the {\em uncertainty} of the domain and find a sequence of actions 
that still reach the goal state.

Let us consider again for example, the {\em Vacuum World} domain of Figure \ref{fig:vacuum}, where the environment 
described is fully observable. At any state it is possible to know where the vacuum cleaner is located or it is possible to 
know exactly whether a room is clean or dirty. Similarly, the actions of the vacuum cleaner are deterministic and therefore, 
the state resulting from the application of an action is known. Let us consider, for example, the state where both rooms are 
dirty and the vacuum cleaner is located in the left room. If the the {\em Suck} operation is applied to this state, then the (only)
successor state is the one with left room clean, the right room dirty and the vacuum cleaner still located in the left room.
Such a simple problem can be made more "realistic" and more challenging if one or more assumptions are removed. Let us 
suppose to remove the assumption about the {\em full observability} of the environment and that it is not possible to 
know whether the rooms are clean or dirty. In such a case, it is necessary to find a sequence of actions that, independently 
from the actual state of the rooms, allows the system to reach a state where certainly both rooms are clean.
Moreover, classical planning approaches have an {\em implicit} representation of {\em time}. Actions are supposed to be 
instantaneous, which means that the effects of an action become true as soon as the action is applied. States and/or goals 
are not supposed to have a {\em temporal extension} such that they hold only for a limited temporal interval, or that they must 
be achieved within a known temporal bound. Again this is a significant simplification in real-world contexts where time, 
{\em temporal constraints} (\eg\ deadlines for goal achievement) and {\em concurrency} (\eg\ a limit on the number of jobs 
that a machine can perform simultaneously) represent strong requirements that must be satisfied by plans.

There are several \pddl-based planning systems \eg\ \satplan\ \cite{satplan}, 
\ff\ \cite{ff}, \lpg\ \cite{lpg} or \lama\ \cite{lama}, that have shown excellent solving capabilities during the International 
Planning Competitions. However, all the assumptions described above limit the {\em expressivity} of classical planning systems 
and their efficacy to address real-world problems. 
Consequently several planning approaches have been developed, with the intention of overcoming these limitations 
by removing one or more of the simplifying assumptions described above. 
In particular, this work focuses on {\em Temporal Planning} which represents the "class" of planning approaches capable of 
representing information and constraints that concern the temporal features of the domain.
These kind of systems realizes problem solvers that make both planning and scheduling decisions during the solving 
process. Timeline-based planning belongs to this class of planning techniques and it will be further discussed in the next 
chapter. The following sections provide a brief description of the key modeling features of Temporal Planning, a brief 
description of \ppddl\ \cite{pddl21}, the {\em temporal} extension of \pddl, and other hybrid approaches that present some
common features with timeline-based planning like \anml\ \cite{anml}, \fape\ \cite{fape} and \chimp\ \cite{chimp}.

\section{Temporal Planning}
\label{sec:temporal-planning}
The primary distinct characteristic of temporal planning paradigms is that they synthesize plans by combining
{\em causal} reasoning with reasoning about {\em time} and {\em resource}s. They overcome 
the traditional division between planning and scheduling technologies. In this context, planning is intended
as the generation of a system behaviour that satisfies certain desired conditions over a prefixed 
temporal horizon. Therefore, planning is not only the process of deciding {\em which} actions to perform
in order to satisfy some desired conditions, but also deciding {\em when} to execute these actions in order
to obtain some desired behavior of the system. Indeed, temporal planning systems try to {\em integrate} {\em planning} 
and {\em scheduling} in a unified solving process.

\subsection{\ppddl}
\ppddl\ \cite{pddl21} has been designed to allow \pddl-based systems to model and solve more realistic domains by 
introducing the capability of dealing  with time. There are several planning systems that rely on this language, \eg\ 
\optic\ \cite{optic}, \colin\ \cite{colin} \popf\ \cite{popf}, which also maintains backward compatibility with \pddl. Existing
\pddl\ domains are valid \ppddl\ domains and valid \pddl\ plans are valid \ppddl\ plans.
A relevant contribution of \ppddl\ is the introduction of {\em discretized} durative actions with {\em temporally annotated} 
conditions and effects. 
Conditions and effects must be temporally annotated in order to specify {\em when} a particular proposition must 
{\em hold}. Specifically, a proposition (\ie\ a condition or an effect) can hold at the {\em start} of the 
interval of the action (\ie\ the time point at which the action is applied), at the {\em end} of the interval (\ie\ the time point
at which the effects of the action are asserted) or over the entire interval (\ie\ invariant over the duration of the action).
The annotation of an effect specifies whether the related effect of the action is {\em instantaneous} (\ie\ the effect becomes 
true as soon as the action is applied) or {\em delayed} (\ie\ the effect becomes true when the action finishes).
The code below shows a simple example of a durative action for loading a truck from the Dock-Worker Robots domain 
described in \cite{automated}.

\begin{code}
(:durative-action load-truck
   :parameters (?t - truck) 
            (?l - location)
            (?o - cargo)
            (?c - crane)
   :duration (= ?duration 5)
   :condition (and 
            (at start (at ?t ?l))
            (at start (at ?o ?l))
            (at start (empty ?c))
            (over all (at ?t ?l))
            (at end (holding ?c ?o)))
   :effect (and 
            (at end (in ?o ?t))
            (at start (holding ?c ?o))
            (at start (not (at ?o ?l)))
            (at end (not (holding ?c ?o))))
)
\end{code}

Invariant conditions of a durative action hold over the entire duration of the action and are specified by means of the {\em 
over all} keyword (see the code above). It is worth observing that, the {\em over all} keyword excludes the start point and 
the end point of the action interval which is considered as an {\em open temporal interval}. If a particular preposition $p$ 
must hold at the start, at the end and also during the entire duration of the action, it must be specified with three temporal 
constraints, \ie\ {\em (at start p)}, {\em (over all p)} and {\em (at end p)}.


%
%
%
%
%
\subsection{Hybrid Planning approaches}
There are other languages and planning frameworks that integrate causal and temporal reasoning without directly extending 
\pddl. An interesting planning language is the {\em Action Notation Modeling Language} (\anml) \cite{anml}. \anml\ has been 
introduced  as an alternative to existing (temporal) planning languages like \ppddl, the \ixtet\ language or \nddl\ (the \europa\ 
planning language). 
Broadly speaking \anml\ represents an high-level language whose aim is to uniformly support generative and 
\htn\ planning models and provide a clear and well-defined semantics compatible with \pddl\ family of languages. 
\anml\ relies on a strong notation of action and state, provides constructs for expressing common forms of action conditions
and effects, supports rich temporal constraints and uses a variable/value representation.

\begin{code}
action Navigate (location from, to) \{
   duration := 5 ;
   [all] \{ arm == stowed ;
               position == from :-> to ;
               batterycharge :consumes 2.0 
         \}
\}
\end{code}

The code above shows an example of an high-level navigation action for a rover expressed in \anml. The action has two 
location parameters of type {\em from} and {\em to} and a fixed duration (5 time units). The temporal qualifier {\em [all]} 
means that the related statements (\ie\ the statements contained by the adjacent block of code) are valid all along the 
duration of the action.
Specifically, the first statement specifies that the arm of the rover is stowed over the entire action. The second statement 
specifies that the position of the rover is {\em from} at the start of the action (a condition), the position is {\em undefined} 
during action execution, and the position is {\em to} at the end of the action (an effect). The last statement 
specifies the amount of energy consumed by the action.

The {\em Flexible Acting and Planning Environment} (\fape) is a recently introduced planning framework \cite{fape} which 
extends \htn\ planning with temporal reasoning by implementing the \anml\ language.
Another recent planner worth to be considered is \chimp\ \cite{chimp}. \chimp\ relies on its own modeling language and 
extends \htn\ planning domain representation with temporal representation by leveraging the functionalities of the 
{\em meta-csp} \cite{metacsp}.



%


\ifpdf
    \graphicspath{{chapters/3_timeline/figures/PNG/}{chapters/3_timeline/figures/PDF/}{chapters/3_timeline/figures/}}
\else
    \graphicspath{{chapters/3_timeline/figures/EPS/}{chapters/3_timeline/figures/}}
\fi


%
%
%
\chapter{Timeline-based Planning in a Nutshell}
\label{chap:timelines}
\lettrine[lines=2]{T}{he timeline-based approach} is a Temporal Planning paradigm introduced in early 90's \cite{hsts-94}, which takes 
inspiration from classical control theory. The main distinct factor is the centrality of {\em time} in the representation formalism. 
Unlike classical approaches, timeline-based planning puts {\em time} to the center of the solving approach by dealing with 
{\em concurrency}, {\em temporal constraints} and {\em flexible durations}. Timeline-based planning realizes a sort of {\em hybrid} 
representation and reasoning framework which allows a solver to "easily" interleave planning and scheduling decisions. This 
{\em hybrid view} of planning is one of the key characteristic for successfully addressing real-world problems. 
Timeline-based solvers have been successfully applied in real-world contexts (especially in space-like contexts) \cite{hsts-92}, 
\cite{Jonsson2000}, \cite{ICAPS07-mexar}.

The {\em world model} of a timeline-based application is characterized by a set of features that must be controlled over time in order 
to realize a complex behavior/task of a particular system to control. A complex system (\eg\ a planetary exploration
rover) is modeled by identifying a set of features that are relevant from the control perspective (\eg\ the stereo camera or the 
communication facility). Each feature is modeled in terms of the {\em values} it may assume over time and their related temporal 
durations. The temporal evolution of a feature is represented as a {\em timeline} which consists of an ordered sequence of 
{\em valued temporal intervals}, usually called {\em tokens}. 
These tokens describe the behavior of the feature within a given {\em temporal horizon}. In addition to the description of the 
features, the model may also specify {\em domain rules} that allow to further constrain the temporal behaviors of the features 
through temporal constraints. Such rules are necessary to achieve high-level goals (\eg\ take and communicate pictures of a 
target) by {\em coordinating} different features properly. 
For example, a rule may require that a particular value of a feature occurs {\em during} a known temporal interval or 
that a token of a timeline must always occur {\em before} a particular token of another timeline. 
Thus, a {\em timeline-based plan} consists of a set of {\em timelines} and that must satisfy all the temporal constraints of 
the domain in order to be {\em valid}.

In timeline-based planning, unlike classical planning, there is not a clear distinction between {\em states} and {\em actions}. A 
valued temporal interval may represent either an action or a state the related feature must perform or assume over a 
particular temporal interval. Similarly {\em planning goals} do not represent simply states or conditions that must be achieved. 
Rather, a planning goal may be either a value that a particular feature is supposed to assume during a certain temporal interval, 
or a complex task (\eg\ take a picture of a target) that must be performed within a given time.
The solving process acts on an initial set of partially specified timelines representing the initial known {\em facts} about the 
world. The process {\em completes} the behaviors of these timelines by iteratively adding values and temporal constraints 
according to desired requirements (including planning goals). 
Thus, timeline-based planners realize a {\em behavior-based} approach to planning, whose focus is on constraining the 
temporal evolutions of the system rather than synthesizing a sequence of actions that allow to achieve a desired goal state.

There are several timeline-based systems that have been introduced into literature and successfully applied to real-world
problems (especially in space-like contexts). \europa\ \cite{europa2012} developed by \nasa, \ixtet\ 
\cite{ghallab1994representation} developed at \laas, and \apsi\ \cite{apsigoac11} developed for \esa, represent some of 
the most known existing frameworks in this field. The next sections provide a brief description of the most relevant features of 
these timeline-based planning frameworks.

\section{\europa}
The \europa\  framework \cite{europa2012} relies on {\em Constraint-based Temporal Planning} (\cbtp) \cite{FrankJonsson-03} 
which is a Temporal Planning formalism successfully applied in many space application contexts  by \nasa.
The \cbtp\ modeling approach focuses on the temporal behaviour of the system we want to control and not just on the 
causality relationships. Therefore, a {\em complex system} (\eg\ a planetary exploration rover) is modelled by identifying a 
set of relevant components that can independently evolve over time.

A component models a physical or logical feature of the system to be controlled by specifying a (finite) set of mutually exclusive activities the related feature may assume over time. An {\em activity} is an atomic formula of the form:
$$A(x_{1},..., x_{n}, st_{A}, et_{A}, \delta)$$
where (i) $A$ is a predicate representing a particular condition of the world, (ii) $\vec{x} = \{x_{1}, ..., x_{n}\}$ are numerical or 
symbolic \emph{parameters} of activities, (iii) $st_{A}$ and $et_{A}$ are temporal variables representing respectively the activity 
start and end times and (iv) $\delta = [\delta_{min}, \delta_{max}]$ is an interval representing lower and upper bounds of 
{\em activity's duration}. 

Let $Act = \{A_1(\vec{x}_1, \delta_1), ..., A_k)(\vec{x}_k, \delta_k)\}$ be the set of activities. In \cbtp\ formalism a component 
$C_i$ is defined by a subset $Act_i = \{A_{i,1}, ..., A_{i,m}\}$ of $Act$, where activities $A_{i,j}$ represent possible states 
or actions of the component $C_i$. Components statically describe the possible temporal evolutions of the elements of the 
system. 
However, it is necessary to specify additional constraints in order to coordinate system's element and guarantee the 
overall system safeness. \cbtp\ considers two types of constraints:
(i) {\em Codesignation Constraints} can impose equalities or inequalities between the parameters of activities;
(ii) {\em Temporal Constraints} can model temporal constraints between activities by expressing either {\em interval-based} 
or {\em point-based} temporal predicates.

In general, \cbtp\ models temporal constraints by extending the {\em qualitative} temporal interval relationships defined in \cite{allen}
with {\em quantitative} information. Namely, the basic temporal relations between intervals are enriched with metric 
information, \ie\ lower and upper bounds of the distances between temporal intervals. 
For example, the relation {\em A before [10, 20] B} states that the interval {\em A} must precede interval {\em B} not less 
than 10 time units and not more than 20 time units. 

The causal and temporal constraints of the system are modeled by means of dedicated rules, called {\em compatibilities}, 
that specify interactions between a particular activity of a component and other activities that can either belong to the same 
component ({\em internal compatibility}) or to a different component ({\em external compatibility}).
Compatibilities describe how a particular activity (the {\em master}) is related to other activities (the {\em slave}) by 
specifying a set of {\em codesignation} and/or {\em temporal} constraints that must be satisfied in order to build {\em valid} 
plans. {\em Conditional compatibilities} can be defined by means of {\em guard constraints} that "extend" the conditions under 
which the related compatibility can be applied. If the guard constraints of a compatibility are satisfied, then the related temporal 
and/or codesignation constraints can be applied. Given a set of activities $Act$, the compatibility for an activity 
$A_{i}(\vec{x_{i}}, st{A_{i}}, et_{A_{i}}, \delta_{i}) \in Act$ is defined as
$$C[A_{i}]: G(\vec{y}) \rightarrow T(A_{i}, A_{j}, ..., A_{k}) \land P(\vec{x_{i}}, \vec{x_{j}}, ..., \vec{x_{k}}) $$
where (i) $G(\vec{y}) \equiv g_{1}(\gamma_{1}) \land ... \land g_{m}(\gamma_{m}))$ is a conjunction of guard constraints, (ii)
$T(A_{i}, A_{j}, ..., A_{k}) \equiv t_{1}(A_{i}, A_{j}) \land ... \land t_{m}(A_{i}, A_{k})$ is a conjunction of temporal constraints
involving the activities $A_{i}, A_{j}, ..., A_{k}$ and (iii) $P(\vec{x_{i}}, \vec{x_{j}}, ..., \vec{x_{k}}) \equiv p_{1}(\vec{x_{i}}, 
\vec{x_{j}}) \land ... \land p_{n}(\vec{x_{i}}, \vec{x_{k}})$ is a conjunction of \emph{codesignation} constraints on variable 
$\vec{x_{i}}$ and variables in $\cup_{t=j}^{k}(\vec{x_{t}})$.

If a compatibility $C[A]$ specifies different constraints according to the different values a particular {\em guard variable}  
$g_{i}(\gamma_{i})$ may assume then, $C[A]$ represents a {\em disjunctive} compatibility. 
Given an activity $A(\vec{x}, st, et, \delta)$, a {\em configuration rule} for $A$ is a conjunction of compatibilities and it is 
defined as
$$R[A(\vec{x}, st, et_{A}, \delta)] = C_{1}[A] \land ... \land C_{n}[A]$$
The code below shows some compatibilities and configuration rules for a classical planning problem concerning the 
control of a planetary exploration rover.

\begin{code}
R[Unstow()] = \{
   [meets Place(rock) \(\land\) met\_by Stowed()]
\}

R[Place(\(rock_b\))]   =   \{
   [meets Use(inst, \(rock_u\)) \(\land\) (\(rock_u = rock_b\))] \(\land\)
   [met\_by Unstow()] \(\land\)
   [contained\_by MobilitySystem.At(\(rock_a\)) \(\land\) (\(rock_a = rock_b\))]
\}

R[Use(inst, \(rock_b\))] = \{
   [\(\gamma\) = 0 \(\to\) meets Stow()
    \(\gamma\) = 1 \(\to\) meets Place(\(rock_p\) \(\land\) (\(rock_p \neq\ rock_b\))] \(\land\)
   [met\_by Place(\(rock_p\)) \(\land\) (\(rock_p = rock_b\))] \(\land\)
   [contained\_by MobilitySystem.At(\(rock_a\)) \(\land\) (\(rock_a = rock_b\))]
\}
\end{code}

Given the elements described above, a {\em planning domain} $D$ is defined by a set of components 
$C[D] = \{C_{1}, ..., C_{n}\}$, a set of activities $Act$ associated with each component and a set of {\em evolution rules} 
$R[D] = \{R[A_{1}], ..., R[A_{m}]]\}$, the domain contains an evolution rule $R[A_{i}]$ for each activity $A_{i} \in Act$.
The \cbtp\ planning  process aims at building a {\em valid} description of the {\em temporal behaviors} of the components within a 
temporal {\em horizon} where {\em goal} activities are scheduled at proper times. Thus, a {\em planning problem}
consists of a {\em planning horizon} and an {\em initial configuration} which (partially) describes the behaviors of the 
components. A solution plan is represented by a temporal execution trace which specifies for each time point, the activity 
the components are supposed to execute.

\section{\ixtet}
\ixtet\ \cite{ghallab1994representation} is a temporal planning system which tries to integrate plan generation and 
scheduling into the same planning process. Some of the most important features of the \ixtet\ planning paradigm are: 
(i) an explicit representation of time with different types of metric constraints between time points; 
(ii) a powerful representation of the world through multi-valued attributes; 
(iii) the management of a large range of resource types ({\em unsharable, sharable, consumable} and {\em producible}); 
(iv) a task formalism allowing for the representation of complex macro-operators.

Properties of the world are described by a set of {\em multi-valued state attributes} and a set of {\em resource 
attributes}. A state attribute describes a particular feature of the domain as a key-mapping from some finite 
domains into a finite range (the {\em value} of the attribute). The code below shows an example of a domain feature 
modeling the possible location of a robot.

\begin{code}
attribute position(?robot) \{
   ?robot \(\in\) \{robot1, robot2\};
   ?value \(\in\) \{RoomM, LabRoom1, LabRoom2\}; 
\}
\end{code}

A resource is defined as any substance, or set of objects whose cost or availability induces constraints on the actions that use 
them. So a resource can be either a single item with unit capacity (\ie\ an {\em unsharable} resource) or an aggregate resource 
that can be shared simultaneously between different actions without violating its maximal capacity constraint.

\begin{code}
resource robots(?robot) \{
   \(?robot \in\ \{robot1, robot2\}\);
   capacity = 1;
\}

resource paper_on_robot() \{
   capacity = 3;
\}
\end{code}

\ixtet\ defines different types of state attributes that can classified as: (i) {\em rigid attributes} (or atemporal) 
representing attributes whose value does not change over time (they express a structural relationship between their 
arguments); (ii) {\em flexible attributes} (or fluents) representing attributes whose value may change over time.
Flexible attributes may be further classified in: (i) {\em controllable attributes} representing attribute whose change 
of values can be planned for (but they can even change independently from the planning system); (ii) {\em contingent
attributes} representing attributes whose changes of values cannot be controlled.

Moreover, \ixtet\ relies on a reified logic formalism where fluents (\ie\ {\em flexible} attributes) are temporally qualified
by the {\em hold} and the {\em event} (temporal) predicates.\\
The {\em hold} predicate
$$hold(att(x_1, ...): v, (t_1, t_2))$$
asserts the (temporal) persistence of the value of state attribute $att(x_1, ...)$ to $v$ for each $t: t_1 \leq\ t < t_2$.\\
The {\em event} predicate
$$event(att(x_1, ...): (v_1, v_2), t)$$
asserts the {\em instantaneous} change of the value of {\em att($x_1$, ...)} from $v_1$ to $v_2$ occurred at time {\em t}.

Similarly, resource availability profile and the resource usage by the different operators are described by means of 
{\em use}, {\em consume} and {\em produce} predicates.\\
The {\em use} predicate 
$$use(typ(r): q, (t_1, t_2))$$
asserts the borrowing of an integer quantity $q$ of resource $typ(r)$ on the temporal interval $[t_1, t_2]$.\\
The {\em consume} predicate
$$consume(typ(r): q, t)$$
asserts that a quantity $q$ of resource $typ(r)$ is consumed at time $t$.\\
The {\em produce} predicate 
$$produce(typ(r): q, t)$$
asserts that a quantity $q$ of resource $typ(r)$ is produced at time $t$.

Temporal data representation and storage is managed by the {\em time-map manager} which relies on time-points as 
elementary primitives \cite{dechter1991}. Time is considered as a linearly ordered discrete set of instants. Time-points are 
seen as symbolic variables on which temporal constraints can be posted. \ixtet\ handles both {\em symbolic constraints} 
and {\em numeric constraints} expressed as a bounded interval $[I^{-}, I^{+}]$ on the temporal distance between time points.
The {\em time-map manager} is responsible for propagating constraints on time-points to check the {\em global} consistency
of the network and to answer queries about the relative position of time-points.

Planning operators are represented by means of a {\em hierarchy} of {\em task}s. A {\em task} is a temporal 
structure composed of: (i) a set of {\em sub-task}s; (ii) a set of events describing the changes of the world the 
task causes; (iii) a set of assertions on state attributes to express the required conditions or the protection of 
some fact between two task events; (iv) a set of resource usage; (v) a set of temporal and instantiation constraints
binding the different time-points and variables of the task.
{\em Tasks} are deterministic operators without ramification effects that may also refer to other {\em sub-tasks} in order to 
express macro-operators. The code below shows an example of {\em elementary task} (\ie\ a task without sub-tasks) for a 
robot in charge of the maintenance of a laboratory consisting in putting paper in a machine when it is out of paper:

\begin{code}
task feed_machine(?machine) (start, end) \{
   variable ?room;
   place(?machine, ?room);
   hold(position(robot): ?room, (start, end));
   event(machine_state: (out_of_paper, ok), end);
   consume(paper_on_robot(): 1, end);
   produce(trunk_size(): 1, end);
   (end - start) in [00:01:00, 00:02:00]; 
\}
\end{code}

The initial plan is a particular task that describes a problem scenario by specifying: (i) the initial values for the set of 
instantiated state attributes (as a set of explained events); (ii) the expected changes on some contingent state
attributes that will not be controlled by the planner (as a set of explained events); (ii) the expected availability
profile of the resources (as a set of uses); (iv) the goals that must be achieved (usually, as a set of assertions).

\section{\apsi}
\apsi\ \cite{apsigoac11} is a software framework developed for \esa\ whose aim is to support the design and development 
of \ps\ applications by leveraging the timeline-based approach. The \apsi\ framework provides the designer with a 
ready-to-use software library for modeling planning and scheduling concepts in the form of timelines.
Specifically, \apsi\ relies on the same modeling assumptions of \hsts\ \cite{hsts-94} and therefore, a complex system is modeled 
by identifying a set of relevant features to control over time. The \apsi\ framework makes available the modeling language and 
the software functionalities needed to model timeline-based domains in shape of {\em multi-valued state variables} and 
{\em synchronization rules}.

Multi-valued state variables model the features of the domain by describing their allowed temporal behaviors. State variables 
model domain features by specifying the values, the related feature may assume over time, together with the allowed durations 
and transitions. Thus, a state variable $x$ is defined as the tuple
$$x = \left(V, D, T\right)$$
where (i) $V$ is the set of values the variable $x$ can assume over time, (ii) $D: V \rightarrow\mathbb{R}\times\mathbb{R}$ is
a {\em duration function} specifying for each value $v \in\ V$ the minimum and maximum duration and (iii) $T: V \rightarrow\ 2^{V}$ 
is a {\em transition function} specifying for each value $v \in\ V$ the set of allowed {\em successors}. State variables specify 
causal and temporal constraints of the {\em single} features of a planning problem. 
They specify {\em local rules} that allow a planning system to build the timelines of the features composing  the domain.
Given a state variable $x$, a {\em timeline} describes the sequence of values the variable assumes over time by specifying a 
sequence of {\em valued temporal intervals} called {\em tokens}. A token is defined as the tuple
$$x_i = \left(v_j, [s_i, s'_i], [e_i, e'_i]\right)$$
where $[s_i, s'_i]$ and $[e_i, e'_i]$ represent respectively the flexible start and end of the temporal interval during which the 
variable $x$ is supposed to assume the value $v_j \in\ V$.

Synchronization rules model causal and temporal constraints of a planning domain by specifying {\em global} relations 
between tokens of different variables. In general, whenever a particular token $x_i$ occurs on a timeline (\ie\ the 
trigger) a synchronization rule specifies a set of different tokens (\ie\ the targets) that must occur on other timelines and a 
set of temporal constraints between the trigger and targets of the rule that must {\em hold} in order to build valid temporal 
behaviors. Indeed, synchronization rules allow the planning system to further constrain the temporal behaviors of the state 
variables in order to build timelines that satisfy some desired {\em planning goals}.
Temporal constraints of synchronization rules are modeled by extending the {\em qualitative} relationships of the 
Allen's interval algebra \cite{allen}, with {\em quantitative} information.

It is worth observing that \apsi, unlike other timeline-based frameworks (\eg\ the \europa\ and \ixtet\ frameworks 
mentioned above), is not a planner but a development library for designing planning applications. In this regard, 
\omps\ \cite{omps08acs} represents a {\em domain-dependent} timeline-based solver which has been developed on-top of 
the \apsi\ modeling functionalities and successfully applied in space-exploration scenario \cite{goac11}.

%
%
%
%
%
%
\section{Temporal Formalisms}
Temporal Planners rely on expressive temporal formalisms that allow these paradigms to deal with time and temporal 
constraints. Many timeline-based systems (including \europa, \ixtet\ and \apsi) model temporal information about plans by 
extending the Allen's interval algebra \cite{allen} in order to represent expressive temporal relations between the temporal 
elements of a plan.

Temporal information represents additional knowledge the planner must properly managed during the solving 
process. Thus, timeline-based planners must encapsulate temporal reasoning mechanisms that process 
temporal information in order to {\em verify} the (temporal) consistency of plans. Temporal reasoning mechanisms are 
usually implemented by leveraging the formalism of {\em Temporal Networks} \cite{dechter1991} which represents a 
flexible representation of temporal data as a network of time points (\ie\ the nodes of the network) and {\em distance 
constraints} between time points (\ie\ the edges of the network).

\subsection{The Simple Temporal Problem}
The {\em Simple Temporal Problem} (STP) is a well-known formalism introduced in \cite{dechter1991} which consists of a 
set of {\em events} that may occur over known temporal intervals and a set of {\em requirement constraints} that specify 
{\em distance constraints} on the temporal occurrences of pairs of events. The problem is to find a {\em temporal allocation} 
of the events satisfying all the requirement constraints (\ie\ the distance constraint). Namely, temporal reasoning mechanisms 
try to find an assignment of events to {\em time points} such that all the temporal constraints are satisfied. This concept is 
known as {\em temporal consistency} and is central to STPs.

Below is the description of a simple scenario taken from \cite{dechter1991}, representing an example of the type of problems 
and inference the STP formalism can support.
\begin{example}
\label{ex:stn}
John goes to work by car (30-40 minutes). Fred goes to work in a carpool (40-50 minutes). Today John left home between 
7:10 and 7:20, and Fred arrived at work between 8:00 and 8:10. We also know that John arrived at work about 10-20 minutes 
after Fred left home. We wish to answer queries such as: "Is the information in the story consistent?", "What are the 
possible times at which Fred left home?", and so on.
\end{example}
\begin{figure}
\centering
\includegraphics[width=.8\textwidth]{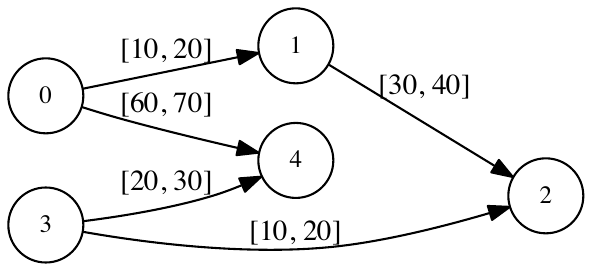}
\caption{\small{A graph representation of the STP problem described in Example \ref{ex:stn}}}
\label{fig:stn}
\end{figure}
Figure \ref{fig:stn} shows the STP problem of Example \ref{ex:stn} in graph form. When STPs are shown as graphs where nodes 
represent events and edges represent requirement constraints, they are called {\em Simple Temporal Networks} (STNs). 
The node "0" of Figure \ref{fig:stn} represents the {\em temporal origin} of the plan/problem, the "absolute" time 7:00 with 
respect to Example \ref{ex:stn}. Node "1" in Figure \ref{fig:stn} is associated with the event representing the time at which 
John leaves home. The edge between node "0" and node "1" labeled "[10, 20]" models the fact "John left home between 7:10 
and 7:20" as a distance constraint (\ie\ a requirement constraint)  between the two related events. 
A strong limitation of the STP formalism concerns disjunctive constraints. STP cannot represent and therefore, cannot reason 
about disjunctive temporal intervals on events. Considering Example \ref{ex:stn}, STP cannot model {\em disjunctive 
assertions} like {\em "John goes to work either by car (30-40 minutes), or by bus (at least 60 minutes)"}. 
Disjunctive assertions represent alternative plans the planning process may generate accordingly by branching the search space.

From a planning perspective, the temporal part of the plan can be reduced to a STP by modeling the start and end times of 
the activities of the plan (\eg\ the start and end times of the tokens of a timeline) as {\em events} of the STP. Temporal 
relations and/or duration constraints concerning the activities/actions of the plan can be easily translated in the STP as 
one or more requirement constraints involving events related to the start/end times of the activities of the plan.
Thus a planning system can leverage the STP formalism to post ordering constraints between activities during the solving 
process and check the temporal assignment of the activities of the plan. Namely, a planning system can check the 
(temporal) consistency of a plan by verifying the existence of a valid {\em schedule} of all the activities.

\subsection{The Simple Temporal Problem with Uncertainty}
STP makes the assumption that all the events of the plan are {\em controllable}. It means that the planning system can 
{\em decide} the temporal allocation (\ie\ the schedule) of all the events. However, this is not always possible in real-world 
settings. Indeed, the activities of a plan usually model real-world tasks/actions whose durations can be affected by exogenous 
factors and therefore, the planning system cannot decide the temporal allocation of these activities (\eg\ the planner can decide 
the start time of the execution of an action but not the end time). Such activities are called {\em uncontrollable}. Thus, a more 
expressive temporal formalism is the {\em Simple Temporal Problem with Uncertainty} (STPU). 

The STNU formalism takes into account both {\em controllable} and {\em uncontrollable} events. In this formalism an event 
is considered {\em uncontrollable} if it is the target of {\em contingent constraints} that are typically used to model 
uncontrollable durations of the activities/actions of the plan. The key point of STPUs is that {\em temporal consistency} 
is not sufficient to solve real-world problems. {\em Temporal uncertainty} introduces the additional problem of deciding how 
to schedule controllable events according to the observed/possible temporal occurrences of uncontrollable events, in order 
to complete the execution of the plan. Such a problem is called the {\em controllability problem} which has been fairly 
investigated in the literature \cite{VidalF99,morris01}. Broadly speaking, three different types of controllability ({\em weak}, 
{\em strong} and {\em dynamic controllability}) have been defined according to the different assumptions made on 
the uncontrollable events of a plan. Planning systems may leverage the STPU formalism to generate plans with some 
desired properties concerning the controllability of generated plans (\ie\ properties concerning the execution of the 
generated plans in the real-world).


%


\ifpdf
    \graphicspath{{chapters/4_formalization/figures/PNG/}{chapters/4_formalization/figures/PDF/}{chapters/4_formalization/figures/}}
\else
    \graphicspath{{chapters/4_formalization/figures/EPS/}{chapters/4_formalization/figures/}}
\fi


%
%
%
\chapter{Flexible Timeline-based Planning with Uncertainty}
\label{chap:formalization}
\lettrine[lines=2]{D}{espite} the practical success of timeline-based planning, formal frameworks characterizing this paradigm 
have been proposed only recently. There is a multitude of software frameworks that have been realized and introduced in the 
literature, each of which applies its own {\em interpretation} of timeline-based planning.
In such a context, it is not easy to evaluate the modeling  and solving capabilities of different timeline-based planning 
systems. It is not even easy to define {\em benchmarking domains} to compare timeline-based systems, or to {\em open} 
the assessment to other planning techniques.
 
This chapter describes a complete and comprehensive formal characterization of the 
timeline-based approach which has been introduced in \cite{cialdea2015planning}. The proposed formalization aims 
at defining a clear semantics of the main planning concepts by taking into account the features of the most 
known timeline-based planning frameworks.
In addition, the formalization takes into account {\em temporal uncertainty} which is particularly relevant in 
real-world domains where not everything is controllable. Indeed, the execution process is not completely 
under the control of the executive system. Exogenous events can affect or even prevent the complete and 
successful execution of generated plans. Thus, the capability of representing and dealing with {\em temporal 
uncertainty} and {\em controllability properties} at both planning and execution time is crucial to deploy effective 
timeline-based applications in real-world scenarios.

\section{A Running Example: The \rover\ Domain}
\label{sec:rover}
In order to support the formal definitions given below, a simple case study will be used as a running example. The 
domain takes inspiration from a typical scenario of \ai-based control for a single autonomous agent. 

The \rover\ domain consists of an exploration rover which can autonomously navigate 
a (partially) unknown environment, take samples of some {\em targets} (e.g. rocks) and 
communicate scientific data to a satellite. An exploration rover is a complex system endowed with 
several devices that must be properly controlled in order to achieve the desired objectives. 
A {\em navigation facility} allows the rover to move and {\em explore} the environment. 
A dedicated {\em instrument facility} allows the rover to take samples of targets that must be analyzed. 
A {\em communication facility} allows the rover to send data acquired from sampled targets to a satellite 
whose {\em orbit} is known.

A mission goal requires the rover to move towards a desired target, take a sample of it
and communicate gathered data when possible. All the features that compose the rover must be coordinated 
properly in order to realize the complex behavior needed to satisfy mission goals. 
Thus, a set of {\em operative constraints} must be satisfied. For instance, communication of data must be 
performed while the rover is still and during some known {\em communication windows} that represent 
temporal intervals during which the target satellite is {\em visible}. Another operative constraint requires 
that the instrument facility must be set in a {\em safe} position/configuration while the rover is moving.

\section{Domain Specification}
The timeline-based  approach to planning pursues the general idea that planning and scheduling  for 
controlling complex physical systems consist of the synthesis of desired temporal behaviors (or {\em timelines}). 
According to this paradigm, a domain is modeled as a set of features with an associated set of temporal functions 
on a finite set of values.
The time-varying features are usually called {\em multi-valued state variables} \cite{hsts-94}. Like in classical control 
theory, the evolution of the features is described by some causal laws and limited by domain constraints. 
These are modeled in a {\em domain specification}.

The task of a planner is to find a sequence of decisions that brings the timelines into a final desired set, 
satisfying the domain specification and special conditions called {\em goals}. Causal and temporal constraints 
specify which value transitions are allowed, the minimal and maximal duration of each valued interval and 
(so-called) {\em synchronization constraints} between different state variables.  
Moreover, a domain specification must take into account the {\em temporal uncertainty} of planning domains in 
order to model more realistic problems. In particular, two sources of uncertainty are considered. 

On the one hand, the evolution of some components of the domain may be completely outside the control of 
the system. What the planner and the executive know about them is only what is specified in the underlying planning 
problem. On the other hand, some events may be partially controllable. In this case, the planner and the executive can 
decide when to start an activity, but they cannot fix the duration of the activity. 
According to this characterization, two types of state variables constitute a planning domain: the {\em planned variables}
model the controllable or partially controllable features of a domain; the {\em external variables} model the uncontrollable 
features of a domain. Thus, the planning system or the executive must respectively make planning and execution 
decisions, without changing the behavior of external variables or making hypothesis on the actual duration of partially 
controllable features. 

For the sake of generality, temporal instants and durations are taken from an infinite set of non-negative numbers 
$\tp$, including $0$. For instance, $\tp$ can be the set of natural numbers $\Nat$ (in a discrete time framework), 
as well as the non-negative  real numbers $\reals$. Sometimes, $\infty$ is given as an upper bound to allowed 
numeric values, with the meaning that $t<\infty$ for every $t\in\tp$.
The notation $\tp^\infty$ will be used to denote $\tp\cup\{\infty\}$, $\tp_{> 0}= \tp -\{0\}$ and $\tp^\infty {> 0}
= \tp^\infty -\{0\}$. When dealing with temporal intervals, if $s,e\in\tp$, the (closed) interval $[s,e]$ denotes the set of 
time points $\{t\mid s\leq t\leq e\}$.

\subsection{State Variables}
A state variable $x$ is characterized by four components:  the set $V$ of  values the  variable may assume, 
a function $T$ mapping each value  $v \in V$ to the set of values that are allowed to follow $v$, a function 
$\gamma$  tagging each value with information about its controllability, and a function $D$ which may 
set upper and lower bounds on the duration of each variable value.
\begin{definition}
\label{variables}
{\bf [State Variable]}
A \defi{state variable}   $x$, where $x$ is a unique identifier, called the {\em variable name}, is a tuple 
$(V,T,\gamma,D)$, where:
\begin{enumerate}
\item $V$, also denoted by $\values(x)$, is a non-empty set, whose elements are the state variable {\em values}.
\item $T: V\rightarrow 2^V$ is a total function, called the state variable {\em value transition function}.
\item $\gamma:V\rightarrow \{c,u\}$ is a total function, called the {\em controllability tagging function}; 
$\gamma(v)$ is  the {\em controllability tag} of the value $v$. If $\gamma(v)=c$, then $v$ is a {\em controllable 
value}, and  if $\gamma(v)=u$, then $v$ is {\em uncontrollable}.
\item $D: V \rightarrow\ \tp\times \tp^\infty$ is a total function such that $D(v)=(d_{min},d_{max})$ for some 
$d_{min}\geq 0$ and $d_{max}\geq d_{min}$, and if $\gamma(v)=u$, then $d_{min}>0$ and $d_{max}\neq\ \infty$;
$D$ is called the state variable {\em duration function}.
\end{enumerate}
\end{definition}
If $\gamma(v)=c$, then the planning or executive system can control the value $v$ and can decide the actual 
duration of related activities (\eg\ the executive can decide when to start and end the execution of these activities).
If $\gamma(v)=u$, then the planning or executive system cannot control the value $v$ and cannot decide 
the actual duration of related activities. The behaviors of these activities are under the control of the environment.

The intuition behind the duration function is that if $D(v)=(d_{min},d_{max})$, then the duration of each interval in 
which $x$ has the value $v$  is included between $d_{min}$ and $d_{max}$ inclusive, if $d_{max}\in \tp$; it is 
not shorter than $d_{min}$ and has no upper bound, if $d_{max}=\infty$. 
In practice, existing systems, such as \europa\ \cite{europa2012} and \apsi\ \cite{apsigoac11}, 
allow values to be represented by means of parametrized expressions 
In the present theoretical approach, values are taken to be completely instantiated in order 
to simplify the presentation. This amounts to describing sets and functions  by enumeration and does not diminish 
the expressive power.

In what follows, it is assumed that, whenever a set of state variables $SV$ is considered, for every distinct pair $x$ 
and $y$ in $SV$, $x\neq y$. The set $SV$ is partitioned into two disjoints sets, $SV_P$, containing the {\em planned} 
state variables, and $SV_E$, the set of the {\em external} ones. Every value $v$ of an external state variable is 
uncontrollable, i.e., $\gamma(v)=u$. An external variable represents a component of the 
"external world" that is completely outside the system control: the planner cannot decide when to start or end its 
activities. What is known about an external variable is specified in the planning problem.
On the contrary, a planned state variable represents a component of the system that is under the control of the 
executive. Nevertheless, controllable sub-systems may also have uncontrollable activities (i.e., activities whose 
starting times can be decided by the executive, but their durations and consequently their ending times, are not 
controllable). In other terms, the planner can decide when to start an uncontrollable activity of a planned variable 
(\ie\ when the variable assumes an uncontrollable value),  even if it cannot precisely predict how long it will last.
In general, every time an activity (either controllable or not) is preceded by an uncontrollable one, the system 
cannot control its start time. Indeed, the start time of the activity is affected by the end of the previous activity,
which is uncontrollable.

\begin{figure}
\centering
\includegraphics[width=.9\textwidth]{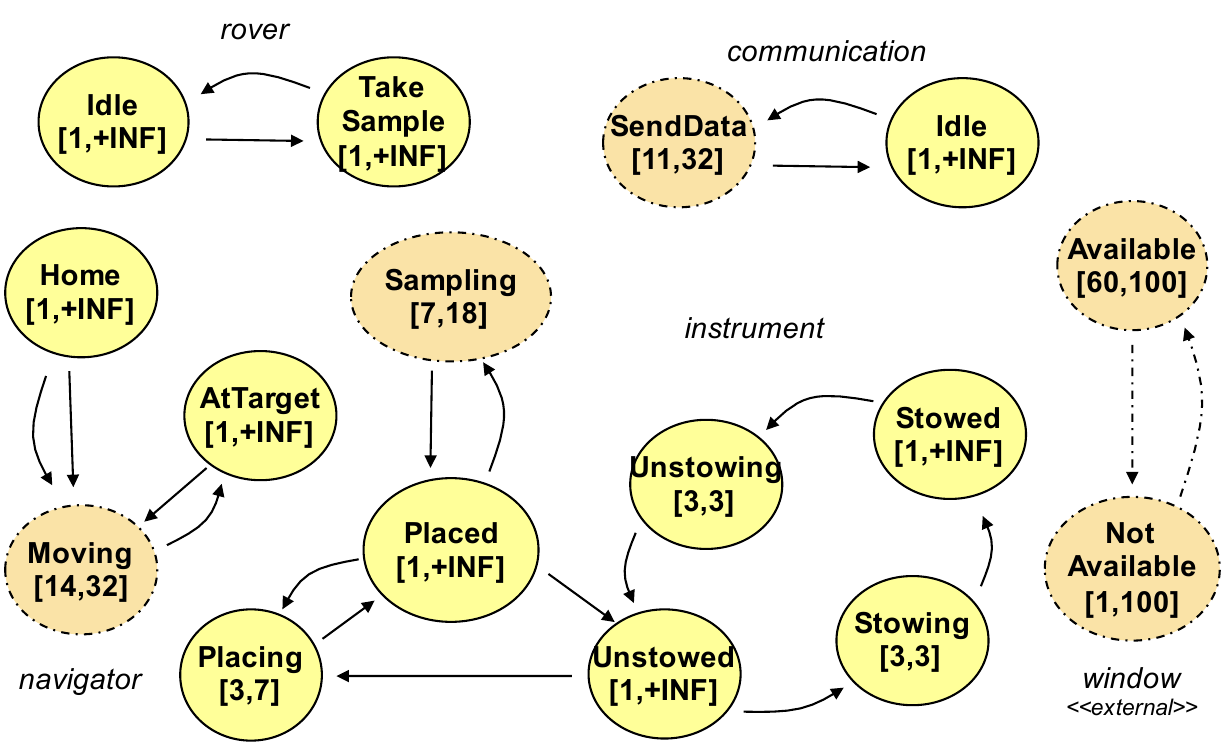}
\caption{\small{State Variable specification for the \rover\ planning domain}}
\label{fig:rover}
\end{figure}

\begin{example}
\label{ex:formal1}
In the considered running example, the timeline-based specification identifies five state variables, that 
will be called $r$ (for "rover"), $nv$ (for "navigator"), $inst$ (for "instrument"), $cm$ (for "communication") 
and $win$ (for "window") whose values, transitions and controllability properties are illustrated in 
Figure \ref{fig:rover} (values with dotted borders represent {\em uncontrollable} values).\\
Therefore, the set of considered state variables is $SV=\{r,nv,inst,cm,win\}$ where $SV_P=\{r\comma nv\comma 
inst\comma cm\}$ are planned state variables, while $SV_E=\{win\}$ is an external one.
For example, the state variable $inst$ models the instrument facility the rover uses to sample targets. The state 
variable can be defined by the typle $inst=(V_{inst}\comma T_{inst}\comma \gamma_{inst}\comma D_{inst})$ 
where:
\begin{itemize}
\item $V_{inst} = \{Stowed$, $Unstowed$, $Stowing$, $Unstowing$, $Placing$, $Placed$, $Sampling\}$;
\item $T_{inst}$ is the value transition function such that
	\begin{itemize}
	\item $T_{inst}(Stowed) = \{Unstowing\}$,
	\item $T_{inst}(Unstowed) = \{Stowing\comma Placing\}$,
	\item $T_{inst}(Stowing) = \{Stowed\}$,
	\item $T_{inst}(Unstowing) = \{Unstowed\}$,
	\item $T_{inst}(Placing) = \{Placed\}$,
	\item $T_{inst}(Placed) = \{Placing\comma Unstowed\comma Sampling\}$,
	\item $T_{Inst}(Sampling) = \{Placed\}$;
	\end{itemize}
\item $\gamma_{inst}$ is the controllability tagging function such that
	\begin{itemize}
	\item $\gamma_{inst}(Stowed) = \gamma_{inst}(Unstowed) = \gamma_{inst}(Stowing) =\\ 
		\gamma_{inst}(Unstowing) = \gamma_{inst}(Placing) = \gamma_{inst}(Placed) = c$,
	\item $\gamma_{inst}(Sampling) = u$;
	\end{itemize}
\item $D_{inst}$ is the value duration function such that
	\begin{itemize}
	\item $D_{inst}(Stowed) = D_{inst}(Unstowed) = D_{inst}(Placed) = (1, \infty)$,
	\item $D_{inst}(Stowing) = D_{inst}(Unstowing) = (3, 3)$,
	\item $D_{inst}(Placing) = (3, 7)$,
	\item $D_{inst}(Sampling) = (7, 18)$.
	\end{itemize}
\end{itemize}
The state variable $win$, instead, is an external variable which models the availability of the communication 
channel during the satellite orbit. It can be defined by the tuple $win=(V_{win}\comma T_{win}\comma 
\gamma_{win}\comma D_{win})$, where:

\begin{itemize}
\item $V_{win} = \{Available\comma NotAvailable\}$; 
\item $\gamma_{win}(Visible) = \gamma_{win}(NotAvailable) = u$;
\item $T_{win}$ is the value transition function such that
	\begin{itemize}
	\item  $T_{win}(Available) = \{NotAvailable\}$
	\item  $T_{win}(NotAvailable) = \{Available\}$;
	\end{itemize}
\item  $D_{win}$ is the duration function such that
	\begin{itemize}
	\item $D_{win}(Available) = (60, 100)$ 
	\item $D_{win}(NotAvailable) = (1, 100)$.
	\end{itemize}
\end{itemize}
\end{example}

\subsubsection*{(Flexible) Timelines}
A timeline represents the temporal evolution of a system component up to a given time. It consists of a
sequence of valued intervals, called \emph{tokens}, each of which represents a time slot in which
the variable assumes a given value. A token represents a temporal interval which determines the 
instant the variable starts executing the related value and the time instant the variable ends executing 
that value. 

However, planning with timelines takes into account {\em time flexibility} by allowing  token durations to 
range within given bounds. It means that the start and end time instants of a token are replaced by 
temporal intervals. Thus, the notion of (flexible) timeline can be defined as follows:

\begin{definition}
\label{ftimeline}
{\bf [Timeline]}
If $x=(V,T,\gamma,D)$ is a state variable, a \defi{token} for the variable $x$ has the form:
$$x^i=(v, [\einf,\esup],[\dinf,\dsup], \gamma(v))$$ 
where $x^i$, for $i\in\Nat$, is the token \defi{name}, $v\in V$, $\einf,\esup,\dinf,\dsup\in\tp$, $\einf\leq \esup$
and $d_{min}\leq \dinf \leq \dsup\leq d_{max}$, for $D(v)=(d_{min},d_{max})$.
The value $\gamma(v)$ is called the token controllability tag; if $\gamma(v)=c$, then the token  is \defi{controllable};  
if $\gamma(v)=u$, then the token is  \defi{uncontrollable}. 
A \defi{timeline} $\mathit{FTL}_x$ for the state variable $x=(V,T,\gamma,D)$ is a finite sequence of tokens 
for $x$, of the form:
\[\begin{array}{cc}
 x^1=(v_1,
[\einf_1,\esup_1],[\dinf_1,\dsup_1],\gamma(v_1)),\\
\dots,\\
x^k=(v_k,[\einf_k,\esup_k],[\dinf_k,\dsup_k],\gamma(v_k)),
\end{array}\]
where for all $i=1\dots k-1$, $v_{i+1}\in T(v_i)$ and $\esup_i\leq \einf_{i+1}$. The interval $[\einf_k,\esup_k]$ in the 
last token is called the \defi{horizon} of the timeline and the number $k$ of tokens making up $\mathit{FTL}_x$ is 
its \defi{length}.
If $x^i=(v, [\einf,\esup],[\dinf,\dsup],\gamma(v))$ is a token in the timeline $\mathit{FTL}_x$, then:
\begin{itemize}
\item $\val(x^i)=v$;
\item $\et(x^i) = [\einf,\esup]$;
\item $\st(x^0) = [0,0]$  and $\st(x^{i+1})=\et(x^i)$;
\item $\duration(x^i) = [\dinf,\dsup]$;
\item with an abuse of notation, $\gamma(x^i)$ denotes the token controllability tag $\gamma(\val(x^i))$.
\end{itemize}
\end{definition}

Intuitively, a token $x^i$ of the above form represents the set of valued intervals starting at some $s\in \st(x^i)$, 
ending at some $e\in\et(x^i)$ and whose durations are in the range $\duration(x^i)$.
The horizon of the timeline is the end time  of its last token.

\begin{example}
\label{ex:formal2}
Let us consider the timeline $\mathit{FTL}_{inst}$ for the state variable $inst$, in the \rover\ domain, made of the
following sequence of tokens: 
\[\begin{array}{l}
	inst^1= (Stowed,[20,28],[20,30],c)\\
	inst^2=(Unstowing,[23,31],[3,3],c)\\
	inst^3=(Unstowed,[50,55],[19,32],c)
\end{array}\]
The horizon of $\mathit{FTL}_{inst}$ is $[50,55]$.\\
An example of {\em non-flexible} timeline for the same state variable $inst$ is made of the 
following sequence of tokens:
\[\begin{array}{l}
	inst^1= (Stowed,[25,25],[20,30],c)\\
	inst^2=(Unstowing,[28,28],[3,3],c)\\
	inst^3=(Unstowed,[50,50],[19,32],c)
\end{array}\]
and its horizon is $[50,50]$. 
\end{example}

It is worth pointing out that often in the literature (e.g., \cite{omps08acs}), a flexible token contains also a 
{\em start} interval. However, once a token $x^i$ is embedded in a timeline, the time interval to which its 
start point belongs ($\st(x^{i})$) can be easily computed as shown in the definition above. Thus, including 
it as part of the token itself is redundant.

On the contrary, duration restrictions alone would be inadequate to precisely identify when the valued 
intervals represented by a given token must begin and end. As a matter of fact, duration and end time 
bounds interact when determining which legal values a token end time may assume.
Let us assume, for instance, that the duration of a given token $x^i$ is $[20,30]$ and that one may 
compute, from the durations of the previous tokens, that its start time  is $[40,50]$.
One can then infer that the end points of the valued intervals it represents are necessarily in the 
range $[60,80]=[40+20,50+30]$. However, it may be the case that a stricter end time is required, for instance
$[65,75]$. 
In this case, starting $x^i$ at $50$ and ending it at $80$, though respecting the duration bounds, would not 
be a legal value to "execute" the token, since $80\not\in [65,75]$. So, differently from the case of non-flexible 
timelines, durations alone are not sufficient to suitably represent tokens.
Analogously, end time bounds do not capture all the necessary information: the above described token 
$x^i$ does not represent a valued interval starting at $40$ and ending at $75$, even though it
respects the start and end time bounds, it violates the duration constraint. 
 
Controllability tags are part of token structures for a different reason. Although $\gamma(x^i)$ is equal to 
$\gamma(\val(x^i))$, such  information is included in the token $x^i$ 
with the aim of having a self-contained representation of flexible plans, encapsulating all the relevant execution 
information. This allows the executive system to handle  plans  with no need of considering  also the
description of the state variables.
When considering a set $\ftl$ of  timelines for the state variables in $SV$, it is always assumed
that it contains exactly one timeline for each element of  $SV$.

\subsubsection*{Schedules}
A {\em scheduled timeline} is a  particular case where each token has a singleton $[t,t]$ as its end 
time, \ie\ the end times are all fixed. A \defi{schedule} of a timeline $\mathit{FTL}_x$ is essentially 
obtained from $\mathit{FTL}_x$ by narrowing down to singletons (\ie\ time points) the token end times.

The schedule of a token corresponds to one of the valued intervals it represents ({i.e.}, it is obtained by 
choosing an exact end point in the allowed interval, without changing its duration bounds).
A scheduled timeline is a sequence of scheduled tokens satisfying the duration requirements.
Tokens, timelines and sets of timelines represent the set of their schedules. 

In general, $\mathit{STL}_x$ and $\tl$ will be used as meta-variables for scheduled timelines and sets of 
scheduled timelines, respectively, while $\mathit{FTL}_x$ and $\ftl$ as meta-variables for generic (flexible) 
timelines and sets of timelines.
In what follows, an interval of the form $[t,t]$, consisting of a single time point,  will be identified with the time 
point $t$ (and, with an abuse of notation, singleton intervals are allowed as operands of additions, subtractions, 
comparison operators, etc.).

\begin{definition}
{\bf [Scheduled]}
A \defi{scheduled} token is a token of the form $$x^i=(v,[t,t],[\dinf,\dsup],\gamma(v))$$ (or succinctly 
$x^i=(v,t,[\dinf,\dsup],\gamma(v))$).
A \defi{schedule} of a  token $x^i=(v\comma [\einf,\esup]\comma[\dinf,\dsup],\gamma(v))$ is a scheduled 
token  $x^i=(v,t,[\dinf,\dsup],\gamma(v))$, where  $\einf\leq t\leq \esup$.\\
A \defi{scheduled timeline} $\mathit{STL}_x$ is a timeline consisting only of scheduled tokens and such that if 
$k$ is the timeline length, then: for all $1\leq i\leq k$, if $\duration(x^i)=[\dinf_i,\dsup_i]$, then 
$\dinf_i\leq\et(x^i)-\st(x^i)\leq \dsup_i$.\\
A scheduled timeline $\mathit{STL}_x $ for the state variable $x$ is a schedule of $\mathit{FTL}_x $ if
$\mathit{STL}_x $ and $\mathit{FTL}_x $ have the same length $k$, and for all $i$, $1\leq i\leq k$, the 
token $x^i$ of  $\mathit{STL}_x $ is a schedule of the token $x^i$  of $\mathit{FTL}_x $.\\
Let $\ftl$ be a set of  timelines for the state variables in $SV$.  A schedule  $\tl$ of $\ftl$ is a set of
scheduled timelines for the state variables in $SV$, where each $\mathit{STL}_x\in \tl$ is a schedule of 
the timeline $\mathit{FTL}_x \in\ftl$.
\end{definition}

In simple terms, a scheduled timeline is a timeline where every end time is a singleton respecting the 
duration bounds. A schedule of a timeline  is a way of assigning values to each token end time, so that both 
duration and end time bounds are respected. Tokens, timelines, and sets of timelines represent the set of their 
respective schedules.

\begin{example}
\label{ex:formal3}
Let us consider the  flexible timeline $\mathit{FTL}_{inst}$ of example \ref{ex:formal2}:
\[\begin{array}{ll}
\mathit{FTL}_{inst}=&inst^1= (Stowed,[20,28],[20,30],c)\\
& inst^2=(Unstowing,[23,31],[3,3],c)\\
& inst^3=(Unstowed,[50,55],[19,32],c)\\
\end{array}\]
It is worth pointing out that, since the start time of the first token of a timeline is $[0,0]$, its end time bounds 
are usually equal to its duration bounds, but, like this example shows, it is not necessarily so.\\ 
When the two intervals differ, the end point of the corresponding first token in any schedule of the timeline 
belongs to their intersection. Each schedule of the timeline $\mathit{FTL}_{inst} $ represents a series of choices for
the token  end points, within the allowed intervals and respecting the allowed durations.\\
For instance, the following timeline is a schedule of $\mathit{FTL}_{inst}$:
\[\begin{array}{ll}
\mathit{STL}_{inst} = & inst^1= (Stowed,25,[20,30],c)\\
&inst^2=(Unstowing,28,[3,3],c)\\
&inst^3=(Unstowed,51,[19,32],c)\\
\end{array}\]
In fact, it satisfies all of  the endpoint and duration bounds in $\mathit{FTL}_{inst}$.\\
Clearly, not every sequence of scheduled tokens is a scheduled timeline.
For instance, the sequence of tokens obtained from $\mathit{STL}_{inst}$ 
by replacing the token $inst^2$ with $inst^2=(Unstowing, 31, [3,3], c)$
is not a scheduled timeline at all, since it does not satisfy the duration constraints for $inst^2$:
$\et(inst^{2}) - \st(inst^{2}) = 31 - 25 = 6 > 3$.\\
Let us now  consider the scheduled timeline $\mathit{STL}'_{inst}$ obtained from $\mathit{STL}_{inst}$ by replacing 
the token $inst^3$ with $inst^3=(Unstowed,60,[19,32],c)$. Although $\mathit{STL}'_{inst}$ satisfies all the duration 
bounds, it is not a schedule of $\mathit{FTL}_{inst}$, since the end time of $inst^3$ is not in the allowed interval 
$[50,55]$ of $inst^3$ in $\mathit{FTL}_{inst}$. 
\end{example}

\subsection{Restricting the Behavior of State Variables}
The behavior of state variables may be restricted by requiring that time intervals with given state variable values 
satisfy some temporal constraints. For instance, in the \rover\ sample domain, data can be communicated only 
when the communication channel is available. 
In other terms, for every token $cm^i$ in the timeline for the state variable $cm$ having the value $Communicating$, 
there must exist a token in the timeline for $win$, with the value $Available$ and bearing a given temporal relation 
with $inst^i$. 
This type of relations are expressed by means of {\em synchronization rules} that complete the definition of all the 
components of a domain specification.

\subsubsection*{Temporal Relations}
As a first step, the set of allowed temporal relations is introduced. They are  either relations between
two intervals or relations between an interval and a time point.
In particular, this work considers {\em quantitative temporal constraints}. For the sake of simplicity, a small 
set of primitive relations is chosen, all of which are parametrized by a (single) temporal interval.

\begin{definition}
\label{def:trel}
{\bf [Temporal Relation]}
 A \defi{temporal relation  between intervals} is an expression of the form $A$ $\rel_{[lb,ub]}$ $B$, where 
 $A$ $=$ $[s_A,e_A]$ and $B$ $=$ $[s_B,e_B]$ are time intervals, with $s_A\comma e_A\comma s_B\comma e_B\in\tp$,
$\rel_{[lb,ub]}\in\Rel$ $=$ $\{\sbs\comma \ebe\comma \sbe\comma \ebs\}$,  $lb\in\tp$ and $ub\in\tp^\infty$. 
The following table defines when a relation $A$ $\rel_{[lb,ub]}$ $B$ holds:

\begin{center}
	\begin{tabular}{r | l}
	\hline
	{\em the relation} & {\em holds if}\\
	\hline
	$A\sbs_{[lb, ub]}B$ & $lb\leq s_B - s_A\leq ub$\\[.2cm]
	$A\ebe_{[lb, ub]}B$ & $lb\leq e_B - e_A\leq ub$\\[.2cm]
	$A\sbe_{[lb,ub]}B$ & $lb\leq e_B - s_A\leq ub$\\[.2cm]
	$A\ebs_{[lb,ub]}B$ & $lb\leq s_B - e_A\leq ub$\\
	\hline
	\end{tabular}
\end{center}
\end{definition}

Other relations can be defined in terms of the primitive relations in Definition \ref{def:trel} (and their converses). 
These relations, like those used by \europa\ \cite{europa2012} and \apsi\ \cite{apsigoac11}, correspond to the quantitative 
extension of Allen's temporal relations \cite{allen}. Thus, the relations in the left most column of Table \ref{table:deftemprel}, 
are meant as abbreviations of the corresponding expressions on their right.

\begin{table}
\centering
	\begin{tabular}{r | l}
	\hline
	{\em the relation} & {\em is defined as}\\ 
	\hline
	$A \meets\ B$ &  $A \ebs_{[0,0]} B$\\[.2cm]
	$A \before_{[lb,ub]} B$ & $A \ebs_{[lb,ub]} B$\\[.2cm]
	$A \overlaps_{[lb_1, ub_1][lb_2, ub_2]} B$ & $A \sbs_{[lb_1 ,ub_1]} B\land$\\
									& $A \ebe_{[lb_2, ub_2]} B\land$\\
									& $B \sbe_{[0,\infty]} A$\\[.2cm]
	$A \equals\ B$ & $A \sbs_{[0, 0]} B\land$\\ 
				& $A \ebe_{[0, 0]} B$\\[.2cm]
	$A \contains_{[lb_1, ub_1][lb_2, ub_2]} B$ & $A \sbs_{[lb_1, ub_1]} B\land$\\
									& $B\ebe_{[lb_2, ub_2]} A$\\[.2cm]
	$A \starts_{[lb, ub]} B$ & $A \sbs_{[0, 0]} B\land$\\
					& $A \ebe_{[lb, ub]} B$\\[.2cm]
	$A \finishes_{[lb, ub]} B$ & $A \sbs_{[lb, ub]} B\land$\\
						& $A \ebe_{[0, 0]} B$\\[.2cm]
	$A \startat\ t$ & $A \startbefore_{[0, 0]} t$\\[.1cm]
	$A \xendat\ t$ & $A \xendbefore_{[0, 0]} t$\\
	\hline
	\end{tabular}
\caption{Defined temporal relations}
\label{table:deftemprel}
\end{table}

Once relations on time intervals are defined, they can be transposed to relations on tokens. The expressions 
used to denote such relations refer to tokens by means of their names.

 \begin{definition}
 \label{def:rels2token}
{\bf [Token Relation]}
Let $x^i$ and $y^j$ be names of tokens belonging to scheduled timelines for the state variables $x$ and $y$, 
respectively, with $\st\left(x^i\right)$ $=$ $s_i$, $\et\left(x^i\right)$ $=$ $e_i$, $\st\left(y^j\right)$ $=$ $s_j$, 
$\et\left(y^j\right)$ $=$ $e_j$.
Moreover, let $t,lb\in\tp$ and $ub\in\tp^\infty$. Expressions of the form $x^i\,\rel_{[lb,ub]}y^j$, for $\rel\in\Rel$,
and $x^i\,\rel_{[lb,ub]}t$, for $\rel\in\tpRel$, are called \defi{relations on tokens}.\\
The relation $x^i$ $\rel_{[lb,ub]}$ $y^j$ holds iff $[s_i,e_i]$ $\rel_{[lb,ub]}$ $[s_j,e_j]$ holds and the relation $x^i$ 
$\rel_{[lb,ub]}$ $t$ holds iff $[s_i,e_i]$ $\rel_{[lb,ub]}$ $t$ holds.
When a relation on two tokens $x^i$ and $y^j$ holds, we also say that the tokens $x^i$ and $y^j$ satisfy the relation, 
and that any set of scheduled timelines that contain $x^i$ and $y^j$ satisfies the relation.
Analogously, if a relation $x^i\,\rel_{[lb,ub]}t$ holds, then the token $x^i$ and any set of scheduled timelines containing 
$x^i$ satisfy the relation.
\end{definition}

\begin{example}
\label{ex:formal4}
Let $\tl=\{\mathit{STL}_{cm},\mathit{STL}_{win}\}$ be a set of timelines for the \rover\ domain, including 
$\mathit{STL}_{r}$, $\mathit{STL}_{inst}$ and $\mathit{STL}_{nav}$, where $\mathit{STL}_{cm}$ contains 
the tokens
\[\begin{array}{l}
	cm^5=(Idle, 100, [1, 43],c)\\
	cm^6=(SendData, 123, [11, 32],u)
\end{array}\]
and $\mathit{STL}_{win}$ contains the tokens
\[\begin{array}{l}
win^1=(NotAvailable, 60, [60, 80],u)\\
win^2=(Available, 130, [50, 90],u)
\end{array}\]
The expressions  $win^2\sbs_{[5,\infty]}cm^6$ and $cm^6\xendbefore_{[30,45]}165$ are relations on tokens and they are 
satisfied by $\tl$.
\end{example}

\subsubsection*{Synchronization Rules}

A synchronization constraint can be informally considered as a statement of the form "for every token \dots there exist 
tokens such that \dots". Namely, it represents a kind of quantified sentence. The formal counterpart of this kind of
assertions makes use of variables: for every $var_0$ \dots there exist $var_1,\dots var_n$ such that \dots. 
The variables used to express synchronizations are called \defi{token variables}. They are taken from 
a (potentially infinite) set $X=\{a_0,a_1,\dots\}$ of names, whose elements are all different from variable names, 
values and numbers. These variables are intended to range over tokens in the considered set of timelines.
 
Making a step forward, it can be observed that synchronization assertions actually use a form of bounded quantification: 
"for all/exist tokens {\em with value $v$} in the timeline for the state variable $x$ \dots". 
Such token variables with  restricted range will be denoted by expressions of the form $a_i[x=v]$, where $a_i$ is a token 
variable, $x$ is a state variable name, and $v\in \values(x)$. Such expression are called \defi{annotated} \defi{token variables}.
The next definition introduces the form of the assertions that can be used to express parametrized relations on tokens.
\begin{definition}
\label{def:pbf}
{\bf [Existential Statement]}
An \defi{atom} is  either the special constant $\top$ or an expression of the form 
$a_i$ $\rel_{[lb,ub]}$ $a_j$ or $a_i$ $\rel'_{[lb,ub]}$ $t$, where $a_i$ and $a_j$ are token variables, $lb, t\in\tp$, 
$ub\in\tp^\infty$, $\rel\in\Rel$, and $\rel'\in\tpRel$.\\
An \defi{existential statement} is an expression of the form
\[\exists\,
a_1 [x_1=v_1] \dots\ a_n [x_n=v_n]\,.\, \C
\]
where
\begin{itemize}
\item[(i)] $a_1,\dots,a_n$ are distinct token variables; 
\item[(ii)] for all $i$ $=$ $1,\dots,n$, $x_i$ is a state variable and $v_i\in\values(x_i)$ (i.e., $a_i [x_i=v_i]$ is an 
annotated token variable);
\item[(iii)] $\C$ is a conjunction of atoms.
\end{itemize}
The \defi{bound variables} of the statement are $a_1,\dots,a_n$ and any variable different from $a_1,\dots,a_n$ possibly 
occurring in $\C$ is said to occur {\em free} in the statement.
\end{definition}

Disjunctions of existential statements constitute the body of synchronization rules.

\begin{definition}
\label{def:sync}
{\bf [Synchronization Rule]}
A \defi{synchronization rule} is an expression of the form
\[a_0[x_0=v_0]\rightarrow {\cal E}_1\vee \dots\vee {\cal E}_k\]
(for $k\geq 1$) where every ${\cal E}_i$ is an existential statement whose bound variables are all different from 
$a_0$ and where only the token variable $a_0$ may occur free.
The left-hand part of the synchronization rule, $a_0[x_0=v_0]$, is called the \defi{trigger} of the rule.\\
A synchronization rule with empty trigger is an expression of the form:
\[\top\rightarrow {\cal E}_1\vee \dots\vee{\cal E}_k\]
(for $k\geq 1$) where every ${\cal E}_i$ is an existential statement with no free variables.
\end{definition}

Intuitively, a synchronization rule with non-empty trigger of the above form requires that, whenever the state variable 
$x_0$ assumes the value $v_0$ in some interval $a_0$, there is at least an existential statement ${\cal E}_i=\exists\,
a_1[x_1=v_1]\dots a_n[x_n=v_n]\,.\, \C$ and tokens $a_i$ ($1\leq i\leq n$) where the variable $x_i$ has the value 
$v_i$, such that $\C$ holds (if $\C=\top$, no temporal relation is required to hold). 
When the trigger is empty, the existence of the intervals $a_i$ and the relations among them have
to hold unconditionally. Synchronization rules with empty triggers are useful to represent \emph{domain invariants}, 
as well as planning goals (both called "facts" in \cite{cimatti2013timelines}). 
The use of token variables (which are absent in \cite{cimatti2013timelines}) allows one to refer to different intervals having 
the same value. Indeed, although the token variables $a_0,\dots,a_n$ are pairwise distinct, multiple occurrences of state 
variable names and values are allowed.

\begin{example}
\label{ex:formal5}
Consider the operational constraint of the \rover\ domain concerning the data communication activity: the rover can 
send data only when the communication channel is available and when it is not moving. 
A synchronization rule expressing this operational constraint is the following:
\begin{center}
\begin{tabular}{l l}
$a_{0}[cm=SendData] \rightarrow$ & $\exists a_{1}[win=Available]\, a_{2}[nav=At].$\\
							& $a_{1} \contains_{[0,\infty] [0,\infty]} a_{0} \land a_{2} \contains_{[0,\infty] [0,\infty]} a_{0}$\\[.1cm]
							&$\vee\, \exists a_{1}[win=Available]\, a_{2}[nav=Home].$\\
							& $a_{1} \contains_{[0,\infty] [0,\infty]} a_{0} \land a_{2} \contains_{[0,\infty] [0,\infty]} a_{0}$
\end{tabular}
\end{center}
According to this rule, whenever the state variable $cm$ assumes the value \emph{SendData} in an interval $a_{0}$, 
the state variable $win$ has the value \emph{Available} in some interval $a_{1}$ containing $a_0$, and the state 
variable $nav$ must have the value {\em At} in some interval $a_2$ containing $a_0$ or the value {\em Home} in some 
interval $a_2$ containing $a_0$. Namely the rover cannot move during communication tasks and communications can 
be performed only if the communication channel is available.
Synchronization rules with empty triggers may be useful to state known facts, such as, for instance:
\begin{center}
\begin{tabular}{l l}
$\top \rightarrow$ & $\exists a_1[nav = Home]. a_1\startat 0$
\end{tabular}
\end{center}
This rule represents the fact that the rover is at the "home" location at the beginning of the mission.
Synchronization rules with empty triggers are also used to represent planning goals, as will be described later on.
\end{example}

The following definition introduces the semantics of synchronization on scheduled timelines. Since  the statement of a 
synchronization rule makes use of token variables, each of them must be "interpreted", {i.e.}, mapped to a token of the 
considered timelines.

\begin{definition}
\label{def:semanticsync}
{\bf [Satisfiability of Synchronization Rules]}
Let $\ftl$ be a set of  timelines for the state variables $SV$. 
A {\em token assignment} for a set of annotated token variables $\{a_1 [x_1=v_1],\dots,a_n [x_n=v_n]\}$ on
$\ftl$ is a function $\varphi$ mapping every $a_i$ to a token of the timeline $\mathit{FTL}_{x_i}\in \ftl$ and such 
that $\val\left(\varphi\left(a_i\right)\right)$ $=$ $v_i$ for all $i=1,\dots,n$.\\
Let $\C=A_1\wedge\dots\wedge\ A_m$ be a conjunction of atoms and $\tl$ a set of scheduled timelines, including a 
timeline for every state variable occurring in $\C$. A token assignment $\varphi$ on $\tl$ satisfies $\C$ if for every  atom 
$A\in \{A_1,\dots, A_m\}$,
\begin{itemize}
\item[(i)] if $A$ $=$ $a_i$ $\rel_{[lb, ub]}$ $a_j$ then the relation $\varphi\left(a_i\right)$ $\rel_{[lb, ub]}$ 
$\varphi\left(a_j\right)$ holds;
\item[(ii)] if $A$ $=$ $a_i$ $\rel_{[lb, ub]}$ $t$, then the relation $\varphi\left(a_i\right)$ $\rel_{[lb, ub]}$ $t$ holds.
\end{itemize}
A token assignment $\varphi$ on $\tl$ satisfies an existential statement of the form 
$$\exists\ a_1 [x_1=v_1]\dots a_n[x_n=v_n]\,. \,\C$$ if $\varphi$ is a token assignment for a set of annotated variables 
including $a_0 [x_0=v_0],\dots, a_n[x_n=v_n]$ and $\varphi$ satisfies $\C$.
Let $$S=a_0[x_0=v_0]\rightarrow {\cal E}_1\vee \dots\vee {\cal E}_k$$ be a synchronization rule. A set of scheduled timelines 
$\tl$ for the state variables $SV$ {\em satisfies the   synchronization rule} $S$ if for every token 
$x_0^{k}$ in $\mathit{STL}_{x_0}\in\tl$ such that $\val(x_0^k)=v_0$, there exists a token assignment $\varphi$ on $\tl$ such 
that $\varphi(a_0)=x_0^{k}$ and $\varphi$ satisfies ${\cal E}_i$ for some $i\in\{1,\dots, k\}$.\\
A set of timelines $\tl$ for the state variables $SV$  satisfies a synchronization rule with empty trigger 
$\top\rightarrow {\cal E}_1\vee \dots\vee {\cal E}_k$ if, for some $i\in\{1,\dots, k\}$, there exists a token assignment 
$\varphi$ on $\tl$ satisfying ${\cal E}_i$.
Let $SV$ be a set of state variables, $\S$  be a set of synchronization rules concerning variables in $SV$ and $\tl$ 
be a set of scheduled  timelines for the state variables in $SV$.
$\tl$ \defi{satisfies the set of synchronizations  $\S$} iff $\tl$ satisfies all the elements of  $\S$.
\end{definition}

\begin{example}
\label{ex:formal6}
Consider the synchronization rule given in Example \ref{ex:formal5}, that constrains the rover to send data only when the 
communication channel is available, and to be still during communication:
\begin{center}
\begin{tabular}{l l}
$a_{0}[cm=SendData] \rightarrow$ & $\exists a_{1}[win=Available]\, a_{2}[nav=At].$\\
							& $a_{1} \contains_{[0,\infty] [0,\infty]} a_{0} \land a_{2} \contains_{[0,\infty] [0,\infty]} a_{0}$\\[.1cm]
							&$\vee\, \exists a_{1}[win=Available]\, a_{2}[nav=Home].$\\
							& $a_{1} \contains_{[0,\infty] [0,\infty]} a_{0} \land a_{2} \contains_{[0,\infty] [0,\infty]} a_{0}$
\end{tabular}
\end{center}
Let $\mathit{STL}_{cm}$ and $\mathit{STL}_{win}$ be two scheduled timelines.\\
Assume that the timeline for the pointing system contains a single token whose value is $SendData$, and has 
the form:
\[\begin{array}{ll}
\mathit{STL}_{cm} =  & \dots\,\\ 
& cm^{i-1} = (Idle, 110, [50, 80], c),\\
& cm^{i} = (SendData, 130, [11, 32], u)
\end{array}\]
In addition, assume that the timeline $\mathit{STL}_{win}$ (for the availability of the communication channel) and 
the timeline $\mathit{STL}_{nav}$ (for the navigation system of the rover) have the forms:
\[\begin{array}{ll}
\mathit{STL}_{win} = & \dots\, \\
& win^{j-1} = (NotAvailable, 80, [1, 100], u),\\
& win^{j} = (Available, 170, [60, 100], u)\\[.2cm]
\mathit{STL}_{nav} = & \dots\, \\
& nav^{k-1} = (Moving, 95, [14, 32], u),\\
& nav^{k} = (At, 185, [34, 95], c)
\end{array}\]
The synchronization rule is satisfied by the scheduled timelines $\{\mathit{STL}_{cm}$, $\mathit{STL}_{win}$, 
$\mathit{STL}_{nav}\}$. In fact, $cm^{i}$ is the only token in $\mathit{STL}_{cm}$ whose value is $SendData$, and the
token assignment $\varphi$, such that $\varphi(a_0)=cm^i$ and $\varphi(a_1)=win^j$, satisfies the existential statement
$\exists\, a_{1}[win=Available]\,. ~a_1\,\contains_{[0,\infty] [0,\infty]}a_0$: 
$\val(\varphi(a_1))=Available$ and $\varphi$ satisfies $a_1\,\contains_{[0,\infty] [0,\infty]}a_0$.
The latter assertion  holds because $\varphi(a_1) \contains_{[0,\infty] [0,\infty]} \varphi(a_0)$ -- {i.e.},
$win^{j}~\contains_{[0,\infty] [0,\infty]}~ cm^{i}$ -- ~holds, since  $[80, 170] ~\contains_{[0,\infty] [0,\infty]} ~[110, 130]$ holds.
\end{example}

\subsection{Planning Domains}
A planning domain is described by specifying a set of state variables and a set of synchronization rules. The formal 
definition of planning domains is given next, together with the notion of a set of scheduled timelines respecting the 
requirements of the domain.

\begin{definition}
\label{def:domain}
{\bf [Planning Domain]} A \defi{planning domain} is a triple $(SV_P,SV_E,\S)$, where:
\begin{itemize}
\item $SV_P$ is a set of planned state variables;
\item $SV_E$ is a set of external state variables (with $SV_P\cap SV_E= \emptyset$); 
\item $\S$ is a set of synchronization rules involving state variables in $SV_P\cup SV_E$.
\end{itemize}
A set of scheduled timelines $\tl$ for the state variables in $SV$ is \defi{valid} with respect to the planning domain 
$\D=(SV_P,SV_E,\S)$ if $SV=SV_P\cup SV_E$ and  $\tl$ satisfies the set of synchronizations $\S$.
\end{definition}

\begin{example}
\label{ex:formal7}
Let us consider the planning domain $\D=(SV_P,SV_E,\S\}$ where $SV_P=\{r,inst,nav,cm\}$, $SV_E=\{win\}$ -- 
for the state variables described in Example \ref{ex:formal1} -- and $\S$ contains the following synchronization
rules, modeling the operational constraints described in Section \ref{sec:rover}:
\begin{center}
\begin{tabular}{r l}
$a_{0}[cm=SendData] \rightarrow$ & $\exists a_{1}[win=Available]\, a_{2}[nav=At].$\\
							& $a_{1} \contains_{[0,\infty] [0,\infty]} a_{0} \land a_{2} \contains_{[0,\infty] [0,\infty]} a_{0}$\\[.1cm]
							&$\vee\, \exists a_{1}[win=Available]\, a_{2}[nav=Home].$\\
							& $a_{1} \contains_{[0,\infty] [0,\infty]} a_{0} \land a_{2} \contains_{[0,\infty] [0,\infty]} a_{0}$\\[.2cm]
$a_{0}[nav=Moving]\rightarrow$ & $\exists\, a_{1}[inst=Stowed]. a_{1} \contains_{[0,\infty] [0,\infty]} a_{0}$\\[.2cm]
$a_{0}[r=TakeSample]\rightarrow$ & $\exists\, a_{1}[inst=Sampling]\, a_{2}[nav=At].$\\ 
							& $a_{0} \contains_{[0,\infty] [0,\infty]} a_{1} \land\, a_{2} \contains_{[0,\infty] [0,\infty]} a_{1}$
\end{tabular}
\end{center}
Let moreover $\tl$ be the set of timelines containing
\[\begin{array}{l l}
\mathit{STL}_{r} = & r^1 = (Idle, 23, [11, 200], c),\\
& r^2 = (TakeSample, 55, [1, 70], c),\\
& r^3 = (Idle, 200, [23, 178], c)
\end{array}\]

\[\begin{array}{l l}
\mathit{STL}_{inst} = & inst^1 = (Stowed, 28, [9, 45],  c),\\ 
& inst^2 = (Unstowing, 31, [3, 3], c),\\ 
& inst^3 = (Unstowed, 32, [1, 30], c),\\
& inst^4 = (Placing, 35, [3, 7], c),\\
& inst^5 = (Sampling, 42, [7, 18], u),\\
& inst^6 = (Unstowed, 200, [1, 200], c)
\end{array}\]

\[\begin{array}{l l}
\mathit{STL}_{nav} = & nav^1 = (Home, 5, [5, 5], c),\\
& nav^2 = (Moving, 27, [19, 37], u),\\ 
& nav^3 = (At, 200, [1, 200], c)
\end{array}\]

\[\begin{array}{l l}
\mathit{STL}_{cm} =  & cm^1= (Idle, 65, [50,80], c),\\
 & cm^2 = (SendData, 83, [11, 32], u),\\
 & cm^3 = (Idle, 200, [1, 200], c)
\end{array}\]

\[\begin{array}{l l}
\mathit{STL}_{win} = & win^1 = (NotAvailable, 54, [23, 88], u),\\
 &  win^2 = (Available, 142, [60, 100], u),\\
 &  win^3 = (NotAvailable, 200, [1, 100], u)
\end{array}\]
The set $\tl$ is valid with respect to the planning domain $\D$: it contains exactly one timeline for each state variable 
in $SV_P\cup SV_E$ and it satisfies all the synchronization rules of the domain.
\end{example}

\section{Flexible Plans}
The main component of a flexible plan is a set $\ftl$ of timelines, representing different sets $\tl_i$ of scheduled timelines.
It may be the case that not every $\tl_i$ satisfies the synchronization rules of the domain. We aim at defining plans so that they
contain all the information needed to execute them, without having to check how the behavior of state variables and timelines is
constrained by the planning domain.\footnote{For the same reason controllability tags are included in token descriptions.} 
Consequently, a flexible plan $\Pi$ must be equipped with additional information in order to guarantee that every set of 
scheduled timelines is valid with respect to the domain specification. Such information represent temporal relations that 
have to hold in order to satisfy the synchronization rules of the domain.

As a schematic example showing why a set of  timelines does not convey enough information to represent a flexible plan, let 
us consider a domain with a synchronization rule $S$ of the form 
$a_0[x=v]\rightarrow \exists a_1[y=v']. \,a_0\,\meets\, a_1$
and  timelines for the state variables $x$ and $y$ containing, respectively, the tokens 
$x^i=(v,[30,50],[20,30],\gamma(v))$ and $y^j$, with $\val(y^j)=v'$ and $\st(y^j)=[30,50]$.
Not every pair of schedules of $x^i$ and $y^j$ satisfies $S$.
Thus, the representation of a flexible plan must also include  information about the relations that must hold between tokens in 
order to satisfy the synchronization rules of the planning domain. In the example above, it would include the relation 
$x^i\,\meets\, y^j$.  In general, a flexible plan includes a set of relations on tokens.  

When there are different ways to satisfy a synchronization rule by the same set $\ftl$ of flexible timelines, there are also 
different (valid) flexible plans with the same set of timelines $\ftl$. 
Thus, a flexible plan represents the set of its instances. 

\begin{definition}
\label{def:flexplan}
{\bf [Flexible Plan]}
A \defi{flexible plan} $\Pi$ is a pair $(\ftl,\R)$, where $\ftl$ is a set of  timelines and $\R$ is a set of relations on tokens 
involving token names in some timelines in $\ftl$.
An \defi{instance} of the flexible plan $\Pi=(\ftl,\R)$ is any schedule of $\ftl$ that  satisfies every relation in $\R$.
\end{definition}

In order to determine when a plan is valid with respect to a planning domain, the semantics of synchronizations  on flexible 
plans must be defined.
Essentially, a plan $\Pi=(\ftl,\R)$ satisfies a synchronization rule $S$ if the constraints represented by $S$ are guaranteed 
to hold for any schedule of $\ftl$ satisfying the relations in $\R$.
In other terms, $\R$ represents a possible way to satisfy $S$. The intuition underlying the formal definition can be explained 
as follows.  

When considering a plan $\Pi=(\ftl,\R)$, a mapping is used to assign the annotated token variables occurring in the 
synchronization to token names occurring in $\ftl$. Let us consider, for instance a rule of the form 
$$a_0 [x=v]\rightarrow \exists a_1[y=v']. a_1\ebs_{[10,20]}a_0.$$
Let us moreover assume that $x^3$ is a token in the timeline for $x$ in $\ftl$ with $\val(x^3)=v$, and that the timeline 
for $y$ in $\ftl$ contains exactly two tokens $y^5$ and $y^8$ having value $v'$. In order for the rule to be satisfied, 
$\ftl$ must be constrained  by requiring that $x^3$ starts from 10 to 20 time units after the end of  either $y^5$ or
$y^8$. The plan $\Pi$ commits to one of the two alternatives: binding $a_1$ to either $y^5$ or $y^8$.  

\begin{example}
\label{ex:formal8}
Let $\Pi=(\ftl,\R)$, where:
\begin{itemize} 
\item $\ftl$ contains the timelines 
\begin{center}
\begin{tabular}{l l}
$\mathit{FTL}_{r}$ = & $r^1 = (Idle, [16, 32], [11, 200], c)$\\
				& $r^2 = (TakeSample, [45, 102], [1, 70], c)$\\
				& $r^3 = (Idle, [200, 200], [23, 178], c)$\\[.2cm]
\end{tabular}
\begin{tabular}{l l}
$\mathit{FTL}_{inst}$ = & $inst^1 = (Stowed, [13, 38], [9, 45], c)$\\
				& $inst^2 = (Unstowing, [16, 41], [3, 3], c)$\\
				& $inst^3 = (Unstowed, [25, 70], [1, 30], c)$\\
				& $inst^4 = (Placing, [28, 75], [3, 7], c)$\\
				& $inst^5 = (Sampling, [33, 90], [7, 18], u)$\\
				& $inst^6 = (Unstowed, [200, 200], [1, 200], c)$\\[.2cm]
\end{tabular}
\begin{tabular}{l l}
$\mathit{FTL}_{nav}$ = & $nav^1 = (Home, [5, 5], [5, 5], c)$\\
				& $nav^2 = (Moving, [24, 35], [19, 37], u)$\\
				& $nav^3 = (At, [200, 200], [1, 200], c)$\\[.2cm]
\end{tabular}
\begin{tabular}{l l}
$\mathit{FTL}_{cm}$ = & $cm^1 = (Idle, [58, 77], [50, 80], c)$\\
				& $cm^2 = (SendData, [70, 105], [11, 38], u)$\\
				& $cm^3 = (Idle, [200, 200], [1, 200], c)$\\[.2cm]
\end{tabular}
\begin{tabular}{l l}
$\mathit{FTL}_{win}$ = & $win^1 = (NotAvailable, [32, 75], [23, 88], u)$\\
				& $win^2 = (Available, [95, 155], [60, 100], u)$\\
				& $win^3 = (NotAvailable, [200, 200], [1, 100], u)$
\end{tabular}
\end{center}
\item $\R$ contains the temporal relations
\begin{center}
\begin{tabular}{c}
$win^2 \contains_{[0,\infty] [0,\infty]} cm^2$\\[.1cm]
$nav^3 \contains_{[0,\infty] [0,\infty]} cm^2$\\[.1cm]
$inst^1 \contains_{[0,\infty] [0, \infty]} nav^2$\\[.1cm]
$r^2 \contains_{[0,\infty] [0,\infty]} inst^5$\\[.1cm]
$nav^3 \contains_{[0,\infty] [0,\infty]} inst^5$\\[.1cm]
$r^2 \before_{[0,\infty]} cm^2$
\end{tabular}
\end{center}
\end{itemize}
$\Pi$ is a flexible plan, and the set $\tl$ of scheduled timelines containing
\begin{center}
\begin{tabular}{l l}
$\mathit{FTL}_{r}$ = & $r^1 = (Idle, 18, [11, 200], c)$\\
				& $r^2 = (TakeSample, 65, [1, 70], c)$\\
				& $r^3 = (Idle, 200, [23, 178], c)$\\[.2cm]
\end{tabular}
\begin{tabular}{l l}
$\mathit{FTL}_{inst}$ = & $inst^1 = (Stowed, 42, [9, 45], c)$\\
				& $inst^2 = (Unstowing, 45, [3, 3], c)$\\
				& $inst^3 = (Unstowed, 46, [1, 30], c)$\\
				& $inst^4 = (Placing, 50, [3, 7], c)$\\
				& $inst^5 = (Sampling, 57, [7, 18], u)$\\
				& $inst^6 = (Unstowed, 200, [1, 200], c)$\\[.2cm]
\end{tabular}
\begin{tabular}{l l}
$\mathit{FTL}_{nav}$ = & $nav^1 = (Home, 5, [5, 5], c)$\\
				& $nav^2 = (Moving, 40, [19, 37], u)$\\
				& $nav^3 = (At, 200, [1, 200], c)$\\[.2cm]
\end{tabular}
\begin{tabular}{l l}
$\mathit{FTL}_{cm}$ = & $cm^1 = (Idle, 70, [50, 80], c)$\\
				& $cm^2 = (SendData, 93, [11, 38], u)$\\
				& $cm^3 = (Idle, 200, [1, 200], c)$\\[.2cm]
\end{tabular}
\begin{tabular}{l l}
$\mathit{FTL}_{win}$ = & $win^1 = (NotAvailable, 63, [23, 88], u)$\\
				& $win^2 = (Available, 125, [60, 100], u)$\\
				& $win^3 = (NotAvailable, 200, [1, 100], u)$
\end{tabular}
\end{center}
is an instance of $\Pi$, since $\{\mathit{STL}_{r}$, $\mathit{STL}_{inst}$, $\mathit{STL}_{nav}$, $\mathit{STL}_{cm}$, 
$\mathit{STL}_{win}\}$ is a schedule of $\ftl$ and it satisfies all the temporal relations  in $\R$.\\
If however, the end time of $inst^5$ in $\mathit{STL}_{inst}$ is replaced by $68$, the so obtained set of scheduled timelines 
is not an instance of $\Pi$, although it is a schedule of $\ftl$, because the relation
$$\{r^2 \contains_{[0,\infty] [0,\infty]} inst^5\}\in\R$$ is not satisfied.
\end{example}

The correspondence between token variables and token names is established by use of a function $\varphi$ mapping 
$a_1$ to either $y^5$ or $y^8$. According to the chosen option, the set of relations $\R$ in the plan contains either 
$y^5\ebs_{[10,20]}x^3$ or $y^8\ebs_{[10,20]}x^3$.

\begin{definition}
\label{def:semanticasync}
{\bf [Plan Satisfiability]}
Let $\C=A_1\wedge\dots \wedge A_m$ be a conjunction of atoms, $\Pi=(\ftl,\R)$ a flexible plan, where $\ftl$ contains 
a timeline for every state variable occurring in $\C$, and $\varphi$ a token assignment on $\ftl$.
The plan $\Pi$ satisfies $\C$  with $\varphi$ if for every atom $A\in \{A_1,\dots, A_m\}$, (i) if $A$ $=$ $a_i$ $\rel_{[lb,ub]}$ 
$a_j$, then $\varphi\left(a_i\right)$ $\rel_{[lb,ub]}$ $\varphi\left(a_j\right)\in\R$, and (ii) if $A$ $=$ $a_i$ $\rel_{[lb,ub]}$ $t$, 
then $\varphi\left(a_i\right)$ $\rel_{[lb,ub]}$ $t\in\R$.
Let $$\calE=\exists\,a_1[x_1=v_1]\dots a_n[x_n=v_n]\,.\, \C$$ be an existential statement and $\varphi$ be a token 
assignment on $\ftl$. The flexible plan $\Pi$ satisfies $\calE$ with $\varphi$ if $\varphi$ is an assignment for a set of 
annotated token variables including $a_0[x_0=v_0],\dots,a_n[x_n=v_n]$ and $\Pi$  satisfies $\C$ with $\varphi$.\\
The plan $\Pi$ satisfies a synchronization rule with non-empty trigger 
$$a_0[x_0=v_0]\rightarrow {\cal E}_1\vee \dots\vee {\cal E}_k$$ if for every flexible token $x_0^m$ of the timeline 
$\mathit{FTL}_{x_0} \in \ftl$ such that $\val(x_0^m)=v_0$, there exists a token assignment $\varphi$  on $\ftl$ such 
that $\varphi(a_0)=x_0^m$ and $\Pi$  satisfies ${\cal E}_i$ with $\varphi$, for some $i\in\{1,\dots,k\}$.\\
The plan $\Pi$   satisfies a synchronization rule with empty trigger 
$$\top\rightarrow {\cal E}_1\vee \dots\vee {\cal E}_k$$ if, for some $i\in\{1,\dots, k\}$, there exists a token assignment 
$\varphi$ on $\ftl$ such that $\Pi$  satisfies ${\cal E}_i$ with $\varphi$.
\end{definition}

\begin{example}
\label{ex:formal9}
Let us consider the flexible timelines $\mathit{FTL}_{cm}$, $\mathit{FTL}_{nav}$ and $\mathit{FTL}_{win}$ 
of Example \ref{ex:formal8}. The flexible plan $\Pi=( \ftl,\R)$, where $\ftl=\{$ $\mathit{FTL}_{cm}$, 
$\mathit{FTL}_{nav}$, $\mathit{FTL}_{inst}$, $\mathit{FTL}_{win}\}$ and $\R$ containing the temporal relations
$$\{win^2 \contains_{[0,\infty] [0,\infty]} cm^2, nav^3 \contains_{[0,\infty] [0,\infty]} cm^2\}$$
satisfies the synchronization rule with trigger {\em SendData}, given in Example \ref{ex:formal7}:
\begin{center}
\begin{tabular}{r l}
$a_0[cm = SendData] \rightarrow$ & $\exists\ a_1[win = Available]\, a_2[nav = At].$\\
							& $a_1 \contains_{[0,\infty] [0,\infty]} a_0 \land\ a_2 \contains_{[0,\infty] [0,\infty]} a_0$
\end{tabular}
\end{center}
Considering the definition of the relation $\contains$ given in Table \ref{table:deftemprel}, the two temporal 
constraints of the synchronization rule can be rewritten as:
$$(a_1 \sbs_{[0,\infty]} a_0 \land\ a_0 \ebe_{[0,\infty]} a_1),$$
$$(a_2 \sbs_{[0,\infty]} a_0 \land\ a_2 \ebe_{[0,\infty]} a_0).$$
The set  $\R$ contains the atoms 
$$\varphi(a_1) \sbs_{[0,\infty]}\varphi(a_0),~\varphi(a_0) \ebe_{[0,\infty]}\varphi(a_1),$$
$$\varphi(a_2) \sbs_{[0,\infty]}\varphi(a_0),~\varphi(a_0) \sbs_{[0,\infty]}\varphi(a_2)$$
for the token assignment $\varphi$ such that $\varphi(a_0)=cm^2$, $\varphi(a_1)=win^2$ and $\varphi(a_2)=nav^3$.
Clearly, there might be schedules of the set of timelines $\ftl$ that do not satisfy all the requirements.
For example the requirement $win^2 \contains_{[0,\infty]}{[0,\infty]} cm^2$ is not satisfied by the schedules where 
$\st(cm^2) = 70$, $\et(cm^2) = 108$, $\st(win^2) = 30$ and $\et(win^2) = 90$, which consequently are not instances 
of the flexible plan $\Pi$.\\
However, if schedules like those in Example \ref{ex:formal8} satisfy all the requirements, then such schedules of $\ftl$
are also instances of $\Pi$.
As a further example showing how a plan commits to a choice among the possibly different ways to satisfy a synchronization 
rule, let us consider a  set $\ftl$ of timelines  and a rule $S$ of the form
\begin{center}
\begin{tabular}{r l}
$a_0 [x=v]\rightarrow$ & $\exists a_1[y=v'].  \,a_1\ebs_{[10,20]}a_0$\\[.1cm]
  				&  $\vee ~\exists a_1[z=v'']. \,a_0\ebs_{[5,\infty]}a_1$
\end{tabular}
\end{center}
Let us moreover assume that $x^3$ and $x^7$ are the only tokens with value $v$ in the timeline $\mathit{FTL}_x$ for $x$ in 
$\ftl$, that the timeline  $\mathit{FTL}_y$ for $y$ in $\ftl$ contains exactly one token $y^5$  having value $v'$, and that
$\mathit{FTL}_z$ contains exactly one token $z^8$ with value $v''$.\\
In order to satisfy the rule $S$:
\begin{enumerate}
\item $\R$ must contain the constraints $$\{y^5\ebs_{[10,20]} x^3, x^3\ebs_{[5,\infty]} z^8\}$$ 
In fact the plan has to satisfy either $\exists a_1[y=v'].a_1\ebs_{[10,20]}a_0$ or $\exists a_1[z=v''].a_0\ebs_{[5,\infty]}a_1$
with a token assignment $\varphi$ such that $\varphi(a_0)=x^3$.
If $\R$ contains $y^5$ $\ebs_{[10,20]}$ $x^3$, then the plan satisfies the existential statement (i)  with $\varphi$, when 
$\varphi(a_1)=y^5$.
If it contains $x^3$ $\ebs_{[5,\infty]}$ $z^8$, then the plan satisfies (ii) with $\varphi$, when $\varphi(a_1)=z^8$.
\item $\R$ must contain the constraints $$y^5\ebs_{[10,20]} x^7, x^7\ebs_{[5,\infty]} z^8$$
The reasoning is the same as above, just replacing $x^7$ for $x^3$.
\end{enumerate}
Therefore, for instance, both plans 
\[\begin{array}{ll}
& (\ftl,\{y^5\ebs_{[10,20]} x^3,x^7\ebs_{[5,\infty]} z^8\})\\
& (\ftl,\{x^3\ebs_{[5,\infty]} y^5,y^5\ebs_{[10,20]} x^7\})
\end{array}\]
satisfy $S$
\end{example}
The notions of plan validity and consistency can now be defined.

\begin{definition}
\label{def:plan-val-cons}
{\bf [Plan Validity]}
A flexible plan $\Pi = \left(\ftl,\R\right)$ is \defi{valid} with respect to a planning domain $\D=(SV_P,SV_E,\S)$ iff:
\begin{enumerate}
\item $\ftl$ is a set of timelines for the state variables $SV=SV_P\cup SV_E$;
\item $\Pi$  satisfies all the synchronization rules in $\S$;
\item for each planned state variable $x=(V,T,\gamma,D)\in SV_P$, and each uncontrollable token $x^i$ 
in $\mathit{FTL}_x\in  \ftl$, if $D(\val(x^i))=(d_{min},d_{max})$ and $\st(x^i)=[s,s']$, then 
$\duration(x^i)$ = $[d_{min}, d_{max}]$ and $\et(x^i)$ = $[s+d_{min},s'+d_{max}]$.
\end{enumerate}
The plan $\Pi$ is \defi{consistent} if there exists at least one instance of $\Pi$.  
\end{definition}

The last condition required for a plan to be valid guarantees that the plan does not make any hypothesis on the 
duration of uncontrollable values of planned variables. The restriction is not applied to external variables, since the 
planner is not allowed to control them at all: their behavior is described in the planning problem as a sort of observation 
of the external world.

It is important to point out that plan consistency is a minimal requirement for a plan to be considered 
meaningful, although, when the domain includes uncontrollable elements, it is not enough to guarantee its executability. 
In this regards, the work \cite{cialdea2015planning} proves a result showing that there exists a set $\Theta$ of flexible plans 
for which an effective consistency check procedure exists, yet every scheduled valid plan is an instance of some flexible 
plan in $\Theta$.  
Intuitively, each plan $\Pi\in \Theta$ is such that the sequence of scheduled tokens, obtained by fixing every token end point 
to the lower bound of the respective end time interval, is an instance of $\Pi$ (i.e., it is a scheduled timeline respecting the 
relations in $\Pi$). The mentioned result implies that, when searching for a consistent plan, it is sufficient to consider candidate 
plans in $\Theta$, respecting the above condition.

\section{Problem Specification}
\label{sec:prob-spec}
In timeline-based planning, a planning problem typically includes   a {\em planning horizon}, {i.e.}, the time 
by which the system behavior has to be planned.
Finally, since the external state variables are not under the system control, the problem must include information about 
their behavior up to the given horizon. Such information is given in the form of a set of flexible timelines and temporal relations 
on their tokens.

\begin{definition}
\label{def:pproblem}
{\bf [Planning Problem]}
A \defi{planning problem} is a  tuple $(\D,\G,\Obs,H)$, where $\D=(SV_P\comma SV_E\comma \S)$ is a planning domain, 
$\G$ a planning goal for $\D$, $H\in\tp_{>0}$ is the {planning horizon}, and $\Obs=(\ftl_E,\R_E)$, where 
\begin{itemize}
\item [(i)] $\ftl_E$ is a set containing exactly one flexible timeline for each external state variable in $SV_E$;
\item [(ii)] the horizon of every timeline in $\ftl_E$ is $[h,h']$ for some $h\geq H$;
\item [(iii)] $\R_E$ is a set of temporal relations on tokens of timelines in $\ftl_E$;
\item[(iv)] $(\ftl_E,\R_E)$ is consistent, i.e., there is at least one schedule of $\ftl_E$ satisfying the relations in $\R_E$.
\end{itemize}
\end{definition}

The pair $\Obs$, called the  \defi{observation}, specifies the behavior of external state variables up to a time point not less 
than the planning horizon. Item (iv) rules out inconsistent observations, i.e., descriptions of the behavior of the external state 
variables with no instances. The pair $(\ftl_E,\R_E)$ can be viewed as a flexible plan.
In particular, even when $\R_E=\O$, the set of timelines $\ftl_E$ must have at least one schedule.

The planner must respect what is specified by the set of timelines $\ftl_E$, without taking any autonomous decision:
this requirement is fulfilled simply when the  timeline for each external variable in the plan is exactly the timeline for the
same state variable in $\ftl_E$. The relations in $\R_E$ represent known facts about the external world.

It is worth pointing out that, since $\ftl_E$ is a set of timelines, the planner knows how the external components evolve, i.e., 
the sequence of activities/states constituting their behavior, the only uncertainty being the duration of such states. 
This rules out, for instance, scenarios where the uncontrollable events might occur an unknown number of times 
within the given horizon.

\begin{example}
\label{ex:formal10}
For instance, a planning problem for our sample domain can be the problem $\Pi=(\D,\G,\Obs,H)$,
where
\begin{itemize}
\item $\D$ is the planning domain of Example \ref{ex:formal1}, i.e., $\D=(SV_P,SV_E,\S\}$ where 
$SV_P=\{r,inst,nav,cm\}$, $SV_E=\{win\}$ and $\S$ contains the synchronization rules of Example \ref{ex:formal7}
\item $\G=(\Gamma ,\Delta)$ -- see Example \ref{ex:formal11} -- where $$\Gamma = \{g_1 = (r, TakeSample), g_2 = (cm, SendData)\}$$ and 
$$\Delta = g_1 \before_{[0, 65]} g_2$$
\item $\Obs = (\{\mathit{FTL}_{win}\}, \R_{E})$, where 
\begin{center}
\begin{tabular}{r l}
$\mathit{FTL}_{win}$ = & $win^1 = (NotAvailable, [32, 75], [23, 88], u)$\\
				& $win^2 = (Available, [95, 155], [60, 100], u)$\\
				& $win^3 = (NotAvailable, [200, 200], [1, 100], u)$
\end{tabular}
\end{center}
\item $H=200$.  
\end{itemize}
\end{example}

\subsection{Planning Goals}
A planning problem includes the description of the underlying planning domain and of a desired goal to be 
accomplished. This work considers {\em temporally extended goals}: a planning goal specifies that some planned 
variables have to assume some given values in some intervals, possibly satisfying some temporal relations. 
Disjunctive goals are also allowed.

\begin{definition}
\label{def:goal}
{\bf [Planning Goal]}
A \defi{planning goal} $\G$ for a  domain $\D=(SV_P,SV_E,\S)$ is a pair $(\Gamma,\Delta)$, where:
\begin{itemize}
\item[(i)] $\Gamma$ is a set of \defi{accomplishment goals}, {i.e.}, expressions of the form $g=(x,v)$, where $g$ is a 
token variable, called the goal name, $x\in SV_P$, and  $v\in \values(x)$;
\item[(ii)] $\Delta$, called a \defi{relational goal}, is a disjunction $\calD_1\vee\dots\vee\calD_k$, where each $\calD_i$ 
is a conjuntion of atoms containing  only  goal names occurring in $\Gamma$.
\end{itemize}
A planning goal $\G=(\Gamma,\Delta)$, with $\Gamma=\{g_1=(x_1,v_1)\comma\dots\comma g_n=(x_n\comma v_n)\}$
and $\Delta=\calD_1\vee\dots\vee\calD_k$, is represented by a synchronization rule $S_{\cal G}$ with empty trigger, of 
the form:
\begin{center}
\begin{tabular}{r l}
$\top\rightarrow$ & $\exists\, g_1[x_1 = v_1]\dots\ g_n[x_n = v_n].\,\calD_1$\\
				& $\vee\dots\vee$\\
				& $\exists\, g_1[x_1 = v_1]\dots g_n[x_n = v_n].\,\calD_k$
\end{tabular}
\end{center}
\end{definition}

It is worth pointing out that restrictions on the start and end intervals of a given goal (like in 
\cite{cimatti2013timelines,iaai09mrspock}) can be expressed by means of relational goals. In particular, if the start
point of a given goal $g$ is required to be in the interval $[\binf,\bsup]$ and its end point in $[\einf,\esup]$, then such
restrictions can be  expressed by the  relational goal $(g\,\startafter_{[0,\bsup-\binf]}\binf)\wedge 
(g\,\xendafter_{[0,\esup-\einf]}\einf)$.

\begin{example}
\label{ex:formal11}
A simple planning goal for the \rover\ domain may be that, in order to accomplish the mission, the rover has to 
take a sample of a target to be analyzed and then communicate the scientific results no later than 65 time units 
after the completion of the sampling task. Such a goal is is represented by the pair $(\Gamma ,\Delta)$, where 
$$\Gamma =\{g_{1} = (r, TakeSample), g_{2} = (cm, SendData)\}$$ and
$$\Delta =g_{1} \before_{[0,65]} g_2$$ 
which can be turned into the synchronization rule
\begin{center}
\begin{tabular}{r l}
$\top\rightarrow$ & $\exists\, g_1[r = TakeSample]\, g_2[cm = SendData].$\\
				& $g_1 \before_{[0, 65]} g_2$
\end{tabular}
\end{center}
Analogously, if the rover has to come back "home" (a known initial position) to complete the mission and we
want to specify an alternative ordering constraint for the communication task, {i.e.}, the rover may communicate 
scientific data either before going back "home" or immediately after, the goal is the goal is $\G=( \Gamma ,\Delta)$, 
where 
$$\Gamma =\{g_1 = (r, TakeSample), g_2 = (cm, SendData), g_3 = (nav, Home)\},$$ 
and 
$$\Delta = (g_1\before_{[0,\infty]} g_3 \land\ g_3\meets g_2) \vee (g_1\before_{[0, 65]} g_2 \land\ g_2 \before_{[0,\infty]} g_3).$$
The  corresponding synchronization rule is:
\begin{center}
\begin{tabular}{r l}
$\top\rightarrow$ & $\exists\, g_1[r = TakeSample]\, g_2[cm, SendData]\, g_3[nav, Home].$\\
				& $(g_1\before_{[0,\infty]} g_3 \land\ g_3\meets g_2)$\\[.1cm]
				& $\vee\, \exists\, g_1[r = TakeSample]\, g_2[cm = SendData]\, g_3[nav, Home].$\\
				& $(g_1\before_{[0, 65]} g_2 \land\ g_2 \before_{[0,\infty]} g_3)$
\end{tabular}
\end{center}
\end{example}

The next definition introduces the notion of goal fulfilment for scheduled timelines.

\begin{definition}
\label{def:goal-fulfillment} 
{\bf [Goal Satisfiability]}
A set of scheduled timelines $\tl$ fulfils the planning goal $\G$ if it satisfies the  synchronization rule $S_{\cal G}$ 
representing $\G$.
\end{definition}

\subsection{Solution Plans}
\begin{definition}\label{solutionPlan}
{\bf [Solution Plan]}
Let $\P=(\D,\G,\Obs,H)$ be a planning problem and $\Pi=(\ftl,\R)$ be a flexible plan.  
$\Pi$ is a \defi{flexible solution plan} for $\P$ if:
\begin{enumerate}
\item \label{cond:uno} for every planned state variable $x$, the horizon of $\mathit{FTL}_x\in\ftl$ is $[H,H]$;
\item \label{cond:due} $\Pi$ is valid with respect to $\D$;
\item \label{cond:tre} $\Pi$ satisfies the synchronization rule $S_{{\cal G}}$ representing $\G$;
\item \label{cond:quattro} If $\Obs=(\ftl_E,\R_E)$, then $\ftl_E\subseteq\ftl$.
\end{enumerate}
\end{definition}

The first condition above guarantees that the behavior of the planned state variables is determined exactly up to 
the horizon of the planning problem, henceforth (condition \ref{cond:tre}) all the planning goals are achieved in due time.
It is worth pointing out that condition \ref{cond:uno} implies that the last token of each planned timeline must be controllable.
Condition \ref{cond:quattro} ensures that the plan does not make any assumption on external variables, except for what is 
implied by the state variable definition and the observation.

\begin{example}
\label{ex:formal12}
Let us consider, for instance, the problem $\P$ of Example \ref{ex:formal10} and the flexible plan $\Pi=(\ftl,\R)$, 
where:
\begin{itemize}
\item $\ftl=\{\mathit{FTL}_{r},\mathit{FTL}_{inst},\mathit{FTL}_{nav},\mathit{FTL}_{cm},\mathit{FTL}_{win}\}$, where 
the timelines are those of the Example \ref{ex:formal8}:
\begin{center}
\begin{tabular}{l l}
$\mathit{FTL}_{r}$ = & $r^1 = (Idle, [16, 32], [11, 200], c)$\\
				& $r^2 = (TakeSample, [45, 102], [1, 70], c)$\\
				& $r^3 = (Idle, [200, 200], [23, 178], c)$\\[.2cm]
\end{tabular}
\begin{tabular}{l l}
$\mathit{FTL}_{inst}$ = & $inst^1 = (Stowed, [13, 38], [9, 45], c)$\\
				& $inst^2 = (Unstowing, [16, 41], [3, 3], c)$\\
				& $inst^3 = (Unstowed, [25, 70], [1, 30], c)$\\
				& $inst^4 = (Placing, [28, 75], [3, 7], c)$\\
				& $inst^5 = (Sampling, [33, 90], [7, 18], u)$\\
				& $inst^6 = (Unstowed, [200, 200], [1, 200], c)$\\[.2cm]
\end{tabular}
\begin{tabular}{l l}
$\mathit{FTL}_{nav}$ = & $nav^1 = (Home, [5, 5], [5, 5], c)$\\
				& $nav^2 = (Moving, [24, 35], [19, 37], u)$\\
				& $nav^3 = (At, [200, 200], [1, 200], c)$\\[.2cm]
\end{tabular}
\begin{tabular}{l l}
$\mathit{FTL}_{cm}$ = & $cm^1 = (Idle, [58, 77], [50, 80], c)$\\
				& $cm^2 = (SendData, [70, 105], [11, 38], u)$\\
				& $cm^3 = (Idle, [200, 200], [1, 200], c)$\\[.2cm]
\end{tabular}
\begin{tabular}{l l}
$\mathit{FTL}_{win}$ = & $win^1 = (NotAvailable, [32, 75], [23, 88], u)$\\
				& $win^2 = (Available, [95, 155], [60, 100], u)$\\
				& $win^3 = (NotAvailable, [200, 200], [1, 100], u)$
\end{tabular}
\end{center}
\item $\R$ contains the two relations on tokens
\begin{center}
\begin{tabular}{c}
$win^2 \contains_{[0,\infty] [0,\infty]} cm^2$\\[.1cm]
$nav^3 \contains_{[0,\infty] [0,\infty]} cm^2$\\[.1cm]
$inst^1 \contains_{[0,\infty] [0, \infty]} nav^2$\\[.1cm]
$r^2 \contains_{[0,\infty] [0,\infty]} inst^5$\\[.1cm]
$nav^3 \contains_{[0,\infty] [0,\infty]} inst^5$\\[.1cm]
$r^2 \before_{[0,\infty]} cm^2$
\end{tabular}
\end{center}
\end{itemize}
The plan $\Pi$ is a flexible solution plan for the planning problem $\P$ because:
\begin{itemize}
\item the horizon is 200 for all the timelines of the planned variables;
\item $\Pi$ is valid with respect to $\D$:
\begin{itemize}
\item $\ftl$ contains the timelines for $r$, $inst$, $nav$, $cm$ and $win$;
\item $\Pi$ satisfies the synchronization rule of the domain; 
%
%
\item all uncontrollable tokens satisfy the duration constraints of the related values
\end{itemize}
\item $\Pi$ satisfies the synchronization rule $S_{{\cal G}}$ representing $\G$: the tokens $pm^3$ 
and $pm^6$ have values $Science$ and $Comm$, respectively, and $\R$ contains the relation 
$pm^ 3 \before_{ [0,65]} pm^6$.
\item $\mathit{FTL}_{gv}\in \ftl$.
\end{itemize}
\end{example}

The next result proves that information encoded by a  flexible solution plan $\Pi$ for a given planning problem 
is sufficient to ensure that every instance of $\Pi$ is  valid with respect to the  planning domain and it fulfils the goal.
Although the proof of this result is a straightforward consequence of the definitions, it deserves to be stated explicitly, 
since flexible plans without such a property would be meaningless.

\begin{theorem}
\label{th:formal1} 
If the plan $\Pi $ is a flexible solution plan for the problem $$\P=(\D\comma \G\comma \Obs\comma H),$$
then every  instance of $\Pi$ is valid with respect to $\D$ and fulfils the goal  $\G$.
\end{theorem}


%


\ifpdf
    \graphicspath{{chapters/5_epsl/figures/PNG/}{chapters/5_epsl/figures/PDF/}{chapters/5_epsl/figures/}}
\else
    \graphicspath{{chapters/5_epsl/figures/EPS/}{chapters/5_epsl/figures/}}
\fi


%
%
%
%
\chapter{The Extensible Planning and Scheduling Library}
\label{chap:epsl}
\lettrine[lines=2]{T}{imeline-based applications} are problem solvers capable of taking into account several features
of a problem as well as integrating different techniques into the reasoning process (\eg\ planning and scheduling integration). 
The design and implementation choices made to realize this kind of applications are really complex and closely 
connected to the specific characteristics of the particular problem to address. These choices are difficult to replicate and 
therefore it is not easy to leverage {\em past experience} and deploy existing applications to different contexts. Typically, 
it is necessary to start developing new applications from scratch in order to solve new types of problem. 

The {\em Extensible Planning and Scheduling Library} (\epsl) \cite{epsl-aixia15,epsl-rcra13} is the result of a research effort 
which aims at realizing a general purpose timeline-based framework capable of supporting the design of \ps\ applications. 
The modeling features of \epsl\ concerning the representation and management of timelines take inspiration from \apsi.
Nevertheless, according to the formalization described in \cite{cialdea2015planning}, \epsl\ extends the \apsi\ representation 
by introducing {\em temporal uncertainty} in shape of {\em uncontrollable activities} and {\em external features} of a 
planning domain.
Temporal uncertainty allows \epsl\ to address planning problems in which not all the features of the domain are under the 
control of the system. Thus, \epsl-based solvers generate, if possible, plans with some desired property concerning 
{\em temporal controllability} \cite{morris01, VidalF99}. The {\em validity} of a plan with respect to the domain
specification does not represent a sufficient condition to guarantee its {\em executability} in the real-world. Namely, the 
{\em uncontrollable dynamics} of the environment may prevent the complete and correct execution of plans. Thus, from the 
planning perspective, it is important to generate plans with some {\em minimum} controllability properties. \epsl\ takes into 
account the {\em pseudo-controllability} property which represents a necessary but not sufficient condition for {\em dynamic 
controllability} \cite{morris01} of plans.

Broadly speaking, the solving approach is a general {\em plan refinement} procedure which iteratively detects a set of 
flaws on the current plan and selects the {\em most promising} flaw to solve according to certain {\em evaluation criteria}.
The actual behavior of the solving procedure is determined by the specific {\em configuration} of an \epsl-based solver in terms 
of the particular strategy and the particular {\em evaluation criteria} applied during the search. In this context, the \epsl\ modular 
architecture allows to find the planner configuration which {\em best} meets the specific features of the problem to address. 
In particular, this work presents a modeling and solving approach which allows to realize a hierarchical reasoning with timelines 
\cite{epsl-aixia15}.

\section{The Modeling Language}
\epsl\ takes inspiration from the \apsi\ representation functionalities and uses the {\em Domain Description Language} (\ddl) 
which is the modeling language that also \apsi\ uses to model planning domains. \ddl\ is a structured modeling language 
introduced in \cite{ddl1-book-96}. It provides the syntactic elements needed to describe timeline-based domains, 
\ie\ {\em state variables} and {\em synchronization rules}. In addition, the {\em Problem Description Language} (\pdl) is a language 
dedicated to describe problem instances.

Thus, an \epsl-based planner takes as {\em input}, a \ddl\ and a \pdl\ files representing respectively the description of a 
timeline-based domain and the description of a particular problem to solve. Given such input, an \epsl-based planner 
generates (if possible) a valid solution plan as {\em output}. In particular, \epsl\ relies on an extended syntax of \ddl\ in order to 
comply with the formal characterization of timelines introduced in \cite{cialdea2015planning}. 
Specifically, \epsl\ introduces the syntactic constructs that allow users to specify controllability properties of state variable values 
(\ie\ whether a value is {\em controllable} or {\em uncontrollable}) and the type of modeled state variables (\ie\ whether a state variable 
is {\em planned} or {\em external}). Thus, the following sections provide a detailed description of the extended \ddl/\pdl\ language by 
exploiting the \rover\ planning domain introduced in Chapter \ref{chap:formalization}.

\subsection{The Domain Description Language}
The Domain Description Language (\ddl) is the language \epsl\ uses for domain modeling. The \ddl\ provides the syntactic
constructs needed to describe state variables, synchronization rules and all the information needed to characterize a 
timeline-based planning domain. This section introduces the "extended" \ddl\ syntax by describing a timeline-based model 
designed for the \rover\ planning domain. In general, a \ddl\ model is composed by the following parts: 
(i) general declaration; 
(ii) state variable specification; 
(iii) component specification; 
(iv) synchronization specification.

The code below shows the general domain declaration for the \rover\ planning domain. It declares the 
name of the planning domain, the temporal horizon and the types of parameters that the values of the 
components may assume. Specifically, the {\em location} parameter models the set of physical locations the rover
can move to. In this specific domain the possible locations are discretized and modeled as symbols rather than as 
coordinates on a two-dimensional or three-dimensional space. Thus, {\em location} is an {\em enumeration parameter} 
whose values are included in a discrete set of symbols.
The {\em file} parameter models the data files that the rover can communicate to share scientific information. 
The parameter is modeled as a {\em numeric parameter} whose values range within the interval [0, 100].

\begin{code}
{\bf DOMAIN} Rover 
\{   
   {\bf TEMPORAL_MODULE} tm = [0, 100];
   
   {\bf PAR_TYPE} {\em EnumerationParameter} location = \{
      home, location1, location2, location3, location4
   \}
   
   {\bf PAR_TYPE} {\em NumericParameter} file = [0, 100];
   
   ...
\}
\end{code}

The state variable specification of a \ddl\ model describes the features of the planning domain that must be controlled 
over time and their possible temporal evolutions. Considering the \rover\ example, the domain specification 
must model the {\em navigation} facility that allows the rover to move, the {\em communication facility} that allows the rover 
to communicate scientific data and the {\em instrument} that allows the rover to take samples to be analyzed. In addition, the 
domain specification must model the available {\em communication windows} during which the rover may actually send data.

The following code shows the state variable declaration modeling the {\em functional} and {\em abstract} behavior 
of the whole system to control, \ie\ the planetary exploration rover. 
The {\em RoverType} state variable models the high-level tasks or states the rover may perform or assume over 
time. The value {\em Idle()} represents the fact that the rover is not operating and therefore it can receive goals, \ie\ 
tasks to be performed.
The value {\em TakeSample(?location, ?file)} represents a {\em mission goal} which asks the rover to take a sample at a
specific {\em ?location}, perform some analysis on the gathered samples and communicate data (\ie\ {\em ?file}) to the 
ground station. Both values are controllable and the related duration constraints do not bound them to some intervals. 
Consequently, the planner can dynamically decide the actual duration of these values according to the specific needs of the 
generated plans.

\begin{code}
{\bf COMP_TYPE} {\bf StateVariable} RoverType (
   Idle(), 
   TakeSample(location, file))
\{
   {\bf VALUE} Idle() {\bf [1, +INF]}
   {\bf MEETS} \{
      TakeSample(?location, ?file);
   \}
   
   {\bf VALUE} TakeSample(?location, ?file) {\bf [1, +INF]}
   {\bf MEETS} \{
      Idle();
   \}
\}
\end{code}

The code below shows the state variable specification for the navigation facility of the rover. The 
{\em NavigationType} state variable models the states and actions that the navigation module of the rover
can assume or perform over time. 
The value {\em At(?location)} models the fact that the rover is still at a known {\em ?location}. 
This value is controllable and represents a {\em temporally stable state} because the related 
duration constraint does not specify a bound for the value. The transition function requires that a value 
{\em GoingTo(?destination)} follows a value {\em At(?location)} and that the parameter constraint 
{\em ?location != ?destination} holds. The parameter constraint guarantees that the rover does not 
move in case that the current location coincides with the desired destination.

The value {\em GoingTo(?location)} models the moving action of the rover. In this case the value is 
{\em uncontrollable} and a duration bound is specified {\em [5, 11]}. Thus, the system knows the 
estimated duration of the action, but it cannot completely control it. Namely, the rover can decide when 
to start moving towards a destination, but it cannot predict the time needed to reach the destination. 
Indeed, external factors, such as obstacles or the features of the ground, may affect the speed of the rover 
 and therefore the time the rover takes to complete the {\em GoingTo(?location)} activity/task.
Moreover, the related transition constraint requires that a value {\em At(?destination)} follows a value
{\em GoingTo(?location)} and that the parameter constraint {\em ?destination = ?location} holds. 
Namely, the the rover is actually located at {\em ?location} (\ie\ the desired destination) after the 
successful execution of the{\em GoingTo(?location)}.

\begin{code}
{\bf COMP_TYPE} {\bf StateVariable} NavigationType (
   At(location), 
   GoingTo(location))
\{ 
   {\bf VALUE} At(?location) {\bf [1, +INF]}
   {\bf MEETS} \{
      GoingTo(?destination);
      ?location != ?destination;
   \}
   
   {\bf VALUE} {\bf uncontrollable} GoingTo(?location) {\bf [5, 11]}
   {\bf MEETS} \{
      At(?destination);
      ?destination = ?location;
   \}
\}
\end{code}

The next block of code describes the state variable modeling the instrument payload of the rover.
The {\em InstrumentType} state variable models the position the instrument assumes and the operations it may 
perform over time. It is a {\em planned variable} with both controllable and uncontrollable values. 
The values {\em Unstowed()} and {\em Stowed()} are {\em temporally stable values} that model respectively the idle 
state of the instrument and the operating state of the instrument. Namely, the instrument is not operative and cannot be 
used to take samples during {\em Stowed()}. Conversely, the instrument is operative and ready to take samples during 
{\em Unstowed()} . The transitions between the two states are modeled through values {\em Unstowing()} and {\em Stowing()}.
They are both {\em controllable} values and their durations are fixed. Thus the system knows
exactly how long the instrument takes to change from an idle state to an operative state and vice versa.

After {\em activation} through {\em unstowing}, the instrument must be placed over a target in order to take samples. Thus, 
the value {\em Placing(?location)} similarly to the {\em GoingTo(?location)} value,
models the transition between the {\em Unstowed} and the {\em Placed(?location)} values. However, the {\em Placed(?location)}, 
differently from the {\em GoingTo(?location)} value,  is modeled as a {\em controllable} value with flexible duration. 
Indeed, the instrument is supposed to move among the reachable position without finding obstacles or any sort of {\em external 
elements} that may slow-down or even prevent the motion. Thus, it can be modeled as a {\em controllable process} whose flexible 
duration is determined by the minimum and maximum time needed by the instrument to reach the possible targets.

The value {\em Placed(?location)} models the fact that the instrument has been actually placed on a target and therefore it is 
ready to perform any operation on it. The {\em Sampling} value models the operation which allows the instrument 
to take samples of the target it is placed on. Such operation is {\em uncontrollable} because the time the instrument 
takes to take samples is affected by the particular shape and size of the target. This is a source of {\em uncertainty} of the 
environment  and therefore the sampling operation is modeled as an {\em uncontrollable} process with an estimated lower and 
upper bounds for the duration.

\begin{code}
{\bf COMP_TYPE} {\bf StateVariable} InstrumentType (
   Unstowed(),
   Stowing(),
   Stowed(),
   Unstowing(),
   Placing(location),
   Placec(location),
   Sampling(location))
\{
   {\bf VALUE} Unstowed() {\bf [1, +INF]}
   {\bf MEETS} \{
      Stowing();
      Placing(?location);
   \}
   
   {\bf VALUE} Stowing() {\bf [3, 3]}
   {\bf MEETS} \{
      Stowed();
   \}
   
   {\bf VALUE} Stowed() {\bf [1, +INF]}
   {\bf MEETS} \{
      Unstowing();
   \}
   
   {\bf VALUE} Unstowing() {\bf [3, 3]}
   {\bf MEETS} \{
      Unstowed();
   \}
   
   {\bf VALUE} Placing(?location) {\bf [3, 7]}
   {\bf MEETS} \{
      Placed(?target);
      ?target = ?location;
   \}
   
   {\bf VALUE} Placed(?location) {\bf [1, +INF]}
   {\bf MEETS} \{
      Sampling(?target);
      ?target = ?location;
      Placing(?newTarget);
      ?newTarget != ?target;
      Unstowed();
   \}
   
   {\bf VALUE} {\bf uncontrollable} Sampling(?target) {\bf [5, 18]}
   {\bf MEETS} \{
      Placed(?location);
      ?location = ?target;
   \}
\}
\end{code}

The communication facility of the rover is modeled by means of the {\em CommType} state variable. As the code below shows, 
the variable is composed by two values only. The {\em Idle} value models the fact that the communication
facility is available for communicating data. The {\em SendData} value models the fact that the communication facility is 
actually sending data and cannot be used for other operation until the data transfer is complete. The communication task is 
modeled as an {\em uncontrollable} process whose actual duration is affected by {\em external factors} like the 
{\em quality} of the communication signal available and the size of the amount of data to be transferred. Thus, the time the 
rover takes to send data cannot be decided and therefore the related value is uncontrollable.

\begin{code}
{\bf COMP_TYPE} {\bf StateVariable} CommType (
   Idle(),
   SendData(file))
\{
   {\bf VALUE} Idle() {\bf [1, +INF]}
   {\bf MEETS} \{
      SendData(?file);
   \}
   
   {\bf VALUE} {\bf uncontrollable} SendData(?file) {\bf [11, 32]}
   {\bf MEETS} \{
   	Idle();
   \}
\}
\end{code}

Finally, the availability of the communication channel during the mission of the rover is modeled by means of the 
{\em WindowType} {\em external} state variable. All the values of an external variable are uncontrollable by 
definition and therefore the {\em uncontrollable} tag is not needed. The {\em Available} value models the fact that the 
communication channel is supposed to be available and data can be actually transferred during the related (flexible) 
temporal interval. Conversely, the {\em NotAvailable} value models the fact that the communication channel is supposed 
to be not available and no data can be transferred during the related (flexible) temporal interval.

As formally described in Chapter \ref{chap:formalization}, external variables model features of the environment that are 
completely outside the control of the system. However, these features are relevant from the control and planning perspectives 
because their behaviors may directly or indirectly affect the behavior of the system. 
In this specific case, the availability of the communication channel affects the {\em scheduling} of the communication tasks 
of the rover. Clearly, the rover can neither decide or make hypothesis on the availability of the signal. Thus, according to 
Section \ref{sec:prob-spec}, the behaviors of these features must be {\em known} and provided with the {\em observations} 
of the problem specification.

\begin{code}
{\bf COMP_TYPE} {\bf StateVariable} {\bf external} WindowType (
   Available(),
   NotAvailable())
\{
   {\bf VALUE} Available() {\bf [1, +INF]}
   {\bf MEETS} \{
      NotAvailable();
   \}
   
   {\bf VALUE} NotAvailable() {\bf [1, +INF]}
   {\bf MEETS} \{
      Available();
   \}
\}
\end{code}

After state variable declaration, the {\em component specification} of a \ddl\ model aims at declaring the set of 
state variable instances that constitute the planning domain. The components of the \ddl\ represent the instantiated 
data structure the planning system actually deals with in order to find plans. 
In this case, the \ddl\ specification is composed by a component for each type of state state variable 
defined. The {\em RoverController} component represents an instance of the {\em RoverType} state variable. The 
{\em Navigation} component represents an instance of the {\em NavigationType} state variable. The {\em Instrument}
component represents  an instance of the {\em InstrumentType} state variable. The {\em Communication} component 
represents an instance of the {\em CommType} state variable. Finally, the {\em Channel} component represents an 
instance of the {\em WindowType} {\em external} state variable.

\begin{code}
{\bf COMPONENT} RoverController : RoverType;
{\bf COMPONENT} Navigation : NavigationType;
{\bf COMPONENT} Instrument : InstrumentType;
{\bf COMPONENT} Communication : CommType;
{\bf COMPONENT} Channel : WindowType;
\end{code}

The last part of a \ddl\ planning model concerns the synchronization rule specification. While the state variable specification 
constrains the behaviors of the single features of the domain, synchronization rules specify {\em global} constraints aiming at
coordinating domain components in order to realize {\em complex behaviors} of the system.
The key point of the timeline-based modeling is that synchronization rules, differently from actions of {\em classical planning}, 
do not explicitly consider {\em causal} relationships among {\em tokens} of the timelines. Such rules "simply" specify additional 
constraints on the behaviors of state variables the planner must take into account when building the related timelines.
The following code shows the synchronization rules defined for the \rover\ domain. These rules model the 
{\em operational constraints} introduced in Section \ref{sec:rover} that allow the rover to successfully complete the mission. 

The rule defined on the value {\em TakeSample} of component {\em RoverController} models the operational requirements
the rover must follow to complete a {\em mission goal}. Specifically, the rule requires the rover to be located at the target  
position when performing sampling operations. 
Then, when the {\em TakeSample} task is completed the rover must send resulting data through the communication 
facility. According to these rules, the following temporal constraints must hold in the generated plans:
\[\begin{array}{l}
TakeSample(?t, ?f) \during_{[0,\infty] [0,\infty]} Navigation.At(?t)\\
TakeSample(?t, ?f) \contains_{[0,\infty] [0,\infty]} Instrument.Sampling(?t)\\
TakeSample(?t, ?f) \before_{[0,\infty]} Communication.SendData(?f)
\end{array}\]

Similarly, the synchronization rule on the value {\em SendData} of the component {\em Communication} models the 
operational requirements the rover must follow to successfully send data to the satellite. Specifically, the rule requires the 
rover to be still for the entire duration of the communication which in turn must be performed when the channel is available. 
Thus, the planner must generate plans that satisfy the following temporal constraints:
\[\begin{array}{l}
SendData(?f) \during_{[0,\infty] [0,\infty]} Navigation.At(?location)\\
SendData(?f) \during_{[0,\infty] [0,\infty]} Channel.Available()
\end{array}\]

Finally, in addition to operational requirements, sycnhronization rules may also model {\em safety constraints}. Namely,
constraints needed to guarantee the {\em safety} of the system but not essential for the mission. The synchronization 
rule on value {\em GoingTo} of the {\em Navigation} component represents an example of such constraints. The rule 
requires the rover to keep the instrument stowed while moving in order to avoid collisions and preserve the safety of 
the device. Thus, every time the rover moves between two locations, following temporal constraint must hold:
\[\begin{array}{l}
Navigation.GoingTo(?t) \during_{[0,\infty] [0,\infty]} Instrument.Stowed()
\end{array}\]

\begin{code}
{\bf SYNCHRONIZE} RoverController
\{
   {\bf VALUE} TakeSample(?target, ?file)
   \{
      cd0 Navigation.At(?location);
      cd1 Instrument.Sampling(?target1);
      cd2 Communication.SendData(?file2);
      
      {\bf DURING [0, +INF] [0, +INF]} cd0;
      {\bf CONTAINS [0, +INF] [0, +INF]} cd1;
      {\bf BEFORE [0, +INF]} cd2;
      
      ?target1 = ?target;
      ?file2 = ?file;
   \}
\}

{\bf SYNCHRONIZE} Communication
\{
   {\bf VALUE} SendData(?file)
   \{
      cd0 Channel.Available();
      cd1 Navigation.At(?location);
      
      {\bf DURING [0, +INF] [0, +INF]} cd0;
      {\bf DURING [0, +INF] [0, +INF]} cd1;
   \}
\}

{\bf SYNCHRONIZE} Navigation
\{
   {\bf VALUE} GoingTo(?destination)
   \{
      cd0 Instrument.Stowed();
      
      {\bf DURING [0, +INF] [0, +INF]} cd0;
   \}
\}
\end{code}

\subsection{The Problem Description Language}
The {\em Problem Description Language} (\pdl) is the language \epsl\ uses for problem modeling. Given a domain 
specification, the \pdl\ provides the syntax constructs needed to specify known {\em facts} about the {\em world}, the
{\em observations} concerning the external variables of the domain (if any) and {\em planning goals}.

The following block of code represents a part of problem specification which declares the name of the problem instance, 
the planning domain it relies on and a set of known facts. In timeline-based planning, facts represent a partial description 
of the timelines of the domain components. Namely, they partially constrain the temporal behaviors of components by 
specifying a set of values the related variables can assume within known temporal bounds. Thus, the planning 
system builds the temporal behaviors of domain components by taking into account also the related {\em known facts} of the 
\pdl.

These facts represent tokens on domain timelines and characterize the initial (partial) plan of the planning process. Namely, 
such tokens (partially) constrain the temporal behaviors of domain components and the solution plan must be built accordingly.
For example, the token {\em f0} in the block of code below specifies the starting position (\ie\ the {\em home} location) of rover 
mission. Similarly, facts {\em f1} and {\em f2} specify respectively that the instrument and the communication facility of the rover 
are {\em stowed} and in {\em idle} state when the mission starts. 

\begin{code}
{\bf PROBLEM} Rover_1task ({\bf DOMAIN} Rover)
\{
   f0 {\bf fact} Navigation.At(?startLocation) {\bf AT [0, 0] [1, +INF] [1, +INF]};
   f1 {\bf fact} Instrument.Stowed() {\bf AT [0, 0] [1, +INF] [1, +INF]};
   f2 {\bf fact} Communication.Idle() {\bf AT [0, 0] [1,+INF] [1,+INF]};
   
   ... 
   
   ?startLocation = home;
\}
\end{code}

If a planning domain contains {\em external variables} the \pdl\ must specify the related {\em observations}. According 
to Definition \ref{def:pproblem}, observations must completely describe the temporal behaviors of external variables. Namely, 
they must specify the complete sequence of tokens that compose the timelines. 
The code below shows an example of observations for the external variables of the \rover\ planning domain. Specifically, the 
observations describe the sequence of (flexible) tokens that compose the timeline of the {\em Channel} component which 
models the availability of the communication signal during the mission.

\begin{code}
o1 {\bf fact} Channel.NotAvailable() {\bf AT [0, 0] [25, 30] [25, 30]};
o2 {\bf fact} Channel.Available() {\bf AT [25, 30] [80, 85] [55, 60]};
o3 {\bf fact} Channel.NotAvailable() {\bf AT [80, 85] [100, 100] [15, 20]};
\end{code}

Finally, the \pdl\ file contains the goal specification that, similarly to facts, represent a set of constraints concerning the temporal 
behaviors of domain components. Given such constraints (\ie\ facts and goals), the planning system must build valid temporal behaviors 
(\ie\ timelines) in order to complete a mission. Namely, the solving process is triggered by planning goals that represent requirements 
that solution plans must satisfy. The code below represents an example of a {\em planning goal} for the \rover\ domain. In this 
example, the goal {\em g0} requires the rover to perform a {\em TakeSample} task within some temporal bounds. 

\begin{code}
g0 {\bf goal} Rover.TakeSample(?tl, ?f) {\bf AT [0, 35] [22, 65] [1, 45]};
   
?tl = location5;
?f = 1;
\end{code}

\section{Architectural Overview}
\begin{figure}[ht]
\centering
\includegraphics[width=\textwidth]{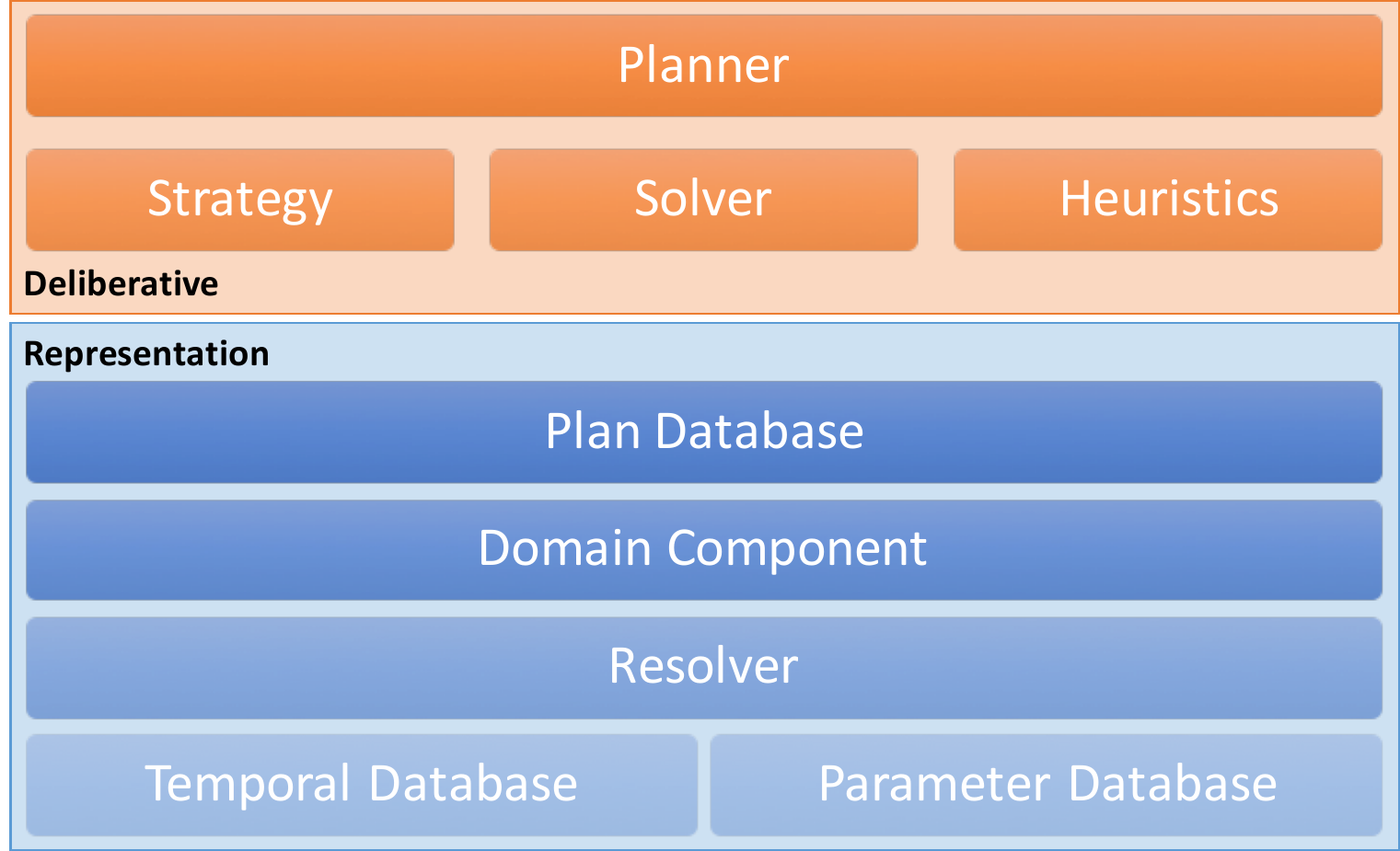}
\caption{\small{The layered architecture of the \epsl\ framework}}
\label{fig:epsl-archi}
\end{figure}

\epsl\ defines a flexible software framework to support the design and development of timeline-based applications. 
It is organized according to the {\em Multilayered architectural pattern}\footnote{A {\em multi-layered architecture} is a software 
architecture that uses many layers for allocating the different responsibilities of a software 
product} \cite{pattern}. The framework addresses the design and development of a timeline-based application from different 
perspectives ranging from the temporal reasoning mechanism to the definition of search strategies and heuristics. 
Each layer provides a set of ready-to-use algorithms and data structures that can be combined together to develop new 
planning instances. The {\em modularity} of the architecture allows users to easily integrate new features and improve the reasoning 
and representation capabilities of the framework.
Figure \ref{fig:epsl-archi} shows the main architectural elements that compose the \epsl\ framework. In general the system 
is composed of two {\em macro layers}, (i) the {\em Representation} layer and (ii) the {\em Deliberative} layer. 

\subsection{Representation Framework}
The {\em Representation} layer is responsible for encapsulating and providing all the information and functionalities 
the solver needs to build timeline-based plans. The {\em Temporal DataBase} provides {\em temporal reasoning mechanisms} 
for checking the temporal consistency and {\em inferring} additional {\em knowledge} about the temporal relations of the plan. 
Similarly, the {\em Parameter DataBase} provides {\em CSP-based reasoning mechanism} for managing variable declaration 
and constraint propagation.

On top of these mechanisms {\em resolvers} and {\em components} provide a set of ready-to-use data structures and algorithms.
These elements encapsulate the functionalities needed to manage timelines and timeline-based plans. 
In particular {\em resolvers} are the basic architectural elements for building timeline-based plans. They encapsulate the 
logic for managing {\em plan flaws} during the planning process. A {\em flaw} represents a particular issue concerning the plan 
which must be solved in order to find a solution. There are two types of flaws that must be managed: (i) {\em planning goals} represent 
flaws affecting the {\em completion} of the plan; (ii) {\em threats} represent flaws affecting the {\em validity} of the plan. 
Thus, each {\em resolver} is a dedicated algorithm responsible for detecting and solving a specific type of flaw.

{\em Domain Components} represent data structures modeling the different types of features of a planning domain. Specifically, each 
component aggregates a set of {\em resolvers} that determine the resulting behavior in terms of possible flaws that may concern the 
particular feature of the domain. Namely, the set of resolvers related to a component determine the conditions that 
must be solved in order to build {\em valid temporal behaviors} of the related feature (\ie\ the timelines).
Given this structure, the representation capability of the \epsl\ framework can be "easily" extended by introducing new 
domain components and new resolvers for building the related temporal behaviors. The higher the number of types of 
resolvers and components available the more the {\em expressivity} of the framework.

\begin{figure}[ht]
\centering
\includegraphics[width=\textwidth]{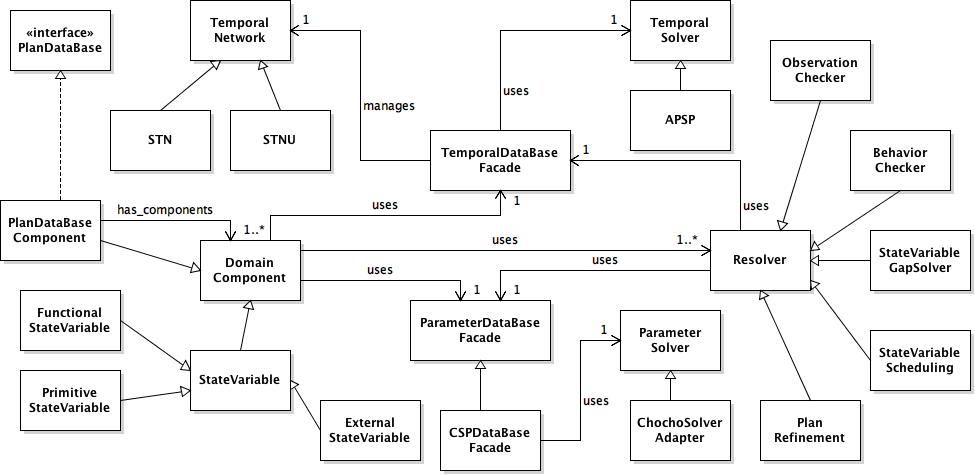}
\caption{\small{The structure of the plan database in the \epsl\ framework architecture}}
\label{fig:epsl-pdb}
\end{figure}

All the functionalities and information of the {\em Representation} layer are made available through the {\em PlanDataBase 
interface} which is a compound element encapsulating the complexity of plan management. Figure \ref{fig:epsl-pdb} provides 
a more detailed representation of the elements that compose the plan database and their relationships.

{\em DomainComponents} model the different types of feature the planning system can reason about. The {\em PlanDataBaseComponent} is a 
particular type of component which encapsulates other components of the domain (see the {\em Composite} design pattern 
\cite{gof}) and provides access to the information of the plan through the {\em PlanDataBase interface}. 
The {\em PlanDataBaseComponent} uses the {\em PlanRefinement} resolver which is responsible for managing 
{\em planning goals} during the solving process. Specifically, the resolver provides functionalities for managing  the 
{\em expansion} and/or {\em unification} of goals during plan refinement. {\em Goal expansion} refines the plan by 
creating and adding new {\em tokens} to timelines and applies the related {\em synchronization rules} that decompose
the goal into a subset of {\em sub-goals}. 
{\em Goal unification} refines the plan by "merging" goals with already existing and compatible tokens on the timelines. 
Thus, unification does not require goal decomposition and the consequent generation of sub-goals.

{\em StateVariables} represent the basic data structures modeling the features of a planning domain. \epsl\ complies with the
formalization described in Chapter \ref{chap:formalization}. Therefore, state variables encapsulate information concerning the 
values the related feature may assume over time, the allowed transitions, their flexible durations and controllability properties. 
There are three types of state variables available: 
(i) {\em ExternalStateVariable}; 
(ii) {\em FunctionalStateVariable}; 
(iii) {\em PrimitiveStateVariable}. 
{\em ExternalStateVariables} model features of the domain that are {\em completely uncontrollable}. They use the 
{\em ObservationChecker} resolver which is responsible for verifying the observations provided as input 
through the problem specification. The resolver verifies that the observations represent a {\em complete} and {\em valid} 
temporal behavior satisfying the domain constraints of the related external variable (\ie\ the value transition function).

{\em FunctionalStateVariables} and {\em PrimitiveStateVariables} are both planned variables. The former type of state variable 
models complex tasks/activities of the problem that must be further decomposed through {\em synchronization rules}. 
The latter type of state variable models tasks/activities that can be directly executed by the system. 
These two types of state variable use {\em StateVariableGapSolver} and {\em StateVariableScheduling} resolvers to build 
{\em timelines}. {\em StateVariableScheduling} resolvers are responsible for handling {\em scheduling threats} of the plan in 
order to avoid temporally overlapping tokens on the timelines.
{\em StateVariableGapSolver} resolvers are responsible for handling {\em gap threats} of the plan in order to avoid 
"empty" temporal intervals on the timelines. Finally, the {\em BehaviorChecker} resolver is responsible for 
checking the resulting temporal behaviors with respect to the value transition function of the related state variable specification.

\subsection{Problem Solving}
The {\em Deliberative} layer in Figure \ref{fig:epsl-archi} deals with problem resolution. It relies on the representation 
functionalities of the {\em Representation} layer and encapsulates the logic for solving timeline-based problems. In particular, 
it provides a set of ready-to-use search strategies, heuristics and solving algorithms that can be composed in order to define 
new planning instances. Indeed, the key point of \epsl\ flexibility is the interpretation of a planner as a ``modular'' solver 
which carries out the reasoning process by combining several elements.

\begin{figure}[ht]
\centering
\includegraphics[width=\textwidth]{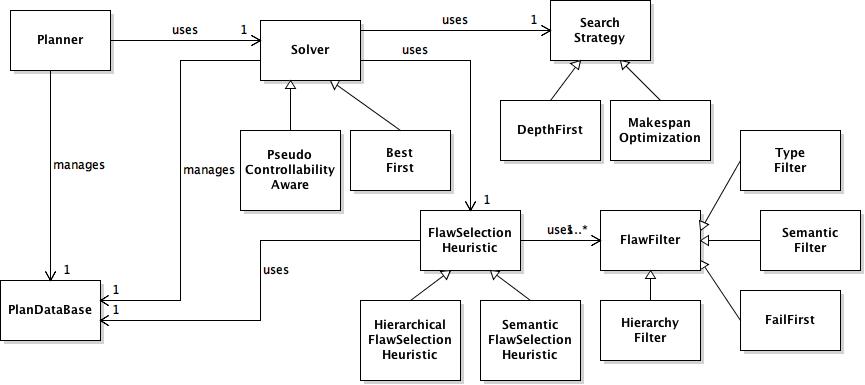}
\caption{\small{The structure of a planner in the \epsl\ framework architecture}}
\label{fig:epsl-planner}
\end{figure}

The {\em Solver} encapsulates the particular structure of the reasoning process. Broadly speaking, the reasoning process 
is a general plan refinement algorithm which is supported by a {\em Strategy} and {\em Heuristics} encapsulating some criteria to 
guide the search. The former provides criteria for managing the {\em fringe} of the search space and selecting the "best" 
(partial)plan to {\em expand} next. The latter encapsulates criteria for managing the {\em flaws} of the plan and selecting the 
{\em most promising} flaws to solve.
Figure \ref{fig:epsl-planner} shows the main architectural elements that compose an \epsl-based planner and their relationships.

Similarly to classical planning, the planning process can be summarized as the search of a solution plan with some desired 
features on a space of possible plans. The {\em search tree} is composed by a set of {\em search nodes} that encapsulates 
a particular plan representing a possible status of the system in terms of temporal behaviors. The {\em fringe} of the search 
tree consists of the subset of nodes not visited yet. 
The {\em SearchStrategy} is responsible for managing the fringe of the search space by encapsulating some particular criteria 
for assessing the nodes composing the fringe. The planning process finds the "best" solution according to the particular search 
strategy used. The \epsl\ framework provides the user with two search strategies. The {\em DepthFirst} strategy 
(DFS) realizes a {\em blind search} where the planning process is simply guided towards the last-generated nodes of 
the search (\ie\ the deepest ones in the search space). The {\em MakespanOptimization} strategy analyzes the temporal 
information of different pans in order to find the solution plan with the minimum makespan. Namely, this strategy tries to 
generates the most temporal efficient plan possible.

The {\em FlawSelectionHeuristic} encapsulates the logic for managing and assessing flaws of a plan detected during the 
refinement process. Specifically, a {\em FlawSelectionHeuristic} relies on a set of {\em FlawFilters} that select the "most 
relevant" flaws the planning process must consider for plan refinement. Each {\em FlawFilter} encapsulates a particular 
criterion for assessing flaws of a plan. The {\em TypeFilter} and the 
{\em FailFirstFilter} represent two straightforward selection criteria that take into 
account information concerning the particular flaws detected. The former filters flaws according to their type. The filter 
structures the planning process by requiring to address all {\em planning goals} first, than, {\em scheduling threats} and 
lastly, {\em gap threats}. The latter encapsulates the {\em fail-first principle} of constraint programming according to which the 
flaws with the minimum number of available solutions are selected first (\ie\ the {\em hardest flaws to solve}).
{\em HierarchyFilter} and {\em SemanticFilter} represents more complex selection criteria that will be explained in more detail 
in the next section. Broadly speaking, these two types of filters want to provide selection criteria that make decisions according 
to relationships and features that can be extracted by analzying the domain specification.

\subsection{The General Solving Procedure}
The {\em Solver} element of Figure \ref{fig:epsl-planner} manages the structure of the planning process by "coordinating" the 
particular strategy and the particular heuristics used to search solutions. In general, \epsl-based planners follow a general 
plan refinement search procedures where an initial plan is iteratively refined by solving flaws until a solution plan is found. 
There are two available implementations the \epsl\ framework provides. The {\em BestFirst} solver represents a simple 
implementation of the planning process, which returns the "best" solution found according to the particular search strategy 
adopted. 
The {\em PseudoControllabilityAware} extends the behavior of the "best-first" solver by adding the assessment of the 
{\em pseudo-controllability} property of the plan during the planning process. Specifically, the {\em PseudoControllabilityAware} 
solver returns a pseudo-controllable plan if possible, or the "best" non-pseudo-controllable plan (if pseudo-controllability cannot be satisfied). 
Before entering into the details of the motivations for pseudo-controllable plans and its implications with respect to the 
planning process, 
this section provides a detailed description of the {\em best-first} solving procedure 
of an \epsl-based planner.

\begin{algorithm}
\small
\caption{The \epsl\ general solving procedure}
\label{alg:epsl-solving}
\begin{algorithmic}[1]
\Function{solve}{$\mathcal{P}$, $\mathcal{S}$, $\mathcal{H}$}
	\State // initialize the plan database
	\State $\pi \leftarrow InitialPlan\left(\mathcal{P}\right)$
	\State // initialize the fringe
	\State $F \leftarrow \emptyset $
	\State // check if the current plan is complete and flaw-free
	\While{$\neg IsSolution\left(\pi\right)$}		
		\State // detect the flaws of the current plan
		\State $\Phi^0 = \{\phi_1, ..., \phi_k\} \leftarrow DetectFlaws\left(\pi\right)$
		\State // apply the heuristic to filter detected flaws
		\State $\Phi^{*} = \{\phi^{*}_1, ..., \phi^{*}_m\} \leftarrow SelectFlaws\left(\Phi^0, \mathcal{H}\right)$
		\State // compute possible plan refinements
		\For {$\phi^{*}_i \in\ \Phi^{*}$}
			\State // compute flaw's solutions
			\State $N_{\phi^{*}_i} = \{n_1, ..., n_t\} \leftarrow HandleFlaw\left(\phi^{*}_i, \pi\right)$
			\State // check if the current flaw can be solved
			\If {$N_{\phi^{*}_i} = \emptyset$}
				\State // unsolvable flaw found
				\State $Backtrack(\pi, Dequeue(F))$
			\EndIf
			\For {$n_j \in\ N_{\phi^{*}_i}$}
				\State // expand the search space with possible plan refinements
				\State $fringe \leftarrow\ Enqueue\left(n_j, \mathcal{S}\right)$
			\EndFor
		\EndFor
		\State // check the fringe of the search space
		\If {$IsEmpty\left(F_{\neg pc}\right)$}
			\State // search failure
			\Return $Failure$
		\EndIf
		\State // refine the plan
		\State $\pi \leftarrow\ Refine(\pi, Dequeue(F))$
	\EndWhile
	\State // get solution plan
	\State \Return $\pi$
\EndFunction
\end{algorithmic}
\end{algorithm}

Algorithm \ref{alg:epsl-solving} describes the abstract solving procedure of an \epsl-based planner. Basically, the reasoning process 
performs a {\em plan refinement procedure} which iteratively \emph{refines} the current plan $\pi$ by detecting and solving flaws.
\epsl\ instantiates the planner solving process over the tuple $\langle\mathcal{P}, \mathcal{S}, \mathcal{H} \rangle$ where $\mathcal{P}$ 
is the specification of a timeline-based problem to solve, $\mathcal{S}$ is the search strategy the planner uses to expand the search space,
and $\mathcal{H}$ is the heuristic function the planner uses to select the most promising flaw to be solved.

The plan database $\pi$ is initialized on the problem description $\mathcal{P}$ (row 2) and the procedure iteratively
refines the plan until a solution or a failure is detected (rows 6-32). At each iteration (rows 8-25) the procedure analyzes the current
plan database $\pi$ by detecting flaws that must be solved $\phi^{0}(\pi)$ (row 9). Then the set of detected flaws $\Phi^{0}$ is filtered
by applying the selected heuristic function $\mathcal{H}$ and the subset of {\em equivalent} flaws is extracted $\Phi^{*} \subseteq\
 \phi^{0}$ (row 11). Each flaw $\phi^{*}_{i} \in\ \Phi^{*}$ may have one or more solution $N_{\phi^{*}_{i}} = \{n_1, ..., n_t\}$ (row 15). 
If no solution is found for a particular  flaw $|N_{\phi^{*}_i}| = 0$ then the flaw is {\em unsolvable} and backtrack is needed (rows 
17-19). Otherwise each available solution represents a branch of the search and is added to the fringe (rows 21-24).
The iteration ends by selecting a node from the fringe and refining the plan $\pi$ accordingly (row 31). The search goes on until a plan 
with no flaws is found, \ie\ a solution plan (row 7).

\subsubsection*{Flaw Filtering}
In general, flaw selection is not a {\em backtracking point} of the search but it can strongly affect the performance of the 
planning process. From the search point of view, each solution of a flaw determines a branch of the search tree. 
Thus a "good" selection of the next flaw to solve can {\em prune} the search space by cutting off branches that would lead 
to {\em unnecessary} or {\em redundant} refinements of the plan. 
Considering a particular branch of the search tree, a {\em FlawSelectionHeueristic} determines an "ordering" in flaw 
resolution the planner must follow in order to reduce the {\em branching factor} and avoid an exhaustive expansion of the search tree.
The \epsl\ framework provides {\em FlawSelectionHeuristics} in shape of a {\em pipeline} of {\em FlawFilters} which is structured as follows:

\begin{center}
$
\phi^{0} ...
\xrightarrow{f_{i}(\pi, \phi^{i-1})} \phi^{i} 
\xrightarrow{f_{i+1}(\pi, \phi^{i})} \phi^{i+1} ...
\xrightarrow{f_{k}(\pi, \phi^{k-1})} \phi^{*} \subseteq\ \phi^{0}
$
\end{center}

The {\em pipeline} represents a quite {\em flexible} structure which can be easily adapted or extended by taking into account 
different types of filters and also different combinations of the same set of filters. Each specific configuration of the 
{\em pipeline} results in a different behavior of the solving process. Given a partial plan $\pi$ with an initial set of flaws $\phi^{0}$, 
a {\em sequence} of filters $f_i$, with $i=1,..., k$, is applied in order to extract the subset of relevant flaws to consider for 
plan refinement, $\phi^{*} \subseteq\ \phi^{0}$. 

The flaws composing the last set represent {\em equivalent choices} from the heuristic point of view, therefore they are all taken 
into account for plan refinement. The {\em amount of information} a particular heuristics function provides to the search can 
be estimated by checking the number of flaws actually filtered with respect to the initial set. If the set of flaws obtained by 
the application of a heuristic function $h(\pi)$ is equal to the initial set, \ie\ $\phi^{*} = \phi^{0}$, then it is possible to argue that 
the heuristic function $h(\pi)$ is {\em not informed}. Indeed, in such a case the heuristics does not provide any useful information 
to problem resolution, therefore the resulting behavior of the solving process is a {\em blind search}.

\section{Looking for Pseudo-controllable Plans}
As discussed previously, \epsl\ relies on the formal characterization of the {\em temporal uncertainty} and {\em controllability 
problem} with respect to timelines given in \cite{cialdea2015planning}. Therefore the framework can represent and reason 
about the {\em temporal uncertainty} of the planning domain  by taking into account the {\em uncontrollable values} and 
{\em external features} in order to generate plans with some desired properties concerning their {\em execution}.
In general, the execution of a plan is a complex and hard task which requires the system to actually interact with the 
environment.  There are many factors that may affect the executive process and even prevent the complete execution of the 
plan. The system must be able to handle the {\em observations} concerning the {\em uncontrollable dynamics of the 
environment} in order to dynamically {\em adapt} the plan as needed and complete the execution. 

Observations allow the system to check whether the execution is diverging from the expected plan or not. If the execution is 
diverging, then the system must manage the plan and react to exogenous events accordingly. In the best case observations 
comply with the expected plan and no change is needed. Sometimes instead, it may happen that the system must dynamically 
{\em adapt} the ongoing plan to the observations in order to proceed with the execution (\eg\ a delay of the execution of an
uncontrollable activity which propagates to the not executed portion of the plan). In the worst case, the observations and the 
plan are incompatible and {\em replanning} is required. It means that the execution is stopped in order to produce a new plan 
starting from the {\em current situation}. Once the deliberative process has generated the plan, the execution can start 
over again.

In this regard, a {\em robust executive system} must cope with the {\em uncertainty} of the environment and complete the 
process by adapting the plan to any {\em expected} exogenous event and perform replanning only if needed. 
There are many works in the literature that take into account {\em temporal uncertainty} and study the {\em controllability 
property} of a plan \cite{VidalF99,morris01,ker09,cialdeaorlandini2015,nilsson-controllability}.
Specifically three types of controllability properties have been defined: 
(i) weak controllability; 
(ii) strong controllability; 
(iii) dynamic controllability. 

{\em Dynamic controllability} is the most relevant property with respect to planning and execution in the real world. Broadly speaking, 
{\em dynamic controllability} concerns with the capability of an executive system to find a valid {\em execution strategy} which takes feasible 
{\em dispatching} decisions (\ie\ it schedules the start of plan's activities) by reasoning only on the {\em past history} and the received 
observations.
It is not easy to deal with these properties during the planning process. They are usually checked with a post-processing step after the 
planning phase and before starting the execution of the plan. With respect to planning, an interesting property worth to be considered, is 
the {\em pseudo-controllability} property. Indeed, the {\em pseudo-controllability} property of a plan represents a necessary 
but not sufficient condition for its {\em dynamic controllability} \cite{morris01}.

The pseudo-controllability property of a plan aims at verifying that the planning process does not make hypotheses on the 
actual duration of the related uncontrollable activities. Specifically, pseudo-controllability verifies that the planning 
process does not {\em reduce} the duration of uncontrollable values of the domain during plan generation. Thus, a timeline-based
plan is pseudo-controllable if all the flexible durations of uncontrollable tokens composing the timelines have not been changed 
with respect to the domain specification.
Although, pseudo-controllability does not convey enough information to assert the dynamic controllability of a plan, it represents
a useful property that can be exploited for {\em validating} the planning domain with respect to temporal uncertainty. Indeed, 
if the planner cannot generate pseudo-controllable plans, then the planner cannot generate dynamically controllable plans.

The {\em PseudoControllabilityAware} solver of the \epsl\ framework (see Figure \ref{fig:epsl-planner}) is responsible 
for generating pseudo-controllable plans (if possible).  In this way, \epsl\ realizes an planning framework capable of taking 
into account (part) of the controllability problem by generating pseudo-controllable plans that can be further investigated. 
Such an {\em integration} between planning and execution, envisages a unified framework which allows plan-based 
controllers to rely on a common representation of the problem. The deliberative and executive capabilities {\em share} 
the same formal representation of the plan enabling a more flexible and effective management of the control process.
This objective represents an ongoing development for the \epsl\ framework which has been partially addressed already as 
Chapters \ref{chap:hrc} shows.

\begin{algorithm}
\small
\caption{The \epsl\ pseudo-controllability aware solving procedure}
\label{alg:epsl-solving-pseudo}
\begin{algorithmic}[1]
\Function{solve\_pc}{$\mathcal{P}$, $\mathcal{S}$, $\mathcal{H}$}
	\State // initialize the plan database
	\State $\pi \leftarrow\ InitialPlan\left(\mathcal{P}\right)$
	\State // initialize "regular" and "non pseudo-controllable" fringe
	\State $F_{pc} \leftarrow\ \emptyset $
	\State $F_{\neg pc} \leftarrow\ \emptyset$
	\State // check if the current plan is complete and flaw-free
	\While{$\neg IsSolution\left(\pi\right)$}		
		\State // get uncontrollable values of the plan 
		\State $U = \{u_1,..., u_n \} \leftarrow\ GetUncertainty\left(\pi\right)$
		\State // check durations of uncontrollable values
		\If {$\neg Squeezed(U)$}
			\State // detect the flaws of the current plan
			\State $\Phi^0 = \{\phi_1, ..., \phi_k\} \leftarrow\ DetectFlaws\left(\pi\right)$
			\State // apply the heuristic to filter detected flaws
			\State $\Phi^{*} = \{\phi^{*}_1, ..., \phi^{*}_m\} \leftarrow\ SelectFlaws\left(\Phi^0, \mathcal{H}\right)$
			\State // compute possible plan refinements
			\For {$\phi^{*}_i \in\ \Phi^{*}$}
				\State // compute flaw's solutions
				\State $N_{\phi^{*}_i} = \{n_1, ..., n_t\} \leftarrow\ HandleFlaw\left(\phi^{*}_i, \pi\right)$
				\State // check if the current flaw can be solved
				\If {$N_{\phi^{*}_i} = \emptyset$}
					\State $Backtrack(\pi, Dequeue(F_{pc}))$	
				\EndIf
				\For {$n_j \in\ N_{\phi^{*}_i}$}
					\State // expand the search space
					\State $F_{pc} \leftarrow\ Enqueue\left(n_j, \mathcal{S}\right)$
				\EndFor
			\EndFor
		\Else
			\State // non pseudo-controllable plan 
			\State $F_{\neg pc} \leftarrow\ Enqueue\left(makeNode\left(\pi\right), \mathcal{S}\right)$
		\EndIf
		\State // check the fringe of the search space
		\If {$IsEmpty\left(F_{pc}\right) \land\ \neg IsEmpty\left(F_{\neg pc}\right)$}
			\State // try to find a non pseudo-controllable solution
			\State $\pi\ \leftarrow\ Refine\left(\pi, Dequeue\left(F_{\neg pc}\right)\right)$
		\ElsIf {$\neg IsEmpty\left(F_{pc}\right)$}
			\State // go on looking for a pseudo-controllable plan
			\State $\pi\ \leftarrow\ Refine\left(\pi, Dequeue\left(F_{pc}\right)\right)$
		\Else
			\State \Return $Failure$
		\EndIf
	\EndWhile
	\State // get solution plan
	\State \Return $\pi$
\EndFunction
\end{algorithmic}
\end{algorithm}

Algorithm \ref{alg:epsl-solving-pseudo} describes the "extended" solving procedure implemented by {\em PseudoControllabilityAware} 
solver. Similarly to Algorithm \ref{alg:epsl-solving}, the solving procedure is an iterative partial plan refinement which searches 
pseudo-controllable plans. If no pseudo-controllable plans can be found, the procedure tries to find a {\em non pseudo-controllable} 
plan before ending. Thus the procedure returns a failure if neither a pseudo-controllable plan nor a non pseudo-controllable plan have 
been found.
Similarly to Algorithm \ref{alg:epsl-solving}, the procedure is instantiated on the tuple $\langle\mathcal{P}, \mathcal{S}, \mathcal{H} 
\rangle$ whose elements represent the problem description, the search strategy and the flaw selection heuristic respetively.

The plan database $\pi$ is initialized on the problem description $\mathcal{P}$ (row 3). The procedure manages two distinct fringes 
during the search (initialized at row 5 and row 6). The {\em pseudo-controllable fringe} $F_{pc}$ is the fringe used when searching 
for pseudo-controllable plans. The {\em non pseudo-controllable fringe} $F_{\neg pc}$ is the fringe used when no pseudo-controllable
plans have been found by the search. 
The solving procedure iteratively refines the plan until a solution is found (rows 8-47). At each iteration, the solving process checks 
the current plan for pseudo-controllability by analyzing the temporal uncertainty features of the domain (rows 10-12), \ie\ uncontrollable 
and external values. If the flexible duration of uncontrollable values has not been changed (row 12) then the procedure starts 
refining the current plan as in the {\em regular procedure} described previously in Algorithm \ref{alg:epsl-solving}. If the pseudo-controllability
condition does not hold (row 32) then the current plan is placed into the {\em non pseudo-controllable fringe} $F_{\neg pc}$, the procedure
skips the refinement of the plan and the search goes on by extracting the next plan from the fringe (rows 38-40). However, if the 
fringe is empty then no pseudo-controllable plans can be found. Consequently, the search goes on by extracting nodes from the 
{\em non pseudo-controllable fringe} $F_{\neg pc}$ (rows 35-37). In such a case, it means that if a solution exists, then the 
plan is not pseudo-controllable and therefore the plan cannot be {\em dynamically controllable}. 
If both the fringes are empty then the procedure returns a failure (row 42), otherwise the search will end with a solution plan which can 
be either pseudo-controllable or not.

\section{Hierarchical Planning with Timelines}
In general, the design of effective models capable of capturing all the relevant features of a particular system is an issue in plan-based 
controller development. Indeed, a model must capture all (and only) the information about the system and the working environment that are 
really relevant with respect to the objectives of a particular application. Given a particular planning technique, it is not easy to find 
the appropriate abstraction level and structure the model accordingly.
Typically, structured models support the solving process by {\em encoding} domain-specific knowledge about the problem. In this regard, 
hierarchical approaches like \htn\ have been successfully applied especially in real-world scenarios. This section introduces a hierarchy-based 
methodology for the design of timeline-based applications which takes inspiration from \htn\ approaches. In addition, a domain independent 
heuristics capable of leveraging the resulting structure of planning domains is presented.

\subsection{Hierarchical Modeling Approach}
In plan-based control systems the focus is usually on controlling an autonomous agent in order to perform complex tasks in a 
specific working environment (\eg\ an industrial robot in a manufacturing work-cell). An effective timeline-based model must 
capture all the features and the related operational constraints that are relevant with respect to the {\em control problem}. 
The timeline-based model must describe the available capabilities that allow the system to actually interact with the environment 
and perform operations (\eg\ actuators, sensors, tools), the features of the environment that may affect the behavior of the system, 
as well as the resulting high-level tasks (\ie\ complex activities) the system can perform. Thus, it is possible to organize 
the information a timeline-based model must capture in three different levels  of abstraction: 

\begin{itemize}
\item The {\em functional level} concerns the high level tasks the agent can perform. It characterizes the high-level goals the 
timeline-based system can plan for.
\item The {\em primitive level} concerns the {\em internal} elements that compose the agent. It characterizes the capabilities of the 
system in terms of the low-level tasks, or commands the agent's components can directly execute. Namely, the primitive level deals 
with the representation of devices and facilities the system is endowed with (\eg\ actuators or sensors) that can be actually used to 
solve a problem.
\item The {\em external level} concerns the environment the agent must care about. It is orthogonal with respect to other levels 
and characterizes the dynamics of the environment the agent must interact with. Namely external level concerns the element of 
the domain that are outside the control of the agent but whose behavior may affect the outcomes of the activities needed to solve 
a problem.
\end{itemize}

\begin{figure}
\centering
\includegraphics[width=.9\textwidth]{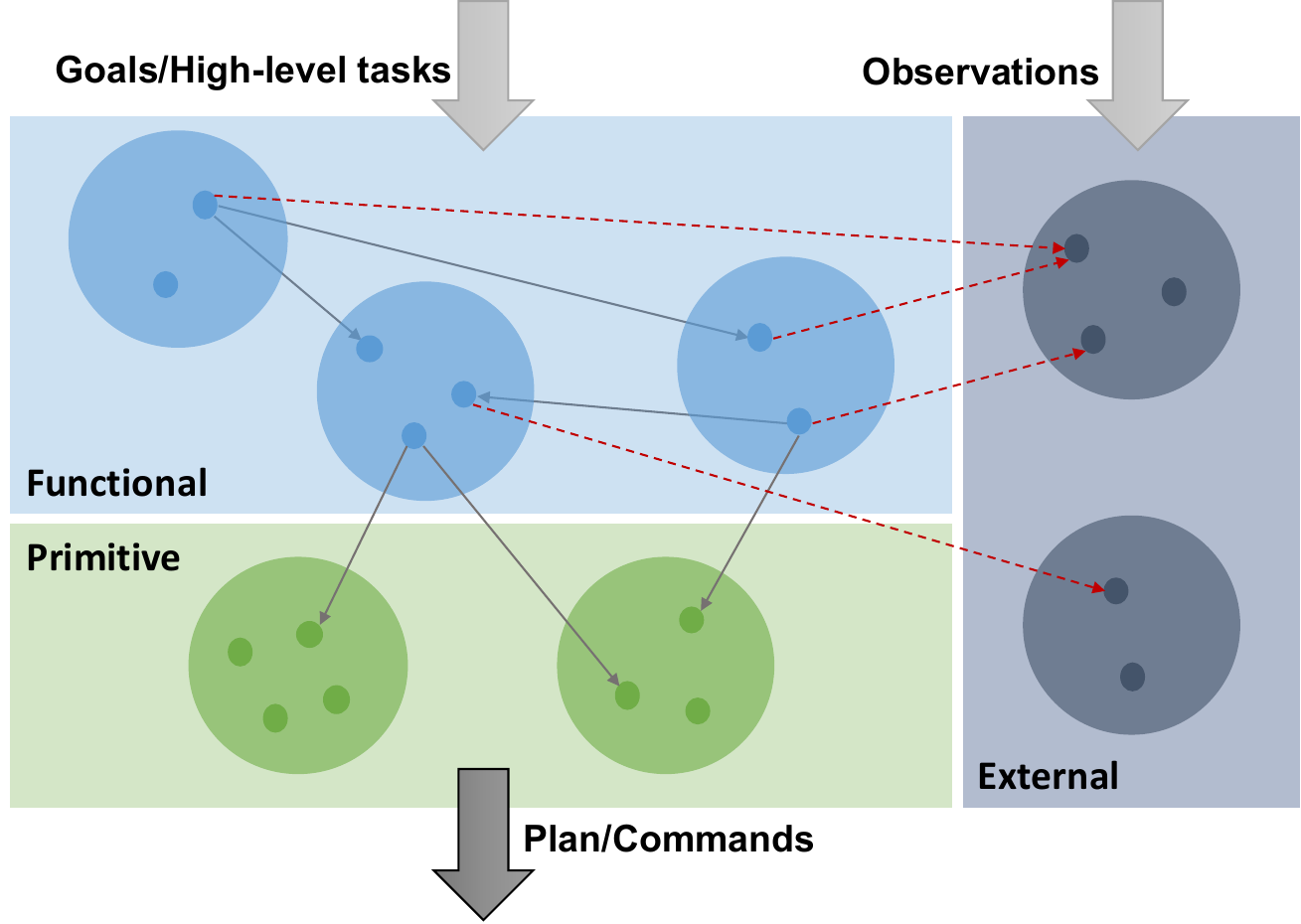}
\caption{\small{Hierarchical modeling of timeline-based domains}}
\label{fig:hierarchical-modeling}
\end{figure}

According to this organization, Figure \ref{fig:hierarchical-modeling} shows the general structure of a timeline-based 
domain. The functional and primitive levels are directly related each other. The external level instead is orthogonal to the 
others as it may have implications at both levels. Within this structure, the model must specify a {\em hierarchical 
decomposition} of high-level tasks in terms of {\em relations} between low-level tasks. Such relationships may require several 
decomposition levels according to the complexity of the considered domain. In any case, the hierarchical decomposition starts with 
a high-level task representing a planning goal, and ends with a set of primitive tasks that can be directly executed by the system.
The arrows in Figure \ref{fig:hierarchical-modeling} represent such a decomposition. Some arrows (the red dotted ones) specify 
relations between functional (or even primitive) values and external values.
In these cases, the arrows represent condition checking rather then decomposition. Namely they represent conditions that must 
hold in the plan with respect to some features of the environment, rather then activities to perform.
Following the three abstraction levels, the envisaged hierarchical modeling approach identifies three types of state variables
composing a timeline-based domain. They are the (i) the {\em functional variables}, (ii) the {\em primitive variables} and (iii) the {\em external 
variables}.

{\em Functional variables} provide a logical representation of the agent in terms of the high-level task the agent can perform, 
notwithstanding its internal composition. The values of this type of variables are {\em controllable} and represents the 
high-level planning goals the related timeline-based system can plan for. 
With regards to the \rover\ domain, the {\em Rover} planned state variable is a functional variable which models the high-level 
tasks the rover can perform. Specifically, it models the rover in terms of the {\em TakeSample} activities that represent the 
planning goals of the problem.

{\em Primitive variables} model a specific physical/logical component of the system. The values of this type of variables 
represent state/actions the related component of the system is actually able to assume/perform over time. These values may 
have bounded flexible durations and may be either controllable or not. If such a value is tagged as {\em uncontrollable}, it means 
that the system cannot decide the actual duration of the related activity during execution (specifically the system can decide 
when to start the activity but not when the activity ends). 
These are the values the planning process looks for in order to check if the pseudo-controllability property of the plan
is satisfied. For example, the {\em SendData} value of {\em Communication} variable is tagged as {\em uncontrollable} because 
the actual duration of the communication activities is affected by factors that are not under the control of the 
rover (\eg\ the size of the data file, or the quality of the communication channel).

{\em External variables} model the features of the environment that are {\em completely uncontrollable} and whose 
behaviors may affect the operations of the system. Namely such variables model conditions that must hold in order to 
successfully carry out the tasks of the system. These variables can only be {\em observed} and the related timelines are 
included into the description of the problem. Therefore, the planning system must adapt the plan to the particular 
observations received (again, without making any hypothesis on their actual durations in order to comply with pseudo-controllability 
property). With regards to the \rover\ planning domain, the generated plan must comply with the observations concerning
the communication channel such that the {\em SendData} activity is performed when the channel is supposed to be available.

Task decomposition is realized by means of {\em synchronization rules} that, like methods in \htn, connect adjacent 
abstraction levels of the domain by specifying relationships between different tasks. Namely, synchronization rules describe 
a top-bottom task decomposition specifying how the high-level tasks (\ie\ functional values) are implemented by the internal 
components of the system (\ie\ the primitive values) and how their execution is related to the environment (\ie\ external values).

\subsection{Building the Dependency Graph}
It is possible to observe that a {\em synchronization rule} basically represents a {\em dependency} between two or more 
variables of the domain. It means that variables affect the temporal behavior of other variables through synchronization
rules and the related temporal constraints between tokens. 
Let us consider a synchronization rule $S_{v_{A,i}}$ which applies to value $i$ of a state variable $A$ ($v_{A,i}$) and contains
a temporal constraint between $v_{A,i}$ and a value $j$ of a state variable $B$ ($v_{B,j}$). Such a temporal constraint
affects the behavior of state variable $B$ and consequently the building process of the related timeline (\ie\ the timeline of 
state variable $B$). Thus, the synchronization rule $S_{v_{A,i}}$ determines a dependency between state variable $A$ and 
state variable $B$. Namely, the tokens that compose timeline $A$ and their {\em temporal allocation} affect the token 
that compose timeline $B$ and their {\em temporal allocation}. According to this observation, it is possible to analyze 
the synchronization rules of a domain specification in order to build a {\em Dependency Graph} (DG) encoding the 
relationships between domain components. Specifically, a DG is a directed graph which provides a {\em relaxed} 
representation of the dependencies between the components of a planning domain. 

\begin{definition}
A {\em dependency graph} DG is a directed acyclic graph defined by the pair $\langle V, R_{d} \rangle$ 
where: (i) $V$ is the set of nodes of the graph representing the components of the planning domain;
(ii) $R_{d}$ is the set of (directed) edges between nodes representing dependency relationships 
between domain components
\end{definition}

\begin{algorithm}
\small
\caption{The Dependency Graph building procedure}
\label{alg:build-dg}
\begin{algorithmic}[1]
\Function{build\_dependency\_graph}{$\Pi$}
	\State // Initialize the DG with the state variables of the domain
	\State $G_{dg} \leftarrow Create\left(GetStateVariables\left(\Pi\right)\right)$
	\State // get synchronization rules
	\State $S = \{..., S_{v_{M,n}}, ...\} \leftarrow GetSynchronizationRules\left(\Pi\right)$
	\For {$S_{v_{A,i}} \in S$}
		\State // check synchronization's constraints
		\For {$r_{k} \in GetConstraints\left(S_{v_{A,i}}\right)$}
			\State // check if reflexive relation
			\If {$\neg IsReflexive\left(r_k\right)$}
				\State // get reference domain component
				\State $C_{s} \leftarrow Reference\left(r_{k}\right)$ 
				\State // get target domain component
				\State $C_{t} \leftarrow Target\left(r_{k}\right)$
				\State // add dependency to the graph
				\State $G_{dg} \leftarrow AddDependency\left(C_{s}, C_{t}\right)$
				\State // look for cycles in the graph
				\If {$HasCycle\left(G_{dg}\right)$}
					\State // remove last added dependency
					\State $G_{dg} \leftarrow RemoveDependency\left(C_{s}, C_{t}\right)$
				\EndIf
			\EndIf
		\EndFor
	\EndFor
	\State // return the computed DG
	\State \Return $G_{dg}$
\EndFunction
\end{algorithmic}
\end{algorithm}

Algorithm \ref{alg:build-dg} describes the building procedure of the DG. The procedure takes as input the planning 
domain and initializes the dependency graph $G_{dg}$ on the set of state variables of the domain (row 3). 
The dependencies between components are generated by analyzing  the synchronization rules of the domain 
(rows 6-24). For each synchronization rule $S_{v_{A,i}}$, where variable $v_{A,i}$ represents the triggerer of the 
synchronization ($A$ the component the value belongs to, $i$ the id of the value), the related temporal 
constraints are taken into account to compute dependencies (rows 8-21). For each temporal constraint
$r_{k}$ the algorithm checks if the relation is reflexive (row 10). If a temporal constraint involves 
two values belonging to the same component (\ie\ it represents a reflexive dependency relation) then the 
constraint is ignored. Otherwise a new dependency is added to the graph concerning the reference and 
the target components of the relations (rows 11-16). 
Every time a new edge is added, the graph is checked for cycles (row 18). If a cycle is detected, it is 
caused by the last added dependency which is discarded and removed from the graph (row 20).
The procedure continues until all synchronization's temporal constraints have been analyzed and the 
resulting dependency graph $G_{dg}$ is returned.

The DG the procedure generates, is acyclic for construction. Indeed, the procedure discards edges 
(\ie\ temporal relations) that introduce cyclic dependencies into the graph. Thus, the DG relaxes the dependency 
relationships of the domain by considering only an acyclic subset of them. A DG is said to be {\em complete} if all 
the dependencies of the domain are modeled. Often, given the hierarchical modeling approach described in the 
previous section, the dependencies of the domain are acyclic. The resulting DG is {\em complete} and encodes the 
{\em hierarchy} of the planning domain which can be easily extracted by analyzing the graph (\eg\ by means of a 
topological sort algorithm if the DG does not contain cycles).

\begin{figure}[ht]
\centering
\begin{minipage}{.65\textwidth}
	\includegraphics[width=.9\textwidth]{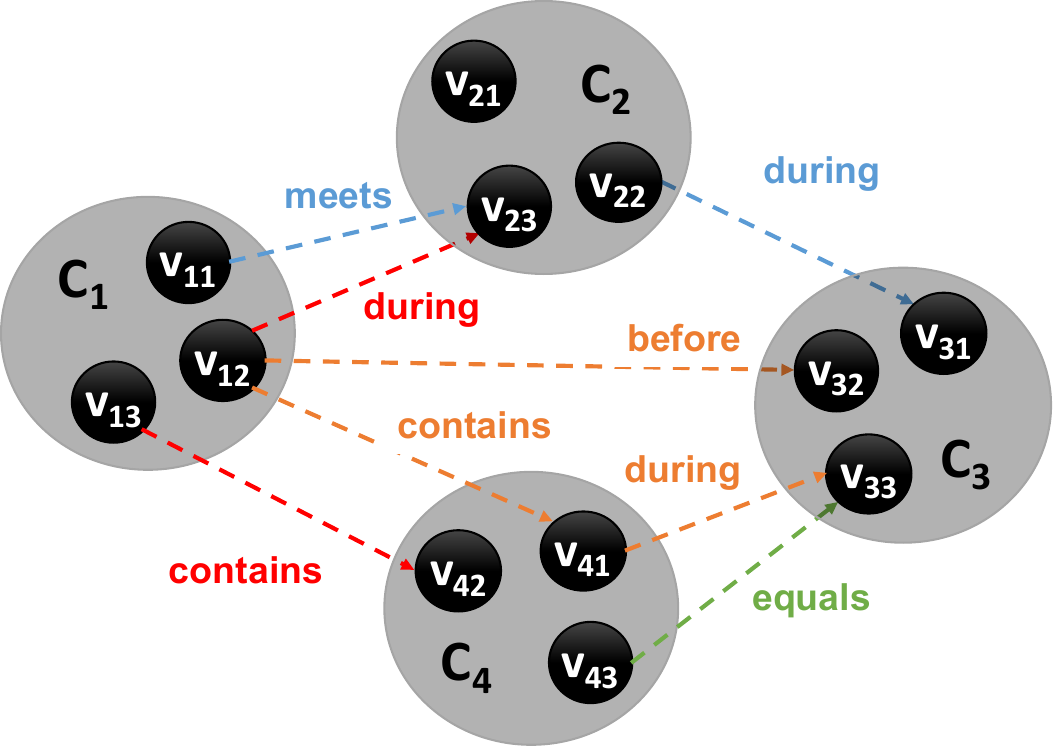}
\end{minipage}
\begin{minipage}{.3\textwidth}
	\includegraphics[width=\textwidth]{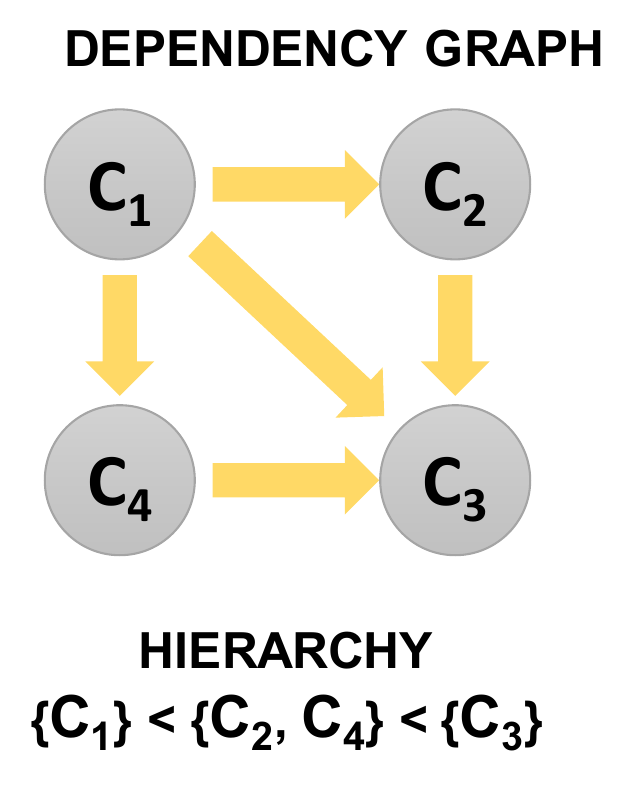}
\end{minipage}
\caption{\small{Extracting domain hierarchy from synchronizations' constraints}}
\label{fig:dg2hierarchy}
\end{figure}

Figure \ref{fig:dg2hierarchy} shows a general example concerning the hierarchy extraction process from domain 
synchronizations. Each synchronization rule specifies temporal constraints that may involve values of different 
state variables. Temporal constraints of different synchronizations can be distinguished according to their color 
in Figure \ref{fig:dg2hierarchy} (\ie\ temporal constraints with the same color belong to the same synchronization 
rule). Each temporal constraint between values of different components $v_{A,i} \in\ C_{A}$ and $v_{B,j} \in\ C_{B}$ 
determines a dependency relation between $C_{A}$ (the source of the relation) and $C_{B}$ (the target of the relation). 
Thus such a constraint is encoded by an edge $r_{A,B} \in R_{d}$ of the DG, where $A, B \in V$. 

Given the DG, it is possible to extract the {\em hierarchy} of the domain as shown in Figure \ref{fig:dg2hierarchy}. Indeed, 
an edge connecting a node $A$ to a node $B$ in the DG, implies that component $B$  depends on component $A$. Thus
the component $A$ is {\em higher} than component $B$ with respect to the hierarchy. Otherwise, If there is not a direct 
path connecting a node $A$ to a node $B$ in the DG, then no implications can be made concerning their relationship. 
In such a case, the related domain components are supposed to be at the same level of the 
hierarchy. Such a hierarchy can guide the solving procedure to search for solutions. In this regard, the {\em HierarchyFilter} 
element of Figure \ref{fig:epsl-planner} encapsulates a flaw selection criterion which selects flaws according 
to the hierarchical level of the component they belong to. The rationale of the selection criteria is that, given a set of 
flaws to solve, the "best" choice is to start solving flaws that belong to the {\em most independent 
component} of the domain. Solving {\em dominant} flaws of the plan may implicitly solve other 
secondary flaws of the plan or even {\em prune} the search space by removing redundant or unfeasible
flaw solutions.


%
%


\graphicspath{{chapters/7_hrc/figures/}}


%
%
%
%
%
\chapter{Planning and Execution with Timelines under Uncertainty}
\label{chap:hrc}
\lettrine[lines=2]{P}{lan generation} is only a part of the problem when controlling a complex system with plan-based technologies 
in \ai. The execution of a plan is a complex process which can fail even if the plan is valid with respect to the domain 
specification. During execution, the system must {\em interact} with the environment, which is {\em uncontrollable} and therefore 
the execution of the activities can be affected by external factors. A {\em robust} executive system must cope with such exogenous 
events and dynamically {\em adapt} the plan accordingly during execution. 
In order to deploy timeline-based applications in real-world scenarios, the \epsl\ planning framework has been extended by 
introducing {\em executive capabilities}. The executive relies on the same semantics of timelines the planning process 
relies on. Thus, the executive leverages information about the {\em temporal uncertainty} of the problem in order to 
properly manage the execution of the plan. In this way, \epsl\ realizes a uniform software framework for planning and execution 
with timelines under uncertainty.
This chapter provides a detailed description of the extended \epsl\ framework and the related approach to execution. Moreover, 
the chapter introduces a real-world manufacturing scenario for Human-Robot Collaboration (\hrc) where \epsl\ and the related 
planning and execution capabilities have been successfully applied.

\section{Model-based Control Architectures}
The classical approach for building model-based controllers relies on the three-layered architecture described in \cite{gat97}. 
These three layers are (starting from the bottom) the {\em functional layer}, the  {\em executive layer} and the {\em planning/scheduling
layer}. Traditional autonomous control architecture follow this structure and the most relevant works concern: 
\ipem\ \cite{ipem1988}, \cpef\ \cite{Myers99}, the \textsc{LAAS} architecture \cite{laas-archi} which relies on the \ixtetexec\ 
\cite{ixtex-exec}, the Remote Agent Experiment \cite{rax-ps-00} and \ase\ \cite{ASE}. Each layer usually requires different 
reasoning and representation technologies. The integration of such different technologies is usually an issue for developing 
this type of controllers. Often, the planning cycle is monolithic making scalability and fast reaction time another issue of this 
type of controllers.

Other approaches like \textsc{CLARAty} \cite{claraty2008}, try to overcome some of these drawbacks using an architecture 
with only two layers. A {\em functional layer} and a {\em decision layer}. The decision layer integrates planning and execution 
through a shared data structure (\ie\ the {\em plan database}) synchronizing planning and execution data that rely on two 
different representations, \ie\ \textsc{CASPER} \cite{Knight01} for planning and \textsc{TDL} \cite{Simmons98} for 
execution. \textsc{CIRCA} \cite{circa-2002} proposes an intelligent controller in hard real-time which leverages reactive 
planning to implement automatic controller synthesis.

\idea\ \cite{idea,bernardini2006modelunified} was the first agent control architecture utilizing a collection of controllers, 
each interleaving planning and 
execution in a common framework. The main drawbacks of \idea\ are the lack of a clear conflict-resolving policy 
between controllers and the lack of an efficient planning algorithm for integrating the current states of controllers. 
The Teleo-Reactive Executive (\trex) \cite{trex} was designed to overcome these restrictions using a collection of 
controllers (called {\em reactors}) implemented as different instances of \europa\ planner \cite{europa2012}. 
The novelty of \trex\ was the capability of realizing a systematic infrastructure which defines the interactions among 
reactors.

\section{Extending the \epsl\ framework with Execution}
The executive system is responsible for managing the execution of timeline-based 
plans by iteratively sending commands to the system and receiving observations 
concerning the actual state of the {\em environment}. Thus, the executive must {\em verify} whether the 
perceived behavior of the world (\ie\ the system and the environment) complies with the expected plan and 
must {\em react} accordingly in case of {\em conflicts}.

\begin{figure}[ht]
\centering
\includegraphics[width=\textwidth]{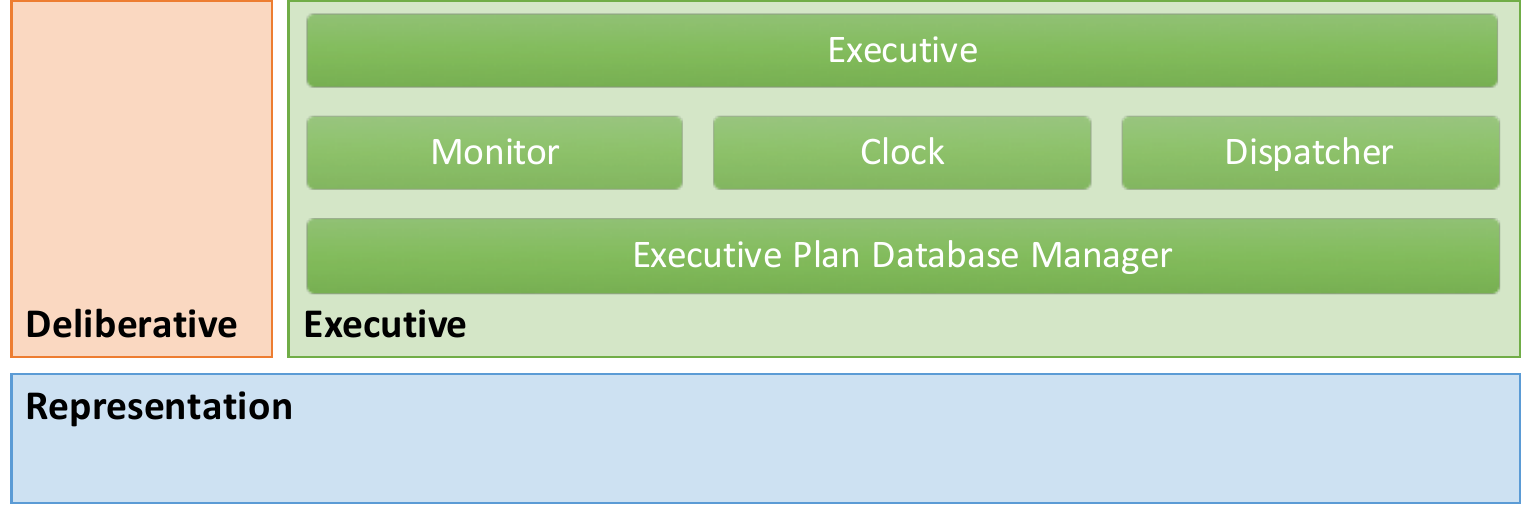}
\caption{\small{The \epsl\ architecture extended with executive capabilities}}
\label{fig:epsl-exec-layers}
\end{figure}

Figure \ref{fig:epsl-exec-layers} shows the extended architecture of the \epsl\ framework with the main 
architectural elements. The executive relies on the same representation functionalities the deliberative 
relies on. Therefore, the system can manage the execution of the activities of the plan according to their
{\em controllability} properties. The obtained \epsl\ framework represents a unified tool 
capable of seamlessly dealing with planning and execution of timelines with uncertainty. 
Plan execution manages particular information representing states and conditions that must be 
monitored during the execution of timelines. The {\em Executive Plan Database Manager} encapsulates
this kind of information by extending the functionalities of the {\em Representation layer}.  Specifically, it
manages information concerning the {\em execution state} of plan's tokens and the related {\em execution
dependencies}. The temporal constraints of a timeline-based plan entail dependencies determining whether 
a token can actually start/end execution or not. Given a timeline-based plan to execute, the {\em Executive Plan 
Database Manager} extracts execution dependencies dynamically and encodes this information into a dedicated
data structure called {\em Execution Dependency Graph} (\edg). 

\begin{figure}[ht]
\centering
\includegraphics[width=\textwidth]{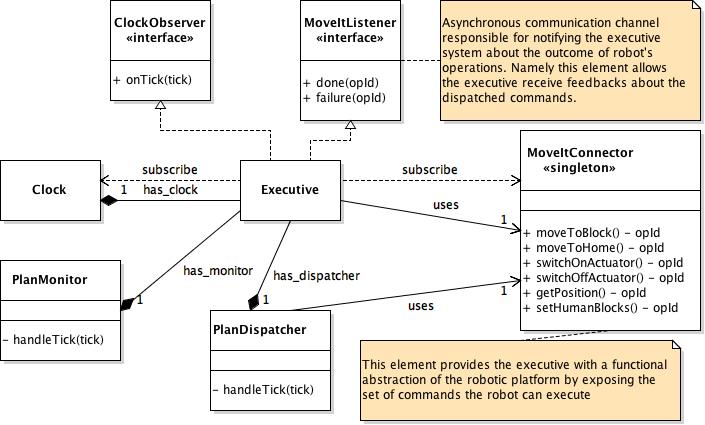}
\caption{\small{The structure of the executive in the \epsl\ framework}}
\label{fig:epsl-exec}
\end{figure}

Figure \ref{fig:epsl-exec} shows the elements composing a general \epsl\ executive and their relationships. In 
particular, the figure shows the additional elements the executive needs to exchange information/signals with 
the specific environment and robotics platform. The {\em MoveItConnector} and the {\em MoveItListener}
represent two elements used within the research project \fbt\ described in Section \ref{sec:fbt}. They encapsulate the 
complexity of the {\em remote communication} with the robot and provide the \epsl\ executive with a set of {\em local
execution services} (see the {\em Proxy} design pattern \cite{gof}). In particular, the {\em MoveItConnector} encapsulates
the set of low-level commands the robot can execute according to the {\em operational interface} of system. 
The {\em MoveItListener} instead, allows the executive to receive asynchronous messages from the system as well as the 
environment concerning the results of execution requests (i.e. feedbacks).

\subsection{The Execution Process}
\label{sec:exec-proc}
The execution process consists of {\em control cycles} whose frequency determines the {\em reactivity} 
of the executive and the advancement of time. Given the temporal horizon of the plan, the execution process 
discretizes the {\em temporal axis} by means of a number of temporal units, called {\em ticks}, according the needed 
frequency. Each {\em control cycle} of the process is associated with a {\em tick} and realizes the execution procedure.
Broadly speaking, the execution procedure is responsible for detecting the actual behavior of the  system ({\em closed-loop} 
architecture), for verifying if the system and also the environment behave as expected from the plan and for for starting the 
execution of the activities of the plan. The procedure is composed by two distinct phases, the {\em synchronization 
phase} and the {\em dispatching phase}. At each tick (\ie\ control cycle) the synchronization phase manages the received 
execution feedbacks/signals in order to build the current status of the system and the environment. 
If the current status is valid with respect to the plan, then the dispatching phase {\em decides} the next activities to 
be executed. Otherwise, if the current status does not fit the plan, an {\em execution failure} is detected and 
{\em replanning} is needed. Indeed, the current plan does not represent the actual status of the system and the environment 
and therefore replanning allows the executive to continue the execution process with a new plan, which has been generated 
according to the {\em observed} status and the executed part of the {\em original plan}.

\begin{algorithm}
\small
\caption{The \epsl\ executive control procedure}
\label{alg:exec}
\begin{algorithmic}[1]
\Function{execute}{$\Pi$, $\mathcal{C}$}
	\State // initialize executive plan database
	\State $\pi_{exec} \leftarrow\ Setup\left(\Pi\right)$ 
	\State // check if execution is complete
	\While{$\neg CanEndExecution\left(\pi_{exec}\right)$}
		\State // wait a clock's signal
		\State $\tau\ \leftarrow\ WaitTick\left(\mathcal{C}\right)$
		\State // handle synchronization phase
		\State $Synchronize\left(\tau, \pi_{exec}\right)$
		\State // handle dispatching phase
		\State $Dispatch\left(\tau, \pi_{exec}\right)$
	\EndWhile
\EndFunction
\end{algorithmic}
\end{algorithm}

Algorithm \ref{alg:exec} describes the general control procedure of the executive and its related 
sub-procedures. The procedure takes as input the plan $\Pi$ to be executed and the clock $\mathcal{C}$
which determines the frequency of the control cycles. 
First of all, the procedure analyzes the plan $\Pi$ in order to identify {\em execution dependencies} among
tokens of the timelines. This information is encapsulated by a dedicated structure $\pi_{exec}$ (row 3) the 
procedure uses during execution. The procedure iteratively executes the plan until all the tokens have been 
executed (rows 5-12). The timing  of the iterations of the procedure is determined by the clock $\mathcal{C}$. 
Indeed, the procedure waits a signal from $\mathcal{C}$ which communicates the current execution time 
$\tau$ (row 7). Then, the procedure checks the status of the execution by calling the 
{\em Synchronize} sub-procedure (row 9) and ends the execution cycle by calling the {\em Dispatch} 
sub-procedure (row 11).

\begin{figure}[ht]
\centering
\includegraphics[width=\textwidth]{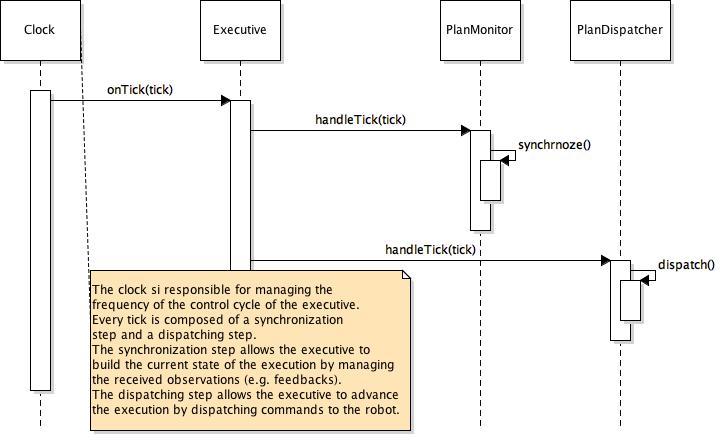}
\caption{\small{The structure and interactions of the executive control cycle}}
\label{fig:epsl-exec-cycle}
\end{figure}

Figure \ref{fig:epsl-exec-cycle} shows the runtime behavior of an \epsl\ executive and its interactions with {\em Clock},
{\em PlanMonitor} and {\em PlanDispatcher} elements shown in Figure \ref{fig:epsl-exec}. The Executive manages the 
structure of the control cycle by coordinating the synchronization and dispatching steps. 
The {\em Clock} iteratively generates control events by sending {\em onTick(tick)} signals to the Executive, according to
the desired frequency. The higher is the frequency of the clock the higher is the reactiveness of the control process. Clearly,
the clock's frequency must be compatible with the {\em response time} of the system.
The {\em PlanDispatcher} and  the {\em PlanMonitor} are the elements responsible for managing respectively the dispatching 
and synchronization steps within the control cycle. Thus, as Figure \ref{fig:epsl-exec-cycle} shows, every time the 
Executive receives a signal from the Clock, it coordinates the PlanMonitor and the PlanDispatcher in order to complete the 
control cycle. Specifically, the Executive calls the PlanMonitor to handle the synchronization step and build the current 
pereceived state of the system and the environment (i.e. the current situation). Then, the Executive calls the PlanDispatcher
to handle the dispatching step according to the current situation and the current execution time.

\subsubsection*{The Synchronization Phase}
\begin{figure}
\centering
\includegraphics[width=\textwidth]{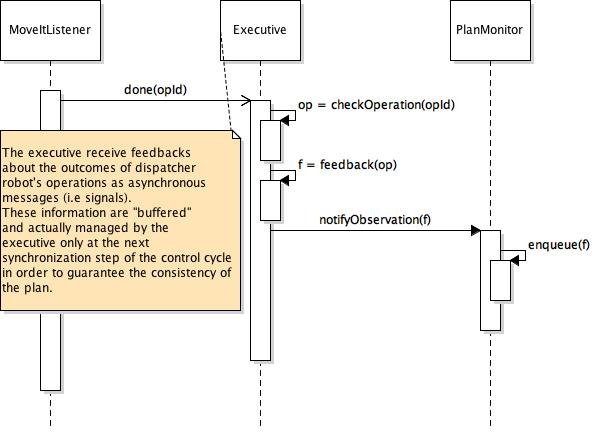}
\caption{\small{Management of the received feedback signals during the control cycle}}
\label{fig:epsl-exec-signals}
\end{figure}

The synchronization phase monitors the execution of the plan by determining if some divergencies occur between the 
expected plan and the observed behavior of the system and the environment. Namely, at each iteration the 
synchronization phase builds the current situation by taking into account the current execution time, the expected plan and 
the feedbacks received during execution. 
Figure \ref{fig:epsl-exec-signals} shows the elements involved within the synchronization phase and their 
interactions. The {\em PlanMonitor} is responsible for propagating observations concerning the actual duration of 
the dispatched activities and detecting discrepancies between the real-world and the plan. The Executive receives 
feedbacks about the successful execution of dispatched commands or failure. The PlanMonitor manages these 
feedbacks in order to detect if the actual duration of tokens comply with the plan. If the feedbacks
comply with the plan then, the status of the related tokens can change from {\em in-execution} to {\em executed}.
Otherwise, an inconsistency is detected (i.e. the current situation does not fit the expected plan) and a {\em failure} 
is notified to the Executive which must react accordingly (e.g. by re-planning).

\begin{algorithm}
\small
\caption{The \epsl\ executive procedure for the synchronization phase}
\label{alg:exec-synchronize}
\begin{algorithmic}[1]
\Function{synchronize}{$\tau$, $\pi_{exec}$}
	 \State // manage observations
	 \State $\mathcal{O} = \{o_1, ..., o_n\} \leftarrow\ GetObservations\left(\pi_{exec}\right)$
	 \For {$o_i \in\ \mathcal{O}$}
	 	\State // propagate the observed end time
	 	\State $\pi_{exec} \leftarrow\ PropagateObservation\left(\tau, o_i\right)$
	 \EndFor
	 \State // check if observations are consistent with the current plan
	 \If {$\neg IsConsistent\left(\pi_{exec}\right)$}
	 	\State // execution failure
	 	\State \Return $Failure$
	 \EndIf
	 \State // manage controllable activities 
	 \State $\mathcal{A} = \{a_i, ..., a_m\} \leftarrow\ GetControllableActivities\left(\pi_{exec}\right)$
	 \For {$a_i \in \mathcal{A}$}
	 	\State // check if activity can end execution
	 	\If {$CanEndExecution\left(\tau, a_i, \pi_{exec}\right)$}
			\State // propagate the decided end time 
		 	\State $\pi_{exec} \leftarrow\ PropagateEndActivity\left(\tau, a_i\right)$
		\EndIf
	 \EndFor
\EndFunction
\end{algorithmic}
\end{algorithm}

\subsubsection*{The Dispatching Phase}
\begin{figure}
\centering
\includegraphics[width=\textwidth]{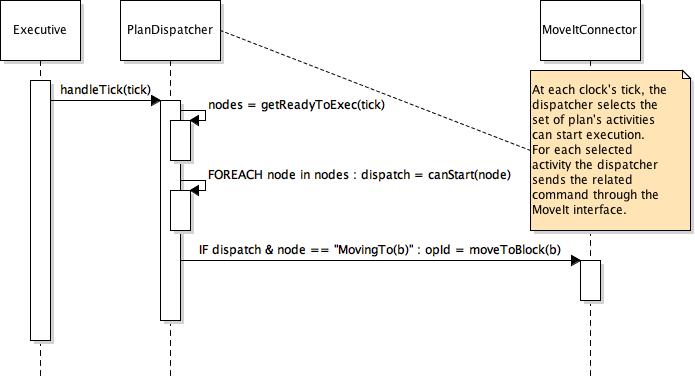}
\caption{\small{Management of the dispatching step during the control cycle}}
\label{fig:epsl-exec-dispatching}
\end{figure}

The dispatching phase manages the actual execution of the plan. Given the current situation and the current 
execution time, the dispatching step analyzes the plan $\pi_{exec}$ in order to find the tokens that can start 
execution and dispatches the related commands to the underlying system. 
Namely, the dispatching step allows the Executive to advance execution by deciding the next tokens to execute. 
Figure \ref{fig:epsl-exec-dispatching} shows the elements involved within the dispatching steps and their 
interactions. The {\em PlanDispatcher} is responsible for making dispatching decisions of plan's tokens. For each 
token, the PlanDispatcher checks the related {\em start condition} by analyzing the token's scheduled time and 
any dependency with other tokens of the plan. If the start condition holds, then the PlanDispathcher can decide to  
start executing the token (\ie\ the dispatcher propagates the scheduled start time into the plan). 
After dispatching, the status of the involved token changes from {\em waiting} to {\em in-execution}.

\begin{algorithm}
\small
\caption{The \epsl\ executive procedure for the dispatching phase}
\label{alg:exec-dispatch}
\begin{algorithmic}[1]
\Function{dispatch}{$\tau$, $\pi_{exec}$}
	 \State // manage the start of (all) plan's activities
	 \State $\mathcal{A} = \{a_i, ..., a_m\} \leftarrow\ GetActivities\left(\pi_{exec}\right)$
	 \For {$a_i \in \mathcal{A}$}
	 	\State // check if activity can start execution
	 	\If {$CanStartExecution\left(\tau, a_i, \pi_{exec}\right)$}
			\State // propagate the decided start time
		 	\State $\pi_{exec} \leftarrow\ PropagateStartActivity\left(\tau, a_i\right)$
			\State // actually dispatch the related command to the robot
			\State $SendCommand\left(a_i\right)$
		\EndIf
	 \EndFor
\EndFunction
\end{algorithmic}
\end{algorithm}

\subsection{Managing the Execution Dependency Graph}
A {\em valid} timeline-based plan consists of a set of timelines whose tokens represent valued temporal intervals satisfying
all the constraints of the domain specification. According to the formal characterization given in Chapter \ref{chap:formalization}, 
each token of a timeline is described by a duration and an end time (interval) satisfying the constraints of the plan. However, 
this information is not sufficient to properly manage plan execution. Temporal relations entail {\em dependencies} among 
tokens of a plan the executive must take into account during execution. 
For example a temporal relation of the form {\em A meets B}, entails that execution of token B must start as soon as execution 
of token A ends. As described in Chapter \ref{chap:epsl}, such dependencies are encoded by the underlying temporal 
network and the {\em inferred} temporal bounds of the related temporal intervals, but the executive must {\em explicitly} model 
these relationships in order to "validate" the plan during execution. Timeline tokens represent flexible intervals and therefore, 
each token is characterized by a {\em start execution condition} and an {\em end execution condition} that allow the executive 
to decide their actual start and end times. 

The {\em Executive Plan Database Manager} of Figure \ref{fig:epsl-exec-layers}, extends information of the plan data-base 
in order to properly manage execution of timelines by means of \edg.
An \edg\ is a data structure encapsulating information about the {\em execution dependencies}, the executive process relies 
on to make decisions about the execution of tokens. An \edg\ is a directed graph built by analyzing temporal 
relationships of the plan being executed. The nodes of the graph represent the tokens of the timelines composing the plan. 
Each node is associated with the current execution status of the related token of the plan. 
The (directed) edges represent {\em execution dependencies} between nodes (\ie\ tokens). The tokens of a plan 
represent flexible temporal intervals and therefore there are two types of edges modeling {\em execution conditions}. 
The {\em start/end execution conditions} model conditions that allow the executive to decide the actual start/end of a token 
during execution.

\begin{definition}
An Execution Dependency Graph (\edg) is a directed graph representing execution dependencies among the tokens of a 
plan. An \edg\ can be formally defined as follows:
$$\langle V, E, \Gamma, \rho, \xi \rangle$$
where:
\begin{itemize}
\item $V$ is the set of nodes of the graph each of which encapsulates a token of the plan.
\item $E = E_{s} \cup\ E_{e} \subseteq\ \{\left(v_{i}, v_{j}\right) : v_{i}, v_{j} \in\ V \land\ v_{i} \neq\ v_{j}\}$ is the set of edges of the 
graph representing execution dependencies between two (distinct) tokens of the timelines. The set $E$ is partitioned into two 
subsets:  
(i) $E_{s}$ contains the edges representing token {\em start dependencies}; (ii) $E_{e}$ contains the edges representing 
token {\em end dependencies}.
\item $\Gamma = \{waiting, starting, inexecution, executed\}$ is a set of constants representing the possible execution status
the tokens of the plan may assume: (i) a token is in {\em waiting} status if the related {\em start conditions} are not satisfied 
and the executive must wait for its execution; (ii) a token is in {\em starting} status if the related {\em start conditions} are 
satisfied and the executive can actually start its execution (this status is particularly relevant for {\em fully uncontrollable} 
tokens as section \ref{sec:edg-uncertainty} will describe); (iii) a token is in {\em inexecution} status if the executive is actually 
executing the token. It means that the executive has started the execution of the token (\ie\ the executive has dispatched the 
start time of the token) but cannot end its execution because the related {\em end conditions} are not satisfied yet; (iv) a token
is in {\em executed} status if its execution is complete. It means that, the {\em end conditions} of the token have been satisfied 
and the executive has ended its execution.
\item $\rho: V \to\Gamma$ is a {\em status function} mapping each node $v_{i} \in\ V$ (\ie\ a token of the plan) 
to its current execution status $\rho(v_{i}) = \gamma_i \in\Gamma$. For example $\rho(v_{i}) = waiting$ means 
that the current status of the token related to $v_{i} \in\ V$, is $waiting$.
\item $\xi: E \to\Gamma$ is a {\em dependency function} mapping each edge of the graph (\ie\ an execution 
dependency) to the required status of the destination node. Specifically, considering {\em start execution} conditions, 
given an edge $\left(v_{i}, v_{j}\right) \in\ E_{s}$, the condition $\xi\left(v_{i}, v_{j}\right) = executed$, means that the 
{\em start condition} of the node $v_{i}$ is satisfied if the status of node $v_{j}$ is $executed$. 
The executive can start the execution of the token related to node $v_{i}$ iff the execution of the token related to 
node $v_{j}$ is ended. The condition $\xi\left(v_{i}, v_{j}\right) = inexecution$ means that the {\em start condition} of 
the node $v_{i}$ is satisfied if the status of node $v_{j}$ is $inexecution$. The executive can start the execution of 
the token related to node $v_{i}$ iff the executive is still executing the token related to node $v_{j}$.
The condition $\xi\left(v_{i}, v_{j}\right) = waiting$ means that the {\em start condition} of the node $v_{i}$ is satisfied if 
the status of node $v_{j}$ is $waiting$. The executive can start the execution of the token related to node $v_{i}$ iff the 
executive has not yet started the execution of the token related to node $v_{j}$.
Finally, the condition $\xi\left(v_{i}, v{j}\right) = starting$ means that the {\em start condition} of the node $v_{i}$ is 
satisfied if the status of node $v_{j}$ is $starting$.
Analogous interpretations hold for {\em end execution} conditions represented by edges $\left(v_{i}, v_{j}\right) \in\ E_{e}$.
\end{itemize}
\end{definition}

The \edg\ is built from the plan before starting execution. The {\em executing conditions}, are dynamically extracted 
from the timeline-based plan by analyzing the temporal relations. Specifically, the graph generation procedure encodes 
Allen's temporal relations \cite{allen} in a set of start and end execution conditions between the involved tokens of the plan 
(\ie\ nodes of the graph).
For example, given {\em A} and {\em B} two tokens of a plan, the temporal relation {\em A during B} can be encoded 
into the \edg\ graph by adding two execution conditions. A {\em start execution condition} asserting that {\em A can start 
execution iff B is "currently" in execution} and, an {\em end execution condition} asserting that {\em A can end execution iff 
B is "currently" in execution}. These two conditions are encoded by two edges, both of which have the node related to the 
token {\em A} as the source, and the node related to the token {\em B} as the target.

\begin{algorithm}
\small
\caption{\edg\ building procedure}
\label{alg:edg-build}
\begin{algorithmic}[1]
\Function{buildExecutionDependencyGraph}{$\Pi$}
	 \State // initialize the \edg\ graph
	 \State $\edg\leftarrow\emptyset$
	 \State // create nodes from the tokens of the timelines
	 \State $\mathcal{FTL} \leftarrow GetTimelines\left(\Pi\right)$
	 \For {$t_i \in\mathcal{FTL}$}
	 	\State // add a new node with the default execution status $waiting \in\Gamma$
		\State $n_{t_i} \leftarrow\ CreateNode\left(t_i, waiting\right)$
		\State $\edg\leftarrow\ AddNode\left(n_{t_i}\right)$
	 \EndFor
	\State // create nodes from the tokens of the observations
	\State $\mathcal{FTL} \leftarrow GetObservations\left(\Pi\right)$
	 \For {$o_i \in\mathcal{FTL}$}
	 	\State // add a new node with the default execution status $waiting \in\Gamma$
		\State $n_{o_i} \leftarrow\ CreateNode\left(o_i, waiting\right)$
		\State $\edg\leftarrow\ AddNode\left(n_{o_i}\right)$
	 \EndFor
	 \State // create edges from the temporal relations of the plan
	 \State $\mathcal{R} \leftarrow\ GetRelations\left(\Pi\right)$
	 \For {$r \in\mathcal{R}$}
	 	\State // encode temporal relation as a set of execution conditions
		\State $\{..., \left(n_{h,i}, n_{h,j}, c_{h,k}\right), ...\}\leftarrow\ GetStartConditions\left(r\right)$
		\State // add start conditions as edges to the graph
		\State $\edg\leftarrow\ addStartConditions\left(\{..., \left(n_{h,i}, n_{h,j}, c_{h,k}\right), ...\}\right)$
	 	\State // encode temporal relation as a set of execution conditions
		\State $\{..., \left(n_{h,i}, n_{h,j}, c_{h,k}\right), ...\}\leftarrow\ GetEndConditions\left(r\right)$
		\State // add end conditions as edges to the graph
		\State $\edg\leftarrow\ addEndConditions\left(\{..., \left(n_{h,i}, n_{h,j}, c_{h,k}\right), ...\}\right)$
	 \EndFor
	 \State \Return $\edg$
\EndFunction
\end{algorithmic}
\end{algorithm}

Algorithm \ref{alg:edg-build} describes the procedure building the \edg\ from the timeline-based plan to be executed. The 
procedure first initializes the graph (row 3). The nodes of the graph represent tokens of the plan with their current execution 
status. The procedure creates a new node for each token of the timelines (rows 5-10) and for each token of the 
external timelines (\ie\ the observations) composing the plan (rows 12-17). The edges of the graph are generated by 
encoding the temporal relations of the plan (rows 19-29). Each temporal relation is "translated" into a set of start and end 
execution conditions the procedure adds to \edg\ as edges (rows 21-28). Each execution condition 
$\left(n_{h,i}, n_{h,j}, c_{h,lk}\right)$ represents an (directed) edge of the graph. The {\em source} node ($n_{h,i}$) represents the 
token the dependency relation refers to. The {\em target} node ($n_{h,j}$) represents the token the source of the relation depends 
on. The {\em condition} ($c_{h,k}$) represents the execution status of the target token enabling the execution of the source token.
The procedure ends by returning a complete \edg.

\subsection{Handling Uncertainty During Execution}
\label{sec:edg-uncertainty}
\begin{figure}[ht]
\begin{tabular}{c c c}
	\includegraphics[width=.3\textwidth]{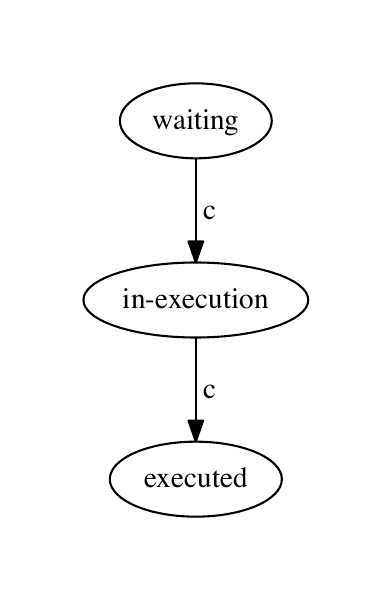} & 
	\includegraphics[width=.3\textwidth]{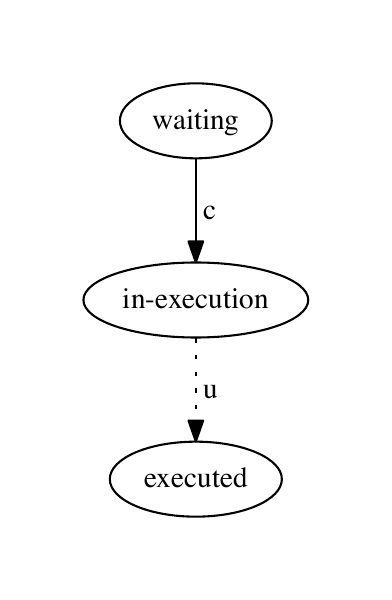} & 
	\includegraphics[width=.3\textwidth]{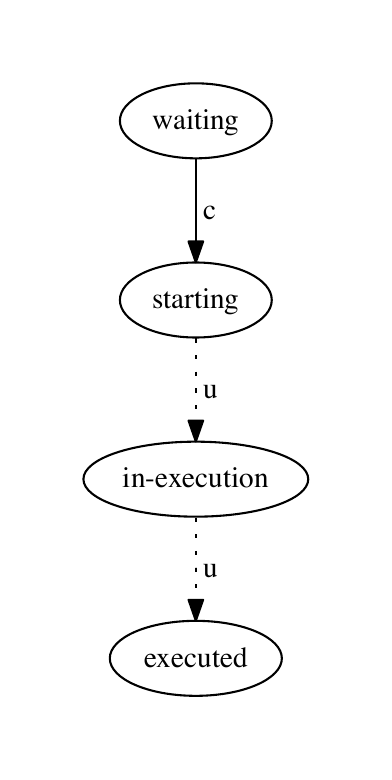}\\
	(a) & (b) & (c)
\end{tabular}
\caption{\small{Different execution state transitions for: (a) {\em controllable tokens}; (b) {\em partially-controllable tokens}; 
(c) {\em fully-uncontrollable tokens}. State transitions tagged with "c" are controllable while transitions tagged with "u" are 
uncontrollable}}
\label{fig:exnode-states}
\end{figure}

An \edg\ encapsulates temporal dependencies between tokens of the timeline-based plans. However, the executive must
also take into account {\em controllability} information concerning the values the tokens of the timelines are related to. 
There are different types of tokens the executive must deal with according to the controllability properties of the related value
of the domain. 
Different types of tokens entail different execution policies and therefore, different state transitions that may be either controllable
or not. Specifically, there are three types of tokens the executive must manage during execution. Figure \ref{fig:exnode-states}
shows the state transitions of the {\em controllable}, {\em partially-controllable} and {\em fully-uncontrollable} tokens.

{\em Controllable tokens} whose state transitions are shown in Figure \ref{fig:exnode-states} (a), are completely under the 
control of the executive. The executive can decide the actual start time of the token and its duration. Thus, the state transition 
between {\em waiting} state and {\em in-execution} state, as well as the state transition between {\em in-execution} state and 
{\em executed} state are both controllable. In this case, the executive can actually dispatch the signals for starting/ending the 
execution of the related command.

{\em Partially-controllable tokens} whose state transitions are shown in Figure \ref{fig:exnode-states} (b), represent tokens
the executive cannot completely control. The executive can decide the start time of this type of tokens and therefore the state 
transition between {\em waiting} state and {\em in-execution} state is controllable. 
However, the actual execution of this type of token is not controllable and therefore, the system can only {\em observe} the 
execution by waiting for a signal from the environment concerning the end of token execution (\ie\ the executive cannot control
the end of the execution of this type of tokens). When the end signal is received, the executive verifies the consistency of the 
plan with respect to the {\em observation} (\ie\ the system checks if the end conditions and the schedule of the token comply 
with the observed behavior) and, the {\em uncontrollable} transition between {\em in-execution} state and {\em executed} state 
is triggered.

{\em Fully-uncontrollable tokens} whose state transitions are shown in Figure \ref{fig:exnode-states} (c), are completely outside
the control of the executive. The executive may suppose when the token is about to start according to its schedule, but cannot
decide its actual start time.  Thus, a state transition between {\em waiting} state and {\em starting} state is 
controllable but it means that the executive is waiting for a signal from the environment which notifies the start of the execution
of the token. The system can only {\em observe} the start of the execution and check if the signal (\ie\ the exogenous event) 
received complies with the plan (\ie\ the start execution dependencies and the schedule of the token are satisfied). When 
the executive receives the start signal from the environment, the {\em uncontrollable} state transition between {\em starting} state
and {\em in-execution} state is triggered. Then, similarly to {\em partially-controllable tokens}, the executive waits the signal 
concerning the end of the execution of the token. When the signal is received, again the executive checks the consistency with 
respect to the plan and the related (end) execution dependencies and the {\em uncontrollable} state transition between the
{\em in-execution} state and the {\em executed} state is triggered.

\subsection{The Importance of Being (Temporally) Robust}
\label{sec:exec-robust}
Timeline-based plans are temporally flexible, hence associated with an envelope of possible execution 
traces. Temporal flexibility allows the executive to be less brittle during execution because the system is able 
to manage the temporal uncertainty of the activities of the plan. The flexible temporal intervals of the tokens 
composing the timelines of the plan allow the executive to "easily" absorb execution delays within the 
specified bounds. 

However, it is not always possible to complete the execution without changing or adapting the plan. Temporal uncertainty and 
{\em uncontrollability features} of the environment may lead to uncontrollable behaviors the timeline-based plan is not able to 
"capture" and therefore, the control system is forced to generate a new plan according to the perceived situation in order to 
complete the execution. Indeed, the plan-based controller relies on a model which tries to describe the (flexible) behavior of the
{\em uncontrollable features} of the domain. Uncontrollability may cause behaviors that do not comply with the plan. 
For example, the execution of an {\em uncontrollable activity} may last longer than expected from the model, or it 
can start later than expected from the plan. In such cases, the executive interrupts the current execution and starts a 
{\em re-planning} phase which tries to generate a new plan from the observed situation.

{\em Re-planning} takes into account the {\em primitive variables} of the domain for building the {\em stable state}  
(\ie\ the problem specification) the planning process starts from in order to generate the new plan. 
The executive analyzes the timelines of {\em primitive variables} by setting the executed tokens with the related 
temporal information as facts of the problem specification. Also, the tokens generated from the observation that caused 
the execution failure are added to the facts of the problem together with their temporal information. 
Moreover, if the executive was executing some uncontrollable tokens when the failure was detected (\eg\ the rover was
moving between two locations), then (supposing the related activities are non-interruptible) the system can wait for the end 
of their execution and add the related facts to the problem specification. Given the resulting problem specification, a new 
plan is generated and the execution can continue starting from the point at which it was interrupted (\ie\ execution failure).

{\em Re-planning} is needed because the execution of these plans is decided on the fly. Without an execution policy, a 
valid plan may fail due to wrong dispatching or environmental conditions (controllability problem \cite{VidalF99}).
It is possible to address this issue in a more robust way by leveraging recent research results exploiting formal methods 
to generate a plan controller suitable for the execution of a flexible temporal plan \cite{ictai13}. 
Namely, \tiga, a model checker for Timed Game Automata (\tga), can be exploited to synthesize robust execution controllers of 
flexible temporal plans. A \tga-based method for the generation of flexible plan controllers  can be integrated within the executive. 
In this case, the \tiga\ engine is embedded within the planning and execution cycle generating plan controllers that guarantee 
a correct and robust execution. This is an important feature of the \fbt\ Task Planner as it enforces a safe plan execution further
enforcing that all the production requirements and human preferences are properly respected.

\section{Human-Robot Collaboration: a Case Study}
\label{sec:fbt}
Human-Robot Collaboration (\hrc) in manufacturing represents an interesting and quite complex application context 
which requires a tight interaction between a human operator and a robotic device (\eg\ a robotic arm) to perform some 
factory operations. 
From the perspective of a plan-based control system, the envisaged environment is composed of two 
{\em autonomous agents} that share the same working environment and may {\em operate independently} or may 
{\em collaborate} by supporting each other. This type of application presents several challenges a plan-based control 
system must cope with in order to control the robot and guarantee a {\em safe} collaboration with the human. 
In general, there are three important features the control system must deal with in order to generate effective plans:

\begin{itemize}
\item {\em Supervision}, to represent and satisfy the production requirements needed to complete the factory processes.
\item {\em Coordination}, to represent the activities the human operator and the robot must perform according to the 
Human-Robot Collaboration settings.
\item {\em Uncertainty}, to manage the {\em temporal uncertainty} about the activities of the human operator that the 
system cannot control.
\end{itemize}

A key enabling feature is the capability to model and manage the {\em temporal uncertainty} concerning the behavior of the
human operator. The human is an active "part" of the environment which is not under the control of the robot. The control 
system must take into account the (expected) behavior of the human in order to properly manage operations of the robot.
Thus, \hrc\ represents a relevant application context to leverage the feature of the timeline-based planning and execution 
framework described above. The following sections deal with the development of an \epsl-based dynamic task planing 
system within the \fbt\ research project, for planning and execution in real-world \hrc\ manufacturing scenarios.

\subsection{The \fbt\ Research Project}
Industrial robots have demonstrated their capability to meet the needs of many application domains, offering accuracy, 
efficiency and flexibility of use. A relevant research challenge is the co-presence of robot and human  in the same 
environment collaborating in a common goal. In general, when robot-worker collaboration is needed, there are a number 
of open issues to be taken into account, first of those is human safety that needs to be enforced in a comprehensive way.  
A key open trend in manufacturing is the design of shared fenceless working spaces in which safe human-robot collaboration 
is seamlessly implemented. 
The \fbt\ research project\footnote{http://www.fourbythree.eu} \cite{etfa16-4x3} aims at designing, building and testing robust and 
configurable robotic solutions capable of collaborating safely and efficiently with human operators in industrial manufacturing 
companies. The overall aim of the project is to create a new generation of robotic solutions, based on innovative hardware 
and software, which present four main characteristics: modularity, safety, usability and efficiency.  The envisaged robot 
services take into account the co-presence of three different actors: humans, robots and the environment.

A human-robot collaboration workcell is a bounded connected space with two agents located in it, a human and a robot 
system, and their associated equipment \cite{marvel2015}. The robot system consists of a robotic arm with its 
tools, its base and possibly additional support equipment. The workcell also includes the workpieces and any other tool 
associated with the targeted task and dedicated safeguards (physical barriers and sensors such as, e.g., monitoring video 
cameras). 
In such a working environment, four different degrees of interaction between a human operator and the robot can be defined 
\cite{helms2002rob}. In all these cases, it is assumed that the robot and the human may need to occupy the same spatial location 
and interact according to different modalities: 
\begin{itemize}
\item {\em Independent}, the human and the robot operate on separate workpieces without collaboration, \ie\ independently 
from each other;
\item {\em Synchronous}, the human and the robot operate on sequential components of the same workpiece, \ie\ one can start 
a task only after the other has completed a preceding task;
\item {\em Simultaneous}, the human and the robot operate on separate tasks on the same workpieces at the same time; 
\item {\em Supportive}, the human and the robot cooperate to complete the processing of a single workpiece, \ie\ they 
work simultaneously on the same task.
\end{itemize}
Different interaction modalities entail the robot to be endowed with different safety (hardware and control) settings while 
executing tasks.

\begin{figure}[ht]
\centering
\includegraphics[width=\textwidth]{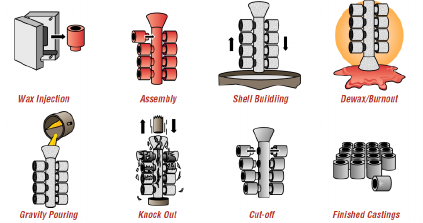}
\caption{\small{The overall ALFA pilot production process}}
\label{fig:process}
\end{figure}

In \fbt\  four different pilots are taken into account covering different production processes, \ie\ assembly/disassembly of parts, 
welding operations, large parts management and machine tending. Among these, the ALFA Pilot is particularly relevant
from the \hrc\ perspective.
This case study corresponds to a production industry (the \textsc{ALFA Precision Casting}\footnote{ALFA is a medium sized 
company producing aluminium parts for different industries for applications that are characterized by low size production batches 
and requiring tight tolerance and dimensional precision.}) which represents a real working scenario with different relevant 
features (\eg\ space sharing, collaboration or interaction needs).
The overall production process (summarized in Fig. \ref{fig:process}) consists of a metal die which is used to produce a wax 
pattern in a injection machine. Once injected, the pattern is taken out of the die.
Several patterns are assembled to create a cluster. The wax assembly is covered with a refractory element, creating a shell 
(this process is called investing). The wax pattern material is removed by the thermal or chemical means. The mould is heated 
to a high temperature to eliminate any residual wax and to induce chemical and physical changes in the refractory cover.
The metal is poured into the refractory mould. Once the mould has cooled down sufficiently, the refractory material is removed 
by impact, vibration, and high pressure water-blasting or chemical dissolution. The casting are then cut and separated from the 
runner system. Other post-casting operations (\eg\ heat treatment, surface treatment or coating, hipping) can be carried out, 
according to customer demands.

Given this production process, the first step (preparation of the die for wax injection and extraction of the pattern from the die) 
has a big impact on the final cost of the product, and it represents a relevant application scenario. Thus, the involvement of 
a collaborative robot has been envisaged to help the operator in the {\em assembly/disassembly} operation.
The operation consists of the following steps: (i) mount the die; (ii) inject the wax; (iii) open the die and 
remove the wax; (iv) repeat the cycle for a new pattern starting back from step (i). The most critical sub-operation is the 
opening of the die because it has a big impact on the quality of the pattern.

\subsection{Assembly/Disassembly Operation}
\begin{figure}[ht]
\centering
\begin{minipage}{.48\textwidth}
	\includegraphics[width=\textwidth]{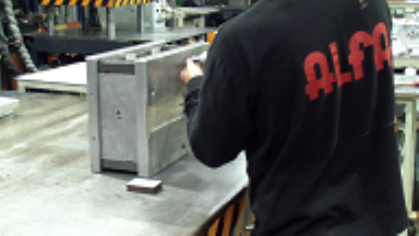}
\end{minipage}
\begin{minipage}{.48\textwidth}
	\includegraphics[width=\textwidth]{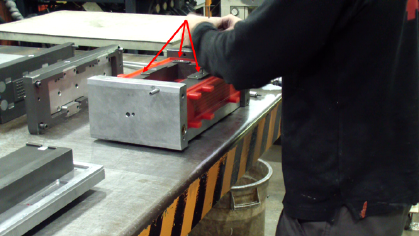}
\end{minipage}
\begin{minipage}{.48\textwidth}
	\includegraphics[width=\textwidth]{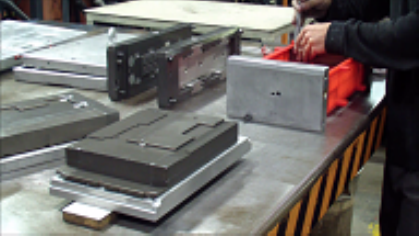}
\end{minipage}
\begin{minipage}{.48\textwidth}
	\includegraphics[width=\textwidth]{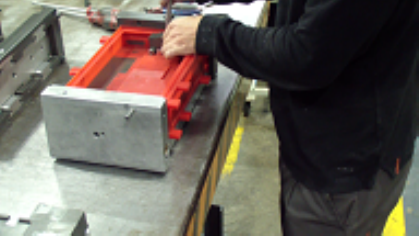}
\end{minipage}
\begin{minipage}{.48\textwidth}
	\includegraphics[width=\textwidth]{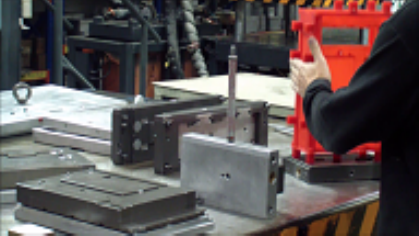}
\end{minipage}
\begin{minipage}{.48\textwidth}
	\includegraphics[width=\textwidth]{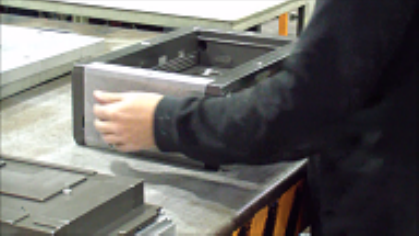}
\end{minipage}
\caption{\small{The manual procedure of the Assembly/Disassembly process of the ALFA pilot}}
\label{fig:manual-operation}
\end{figure}

Due to the small size of the dies and the type of operations done by the worker to remove the metallic parts of the die, 
it is very complex for the robot and the worker to operate on the die simultaneously. Figure \ref{fig:manual-operation} shows 
some of the steps of the overall (manual) operation. However, they can cooperate in the sub-operations concerning the 
assembly/disassembly of the die. 
Once the injection process has finished, the die is taken to the workbench by the worker. The robot and the worker unscrew 
the bolts holding the top cover. There are nine bolts, the robot starts removing those closer to it, and the worker the rest. 
The robot unscrews the bolts on the cover by means of a pneumatic screwdriver. The worker removes the top cover and 
leaves it on the assembly area (a virtual zone that will be used for the re-assembly of the die). 
The worker turns the die to remove the bottom die cover. The robot unscrews the bolts on the bottom cover by means of a 
pneumatic screwdriver. Meanwhile the operator unscrews and removes the threaded pins from the two lateral 
sides to release the inserts. The worker starts removing the metallic inserts from the die and leaves them on the table. 
Meanwhile, the robot tightens the parts to be assembled/reassembled together screwing bolts. The worker re-builds the die. 
The worker and the robot screw the closing covers.
The human and the robot must collaborate to perform assembly/disassembly on the same die by suitably handling different 
parts of the die and screwing/unscrewing bolts. Specifically, the human worker has the role of leader of the process while the robot 
has the role of subordinate with some autonomy. 

\section{Dynamic Task Planning in \fbt}
\label{sec:taskplan}
In \fbt\ and more in general in \hrc\ applications, the envisaged dynamic task 
planning system must realize a {\em human aware planning and execution mechanism} 
capable of allowing a robot to safely interact with an operatore. The control mechanism
must adapt robot plan and motions according to the expected and observed behaviors 
of the related human operator \cite{etfa16,Pellegrinelli2017}.
In this sense, the dynamic task planning system applies and extends the hierarchical 
timeline-based modeling approach by introducing {\em supervision} and {\em coordination} 
issues. 
{\em Supervision} models the operational requirements of the production processes. It models 
the high-level tasks to perform in order to complete the process and the related precedence 
constraints that must be satisfied. {\em Coordination} models the possible decompositions of 
the high-level tasks in low-level tasks the human and the robot can directly perform and the 
possible assignments.
Moreover, the {\em human} is an active element of the environment that cannot be directly controlled by the robot and therefore 
the human is modeled as a variable of the domain whose values (\ie\ the low-level tasks the operator can directly perform) are all 
{\em uncontrollable}. 
The dynamic task planning framework must plan for the tasks the human and the robot must perform by coordinating 
them and by taking into account the {\em temporal uncertainty} of the human. 
Human activities are {\em uncontrollable} and therefore the system must generate and execute plans without making any 
hypothesis on the actual duration of the tasks assigned to the human.

\begin{figure}[ht]
\centering
\includegraphics[width=.9\textwidth]{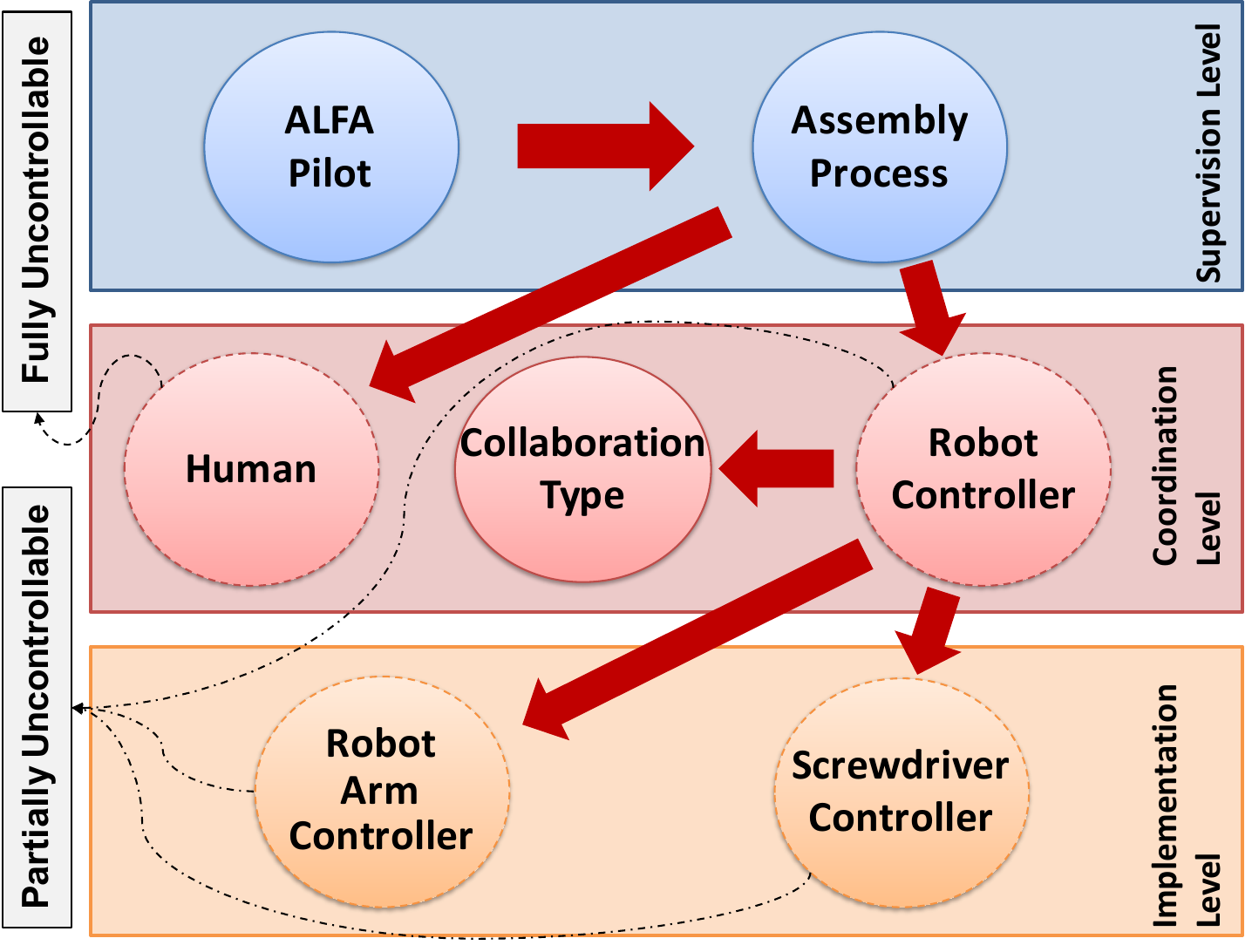}
\caption{\small{The hierarchy of the task planning domain}}
\label{fig:layers}
\end{figure}

\subsection{Task Planning Model for Assembly/Disassembly}
\begin{figure}[ht]
\centering
\includegraphics[width=.8\textwidth]{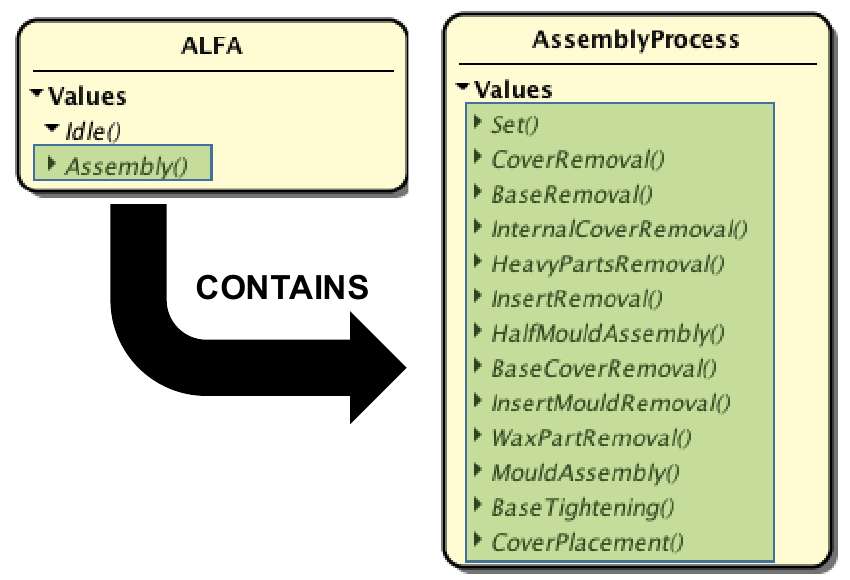}
\caption{\small{Defining the workflow of work-cell operations}}
\label{fig:supervision}
\end{figure}

Figure \ref{fig:layers} shows the hierarchical structure of the control model for the assembly/disassembly production process 
of the ALFA pilot of the project.
The supervision layer represents the elements describing the processes of the work-cell. The {\em ALFA} state variable 
modes the general ALFA pilot and the related processes. Specifically, each value of the state variable represents a specific 
process (\eg\ the {\em Assembly operation}) the human and the robot can perform in the pilot. 
The {\em AssemblyProcess} state variable models the assembly/disassembly operation by specifying the set of 
high-level tasks required. As shown in Figure \ref{fig:supervision} a set of constraints specifies the operational requirements that 
guarantee a correct execution of the process. For example, these requirements may specify ordering constraints between the 
high-level tasks or may specify different procedures for performing the process (\eg\ alternative sequences of high-level tasks).
In this specific case, the operational requirement of the {\em supervision layer} specifies a total ordering among the high-level 
tasks composing the {\em Assembly} process of the case study.

It is worth observing that the control model is not considering coordination features at this abstraction level. Each high-level task
represents a complex procedure which must be further decomposed in (primitive/atomic) low-level tasks the human and the robot 
can directly handle. Some primitive tasks can be performed either by the human or by the robot and it is up to the task
planner deciding who must execute them. Moreover, given a high-level task, the system must coordinate human and robot 
activities according to the type of collaboration desired for the specific collaboration scenario. Different types of collaboration entail
different safety settings and therefore different configurations of the robot for performing tasks.

\begin{figure}[ht]
\centering
\includegraphics[width=.8\textwidth]{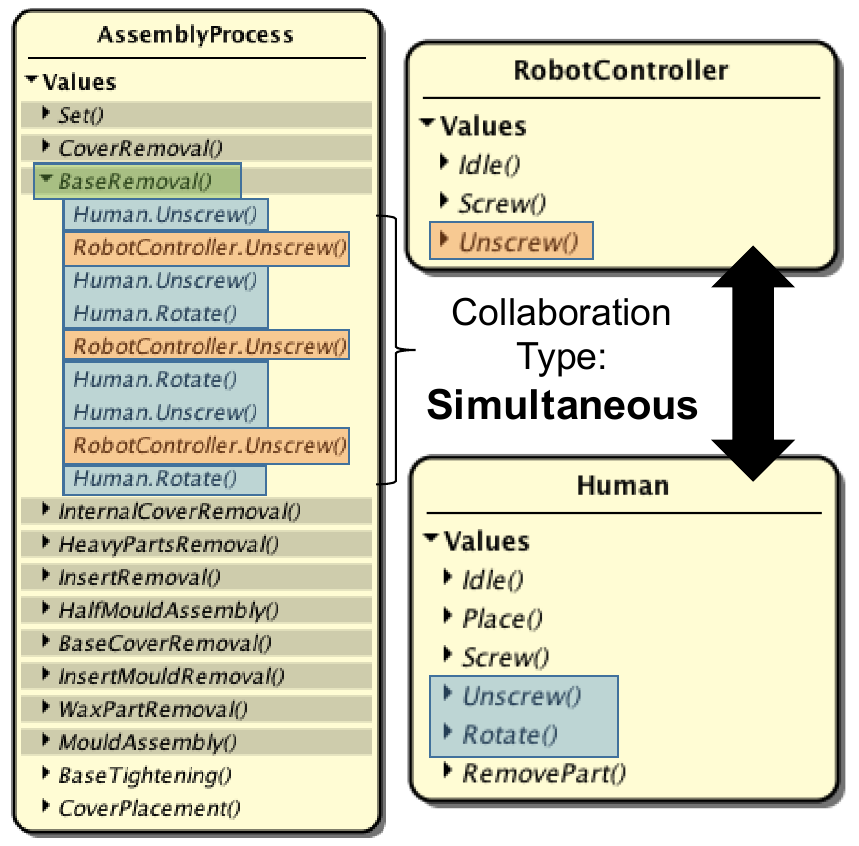}
\caption{\small{Assigning tasks to the robot and the human}}
\label{fig:coordination}
\end{figure}

Figure \ref{fig:coordination} shows an example of coordination requirements between the robot and the human with respect to 
the high-level task named {\em BaseRemoval} of the {\em Assembly} process. The model describes the sequence of low-level 
tasks needed to properly complete the {\em BaseRemoval} task with their assignments. 
The robot and the human simultaneously unscrew the bolts of the base of the die and therefore the type of collaboration required 
is {\em simultaneous} in this specific case (the human and the robot work on the same workpiece while performing different tasks,
\ie\ unscrewing bolts). Again, the control system (and the robot) must be {\em aware of the human} and adapt its tasks according 
to the human-robot collaboration process defined. 

\begin{figure}[ht]
\centering
\includegraphics[width=.45\textwidth]{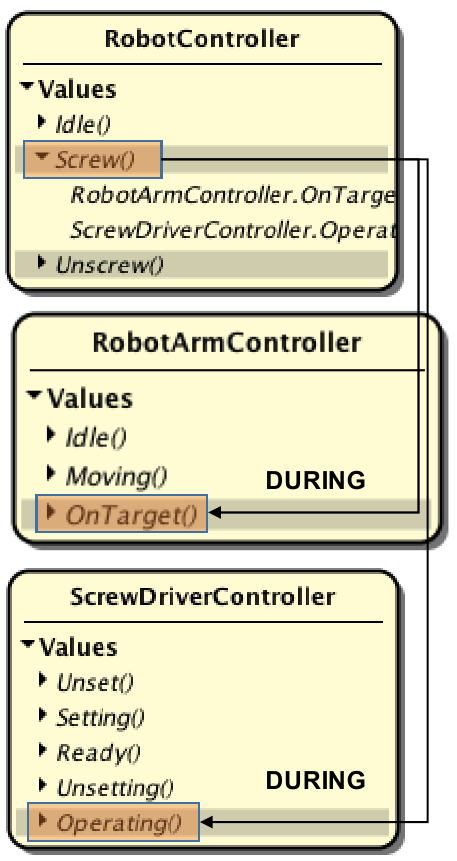}
\caption{\small{Decomposition of the low-level tasks of the robot controller}}
\label{fig:implementation}
\end{figure}

Finally, the low-level tasks of the robot must be further decomposed in order to synthesize the set of {\em commands/signals} 
to be dispatched for execution. For example, the {\em Screw} task of the {\em RobotController} in Figure \ref{fig:coordination}, 
requires to set the arm on the bolt to screw and then activate the tool (\ie\ the screwdriver) in order to actually screw the bolt 
and complete the task. According to this description, the {\em Screw} task must be decomposed in terms of commands that 
allow the robot to assume the desired pose and activate/deactivate the tool. Specifically, the related synchronization rule of the 
model, constrains the behavior of the {\em RobotArmController} and the {\em ScrewDriverController} (see Figure
\ref{fig:implementation}) by specifying the values they must assume (\ie\ tokens) and the related temporal constraints that must 
be satisfied. The {\em OnTarget} value of the {\em RobotArmController} sets the arm on the target bolt. The {\em Operating} value of 
the {\em ScrewDriverController} activates the tool in order to start screwing the bolt. The temporal constraints shown in Figure
\ref{fig:implementation}, allow the arm to keep the position for the entire duration of the task.

\begin{figure}[ht]
\centering
\includegraphics[width=\textwidth]{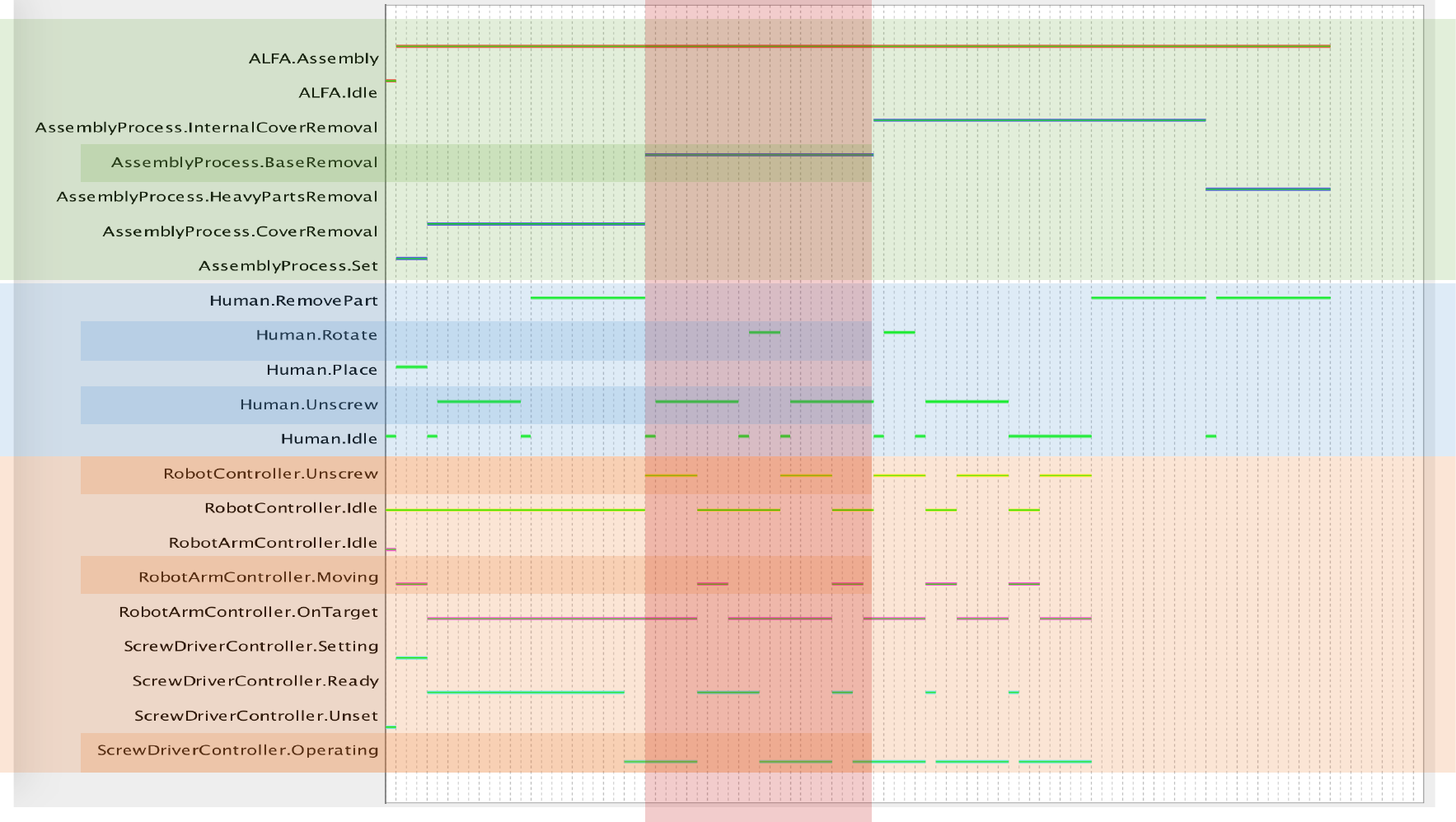}
\caption{\small{The Gantt chart representation of the plan for the ALFA pilot with respect to the earliest 
start time of the related tokens}}
\label{fig:plan}
\end{figure}

Figure \ref{fig:plan} shows an excerpt of a hierarchical timeline-based plan for the {\em Assembly} process of the ALFA 
case study. The horizontal sections (\ie\ bars with different colors) partition the plan according to the hierarchy depicted 
in Figure \ref{fig:layers}. The vertical section (in red) depicts an example of high-level task decomposition and application 
of the synchronization rules of the domain.
Namely, the decomposition of the {\em BaseRemoval} high-level task of the {\em Assembly} process: the {\em BaseRemoval} 
task requires the human operator and the robot to simultaneously unscrew some bolts from two lateral sides of the work-piece,  
then the human should rotate the piece and finally, the operator and the robot unscrew bolts from two lateral sides of the piece.
Figure \ref{fig:plan} shows that the plan satisfies the production requirements of the high-level task. Indeed, a synchronization 
rule requires that the low-level tasks for unscrewing bolts should be executed during the {\em BaseRemoval} task. Moreover, 
the first unscrew tasks must be performed {\em before} the operator rotates the piece, while the second unscrew tasks must be
performed {\em after} the operator rotates the piece.
It is also possible to observe that robot's tasks are further decomposed in order to synthesize a more detailed representation of 
the activities the robot must perform to actually carry out the low-level tasks. For instance, the robot must set the arm on a specific
target and then must activate the tool in order to perform an unscrew operation. Again, in Figure \ref{fig:plan}, a {\em during} 
temporal constraint holds between the {\em Unscrew} low-level task token and the {\em OnTarget} and {\em Operating} tokens.

\subsection{Feasibility Check of the Task Planning Model}
Deliberation time \ie\ the time spent by the dynamic task planning system to generate a plan for the considered production 
process, has been considered as the first key performance indicator to be assessed in order to test the performance of the 
dynamic task planning system.
Thus, with respect to the planning model of the assembly/disassembly process described in the previous section, 
different planning scenarios have been considered by varying the complexity of the dimensions of the problems:
\begin{itemize}
\item {\em Production process complexity} - three different production procedures have been analyzed by taking into 
account an increasing number of tasks needed to complete the assembly/disassembly process: the {\em small} 
procedure consists of 6 tasks; the {\em  medium} procedure consists of 10 tasks; the {\em large} procedure consists 
of 15 tasks.
\item {\em Human-Robot effort} - for each production procedure an increasing involvement of the robot has been 
considered in order to increase the number of tasks the robot must perform to complete the process and consequently 
decrease the effort of the human 
\end{itemize}

\begin{figure}[ht]
\centering
\includegraphics[width=\textwidth]{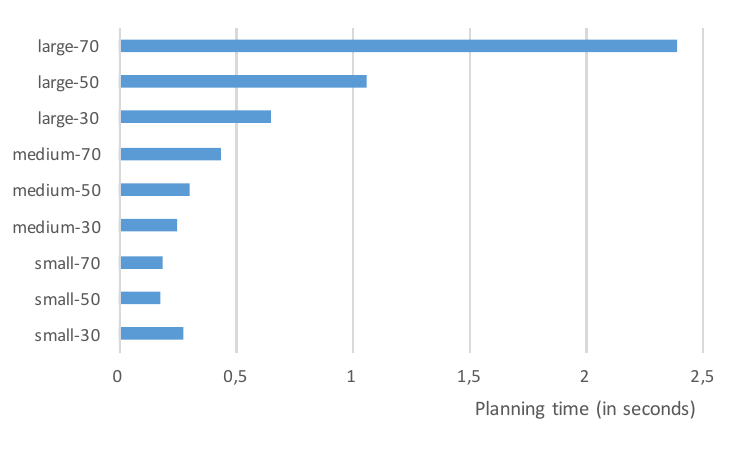}
\caption{\small{Deliberation time on different problems with different assignment policies}}
\label{fig:exec-delib}
\end{figure}

Figure \ref{fig:exec-delib} shows the deliberation times of the dynamic task planning system for the considered 
scenarios. In general, the higher is the number of tasks needed for the process, the higher is the number of tasks 
that can be assigned to the robot and, consequently, the higher is the complexity of the resulting problem with respect 
to deliberation. 
As the results in Figure \ref{fig:exec-delib} show, an increasing  complexity of the scenario entails higher deliberation 
times. Nevertheless, planning costs result to be compatible with the latency of the production environment.
Indeed, the performance is compatible with the latency usually involved in this type of manufacturing 
applications. In particular, the experimentation emphasizes the flexibility of the envisaged approach to planning which 
is capable to adapt the assignment and coordination strategies to different human-robot collaboration settings. 

\subsection{The Dynamic Task Planning Module in Action}
\begin{figure}[ht]
\centering
\includegraphics[width=.9\textwidth]{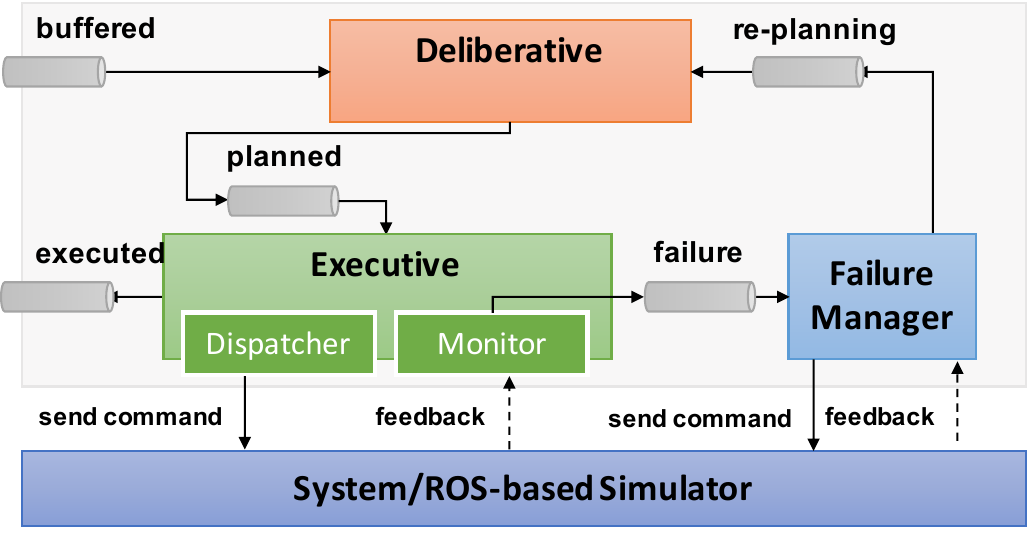}
\caption{\small{Process-view of the dynamic task planning control module}}
\label{fig:fbt-module}
\end{figure}

The dynamic task planning system has been tested on a \ros-based simulator\footnote{{\em "The Robot Operating 
System (\ros) is a collection of tools, libraries, and conventions that aim at simplifying the task of creating complex and 
robust robot behavior across a wide variety of robotic platforms"} - from: http://www.ros.org/about-ros/} which 
provides an implementation of the functional control level of a generic robotic arm.
Figure \ref{fig:fbt-module} shows the process architectural view \cite{4piu1} of the dynamic task planning module describing
the main elements composing the module at runtime (\ie\ the processes) and their interactions. The process architectural view
aims at describing how the control flow is structured and how the deliberative and executive processes (both relying on 
\epsl-based planning and execution capabilities) interact. In particular, this view describes the management of 
{\em plan execution failures} and the related {\em replanning mechanism}. The execution decisions of the dynamic task 
planning module, shown in Figure \ref{fig:fbt-module}, are taken on the fly during execution. Without an execution policy a 
valid plan may fail due to wrong dispatching or environmental conditions (controllability problem \cite{VidalF99}) and 
therefore {\em replanning} is the basic mechanism which allows the module to deal with exogenous events and complete 
the execution of a plan.

The processes of the dynamic task planning module exchange information concerning the task to be managed through 
queues. The different queues of Figure \ref{fig:fbt-module} represent the states composing the {\em task lifecycle} within the 
module. The task lifecycle models the control flow of the dynamic task planning module. The {\em buffered queue} is the 
{\em entry point} of the module, it contains the high-level task requests the module must manage \ie\ requests of performing 
a particular process of the factory (\eg\ the assembly/disassembly process).

The {\em Deliberative} process takes a task request from the {\em buffered queue} and synthesizes a (pseudo-controllable) 
plan for the task. The generated plan represents a suitable set of (low-level) tasks the human and the robot must execute 
according to the desired operational requirements. Thus, the task (\ie\ the high-level task request) is ready for execution 
and therefore the task with the related plan is added to the {\em planned queue}.

The {\em Executive} process takes a task request from the {\em planned queue} and starts executing the related plan by 
sending commands to the system (or a \ros-based simulator) through the {\em Dispatcher} and receiving {\em feedbacks} 
about command execution through the {\em Monitor}. As described in Section \ref{sec:exec-proc}, the {\em Dispatcher} is 
responsible for deciding the start of token execution according to their controllability properties (see Section 
\ref{sec:edg-uncertainty}). The {\em Monitor} is responsible for managing execution feedbacks from the environment (or the
\ros-based simulator) in order to verify the {\em correctness} of the plan with respect to the actual behavior of the 
system. If no inconsistency is detected the execution continues until the plan is ended and the task request, with the resulting 
plan, is added to the {\em executed queue}. Otherwise, if an inconsistency is found then the executive interrupts the execution 
and the task request with the interrupted plan is added to the {\em failure queue}.

Execution fails every time the {\em uncontrollable} dynamics of the environment do not comply with the plan and the "expected"
uncertainty of the domain. As described in Section \ref{sec:exec-robust}, temporal flexibility allows the executive to capture an 
envelope of possible (temporal) behaviors of the environment. Different temporal behaviors can be easily managed by the 
executive by temporally adapting the plan, if such behaviors comply with the model. However, if the observed dynamics
of the environment do not comply with the expected uncertainty then the plan cannot capture these behaviors and {\em replanning}
is needed. For example, human tasks are (fully) uncontrollable and therefore the executive cannot make any hypothesis on their
actual duration. The model provides an estimation of the durations of such tasks in terms of minimum and maximum expected 
duration. The deliberative generates plans according to this estimation ({\em pseudo-controllable} plans), thus the executive can 
execute them if the observed behavior of the environment complies with the model (\ie\ the actual duration of uncontrollable 
tasks comply with the expected durations). Thus, if the observed duration of a human task is higher than the expected maximum, 
then the plan cannot be adapted and a new plan is needed in order to address the real situation.

The {\em Failure Manager} process is responsible for managing the interruption of the execution in order to set a {\em stable 
state} before generating a new plan. The process takes the interrupted task with the related execution trace (\ie\ the executed
portion of the plan) from the {\em failure queue} and interacts with the environment (the \ros-based simulator in this case) in order 
to set the robot in a stable state. As broadly described in Section \ref{sec:exec-robust}, the Failure Manager analyzes the 
timelines concerning the {\em primitive variables} of the domain in order to set the {\em situation} the Deliberative will 
{\em replan} from. 
In this specific case, if the execution is interrupted while the robotic arm is moving, the Failure Manager waits the execution 
feedback of the motion in order to let the Deliberative start replanning with the robotic arm set in a stable position.
Another possible approach would allow the Failure Manager to interrupt the motion and send the commands needed to set the robotic 
arm in a known (initial) position. In general, the logic implemented by the Failure Manager cannot 
be generalized because it is strictly connected to the specific robotic platform considered and the related functional level (\ie\ 
the set of sensing and action primitives available for interacting with the robotic platofrm).

When a stable state is reached \ie\ both the robot and the human are in a known stable state, the Failure Manager leverages 
the execution trace of the interrupted plan and the current situation of the robot and the environment to build the problem 
specification for the new plan. The interrupted task with the related problem specification is added to the {\em replanning 
queue} and the Deliberative starts generating a new plan by fitting the related problem specification.

\subsubsection*{Experimental evaluation on a \ros-based simulator}
\begin{figure}[ht]
\centering
\includegraphics[width=.95\textwidth]{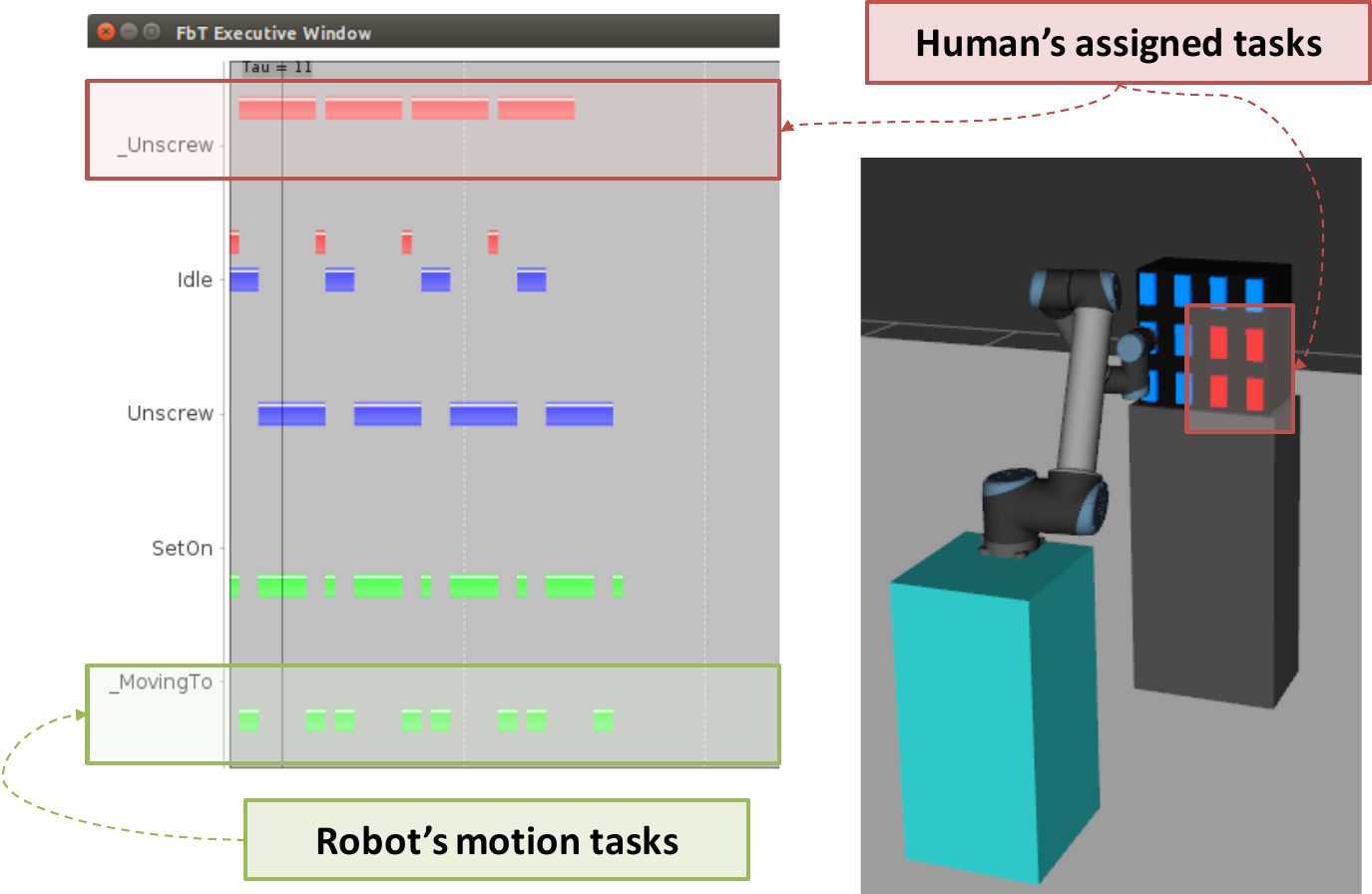}
\caption{\small{\ros-based simulation of the dynamic task planning system}}
\label{fig:exec-sim}
\end{figure}

Figure \ref{fig:exec-sim} shows a screenshot of a simulation for the assembly/disassembly process. The left-hand side of 
Figure \ref{fig:exec-sim} shows a portion of the plan of Figure \ref{fig:plan} during execution. It shows the Gantt chart representing 
the timeline of the Human (the sequence of red tasks), the timeline of the RobotController (the sequence of blue tasks) and the 
timeline of the RobotArmController (the sequence of green tasks).
The right-hand side of Figure \ref{fig:exec-sim} shows a simple 3D model of the workcell composed by a robotic arm and the 
workpiece. The colored blocks of the workpiece represent the bolts the robot and the human are supposed to unscrew
within the assembly/disassembly process. Specifically, the blue blocks represent the bolts the Deliberative has assigned to 
the robot, and the red blocks represent the bolts the Deliberative has assigned to the human.

Simulations have shown the capability of the dynamic task planning system of coordinating the human and 
the robot in order to perform assembly operations. The coordination takes into account the expected duration bounds  of 
the tasks (especially the task of the human) that are assigned by taking into account the {\em makespan}
of the plan. Namely, the dynamic task planning system generates timeline-based plans that minimize the expected 
duration of the overall process and therefore optimize/maximize the throughput of the factory.
Simulations have also shown the capability of the dynamic task planning system of adapting the 
execution of robot tasks to the observed behavior of the human operator. Leveraging temporal flexibility, the system 
easily captures the possible behavior of the human by {\em dispatching} robot tasks accordingly. 
However, if the observed behavior of the human does not comply with the expected one (\ie\ with the model) the 
system cannot proceed with the execution and the plan is {\em interrupted} (execution failure). In this case, simulations 
have shown the capability of the system of generating a new plan through {\em replanning}. Specifically, the Failure Manager 
sets the robot in a stable state by waiting for execution feedbacks of not completed motion tasks or any {\em uninterruptible} 
operation. Then, the Deliberative generates a new plan by taking into account executed tasks and reassigning missing tasks 
to the human and to the robot. Once the new plan has been generated, the Executive resumes plan execution starting with 
the reassigned tasks.


%


\graphicspath{{chapters/8_kbcl/figures/}}


%
%
%
%
\chapter{Knowledge-based Control Loop for Flexible Controllers}
\label{chap:kbcl}
\lettrine[lines=2]{T}{he} ideal robot is an artificial entity (an agent) capable of setting its own goals and of planning 
actions to achieve them. The research community in robotics and AI has been building many types of robots applying 
different techniques but yet is still far from anything ideal. There is still a limited understanding of what are the essential 
characteristics of artificial agent like robots. From a local point of view, robot software or hardware parts and modules, 
like sensors, reasoning engines, etc., present several limitations when compared to the capabilities of similar parts in 
the human being. Similarly, from a global point of view, there is not a clear vision of how to integrate different parts 
together in order to realize an agent capable of autonomously operate in the environment by understanding the current 
situation and "act" accordingly by properly manage the dynamics of the "world".
These philosophical problems are not just theoretical but they are also present in practical and specific areas like industrial
robots. In particular there are several research initiatives that focus on the construction of robots that can quickly adapt 
changes in the production environments \cite{wiendahl2007changable}. Traditional systems, based on centralized or
hierarchical control structures, like the plan-based approach described in the previous chapter, typically require major 
overhauls of their control code when some sort of system adaptation and reconfiguration is required. 

Dynamic working environments like {\em Reconfigurable Manufacturing Systems} (RMSs) \cite{koren1999rms} require
control processes with an high level of flexibility. The actual capabilities of an agent and even the production processes of the 
factory may change quickly in such contexts. Different configurations of the shop-floor or the introduction of different 
production goals may change the type and/or the ways agents carry out their tasks. 
Classical plan-based controllers usually rely on a well-defined and {\em static} model of the world which could 
become obsolete very soon. The domain model would require a great design effort to be as stable as possible and 
a continuous {\em maintenance} which would negatively affect the productivity of the factory. The pursued solution is to 
extend classical plan-based control architectures by introducing {\em knowledge representation and reasoning} mechanisms 
into the control loop. Semantic technologies provide the {\em flexibility} needed to dynamically adapt the control model 
(\ie\ the planning model) to different production settings.
Thus, this chapter presents the {\em Knowledge-based Control Loop} (\kbcl) which proposes an extension to classical 
plan-based control architectures suitable for artificial agents in general and robotics in particular. 
The proposed solution relies on an ontological approach for knowledge classification and management structuring information 
about the capabilities of an agent and the related working environment (\eg\ an industrial robot in a manufacturing environment). 
This chapter describes how the knowledge of the agent is structured and how such knowledge can be exploited to 
dynamically generate a timeline-based planning model used to plan and executive the activities of the agent.

\section{Flexible Plan-based Control Architectures}
The integration of knowledge reasoning and planning is today critical. Indeed, the integration of these two 
technologies involves the manipulation of symbolic information at different levels of abstraction and its translation 
into different structures for controlling the state of the different components of the agent and, consequently, their 
interaction with the environment.

Despite the variety of uses of ontologies in robotic applications, the organization and management of the information 
needed to act at run-time remains an open problem. This is a challenging problem to face in order to develop 
{\em adaptive} autonomous robots. The pursued solution aims at integrating knowledge reasoning and planing 
techniques in order to provide the control process with the flexibility needed to dynamically adapt the control 
model to the actual state of the system and the environment. The proposed approach relies on 
two elements: a foundational ontology which organizes the information, and control process which continuously 
updates data and manages the flow of information needed to plan and execute activities of the agent. 

The ontology defines the structure of the general information the flexible control module must deal with. The 
ontology provides a semantics for the concepts and the general properties characterizing the application domain.
The control process leverages the ontology to generate and manage the {\em knowledge} of the specific agent 
to control. Specifically, the ontology guides the interpretation of data concerning the agent and the environment, 
and allows the control process to dynamically {\em instantiate} such information into a {\em Knowledge Base} (KB)
which describes the specific capability of the agent and the specific working environment. 
On the basis of the obtained KB, the control process can dynamically generate the planning model tailored to the 
actual state of the the actual state of the agent and the related working environment. Then, the control process 
continuously monitor the agent and the environment in order to maintain the KB and also the control model updated.

\subsection{The Manufacturing Case Study}
\begin{figure}[ht]
\centering
\includegraphics[width=.98\textwidth]{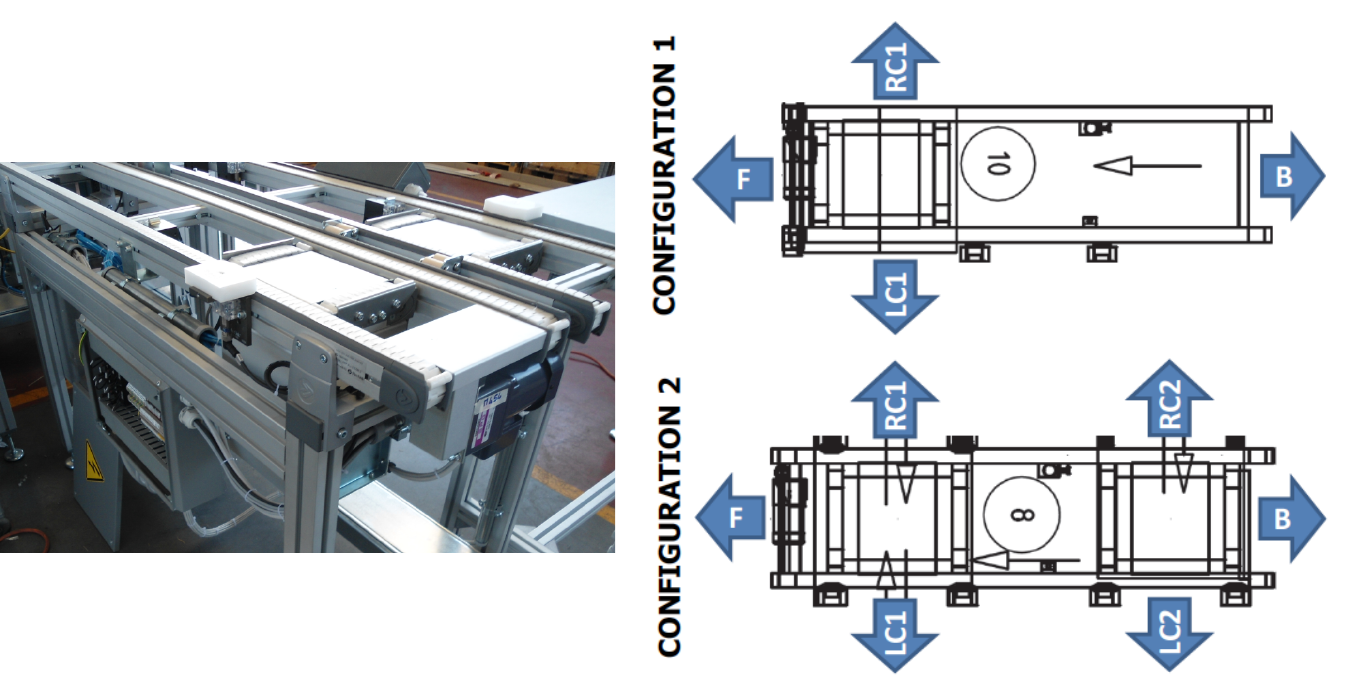}
\caption{\small{A picture of a Transportation Module (on the left) of the RTS and two possible configurations (on the right)}}
\label{fig:module}
\end{figure}

The flexible control architecture described in this chapter has been designed in order to work in the context of a pilot
case from the \gecko\ project \cite{etfa14}. The pilot case consists in a Reconfigurable Manufacturing System (RMS) 
for recycling Printed Circuit Boards (PCB).
The plant is composed of different machines for loading/unloading, testing, repairing and shredding of PCBs and of 
a conveyor system that connects them. The conveyor is implemented through a Reconfigurable Transportation 
System (RTS) composed of a set of reconfigurable mechatronic components, called \emph{Transportation Modules} 
(TM), see Fig. \ref{fig:module}. The goal of the plant is to analyze defective PCBs, to automatically  diagnose their 
faults and, depending on the type of the malfunctions, attempt an automatic repair or send them to waste.
 
The proposed agent architecture is wrapped around each of the TMs hence its functionalities are introduced with more 
details. Each of the TMs combines three Transportation Units (TUs). The units may be unidirectional or bidirectional, 
with bidirectional units enabling also movements from side to side (cross-transfers) from/to other TMs, see Fig. \ref{fig:module}. 
The TM can support two main transfer services, forward and backward, and zero to many cross-transfer services. 
Different configurations can be deployed varying the number of cross-transfers  components and thus enabling multiple I/O 
ports. TMs can be connected back to back to form a set of different conveyor layouts. 

The manufacturing process requires each PCB to be loaded on a fixturing system (a pallet) in order to be transported by 
the TMs and processed by the various machines of the RMS. The transportation system can move one or more pallets 
(i.e., a number of pallets can simultaneously traverse the system) and each pallet can be 
either empty or loaded with a PCB. At each point in time a pallet is associated with a given destination and the RTS 
allows for a number of possible routing solutions. 
The next destination of a pallet carrying a PCB can change over time as operations are executed (e.g., by the test station, 
the shredding station, the loading/unloading cell). The new destination is available only at execution time. 

The \gecko\ proposal was to realize a distributed control infrastructure composed by a {\em community} of autonomous 
agents \cite{etfa14} able to cooperate in order to define the paths the pallets must follow to reach their destinations.
Thus, these paths are to be computed at runtime, according to the actual status and the overall conditions of the shop floor, 
i.e.. no static routes are used to move pallets. The decisions of the coordination algorithm \cite{ijcim15} act as goal injection 
for the planning mechanism of each agent.  Hence according to our pursued abstraction, the plant is a set of TMs endowed 
with independent capabilities to carry on their goals, by analyzing the current situation, synthesizing a planning domain and 
problem, then planning and executing the plan for such goals.

It is worth observing that a plan-based controller can endow an agent with the desired autonomy (i.e., deliberative capabilities), 
but given the particular dynamic nature of RTSs it does not guarantee a continuous control process capable to face all the 
particular situations/configurations. Indeed it is not easy (or hardly possible) to capture all the dynamics of the production 
environment in a unique planning domain. 
The specific capabilities of a TM in the RTS are affected by many factors, e.g., a partial failure of the internal elements of a TM, 
a reconfiguration of the RTS plant or maintenance activities of some TMs of the plant. Thus, it is not always possible to design 
a plan-based controller which is able to efficiently handle all these situations. The higher is the complexity of the planning 
domain the higher is the time needed to synthesize the plans and the latency of the control architecture must be compatible 
with the latency of the plant.

Thus, the key direction in \gecko\ project has been the one of endowing the plan-based controller of a TM (\ie\ an agent) with 
a knowledge reasoning mechanism capable to build the actual state of the production context by dynamically inferring the 
actual capabilities of the TM with respect to the detected configuration of the plant. In this way, the plan-based controller can
automatically generate and continuously maintain updated the timeline-based model of the TM according to the inferred 
knowledge.

\subsection{The Use of Ontologies in Manufacturing}
In robotics and more generally in manufacturing, the use of ontologies is crucial to improve the adaptability and 
the flexibility of classical approaches \cite{turaga2008machine}. In several works, ontologies have been exploited 
to design more autonomous, flexible, adaptive and proactive artificial agents. Since researchers have applied 
ontologies to solve or at least mitigate a variety of problems, applications differ in their assumptions and goals.

In \cite{omrkf-2007}, a Robot knowledge framework (OMRKF) is exploited, OMRKF contains a series of ontology 
layers, includinga robot-centered and a human-centered ontology. Beside a perception layer, needed for the sensory 
data, the system is composed by an object layer (model), a context layer and an activity layer. The framework lacks of a 
foundational approach as can be seen in the object classification where, for example, the "living room" is classified as 
a space region and not as the role of the region (the problem becomes clear by observing that a region of space is 
fixed while the living room can be located in different parts of the building at different times, and can even disappear 
from the building).

Relatively to the connection between the KB and the planning module, the work \cite{hartanto2008fusing} exploits a 
model filtering approach based on a Hierarchical Task Network (HTN). The agent's knowledge of the environment is 
stored in a fixed ontology and some filters on this knowledge are set up. Given a planning task, the system selects 
one of the filters to isolate a suitable subset of the system's knowledge and uses this subset to constrain the plan 
by deleting non-reachable constants. While this technique can be efficient in terms of plan adaptation, the knowledge 
is only filtered, thus cannot be augmented nor modified, not even contextualized to the specific problem. 

Other research projects, like KnowRob \cite{tenorth-beetz-2009} and ORO \cite{lemaignan-et-al-2010} focus on learning and 
symbol grounding and use ontologies for obtaining an action-based knowledge representation able to support cognitive
functionalities. At the ontological level, these knowledge systems show problems similar to those discussed earlier (\eg\ 
functionality is confused with activity so that it is not possible to "discover" new ways to perform a function). 


%
%
%
%
%
%
\section{Knowledge and Plan-based Control in a Loop}
\label{subsec:loop}
\begin{figure}[ht]
\centering
\includegraphics[width=.9\textwidth]{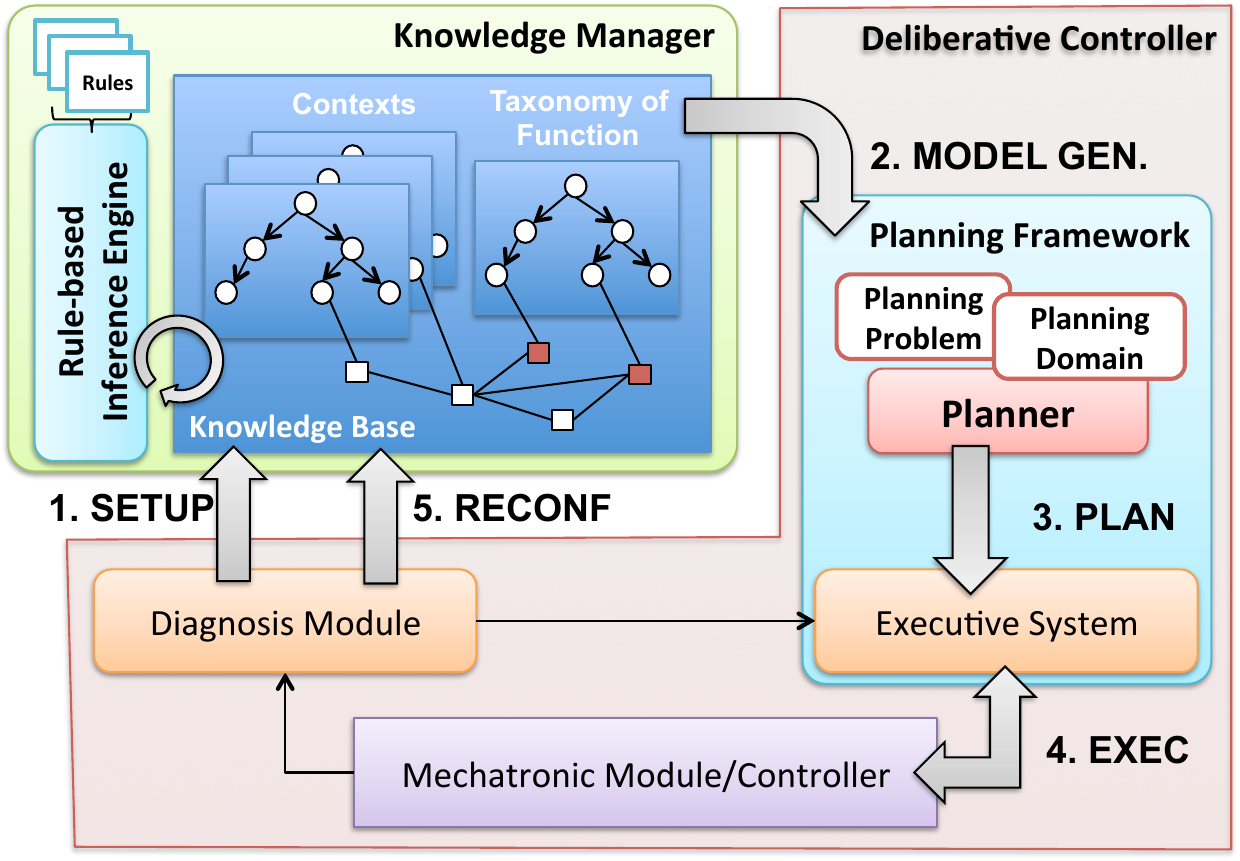}
\caption{\small{The Knowledge-based Control Loop}}
\label{fig:kb-loop}
\end{figure}

The {\em Knowledge-based Control Loop} (\kbcl) represents the envisaged flexible control architecture which integrates 
a knowledge processing mechanism with planning and execution in order to dynamically generated and adapt  
the timeline-based model needed to actually control an agent (\ie\ a TM of the plant in the \gecko\ project). 
Figure \ref{fig:kb-loop} shows the key integration of distinct cognitive functions composing  the architecture. At a higher 
abstraction, the figure shows the integration of two "big boxes" called here {\em Knowledge Manager}, that contains the 
know-how of the agent, and {\em Deliberative Controller} that represents the \epsl-based controller which plans and 
executes the activities of the agent. To make the whole idea operational we need to open the boxes and describe what 
is needed to allow the two functionalities to work together.

The goal of \kbcl\ is to have a coherent and continuous flow of information from the {\em Knowledge Manager} to the 
{\em Deliberarive Controller} and to extend the capabilities of the overall system by exploiting reasoning capabilities.
Following a careful analysis of the reasoning needs, the {\em Knowledge Manager} relies on a suited ontology which models the 
general knowledge of manufacturing environments. 
The ontology contains (i) a classification of relevant information in three distinct {\em Contexts} -- namely {\em Global}, 
{\em Local} and {\em Internal} (see later) -- and (ii) a {\em Taxonomy of Functions} which classifies the set of functions the 
agents can perform according to their effects in the environment (see later).

The {\em Knowledge Manager} exploits the ontology to build and manage the KB of the particular agent to control. 
The KB represents an abstract description of the structure and the capabilities of the agent and also of the production 
environment (from the agent's point of view). Namely, the KB represents the "instantiation" of the general knowledge to the 
particular agent to control. In this context the Rule-based Inference Engine is a specific module which is responsible for 
processing KB information by inferring additional knowledge about the functional capabilities the agent is actually 
able to perform (see later for further details).
Thus, given a TM of the RTS of the case study, the KB contains information concerning the devices that 
compose the TM (e.g. the cross transfers, the conveyor engines, the port sensors), the set of other TMs and/or working 
machines directly connected (i.e. the set of collaborators) and information concerning the whole production 
environment from the agent perspective (e.g. the topology of the shop floor). Then the {\em Inference Engine} analyzes the 
structure of the TM and its collaborators in order to add to the KB information about the set of transportation functions 
the TM is actually able to perform.

The {\em Planning Framework} provides timeline-based deliberative features relying on the planning model generated from 
the KB to actually control the mechatronic device. More specifically, it is a wrapper of the planning and execution system 
employed to provide deliberative capabilities. 
It is responsible to integrate KB information with planning by automatically generating the model of the mechatronic device 
to control. Indeed, planning domain and problem specifications are dynamically generated from the KB and an off-the-shelf 
planning and execution system is activated to synthesize signals for the actuators that control the mechatronic device.
The {\em Mechatronic Module} is the composition of a {\em Control Software} and a {\em Mechatronic Component} 
(e.g., a transportation module, a working machine, etc.).  In our case the control software is based 
on standard reference models (e.g., IEC61499) and each mechatronic component is then represented by dedicated 
hardware/software resources encapsulating the module control logic. 

The \emph{Knowledge-based Control Loop} (KBCL) represents the overall process which allows the integration of the 
elements described above in a unique control infrastructure. The resulting control process enables an agent to dynamically 
represent its capabilities, the detected environmental situation and to infer the set of available functions on which a coherent 
planning model is generated.

\subsection{The Knowledge-based Control Loop at Runtime}
The management of the KB, the generation of the planning domain, the continuous monitoring of the 
information concerning the agent and the environment, represent the rather complex activities the \kbcl\ process 
must properly manage at runtime. In this regard, the \kbcl\ process consists of the following phases: (i) the {\em setup 
phase}; (ii) the {\em model generation phase}; (iii) the {\em plan and execution phase}; (iv) the {\em reconfiguration phase}.

The {\em setup phase} generates the KB of the agent by processing the raw data received from the {\em Mechatronic device}
through a {\em Diagnosis Module}. The resulting KB completely describes the structure of the particular module to control,
the set of TMs the module can cooperate with and the set of functions the module is actually able to perform in order 
to support the production flow. Then, the {\em model generation phase} exploits the KB of the agent to 
generate the timeline-based planning model the {\em Deliberative Controller} needs to actually control the device. 

When the planning domain is ready the {\em planning and execution phase} starts, and the {\em Deliberative Controller}
continuously builds and executes plans. During this phase the \kbcl\ process behaves like classical plan-based control 
systems. The {\em Planning Framework} builds the plan according to some tasks to perform. {\em Soft changes} in 
the plan execution are directly managed by the {\em Deliberative Controller}, \eg\  temporal delays of some 
planned activities. Conversely whenever the {\em Diagnosis Module} detects a structural change of the agent 
and/or of its collaborators \eg\ a total or partial failure of a cross transfer of the TM to control (\ie\ {\em Hard 
changes}), the {\em reconfiguration phase} starts. 

The reconfiguration phase determines a new iteration of the \kbcl\ process cycle. The KB of the agent is updated 
by detecting the new state of the mechatronic device and its production environment as well as inferring the updated set of 
functions the TM can perform according to the new state. As before, once the KB of the agent is complete, the planning model 
of the {\em Deliberative Controller} is also updated with respect to the new state of the module.
It is worth observing that the KB and the planning model are updated only when structural changes that impede the execution 
of the plan are detected.

\section{Modeling Knowledge with Ontology and Contexts}
In a changing environment the agents must coherently share information relevant to the tasks. Thus an ontological analysis 
allows to build reliable systems that exploit different information types and contexts. The aimed generality lead to a structure a 
that is neither tailored to a specific type of agent nor to a specific type of situation. It is not based on an information
model at the enterprise or shop floor level nor developed for some specific type of action.  
The result is a general mechanism to {\em dynamically generate} a high-level description of agent's capabilities and system's 
situations. 

The first result of this analysis is the separation of two layers of information: organizational knowledge and factual knowledge. 
The organization knowledge is the foundational knowledge, i.e., the knowledge about the basic assumptions in the domain like 
the notion of object, agent, production, etc., including their relationships. This knowledge fixes what kind of entities, events and 
interactions there can be in general. Factual knowledge, instead, identifies how the actual scenarios is, out of all the possible 
configurations: which objects are presents and where, which actions are executed and by which agent, which changes occur 
and to which object. Factual knowledge can be extended (without changing the foundational knowledge) as needed, e.g., to 
include knowledge about new devices (tools, machines) or changes in the shop floor layout. Changes in these two parts of the 
knowledge framework follow different principles and have different consequences. By keeping them apart, we can make them 
interoperate covering all the knowledge needed in the production systems \cite{Chandrasegaran2013evolution}.

For the organizational knowledge the proposed approach relies on the foundational ontology \dolce\, the Descriptive Ontology 
for Linguistic and Cognitive Engineering \cite{masolo2002wonderweb}. This is a domain-independent top-level ontology
that has been exploited at different levels in the engineering and industrial domains, e.g.,
\cite{Borgo2014ontological,prestes2013towards,Borgo2004role}.
\dolce\ furnishes the basic structure of the knowledge the \kbcl\ relies on and it will be enriched with domain knowledge, for 
instance adding the notions of artificial agent and engineering function. The knowledge framework available to an agent, 
will be an extension of this ontological system. Since \dolce\ is based on a first-order language with formal semantics, the
ontology and the resulting knowledge base can be exploited via automatic reasoning

\subsection{The \dolce\ Ontology}
The \dolce\ ontology is a formal system built according to an explicit set of philosophical principles that guide its use and 
extension \cite{masolo2002wonderweb}. \dolce\ focuses on particulars, as opposed to universals. Roughly speaking, a 
universal is an entity that is instantiated or concreted by other entities (like the property "being a tool" or "being a production 
process"). A {\em particular}, an element of the category \textsc{particular}, is an entity that is not instantiated by other entities
(like Eiffel Tower in Paris or Barack Obama). \textsc{particular} includes physical entities, abstract entities, events and even 
qualities as shown below.

\begin{figure}
\centering
\includegraphics[width=\textwidth]{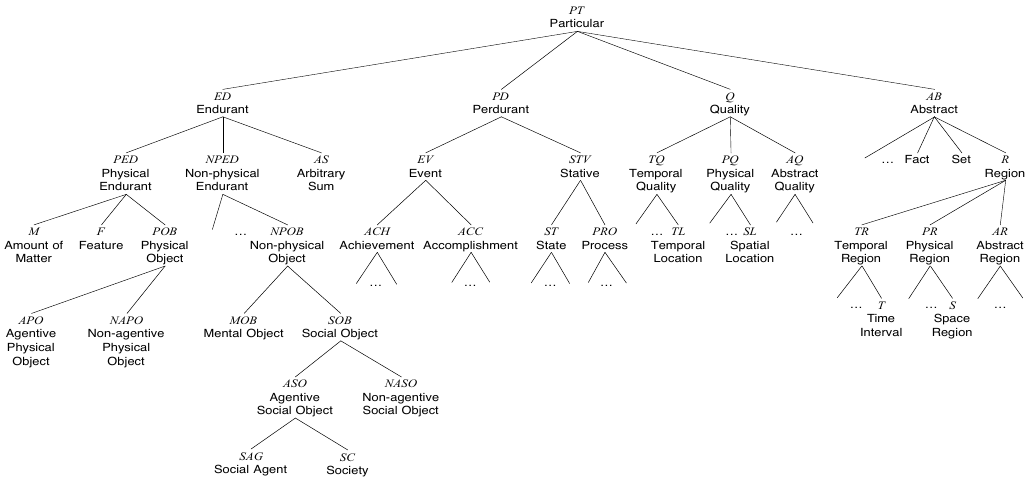}
\caption{\small{The \dolce\ taxonomy of particulars}}
\label{fig:dolce}
\end{figure}

The \dolce\ ontology formalizes the distinction between things like a car and an organization (this category is called 
\textsc{endurant}), and events like transporting by means of a car and resting (category \textsc{perdurant}), see \ref{fig:dolce}.
The term "object" is used in the ontology to capture a notion of unity as suggested by the partition of the category 
\textsc{physical endurant} (a subcategory of \textsc{endurant}) into categories \textsc{amount of matter}, like the plastic
with which a water bottle is made, \textsc{physical object}, like a car, and \textsc{feature}. Features are entities that 
existentially depend on other objects, e.g., a bump on a road or the workspace for a robotic arm. There are other 
two subcategories of \textsc{physical object}, namely, \textsc{agentive physical object}, e.g. a person, and 
\textsc{non-agentive physical object}, e.g., a drill.

\dolce\ also provides a structure for individual qualities (elements of the category \textsc{quality} like the weight of a given 
car), quality types (weight, color and the like), quality spaces (spaces to classify weights, colors, etc.), and quality positions
or {\em qualia} (informally, locations in quality spaces). These, together with measure spaces (where the quality positions 
get associated to a measure system and to numbers), are important to describe and compare devices and processes. The 
exact list of qualities may depend on the entity: {\em shape} and {\em weight} are usually taken as qualities of physical 
endurants, {\em duration} and {\em direction} as qualities of perdurants. An individual quality, e.g., the weight of an hammer, is 
associated with {\em one and only one} entity; it can be understood as the particular way in which the hammer instantiates
the general property "having weight". That individual weight quality is what can be measured when the hammer is put 
on a scale (if we put another hammer, no matter how similar, another individual quality would be measured, i.e., that of the 
second hammer even if the scale indicates exactly the same value). The change of an endurant in time is explained in \dolce\ 
through the change of some of its individual qualities. For example, with the substitution or damaging of a component, the value 
of the weight quality of a car may change.

\dolce's taxonomic structure is depicted in Figure \ref{fig:dolce}. Each node in the graph is a category of the ontology. A category
is a subcategory of another if the latter occurs higher in the graph and there is an edge between the two. \textsc{particular} is
the top category. The direct subcategories of a given category form a partition. In the graph, dots indicate that not all the 
subcategories of that category are listed. Some relations are particularly relevant in this context, e.g., the {\em parthood} 
relation: "x is part of y" (written: \textsc{P(x,y)}), with its cognates the {\em proper part} (written: \textsc{PP(x, y)})
and {\em overlap} relations (written: \textsc{O(x, y)}).
It applies to pairs of endurants (e.g., the joint is part of the robotic arm) as well as to paris of perdurants (e.g., riveting is part
of the assembling process). On endurants parthood has an additional temporal argument since and endurant may loose or gain
parts throughout its existence (e.g., after substituting a switch in a radio, the old switch is not part of the radio). Another important
relation is constitution, indicated by \textsc{K: K(x, y, t)} stands for "entity x constitutes y at time t", e.g., the amount of iron x 
constitutes the robot y at time t (this relation allows to say that part or all the iron x may be substituted over time without changing
the identity of robot y like when substituting a worm component).

%
%
%
%
\subsection{Ontological interpretation of Agents and their Environment}
Recently there has been an increasing interest in the ontological modeling of artificial agents, and robots in particular 
\cite{prestes2013towards}, which led to an IEEE standard (ORA -- Ontologies for Robotics and Automation). 
Today's approaches to robot modeling are interesting but further work si needed. For instance, it is unsatisfactory to take 
the characterization "being a robot" as a role (this is the choice in the IEEE standard ORA) since this implies that robots 
are such only when active, i.e., they appear and disappear by switching them on and off.
While this avoids the problem of characterizing the essence of robots, the choice goes against common-sense. Robots do not 
seem to qualify as agentive entities in the strong sense since they lack intentional states, and it is dubious if they even qualify 
in the weak sense in most cases they have only conventional stimulus-response behavior.
Up to today, any attempt to draw the line between artefactual tools and robots has met important criticisms.
The following sections propose an extension of the \dolce\ ontology to include robots, robotic parts and tools. The goal of this 
extension is to start from the notions of artefact and of agent, as introduced in foundational ontologies, and to propose a way
to descriminate among types of artefacts as needed to model industrial scenarios.

Ontological speaking, following the analysis in \cite{borgo2009artefacts}, a robot is an artefact: it is intentionally 
selected (via construction) and has attributed technical capacities. Technical capacities can vary considerably depending on the 
robots: they can be quite 
limited, like in ant robots, or flexible and multipurpose like in industrial or humanoids robots. Since the focus is on industrial 
settings, thus on robotic arms, transportation modules and the like, the modeled robots are actually technological artefacts
\cite{borgo2014technical}: they are manufactured by following precise production plans and selected via dedicated quality tests. 
Thus, from the formal viewpoint industrial robots can be classified as (technological) artefacts i.e., elements of the 
\textsc{artefact} subcategory of \textsc{non-agentive physical object} \cite{borgo2009artefacts}.

The typical robots in the production scenarios are rational, reactive and may present some degree of autonomy. Today, they are 
rarely adaptive and embedded although these are desirable features. They can also be proactive: they have goals, typically 
provided by the production system to which they belong, and can sometimes choose, or at least reschedule, plans to optimize
their achievements. In short, these robots are artefacts whose behaviors resemble agents' behavior for the same goal(s). Since
this behavior is expected from them, we propose to see a robot as an artefact whose attributed quality is to {\em behave 
agent-like}. It is important to point out that this modeling choice keeps agents and robots apart: a member of  the latter group
just mimics agents. The behavior can range from basic stimulus-response actions to activities controlled by sophisticated 
planning and goal adaptations, depending on what kind of agentivity the robot can behaviorally simulate. This is 
definitely acceptable for today's robots and it does not exclude that future generations of robots might be considered as 
full-fledge agents.

The rest of the section refers to robots as agents. The symbol \textsc{robot} is used for the predicate "being a robot"
and {\em BehSp} for the generic space of behaviors. Specifically, using the language of \dolce\ from 
\cite{masolo2002wonderweb,borgo2009foundational,borgo2009artefacts}, it is possible to formally model the ontological 
status of robots as follows:

\begin{equation}
\label{eq:robot1}
\textsc{robot}(r) \rightarrow\ \textsc{artefact}(r)
\end{equation}

\begin{equation}
\label{eq:robot2}
\begin{aligned}
\textsc{robot}(r) \land\ AttribCap(a) \land\ & \\
qt(a, r) \land\ ql(v, a, t) \rightarrow\ & Loc(v, BehSp)
\end{aligned}
\end{equation}

The first formula says that a robot is an artefact. The second states what distinguishes a robot from other artefacts: the capacity
attributed to the robot ({\em AttribCap(a)} $\land$ {\em qt(a,r)}) has values ({\em ql(v, a, t)}) that belong to the space of behaviors 
({\em Loc(v, BehSp)})\footnote{The existence of quality {\em a} is enforced by formula \ref{eq:robot1} and the theory 
\cite{borgo2009artefacts}. The characterization of the space of behaviors is stil under investigation}.

Robot's parts are themselves artefacts, thus elements of the \textsc{artefact} category. These are typically not robots, so their
attributed qualities are of different types. The main distinction here is between the parts that are components, i.e. that 
constitute the robot like the engines that move the robotic arm structure and the structural pieces that are moved by the 
engines; and the parts that are tools used by the robot like the different types of gripper that can be substituted depending on 
the task to execute. These types of parts are isolated for their functional or structural contribution. There are, of course, also 
arbitrary parts like the upper half of the skeletal frame, which do not have special properties or functionalities and thus are 
not relevant in terms of knowledge and planning.

Components (tools) can be in an active/inactive (available/non-available) state for the robot. Sensors are listed among the
components but the proposed characterization does not distinguish between sensors and actuators since these are
seen as roles of the agent's components (a drill can play both of them at the same or at different times). Finally, an object that 
is a component is such until substituted (or dismantled) while a tool may remain such even if substituted.

In the case of agents, the {\em environment} represents the area of interest in which the agent could act. For artificial agents, the
environment might also include the requirements and specifications about the software components and their development. 
Since the reasoning  mechanism deals with languages and software constraints in terms of contexts, the considered notion 
of environment focuses on the notion of location. 
Thus, at each point in time, the robot's environment is described in terms of robot's location including 
the elements the location contains plus entities that, even though not in the location, can interact (positively or negatively) with 
the robot's activities and goals.

This view is fairly general and assumes that the environment depends on the robot's features as well as on the features of the 
other entities. It is important to point out that the environment can change whenever the robot or its location or the entities there 
change. In the case of production scenarios, the robot's environment can be identified with the collection of physical entities that 
are within a certain range from the robot (where the range may be bounded by physical barriers like floor, walls, ceiling, fences, 
etc). The environment is not necessarily limited to a precise region of space; it includes also entities with which the robot
can interact in some ways (e.g., via wireless communication). 
In ontological terms, the environment is a compound physical object composed by all the physical objects that are within the 
interaction range (workspace) of the robot. The location of the environment corresponds to the location of the objects in the 
environment plus the locations reachable by the robot itself~\footnote{The location is fixed for robots like robotic arms, it is 
parametric (in particular, it may depend on the task) for mobile robots}.

\subsection{Ontology and Engineering of Functions}
The classifications of the robots, the physical entities that may interact with them and their environments take care of the 
"static" part of the problem. Since a robot is supposed to act in order to reach its goals, it must also have the conceptual 
machinery to know what it can do and how, thus to plan its actions. In this regard, reasoning on (engineering) functions
is unavoidable. The formalization of functions in robotics is rarely addressed and is too often confused with the notion of 
action, i.e., the performance of a function.

\begin{figure}
\centering
\includegraphics[width=.9\textwidth]{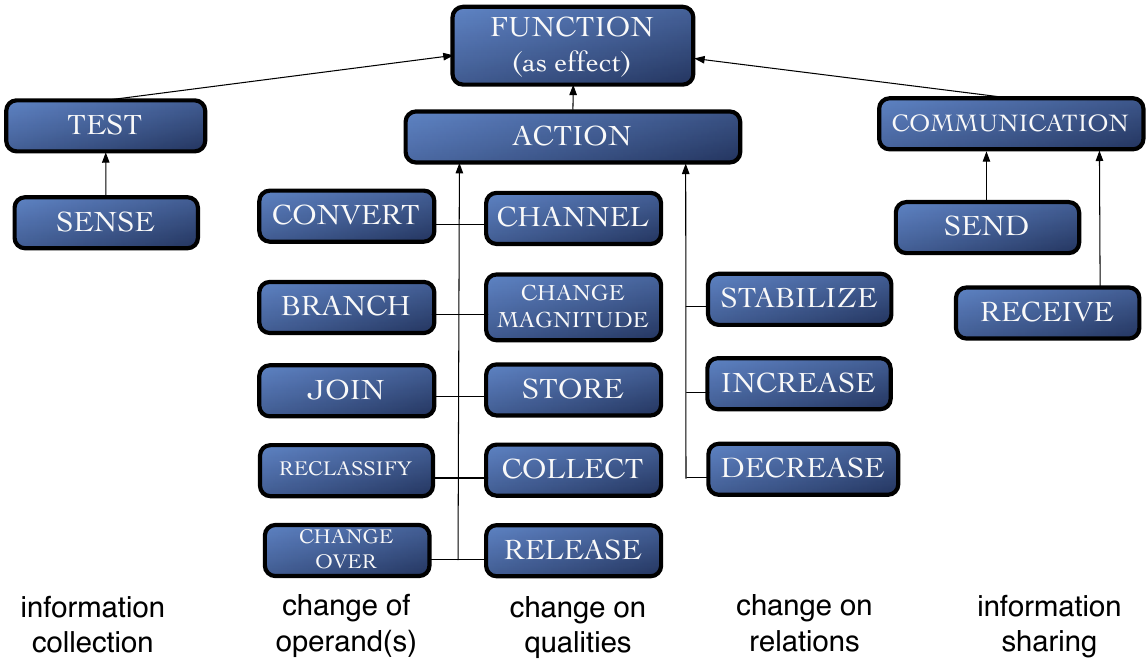}
\caption{\small{The function ontological taxonomy and its rationale}}
\label{fig:OntoFunTaxo}
\end{figure}

To overcome this problem, the proposed approach extends the \dolce\ ontology with an ontology of high-level functions. This 
function ontology is integrated, via \dolce, with the ontology of the robot and robot's parts making it possible to model what a 
robot can do and how. Specifically, the interpretation of functions relies on the notion of function-as-effect (see Figure 
\ref{fig:OntoFunTaxo}) which has been adapted borrowing from well-known functional approaches in engineering design like
the FOCUS/TX \cite{focustx} (for the distinction "what to" vs. "how to" and the notion of behavior), the Functional Basis 
\cite{pahl2007engineering,Hirtz2002} (for the idea of a function list), and the Function Representation \cite{Chandrasekaran2000} 
(for the distinction between environment-centric and device-centric function).
The guiding idea is to make it possible the identification of the high-level function (or sequence of functions) that need to be 
executed to reach a given goal. For this, it can be taken into account the difference between the actual state and the desired 
state, and identify the changes to be made. From this information, the robot can travel the taxonomy to identify the effects of
the high-level functions and find a suitable combination.

Figure \ref{fig:OntoFunTaxo} show the top-level ontological functions organized in five brenches: functions to collect information,
functions to change the operand's qualities, functions to change the qualityrelationships, and functions to share information. For 
instance, "reclassify" stands for the function to change the classification of an operand, e.g. when, after a test, a workpiece is 
classified as malfunctioning; "change-over" applies when, e.g., a robot acts on itself to activate/deactivate some component;
"channel" stands for the moving of an operand (change of its location); "stabilize" for maintaining relational parameters like when 
tuning electronic components to regulate the input-output relationship; "sense" for the operand testing function, i.e., to 
acquire information without altering the status or the qualities of the operand; finally, "send" stands for the function to output
information like a signal that a workpiece is going to be transferred or that a failure occurred.

Of course, this information is not enough since it would model just the {\em ideal} capacities of the robot. Aiming to have a 
robot adapting its plan at {\em run-time}, we have to model the actual capacities of the robot, which implies to take 
malfunctioning and/or missing parts or even deteriorated behaviors into account. This information depends on the capacities 
of self-inspection built-in in the robot as well as on the possibility to compare the ideal action's descriptions and the actual
performances.

\subsection{Context-based Characterization}
An ontology is a conceptual tool used to structure information. Ontologies deal mainly with necessary information 
like the properties that an object must manifest (shape, weight, mass, etc) or the types of event (states, actions, processes
and son on). Factual information, being information that depends on contingent data (like spatio-temporal location, agent's 
setting, goals, etc), is generally characterized at the level of knowledge-bases. While this distinction might not be fully 
justified (and not even sharp), it remains important not to structure the ontology relying on factual knowledge. This principle 
is rarely recognized in applications and in particular in the development of ontologies for industrial application.

The distinction between necessary and contingent information concerns only the development of the ontology structure: it is 
important that factual information finds its place in the factory information system. This allows the system to classify and 
reason on factual information, for example, to understand the actual scenario and possible evolutions, to evaluate optimal 
production plans out of those that are actually possible, and even to establish the status of the resources or maintenance 
schedule. To act in real and evolving scenarios, factual information is thus essential. In the proposed approach, factual 
information is included in the KB (built on top of the ontology) and is organized into main categories called {\em contexts}.
Contextualization enables to manage factual information with an ontologically sound approach. It gives also an advantage 
at the reasoning level: it allows to differentiate types of information depending on their usefulness in reasoning on a situation 
or task. After an ontological analysis based on \cite{Borgo2007,borgo2009foundational}, it is possible to identify three 
contextual models dedicated to factual knowledge, and use them with the ontological framework. In particular, these contexts 
provide the time-dependent information needed to select how to execute high-level functions in the actual scenario. 

The three context types are called global, local and internal, respectively. The {\em global context} collects information 
the agent cannot control nor modify like the shared language of the system, the agents present in the system, the 
system's performance parameters. The {\em local context} collects information on the relationship between the agent and its 
neighbor elements (typically the human and artificial agents directly interacting with it), thus providing a local view of the 
topological setting. Finally, the {\em internal context} collects the information the agent has about itself as well as its capabilities
towards itself (change-over) and towards the environment (communication and manipulation) \cite{borgo2015ontology}.

\subsection{Applying Ontology and Contexts to the Case Study}
Given a particular application like the manufacturing scenario of the case study, it is necessary to define the general 
knowledge the \kbcl\ process must deal with in order to dynamically infer the specific capabilities of an agent and adapt
the control model accordingly. Thus, the \dolce\ ontology has been extended with the type of information needed by 
applying the context-based and the functional characterization described above.

Broadly speaking, the extended ontology aims at characterizing the knowledge concerning the general {\em structure}
of a TM of the plant, the related {\em working environment} and the general {\em functional capabilities}
of TMs in such a context. This information represents the general knowledge (\ie\ the {\em TBox}) a \kbcl\ process instantiates 
according to the specific features of the TM to be controlled, in order to generate the envisaged KB of the TM (\ie\ the {\em ABox}).

\begin{figure}[ht]
\centering
\includegraphics[width=.9\textwidth]{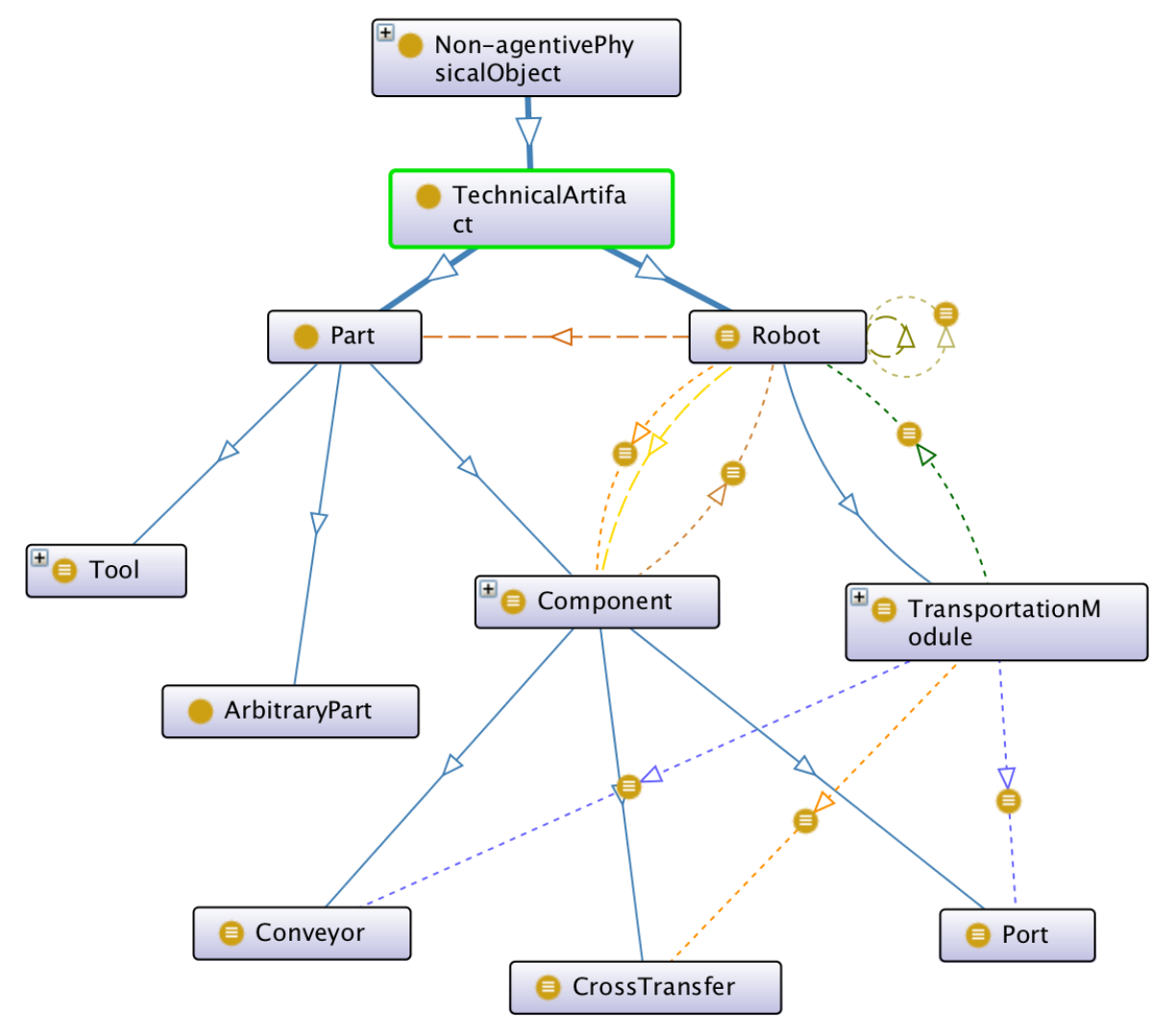}
\caption{\small{Extension of \dolce\ ontology}}
\label{fig:extDolce}
\end{figure}

Figure \ref{fig:extDolce} shows the extension of the \dolce\ taxonomy of particulars with respect to the \textsc{non-agentive 
physical object} category. According to the \dolce\ interpretation of artifacts, robots and robot's parts are modeled as 
subcategorires of \textsc{artifact}, as the taxonomy in Figure \ref{fig:dolce} shows. Robot's parts can be further distinguished 
into robot's components, \ie\ parts that constitute the structure of the robot, and tools. These entities are modeled as 
subcategories of \textsc{part}. They represent artifacts with different attributed qualities with respect to robots as artifacts.
Following \dolce\ interpretation, the taxonomy can be extended by introducing the \textsc{port}, \textsc{conveyor} and 
\textsc{cross transfer} concepts as subcategories of \textsc{component} category, the concept of \textsc{transportation 
module} as subcategory of \textsc{robot} category.

The \textsc{port}, \textsc{conveyor} and \textsc{cross transfer} categories classify the elements characterizing the {\em internal 
structure} of a TM. The \textsc{component} category collects the elements that compose a robot. These components have a 
\textsc{spatial location} within the robot structure (this would not be enforced for tools since they can be external to the robot).
Collaborating components for the Channel function must be spatially connected. In the case of the TM, the {\em internal 
structure} for this kind of functionality is determined by the connections of the components' locations. The choice of modeling
the elements of a TM with different categories rather then using the general \textsc{component} category, relies on the 
different properties these elements bring to implement {\em functional capabilities} (as it will be described in the next sections).
The \textsc{port} category models the structural elements that allow a TM to {\em connect} with other TMs in its local contexts. 
These elements have a {\em communication capacity} which allows a TM to {\em send} and {\em receive} pallets to and 
from other TMs of the plant. The \textsc{conveyor} category models the engine elements that allow a TM to move pallets. They
have a {\em channel capacity} which allows TMs to actually move a pallet between two {\em spatial locations} connected via 
that component. The \textsc{cross transfer} category models engine elements that allow a TM to change its physical 
configuration. They have a {\em change over capacity} which allows a TM to change its internal connections and enable 
the different paths the pallets can follow (internally).

\begin{figure}
\centering
\includegraphics[width=.9\textwidth]{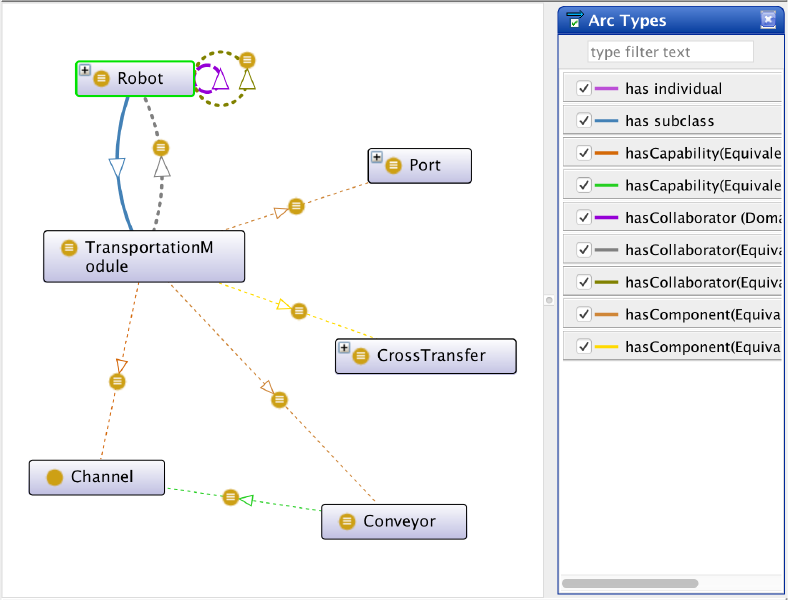}
\caption{\small{The general class axiom for the \textsc{transportation module} category}}
\label{fig:axiom}
\end{figure}

The \textsc{transportation module} category characterizes TMs from a functional point of view. Namely, TMs are modeled as
elements fo the \textsc{robot} category that can perform {\em some} \textsc{channel} functions and that have as components 
{\em some} elements of the \textsc{port} category, {\em some} elements of the \textsc{conveyor} category and {\em some} 
elements of the \textsc{cross transfer} category. 
In the manufacturing environment considered, elements of the \textsc{channel} category are functions that classify changes 
in the {\em spatial location} quality of an operand (i.e., a {\em pallet}). The execution of such a function changes the location
of the pallet from the {\em start location} to the {\em end location}. 
The Figure \ref{fig:axiom} shows a graphical representation of the general class axiom defining the \textsc{transportation 
module} category. 

The {\em (working) environment} of a TM is described in terms of the available {\em collaborators}. A TM collaborates with 
other TMs and machines of the plant by exchanging pallets through their connected ports. Thus, the 
subset of the plant's agents that are directly connected to the TM and with which the TM can actually exchange pallets, 
constitutes the {\em environment} of the TM. In such a context, a collaborator is a relative concept which depends on the 
particular {\em configuration} of the TM considered. It represents a relationship between a TM and the directly connected 
agents. Thus, the concept of \textsc{collaborator} is modeled as a role that an agent, e.g., another TM, plays according 
to its {\em local connections}.

%
%
%
%
%
\section{The Knowledge-Base Life Cycle}
The Knowledge Manager module (KM) in Figure \ref{fig:kb-loop} is responsible for managing the lifecycle of the KB within
the \kbcl\ process. In the specific manufacturing case study, the KB models the particular TM to be controlled by 
specifying its internal structure, its connections with other TMs and the related functional capabilities. The management
of the KB relies on a {\em knowledge processing mechanism} implemented by means of a {\em rule-based inference engine}
which leverages a set of {\em inference rules} to build the KB of the agent.

\begin{figure}
\centering
\includegraphics[width=.9\textwidth]{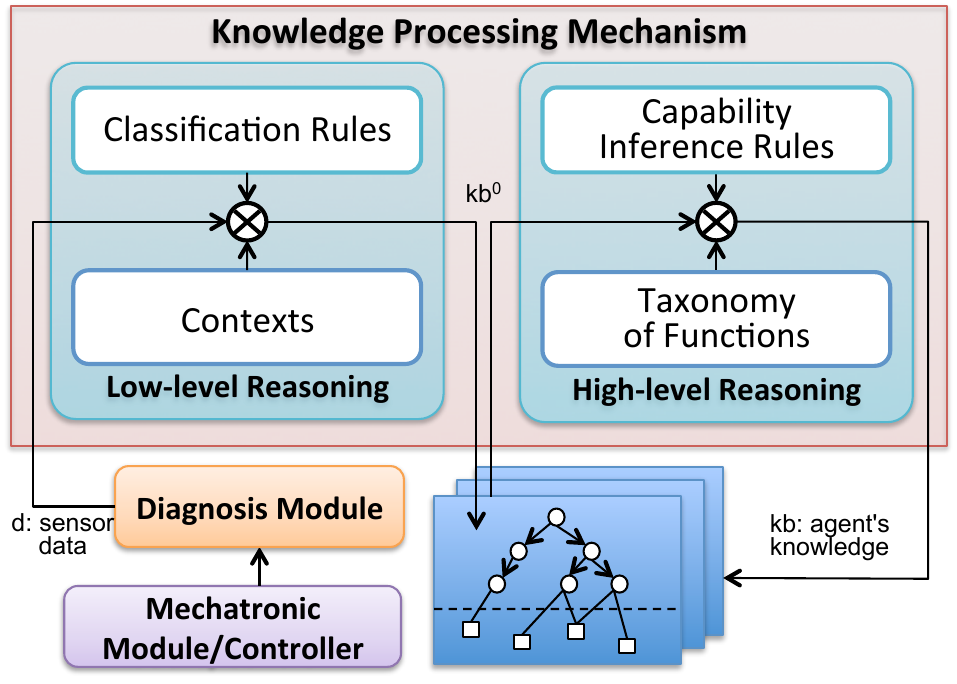}
\caption{\small{The knowledge processing mechanism}}
\label{fig:kb-procflow}
\end{figure}

The knowledge processing mechanism dynamically builds the KB elaborating {\em raw data} received from the 
{\em Diagnosis Module} and infers knowledge concerning the structure, the {\em working environment} and the functional
capabilities of the agent. As Figure \ref{fig:kb-procflow} shows, this mechanism involves two reasoning steps: the (i) the 
{\em low-level reasoning step}, and the (ii) {\em high-level reasoning step}. Specifically, these two steps iteratively refine 
the KB by combining a set of dedicated {\em inference rules} with the general knowledge of contexts and functions 
of the described ontology.

The first reasoning step, called the {\em low-level reasoning}, aims at characterizing the TM in terms of the components
that actually compose the module (e.g., the ports, conveyors, etc.) and its collaborators. It leverages the internal and 
local contexts of the ontology as well as the {\em classification rules} to generate the initial instance of the KB which 
describes the structure of the agent and the related {\em working environment}. Thus, this initial KB describes the 
agent in terms of its internal and local contexts.

The second reasoning step, called the {\em high-level reasoning}, starts from the KB elicited after the previous step and 
generates further knowledge concerning the functional capabilities of the agent. Specifically, the {\em high-level reasoning}
step relies on the {\em taxonomy of functions} and a set of domain-dependent inference rules, called {\em capability 
inference rules}, to complete the knowledge processing mechanism. The KB the {\em high-level reasoning} starts from, 
encodes the particular internal and local context of the agent. The inference mechanism can infer the set of functions the 
agent can actually perform by analyzing its structure and its working environment.

The output of the second reasoning step (and the overall knowledge processing mechanism), is the {\em final} KB which 
encodes a complete description of the structure of the agent, an {\em interpretation} of the {\em working environment} 
from the agent perspective and a description of the related {\em functional capabilities} of the agent. Such knowledge is 
then exploited in the \kbcl\ process to generate the {\em plan-based control model}. The next two subsections provide a 
more detailed discussion of the two reasoning steps constituting the knowledge processing mechanism.

\subsection{The Low-Level Reasoning Step}
The {\em low-level reasoning step} is responsible for inferring information concerning the internal and local contexts of 
the TM. Namely, the result of this inference step is an initial KB describing the operating devices that compose the TM
(i.e., the components) and the available collaborators. It builds an initial version of the KB by classifying data received
from the {\em Diagnosis Module} on the basis of contexts categorization.

\begin{figure}
\centering
\includegraphics[width=.9\textwidth]{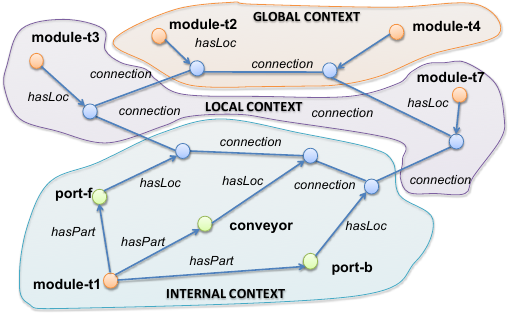}
\caption{\small{Raw data received from the {\em Diagnosis Module}}}
\label{fig:rawdata}
\end{figure}

Input data consists of a set of {\em individuals} representing information about the parts that compose the TM, their 
connections and their capabilities. Figure \ref{fig:rawdata} provides a (partial) graphical representation of a possible set
of individuals (the nodes of the graph) and predicates (the edges of the graph) the knowledge processing mechanism 
receives from the {\em Diagnosis Module}. In particular, the figure shows the different contexts the individuals belong to, 
the reasoning step leverages to provide these data with additional semantics.

Given this set of data, the first rule the {\em low-level reasoning step} applies, aims at identifying the set of operative
parts the TM can actually use to perform functions. These set of operative parts are represented as TM's {\em components}.
According to the ontological interpretation of the \textsc{component} category, a component is a structural part of a robot
which has an {\em operative state} and may have some functional capabilities. The rule follows this {\em functional 
interpretation} of components and therefore, can be formally defined as follows:

\begin{equation}
\label{eq:component}
\begin{aligned}
\textsc{robot}(r) \land\ P(p, r) \land\ & \\
hasCapacity(p, f) \rightarrow\ & \textsc{component}(p)
\end{aligned}
\end{equation}

\noindent
where $P(p, r)$ is a predicate asserting that the part $p$ is a structural element of robot $r$ and 
$hasCapacity(p, f)$ is a predicate asserting that the part $p$ has the functional capacity of performing {\em some} 
function $f$. According to the ontology, being $p$ a structural part of a robot $r$, with the capability of performing some 
function $f$, it is possible to {\em infer} that $p$ is an element of the \textsc{component} category. Consequently, the 
predicate \textsc{component}(p) is true.

The applied ontological approach models the different types of components that may compose a TM as Figure 
\ref{fig:extDolce} shows. These components are modeled according to the different types of functional capabilities
they have. Leveraging this interpretation, it is possible to define two additional rules that infer the specific type of 
component a particular part represents by taking into account the type of the associated functional capability:

\begin{equation}
\label{eq:conveyor}
\begin{aligned}
\textsc{robot}(r) \land\ P(p, r) \land\ & \\
hasCapacity(p, f) \land\ \textsc{channel}(f) \rightarrow\ & \textsc{conveyor}(p)
\end{aligned}
\end{equation}

\begin{equation}
\label{eq:port}
\begin{aligned}
\textsc{robot}(r) \land\ P(p, r) \land\ & \\
hasCapacity(p, f) \land\ \textsc{communication}(f) \rightarrow\ & \textsc{port}(p)
\end{aligned}
\end{equation}

The rules \ref{eq:conveyor} and \ref{eq:port} infer {\em conveyor} and {\em port} components as structural parts of 
a robot, that have {\em channel} and {\em communication} capabilities respectively.

\begin{figure}
\centering
\includegraphics[width=.9\textwidth]{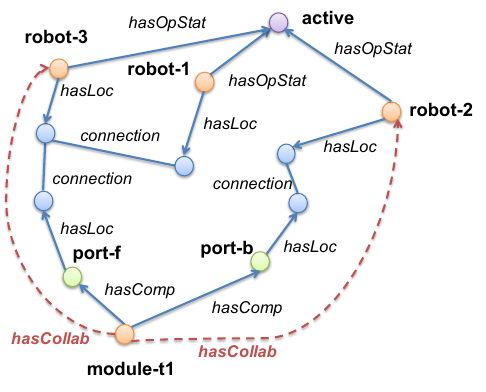}
\caption{\small{Inferring collaborators of a TM}}
\label{fig:infer-collab}
\end{figure}

Given the components of the TM, the {\em low-level reasoning step} is completed by inferring the set of {\em collaborators} 
available. Similarly to components, the collaborators of TM are directly connected TMs of the plant that are in an operative
state and therefore, can actually exchange pallets with the TM. The rule that allow to infer this information can be formally
defined as follows:

\begin{equation}
\label{eq:collaborator}
\begin{aligned}
\textsc{robot}(r) \land\ \textsc{port}(p) \land\ & \\
hasLoc(p, l_p) \land\ P(p,r) \land\ & \\
hasOpStat(p, active) \land\ \textsc{robot}(c) \land\ & \\
hasLoc(c, l_c) \land\ connection(l_p, l_c) \rightarrow\ & hasCollab(r, c)
\end{aligned}
\end{equation}

\noindent
where $connection(l_p, l_c)$ is a predicate asserting that the location of the TM's port $p$ is connected with the robot
$c$. Figure \ref{fig:infer-collab} provides a (simplified) graphical representation of a possible KB resulting from the 
application of rule \ref{eq:collaborator} (the dotted arrows represent the inferred properties concerning collaborators).

\subsection{The High-Level Reasoning Step}
\begin{figure}
\centering
\includegraphics[width=.9\textwidth]{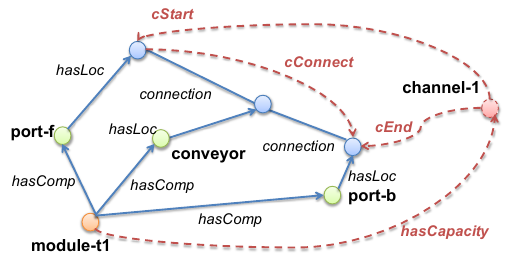}
\caption{\small{Inferring {\em primitive} channels of a TM}}
\label{fig:infer-pchannel}
\end{figure}

The {\em high-level reasoning step} extends the KB elicited at the previous step  to infer the actual capabilities of the 
TM on the basis of its current status and the current production environment.
Given the information concerning components and collaborators of a TM, the first rule the {\em high-level reasoning step}
applies, aims at inferring the {\em primitive channels} the TM can perform according to its internal structure. Indeed, 
operative components can be used by a robot to perform functions. With regard to TMs, a (operative) conveyor allows a 
TM to perform channel functions. According to this interpretation it is possible to define a rule to infer the set of 
{\em primitive channels} a TM can perform as follows:

\begin{equation}
\label{eq:primitive-channel}
\begin{aligned}
\textsc{robot}(r) \land\ \textsc{conveyor}(c_1) \land\ & \\
hasOpStat(c_1, active) \land\ \textsc{component}(c_2) \land\ & \\
\textsc{component}(c_3) \land\ hasLoc(c_1, l_1) \land\ & \\
hasLoc(c_2, l_2) \land\ hasLoc(c_3, l_3) \land\ & \\
connection(l_2, l_1) \land\ connection(l_1, l_3) \rightarrow\ & \textsc{channel}(f) \land\ \\ 
& hasCapacity(r, f) \land\ \\
& cStart(f, l_2) \land\ \\
& cEnd(f, l_3) \land\ \\
& cConnected(l_2, l_3)
\end{aligned}
\end{equation}

\noindent
where \textsc{conveyor}$(c_1) \land\ hasOpStat(c_1, active)$ asserts that $c_1$ is a conveyor component
of the TM whose operative state is {\em active}. Namely, the conveyor $c_1$ is an operative component of the Tm
and therefore, it can be actually used to perform functions. Figure \ref{fig:infer-pchannel} shows a (simplified)
graphical representation of the KB resulting from the application of rule \ref{eq:primitive-channel}. In particular, 
the figure represents predicates (the dotted arrows) and the individual (the "channel-1" onde) inferred and 
added to the KB.

The rationale of rule \ref{eq:primitive-channel} relies on the {\em functional interpretation} of the \textsc{conveyor}
category as the set of components that {\em can perform channel functions}. Thus, if an operative conveyor component
connects two components of the TM through its {\em spatial location}, see the clause $connection(l_2, l_1) \land\ 
connection(l_1, l_3)$ in \ref{eq:primitive-channel}, then the conveyor can perform a {\em primitive channel} function 
between the components' locations. Moreover, the $cConnect(l_2, l_3)$ is a transitive predicate which allows to 
"connect" and compose different channel functions. Indeed, if two spatial locations are connected through 
the $cConnect$ predicate, it means that there exists a composition of {\em primitive channel functions} that connect them.

A {\em primitive channel} involves components of a TM only. However, the channel capabilities the knowledge processing 
mechanism aims at inferring are those that involve the collaborators of a TM. Namely, {\em channel functions} that allow a 
TM to exchange pallets with other TMs of the plant. Such channels are called {\em complex channels} and can be inferred
by applying the following rule:

\begin{equation}
\label{eq:complex-channel}
\begin{aligned}
\textsc{robot}(r) \land\ \textsc{robot}(rc_1) \land\ & \\
\textsc{robot}(rc_2) \land\ hasCollab(r, rc_1) \land & \\
hasLoc(rc_1, rl_1) \land\ hasCollab(r, rc_2) \land\ & \\
hasLoc(rc_2, rl_2) \land\ \textsc{port}(c_1) \land\ & \\
hasOpStat(c_1, active) \land\ hasLoc(c_1, l_1) \land\ & \\
\textsc{port}(c_2) \land\ hasOpStat(c_2, active) \land\ & \\
hasLoc(c_2, l_2) \land\ connection(l_1, rl_1) \land\ & \\
connection(l_2, rl_2) \land\ cConnect(l_1, l_2) \rightarrow\ & \textsc{channel}(f) \land\ \\
& hasCapacity(r, f) \land\ \\
& cStart(f, rl_1) \land \\
& cEnd(f, rl_2)
\end{aligned}
\end{equation}

A key point of the rule \ref{eq:complex-channel} is that a {\em complex channel function} is interpreted as the 
composition of {\em primitive channels} the TM can perform internally. This is a quite flexible and general interpretation
of a {\em channel function}. If one or more parts of a TM stop working (i.e., their operational status changes from 
{\em active} to {\em inactive}), then the TM will not be able to perform the related {\em primitive channels} and therefore, the 
{\em high-level reasoning step} will not be able to infer all the {\em complex channels} that depends on these parts.
Similarly, if new components are added to the TM then, the {\em high-level reasoning step} will be able to inter additional
{\em primitive} and also {complex channels} according to the resulting structure of the TM.

\begin{figure}
\centering
\includegraphics[width=.9\textwidth]{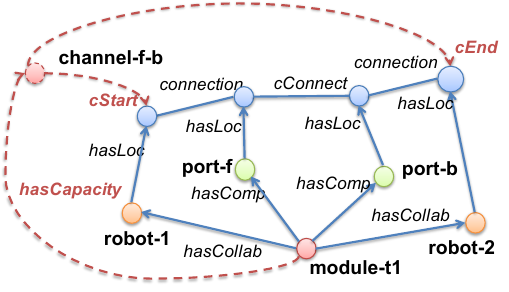}
\caption{\small{Inferring {\em complex} channels of a TM}}
\label{fig:infer-cchannel}
\end{figure}

\section{Generating the (Timeline-based) Control Model}
A key role for the dialogue between the {\em Knowledge Manager} and the {\em Deliberative Controller} is 
played by the {\em Model Generation} process (Step 2 in Figure \ref{fig:kb-loop}). The KB resulting from the 
{\em knowledge processing mechanism} provides an abstract representation of the capabilities, the structure 
and the working environment of the agent. The {\em model generation process} analyzes the KB to generate 
the related control model as a timeline-based planning specification of the agent.

The {\em model generation process} encodes the hierarchical modeling methodology described in 
Chapter \ref{chap:epsl} and builds the control model by leveraging the context-based characterization of the KB. 
The information concerning the {\em global context} and the {\em taxonomy of function} define the 
{\em functional state variables} that provide a {\em functional view} of the agent as a whole. These state 
variable describe the high-level tasks the agent can perform over time. 
The {\em internal context} contains structural information about the agent and therefore it is suited to 
generate the {\em primitive state variables}. These variables describe the behaviors of the physical/logical 
features that compose the agent. 
Usually the values of this type of variables directly correspond to states or actions the related domain 
features may assume or perform over time. 
The {\em local context} manages information concerning the {\em working environment} of the agent and 
therefore it is suited to build the set of {\em external variables} of the model. These variables 
model the {\em collaborating agents} (\ie\ the directly connected TMs of the plant) whose behavior may 
affect the capabilities of the agent, even if not directly controllable.

%
\begin{algorithm}
\small
\caption{\small{The \kbcl\ procedure for generating the planning model}}
\label{alg:kbcl-main}
\begin{algorithmic}[1]
\Function{buildControlModel}{$KB$}
	\State // extract agent's information and initialize the \ps\ model
	\State $agent \gets\ getAgentInformation\left(KB\right)$
	\State $model \gets\ inititalize\left(KB, agent\right)$
	\State // define components of the model
	\State $svs \gets\ buildFunctionalComponents\left(KB, agent\right)$
	\State $svs \gets\ buildPrimitiveComponents\left(KB, agent\right)$
	\State $svs \gets\ buildExternalComponents\left(KB, agent\right)$
	\State // build the set of task decomposition rules
	\State $S \gets\ buildSynchronizationRules\left(KB, agent\right)$
	\State // update the \ps\ model
	\State $model \gets\ update\left(model, svs, S\right)$
	\State \Return $model$
\EndFunction
\end{algorithmic}
\end{algorithm}

Algorithm \ref{alg:kbcl-main} describes the overall procedure of the generation process. Broadly speaking, 
the procedure consists of four specific procedures that analyze different {\em areas} of the knowledge about 
the agent in order to generate different parts of the control model.
The procedure starts by extracting information related to the agent and initializing the \ps\ model (rows 3-4). 
According to the hierarchical timeline-based approach described in Chapter \ref{chap:epsl},  a set of functional, 
primitive and external state variables is generated (rows 6-8). Finally, the hierarchical decomposition of 
functional values (\ie\ values of functional state variables) is described by means of a suitable set of 
generated synchronization rules (row 10). The resulting timeline-based model is then composed and returned 
as the outcome of the procedure (row 12-13).

Thus, the {\em buildControModel} procedure allows the {\em model generation process} to automatically 
build the timeline-based specification by leveraging the knowledge about the agent. 
As described in \cite{icaps16}, every time a change occurs in the KB, a new instance of the {\em model 
generation process} is triggered in order to generate an updated control model of the agent.
The next subsections provide some details about the (sub)procedures the {\em model generation process} 
relies on, and provide an example of a possible timeline-based control model that can be generated for a 
TM of the plant case study.

\subsection{Building State Variables from Contexts}
Algorithms \ref{alg:kbcl-functional}, \ref{alg:kbcl-internal}, \ref{alg:kbcl-external} below describe the information 
extraction procedures that allow the {\em model generation process} to build the functional, primitive and external 
state variables respectively.

\subsubsection*{Building Functional Variables}
Algorithm \ref{alg:kbcl-functional} describes the procedure which builds the {\em functional state variables} of the 
timeline-based control model. The {\em buildFunctionalComponents} procedure generates the set of state 
variables concerning the functional capabilities of the agent. The procedure relies on the set of {\em capabilities} 
the {\em knowledge processing mechanism} has inferred by applying rules \ref{eq:primitive-channel} and 
\ref{eq:complex-channel}. The procedure generates a state variable for each function of the taxonomy 
(see Figure \ref{fig:OntoFunTaxo}) the agent can perform. Namely, given a particular function $f$ of 
the taxonomy, if the KB contains at least one {\em individual} for that function $f$ (i.e., if the {\em knowledge 
processing mechanism} has inferred at least one way for the agent to perform $f$), then a state variable $sv$ 
for $f$ is added to the model. The {\em individuals} of $f$ in the KB represent all the possible implementations 
of $f$ the agent can perform (i.e., all the {\em capabilities} of the agent with respect to $f$). 
Thus, for each {\em inferred} individual of $f$ the procedure adds a value to the related (functional) state 
variable $sv$.

\begin{algorithm}
\small
\caption{The \kbcl\ sub-procedure for generating functional variables}
\label{alg:kbcl-functional}
\begin{algorithmic}[1]
\Function{buildFunctionalComponents}{$KB$, $agent$}
	\State // initialize the list of functional variables
	\State $svs \gets\ \emptyset$
	\State // get types of functions according to the Taxonomy in the KB
	\State $taxonomy \gets\ getTaxonomyOfFunctions\left(KB\right)$
	\ForAll{$function \in\ taxonomy$}
		\State // check if the KB contains individuals of function
		\State $capabilities \gets\ getCapabilities\left(KB, agent, function\right)$
		 \If {$\neg\ IsEmpty\left(capabilities\right)$}
		 	\State // create functional variable
			\State $sv \gets\ createFunctionalVariable\left(function\right)$
			\State // add a value for each "inferred" capability
			\ForAll {$capability \in\ capabilities$}
				\State $sv \gets\ addValue\left(sv, capability\right)$
			\EndFor
			\State // add created state variable
			\State $svs \gets\ addVariable\left(svs, sv\right)$
		 \EndIf
	\EndFor
	\State \Return $svs$
\EndFunction
\end{algorithmic}
\end{algorithm}

The procedure first initializes the set of functional state variables of the domain (row 3). 
Then, it reads the taxonomy of function from the KB and, for each function, checks  the available 
capabilities of the agent (rows 6-20). Given a function, if the KB contains at least one capability for that 
function, then the procedure creates a functional state variable (rows 9-11). Each capability found in the 
KB is modeled as a value of the related state variable (rows 12-15). 
The procedure ends by returning the set of obtained variables.

\subsubsection*{Building Primitive Variables}
Algorithm \ref{alg:kbcl-internal} describes the procedure which builds the {\em primitive state variables} 
of the timeline-based control model. The {\em buildPrimitiveComponents} procedure generates the set 
of state variables concerning the structural components of the agent.
The procedure relies on a {\em functional interpretation} of components as elements that allow the 
agent to perform functions. Thus, according to the inference rules \ref{eq:component}, \ref{eq:conveyor} 
and \ref{eq:port}, the components of an agent are modeled in terms of their {\em capabilities}. 
The procedure adds a {\em primitive state variable} to the model for each component found in 
the KB. According to the inference rule \ref{eq:primitive-channel}, the values of these variables 
represent the {\em primitive functions} of the agent.

\begin{algorithm}
\small
\caption{The \kbcl\ sub-procedure for generating primitive variables}
\label{alg:kbcl-internal}
\begin{algorithmic}[1]
\Function{buildPrimitiveComponents}{$KB$, $agent$}
	\State $svs \gets \emptyset$
	\State // get agent's operative components
	\State $components \gets\ getActiveComponents\left(KB, agent\right)$
	\ForAll{$component \in\ components$}
		\State // check if component can perform some functions
		\State $capabilities \gets\ getCapabilities\left(KB, component\right)$
		\If {$\neg\ IsEmpty\left(capabilities\right)$}
			\State // create primitive variable for component
			\State $sv \gets\ createPrimitiveVariable\left(component\right)$
			\State // check component's functional capabilities
			\ForAll {$capability \in\ capabilities$}
				\State $sv \gets\ addValue\left(sv, function\right)$
			\EndFor
			\State $svs \gets\ addVariable\left(svs, sv\right)$
		\EndIf
	\EndFor
	\State \Return $svs$
\EndFunction
\end{algorithmic}
\end{algorithm}

Algorithm \ref{alg:kbcl-internal} first initializes the set of primitive state variables of the domain (row 2). 
Then, the procedure reads the set of the {\em inferred components} from the KB (row 4). Given a component,
if the KB contains at least one {\em primitive function} the agent can perform through that component, 
then a {\em primitive variable} is created (rows 5 -10). 
The values added to the variable model the capabilities of the related component. Namely, the values model
the primitive functions the agent can perform by means of the considered component (rows 11-16). 
The procedure ends by returning the set of generated state variables.

\subsubsection*{Building External Variables}
Algorithm \ref{alg:kbcl-external} describes the procedure which builds the {\em external state variables} 
of the timeline-based control model. The {\em buildExternalComponents} procedure generates the set 
of state variables concerning the collaborators of the agent. 
The procedure generates a set of state variables representing the {\em collaborators} of the agent. 
Specifically, a state variable is created for each {\em individual} found in the KB that, according to the 
inference rule \ref{eq:collaborator}, has been classified as {\em collaborator}.
The values of these state variables represent the {\em operative states} the collaborators may assume 
over time.

\begin{algorithm}
\small
\caption{The \kbcl\ sub-procedure for generating external variables}
\label{alg:kbcl-external}
\begin{algorithmic}[1]
\Function{buildExternalComponents}{$KB$, $agent$}
	\State $svs \gets \emptyset$
	\State // get agent's collaborators
	\State $collaborators \gets\ getCollaborators\left(KB, agent\right)$
	\ForAll {$collaborator \in\ collaborators$}
		\State // create an external variable to model the collaborator
		\State $sv \gets createExternalVariable\left(collaborator\right)$
		\State // model the possible behaviors of collaborators
		\State $states \gets\ getOperativeStates\left(collaborator\right)$
		\ForAll {$state \in states$}
			\State $sv \gets\ addValue\left(sv, state\right)$
		\EndFor
		\State $svs \gets\ addVariable\left(svs, sv\right)$
	\EndFor
	\State \Return $svs$
\EndFunction
\end{algorithmic}
\end{algorithm}

The procedure first initializes the set of external variables of the domain (row 2). 
Then, the procedure reads the set of inferred collaborators from the KB (row 4). 
For each collaborator found, a state variable is created (rows 5-7) and for each operative state
the collaborator may assume over time, a value is added to the created variable (rows 9-14). 
The procedure ends by returning the set of generated variables.

\subsection{Building Decomposition Rules from Inference Trace}
When all the state variables and their values have been generated, it is necessary to build the {\em synchronization 
rules} of the domain in order to coordinate the temporal behavior of the state variables and achieve the desired goals.
The {\em buildSynchronizationRules} procedure generates the decomposition rules by 
leveraging the {\em inference trace} of the KB. The {\em inference trace} represents {\em intermediate knowledge}  
generated by the application of inference rules. Such knowledge manages {\em intermediate information} 
necessary to complete the {\em knowledge inference mechanism} and therefore build the KB.
For instance, besides {\em primitive channels}, the inference rule \ref{eq:primitive-channel} generates 
{\em cConnect} properties. These properties do not represent specific information about the agent but are 
necessary to generate the set of {\em complex channels}, as shown in the inference rule \ref{eq:complex-channel}.
These properties encode functional dependencies among the components of a TM. In particular, 
they encode these dependencies in terms of {\em primitive channels} needed to {\em implement complex channels}. 

The inferred {\em cConnect} properties can be analyzed in order to build a particular 
data structure, called {\em functional graph}, that correlates functional dependencies among components, 
primitive and complex channels. The graph is built according to the inferred {\em cConnect} properties. 
Thus, the possible {\em implementations} of complex channels can be found by traversing the functional 
graph. This set of information is necessary to build the set of {\em synchronization rules} specifying how
the agent must {\em execute} complex channels. Indeed, synchronization rules are generated by 
analyzing the paths on the functional graph that connect the {\em start} with the {\em end} locations of
complex channels. These paths can be easily expressed in terms of {\em precedence constraints} between
primitive channels of the involved components.

\begin{algorithm}
\small
\caption{The \kbcl\ sub-procedure for generating synchronization rules}
\label{alg:kbcl-synch}
\begin{algorithmic}[1]
\Function{buildSynchronizationRules}{$KB$, $agent$}
	\State $rules \gets\emptyset$
	\State // create the functional graph for channel functions
	\State $graph \gets\ buildChannelFunctionalGraph\left(KB, agent\right)$
	\State // get inferred complex channels 
	\State $channels \gets\ getChannels\left(KB, agent\right)$
	\ForAll{$channel \in\ channels$}
		\State // get available implementations
		\State $implementations \gets\ getImplementation\left(graph, channel\right)$
		\ForAll{$implementation \in\ implementations$}
			\State // create synchronization rule from implementation
			\State $rule \gets\ createSynchronizationRule\left(KB, implementation\right)$
			\State $rules \gets\ addRule(rule)$
		\EndFor
	\EndFor
	\State \Return $rules$
\EndFunction
\end{algorithmic}
\end{algorithm}

Algorithm \ref{alg:kbcl-synch} describes the procedure for building the synchronization rules of the timeline-based
domain with respect to the (complex) channel function the related TM can perform. The procedure first initializes the 
set of rules (row 2) and then analyzes the KB to generate the {\em functional graph} concerning channel functions (row 4).
For each complex channel, the procedure extracts the available implementations from the 
functional graph (rows 6-9). Each implementation encodes a set of temporal constraints between the primitive 
channels of the agent. Thus, given a possible implementation of a complex channel, a new synchronization rule 
is created and added to the model (rows 10-14). The procedure ends by returning the set of generated synchronizations.

\subsection{The Resulting Timeline-based Control Model}
The procedures that have been described in the previous sections encode the model generation process 
which relies on the context-based characterization of the KB. According to this structure, the process 
generates a {\em hierarchical domain specification} modeling the complex functions of the agent in 
terms of the primitive functions that internal components can directly handle according to the status of 
the involved collaborators.

\begin{figure}[ht]
\centering
\includegraphics[width=.95\textwidth]{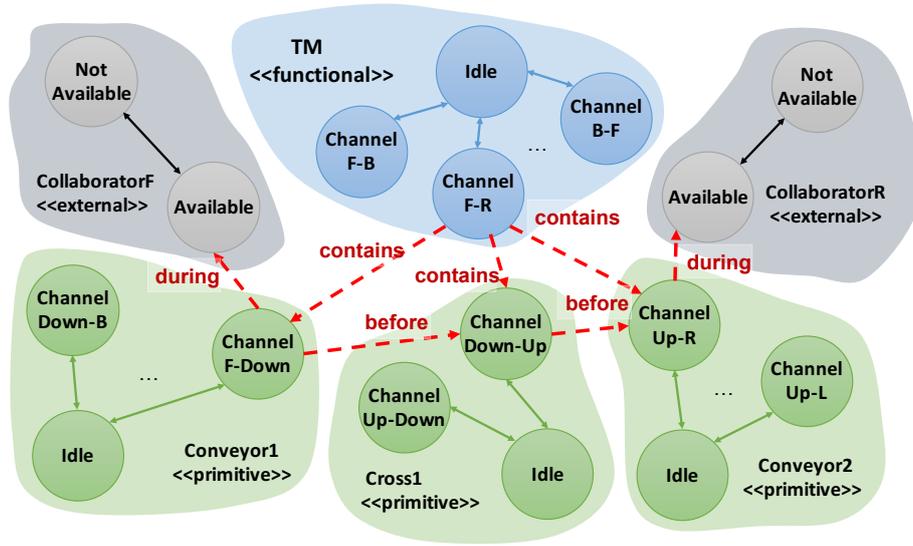}
\caption{\small{A (partial) view of the timeline-based model generated for a TM equipped with one cross-transfer unit only}}
\label{fig:planning-model}
\end{figure}

Figure \ref{fig:planning-model} shows a (partial) example of a timeline-based control model generated 
for a TM composed by one cross-transfer unit only. The model provides a functional characterization of 
the TM according to functional, primitive and external levels cited above. 
The primitive state variables model the {\em active} parts of the TM that can actually perform some 
(primitive) functions. These state variables model the functional capabilities of elements that compose 
the TM.
For example, the component {\em Conveyor1} can perform the primitive channel {\em ChannelF-Down} 
to move a pallet between the location of component {\em PortF} and location {\em Down} of 
component {\em Cross1}. Similarly, the component {\em Cross1} can perform the primitive channel 
{\em ChannelDown-Up} to move a pallet from the location {\em Down} to the location {\em Up} 
of the same component {\em Cross1}.
The external state variables model the inferred collaborators that can directly interact with the 
considered TM. The values of 
these variables  represent the operative states that collaborators may assume over time. 
Figure \ref{fig:planning-model} shows the external state variables concerning two of four collaborators 
available. Specifically, the state variables model the behaviors of {\em CollaboratorF} and {\em CollaboratorR} 
\ie\ the collaborators {\em connected} to the TM through components {\em PortF} and {\em PortR} 
respectively.
The functional state variables model the inferred channel functions the TM can perform by combining 
internal (\ie\ primitive) channel functions. For example, according to this interpretation, {\em ChannelF-R} 
can be seen as the composition of the following primitive channels:

\[
\overbrace{\underbrace{\text{ChannelF-Down}}_\text{Conveyor1} 
\circ\underbrace{\text{ChannelDown-Up}}_\text{Cross1} 
\circ\underbrace{\text{ChannelUp-R}}_\text{Conveyor2}}^\text{ChannelF-R}
\]

Such a composition represents a particular {\em implementation} of the {\em ChannelF-R} function. 
Implementations are modeled by means of synchronization rules that specify a suited set of 
{\em temporal constraints} (the red arrows in Figure \ref{fig:planning-model}).
These temporal constraints encode also the functional dependencies between the TM and its collaborators.
Indeed, {\em CollaboratorF} and {\em CollaboratorR} must be available during the execution of the 
{\em ChannelF-R} function.
The generated timeline-based planning model provides a functional characterization of TMs of the plant 
where planning goals represent functions the considered TM must performpri. These functions are 
described in terms of the {\em atomic} operations (\ie\ primitive functions) the TM is capable to perform
 by means of its components and its available collaborators.

\section{The Knowledge-Based Control Loop in Action}
This section reports on a set of tests on the KBCL with different TM configurations. All the different physical configurations of a 
TM have been considered, from zero to three cross-transfer modules. These configurations are referred to as {\em simple}, 
{\em single}, {\em double} and {\em full}, respectively. Each configuration also entails a different number of connected 
TM neighbors. 
Clearly, the more complex scenario is the one with the highest number of cross-transfers (the full TM) and neighbors.  Also, 
three reconfiguration scenarios ({\em reconf-a, reconf-b} and {\em reconf-c}) have been developed considering different 
external events, namely an increasing number (from 1 to 3) of TM neighbors momentarily unable to exchange pallets, plus 
two scenarios related to internal failures ({\em reconf-d} and {\em reconf-e}) due to a cross-transfer engine failure and to a 
local failure on a specific port. 

The experiments were carried out to evaluate the performance of the following aspects of a TM: (i) the knowledge 
processing mechanism; (ii) the planning model generation; (iii) the synthesis of plans to manage a set of pallet requests. The 
final aim is to evaluate the feasibility of the KBCL approach by showing that the performance are compatible with execution 
latencies of the RMS\footnote{All the experiments have been performed on a workstation endowed with an Intel Core2 Duo 
2.26GHz and 8GB RAM}.

\begin{figure}[ht]
\centering
\includegraphics[width=\textwidth]{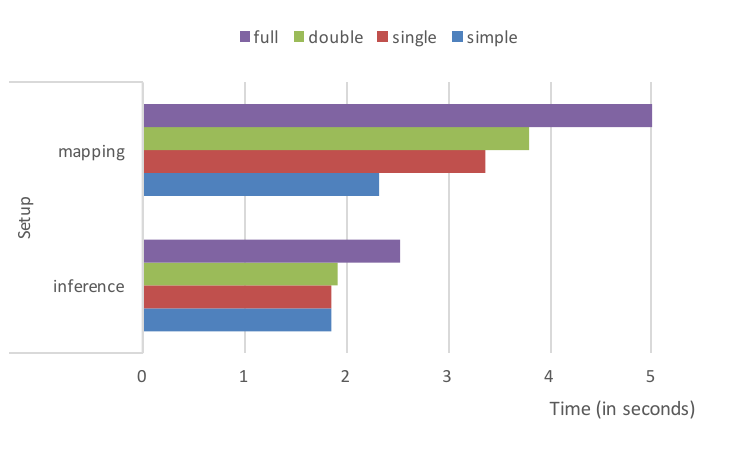}
\caption{\small{KB initial inference and planning domain generation}}
\label{fig:exp-inference}
\end{figure}

Figure \ref{fig:exp-inference} shows the timings in the Setup phase for the KBCL module operation, \ie\ to build the 
KB exploiting the classification and capability inference process (the "inference" side of Figure \ref{fig:exp-inference}), and to generate 
the timeline-based planning specification for the TM (the "mapping" side of Figure \ref{fig:exp-inference}).
On the one hand, the results show that an increase in the complexity of the TM configurations does not entail a degeneration of 
the knowledge processing mechanism: the inference costs are almost constant (around 1.3 secs).
This behavior was expected since the number of instances/relationships in the KB is rather low notwithstanding the 
physical configuration of the TM; thus, the performance of the inference engine deployed here is not particularly affected. 
On the other hand, the model generation is strongly affected spanning from 0.8 secs in the simple configuration, up to a 
maximum of 2.2 seconds in the full configuration.
The model generation process entails a combinatorial effect on the number of instances/relationships needed to generate 
components and synchronizations leading to larger planning models and, thus, to higher process costs.

\begin{figure}[ht]
\centering
\includegraphics[width=\textwidth]{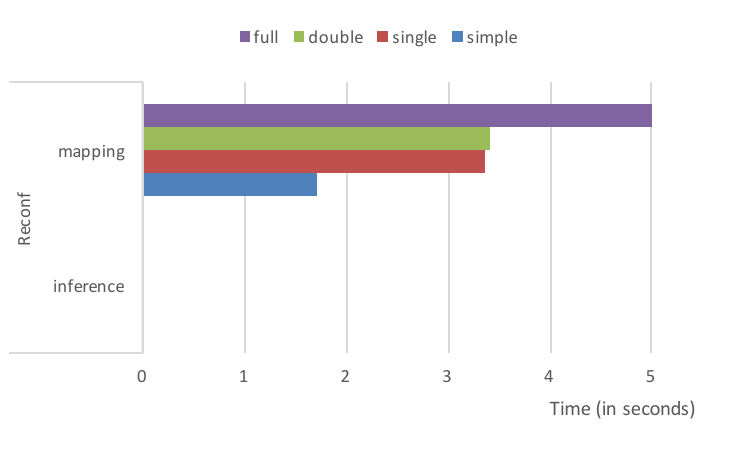}
\caption{\small{KB inference and planning domain generation during \kbcl\ reconfiguration phase}}
\label{fig:exp-inference-reconf}
\end{figure}

When a reconfiguration scenario occurs, the knowledge processing costs are negligible. Among all the considered 
reconfiguration cases (i.e., reconf-a-b-c-d-e), the time spent by the knowledge processing mechanism to (re)infer the 
enabled functionalities is just a few milliseconds. In fact, both the classification and capability inference steps are applied to a 
slightly changed KB and, then, minimal changes in the functionalities can be quickly inferred and represented 
in the new KB. 
For what concerns the planning model generation after a reconfiguration, each reconfiguration scenario (both external and 
internal) leads to a strong reduction of functionalities and, thus, the related costs are relatively small. 
Figure \ref{fig:exp-inference-reconf} shows the time spent to generate the planning model in the full TM configuration (i.e., the 
more complex configuration) is depicted and compared with respect to the time spent in the setup phase. In general, the time 
needed to regenerate the planning model specification is also dependent on reconfiguration scenarios but still negligible.

\begin{figure}[ht]
\centering
\includegraphics[width=\textwidth]{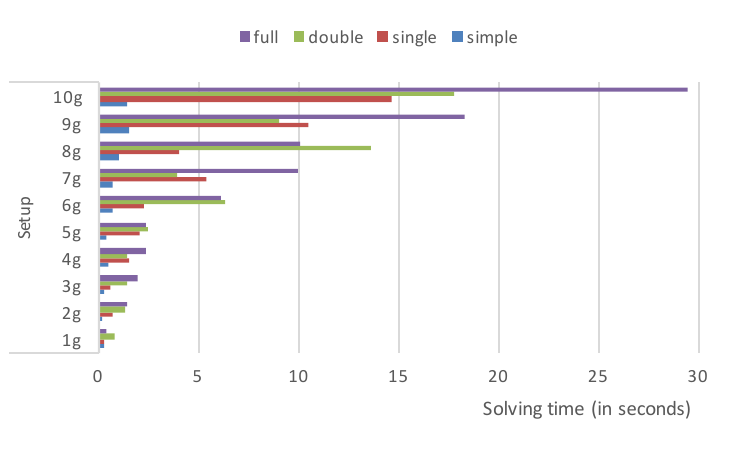}
\caption{\small{Deliberation time with increasing number of goals and different TM configurations during \kbcl\ setup phase}}
\label{fig:exp-planning}
\end{figure}

Finally, we evaluate the planning costs when facing both setup and reconfiguration scenarios.Figure \ref{fig:exp-planning} 
shows the trend of the planning time in the Setup scenario considering all the TM configurations  and an increasing number 
of pallet requests (randomly generated), i.e., planning goals, to be fulfilled.
Planning costs span from few seconds up to nearly 30 seconds when planning for 10 pallet requests within a 15 minutes 
horizon. In general, the more complex the planning model, the harder the plan synthesis problem. Thus, the planning costs 
follow the complexity of the configurations of the specific TM agent. 

The experimental results show the practical feasibility of the KBCL approach in increasingly complex instances of a 
real-world manufacturing case study. The collected data for the initialization (or the update) of a generic agent's KB (considering 
both knowledge processing and model generation) and the cost for planning synthesis have a low impact on its performance 
during operation. In fact, in order to face production periods of 15 minutes --and the management of 10 pallet requests-- no more 
than 5 seconds are required by the Knowledge Manager while less than 30 seconds are required by the Planner to generate 
a suitable plan.  Such performances are compatible with the system latency usually involved in this type of manufacturing 
applications. 

\subsection*{Implementation Notes}
Most of the inferences at runtime are done in the Web Ontology Language (OWL)
version of the KB to exploit primarily the contextual classification and relationships. The ontology editor 
\protege\footnote{http://protege.stanford.edu} has been used fo KB design and testing. For runtime reasoning within the 
Knowledge Manager, the Ontology and RDF APIs and Inference APIs provided by the Apache Jena Software 
Library\footnote{http://jena.apache.org} has been used.


%


\graphicspath{{chapters/9_conclusion/figures/}}


%
%
%
%
%
\chapter{Concluding Remarks}
\label{chap:conclusions}
\lettrine[lines=2]{T}{his thesis} has presented a complete characterization of the 
timeline-based approach ranging from a formalization of timeline-based planning to planning 
and execution of timelines by taking into account temporal uncertainty. After the first 
introductory chapters, Chapter \ref{chap:formalization} presented the formalization which 
defines a clear semantics of the main planning concepts like timelines, state variables, 
plans and goals, and taking into account the domain controllability features. 
Chapter \ref{chap:epsl} introduced \epsl\ a general framework for planning and execution 
with timelines which complies with the formalization and, is therefore capable of dealing with 
temporal uncertainty. The effectiveness of the envisaged approach has been shown by 
applying \epsl\ to real-world manufacturing scenarios within the research projects \fbt\ and 
\gecko\ as described in Chapter \ref{chap:hrc} and \ref{chap:kbcl} respectively. 
Specifically, the \fbt\ project has shown how the envisaged timeline-based approach and the 
\epsl\ capabilities of dealing with temporal uncertainty both at planing and execution time, as 
well as the hierarchical approach, represent an effective solution for controlling a robot in 
scenarios where temporal uncertainty plays a relevant role, like Human-Robot Collaboration 
requiring a tight interaction between a controllable agent (\ie\ the robot) and an uncontrollable 
agent (\ie\ the human). The \gecko\ project has shown some promising results concerning the 
design of a flexible control architecture capable of dynamically inferring the control model by 
integrating knowledge reasoning techniques with timeline-based planning.

The main concern all along this work was not to design the most performing planning algorithm ever
made, but rather the objective was to design and develop new and effective solutions for real-world scenarios.
Thus, the key point has been to understand the features and the problems that must be considered and solved in order to
effectively apply these techniques in real-world applications. For this reason, the importance of a flexible control system 
capable of (robustly) managing and adapting control strategies to the uncontrollable dynamics of the environment has 
come to light within the research projects \fbt\ and \gecko. 

For example, the \fbt\ project shows that flexibility is needed to control the robot and dynamically adapt its behavior 
to the observed behavior of the human. In this case, the pursued solution consists in designing planning and execution 
applications capable of properly dealing with temporal uncertainty at different levels. From the planning point 
of view, the formal characterization of the timeline-based approach introduces the representation 
of the uncontrollable dynamics of the domain in shape of temporal uncertainty. Leveraging this formalization, the general 
hierarchy-based solving procedure of the \epsl\ framework has been extended by introducing temporal uncertainty in order 
to synthesize plans accordingly. In this way, \epsl\ can generate plans that have some desired properties 
(\ie\ the pseudo-controllability property) characterizing their (temporal) robustness at execution time. 
From the execution point of view, once a plan has been generated it must be executed. The \epsl\ framework has 
been extended by introducing executive capabilities that rely on the same representation of the planner. Thus, \epsl\ can 
execute plans by taking into account the controllability properties of the domain and dynamically adapt the execution of the 
plan to the observed behavior of the environment and the related uncontrollable features.

Plan-based control architectures and also the \epsl-based control architecture typically rely on 
a well-defined and {\em static} model of the world. The \gecko\ project shows that another type of flexibility needed in 
real-world scenarios is the capability of dynamically adapting the control model of the plan-based controller to the specific 
configuration/state of the working environment and the robot (\ie\ the agent to control). In this case, the pursued solution 
consists in designing an extended plan-based control architecture which integrates knowledge reasoning and planning. 
Semantic technologies introduce the capability of representing and reasoning about the state of the agent and the 
related working environment. Such a reasoning mechanism, embedded in the control architecture, allows for dynamically 
building and maintaining updated the knowledge concerning the actual functional capabilities of a particular agent. 
Such a knowledge is then exploited to automatically generate a control model that a plan-based controller (\eg\ an \epsl-based 
controller) utilizes to actually plan and execute operations.

\epsl\ represents the main result of this work. It represents a uniform framework for 
planning and execution with timelines under uncertainty. In addition, the application of 
\epsl\ to the research projects \fbt\ and \gecko\ has shown the effectiveness and the flexibility 
of the envisaged approach to solve real-world problems. However this is just a first step, there 
are several aspects to take into account in order to further improve the capabilities of the 
\epsl\ framework.

A short-term objective concerns the introduction of a flexible management of different types 
of resource in \epsl. In general, the objective is to enrich the types of domain element (in addition 
to state variables) the framework can deal with. 
The introduction of different types of resources (\eg\ {\em renewable resources} 
and {\em consumable resources}) would allows \epsl\ and the envisaged timeline-based approach to 
address more realistic problems. This implies also that the solving capabilities of \epsl\ must 
be extended in order to synthesize flexible profiles for the different types of resources considered. 
An initial idea is to leverage temporal flexibility in order to synthesize optimistic temporal profiles of 
resources \cite{laborie2003resources,cesta1997resource,drabble1994resource}.

Timeline-based planning systems, usually rely on a careful engineering of domain together with 
{\em domain-dependent} heuristics in order to control the search process. Nevertheless, there are 
different domain-independent heuristics that have successfully applied in classical planning showing
impressive results \eg\ \cite{ff,graphplan,helmert2011}. Unfortunately, the application/adaption of these 
techniques to timeline-based systems is neither simple nor possible. There are significant differences 
between the timeline-based approach and the classical approach in terms of problem representation 
and resolution that prevent a straightforward adaptation of these heuristics. Thus, an additional 
short-term objective is to investigate the design of domain-independent heuristics by borrowing 
ideas and concepts from classical as well as similar works in the literature 
\eg, \cite{bernardini2008automatically}. The hierarchy-based technique introduced in this 
work represents just an initial step towards the achievement of this research objective.

Also the comparison of \epsl\ with other existing timeline-based systems is a relevant research objective 
to pursue in the near future. More in general, the objective is to compare the timeline-based approach 
with other temporal and hybrid planning approaches. A first contribution is represented by the work
 \cite{umbrico2016steps} which provides an initial comparison between \epsl\ and \europa\ by taking 
 into account modeling and solving features of these two frameworks. The goal is to define a set of 
 {\em benchmarking} problems that can be used to compare modeling and solving capabilities of \epsl\ with 
 \europa, \ixtet\ and other relevant planning systems like \optic\ \cite{optic}, \colin\ \cite{colin}, \fape\ 
 \cite{fape}, \chimp\ \cite{chimp} and \hatp\ \cite{hatp}.

A medium-term objective is to further exploit temporal uncertainty at planning and execution time. 
With respect to planning, the objetive is to enhance the \epsl\ solving procedure in order to 
synthesize {\em dynamically controllable} plans. 
Pseudo-controllabiliy is a useful property but it does not provide enough information about the controllability of a plan. 
Indeed, pseudo-controllability is a necessary but not sufficient property for dynamically 
controllability. Thus, the idea is to further 
analyze information about the temporal uncertainty of the domain during the planning process in order to generate plans with 
more relevant properties characterizing their execution, \ie\ dynamic controllability. 
With respect to execution, the objective is to integrate the synthesis and management of execution strategies \cite{ictai13,ki11}, 
as well as validation and verification techniques \cite{ker09} in order to execute timeline-based plans in a more robust way. 
As it is, the executive takes execution decisions on the fly without reasoning on the overall plan and the observed 
behavior of the environment. The use of an {\em execution strategy} would allow \epsl\ to take more accurate decisions 
and improve the robustness of plan execution.

Finally, a long-term objective is to further investigate the integration of knowledge reasoning techniques with
planning and the automatic synthesis of control models. Specifically the idea is to realize a powerful {\em knowledge 
engineering} tool which leverages semantic technologies in order to allow users that are not expert of planning technologies
 but rather expert of the domain, to use and deploy (timeline-based) planning applications. Knowledge engineering tools provide a 
{\em standard} and expressive interface which allows users to model the particular domain abstracting from the details of 
planning and problem resolution. Then, the resulting knowledge can be exploited in order to dynamically generate the planning 
model used to actually plan and execute operations as shown in Chapter \ref{chap:kbcl}.


%
%
%
%
%
%
%
%
%
\bibliographystyle{apalike}
\bibliography{backmatter/references}
%
%
\end{document}